\documentclass[letterpaper]{article} 
\usepackage{aaai25}  
\usepackage{times}  
\usepackage{helvet}  
\usepackage{courier}  
\usepackage[hyphens]{url}  
\usepackage{graphicx} 
\usepackage{natbib}  
\usepackage{caption} 
\usepackage{algorithm}
\usepackage{algorithmic}

\usepackage[justification=centering]{subfig}
\usepackage{amsmath}
\usepackage{amssymb}
\usepackage{mathtools}
\usepackage{amsthm}
\usepackage{multirow}
\usepackage{booktabs} 
\usepackage{float}

\newcommand{\normal}{\mathcal{N}\left(0, \mathbf{I}\right)}
\newcommand{\diff}[2]{\frac{\mathrm{d} #1}{\mathrm{d} #2}}

\usepackage{newfloat}
\usepackage{listings}
\DeclareCaptionStyle{ruled}{labelfont=normalfont,labelsep=colon,strut=off} 
\floatstyle{ruled}
\newfloat{listing}{tb}{lst}{}
\floatname{listing}{Listing}
\newlength{\subcolumnwidth}

\newcommand{\nextsubcolumn}[1][]{%
  \cr\noalign{\hfill}
  \if\relax\detokenize{#1}\relax\else\hsize=#1\setlength{\subcolumnwidth}{\hsize}\fi
}

\title{Flash Diffusion: Accelerating Any Conditional Diffusion Model for Few Steps Image Generation}

\author{
    Cl\'ement Chadebec\thanks{Corresponding author: [name].[surname]@jasper.ai}, Onur Tasar , Eyal Benaroche\thanks{Work done during an internship at Jasper Research}, Benjamin Aubin
}
\affiliations{
    Jasper Research\\

\vspace{1em}
\includegraphics[width=\linewidth]{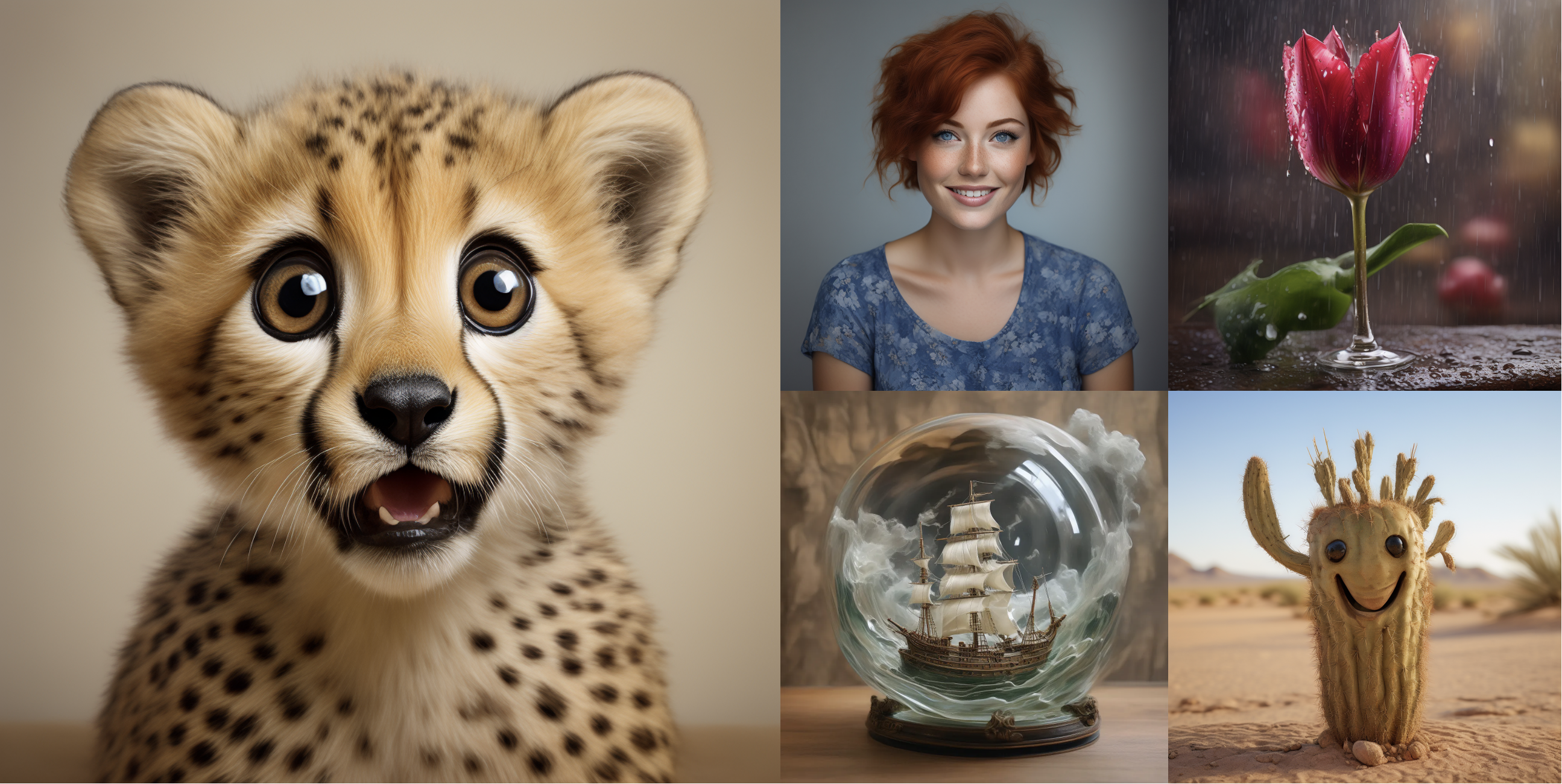}

}

\usepackage{bibentry}

\begin{document}

\maketitle

\begin{abstract}
\end{abstract}
  In this paper, we propose an efficient, fast, and versatile distillation method to accelerate the generation of pre-trained diffusion models. The method reaches state-of-the-art performances in terms of FID and CLIP-Score for few steps image generation on the COCO2014 and COCO2017 datasets, while requiring only several GPU hours of training and fewer trainable parameters than existing methods. In addition to its efficiency, the versatility of the method is also exposed across several tasks such as \emph{text-to-image}, \emph{inpainting}, \emph{face-swapping}, \emph{super-resolution} and using different backbones such as UNet-based denoisers (SD1.5,  SDXL), DiT (Pixart-$\alpha$) and MMDiT (SD3), as well as adapters. In all cases, the method allowed to reduce drastically the number of sampling steps while maintaining very high-quality image generation. 

%
\begin{links}
    \link{Code}{https://github.com/gojasper/flash-diffusion}
\end{links}

\begin{figure*}[t]
  \centering
  \captionsetup[subfigure]{position=below, labelformat = empty}
  \subfloat{\includegraphics[width=1.72in]{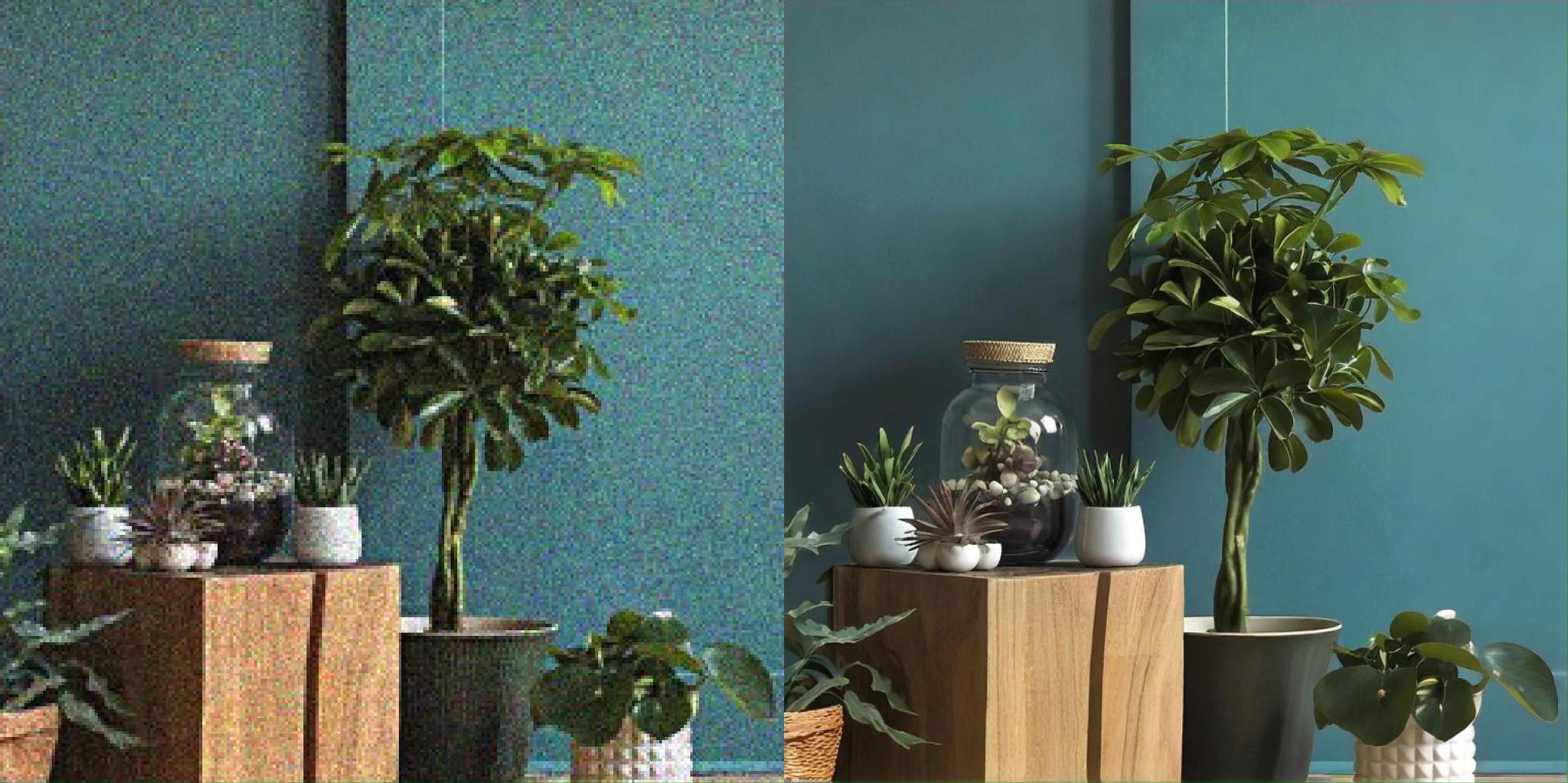}}
  \hskip 0.1em
  \subfloat{\includegraphics[width=1.72in]{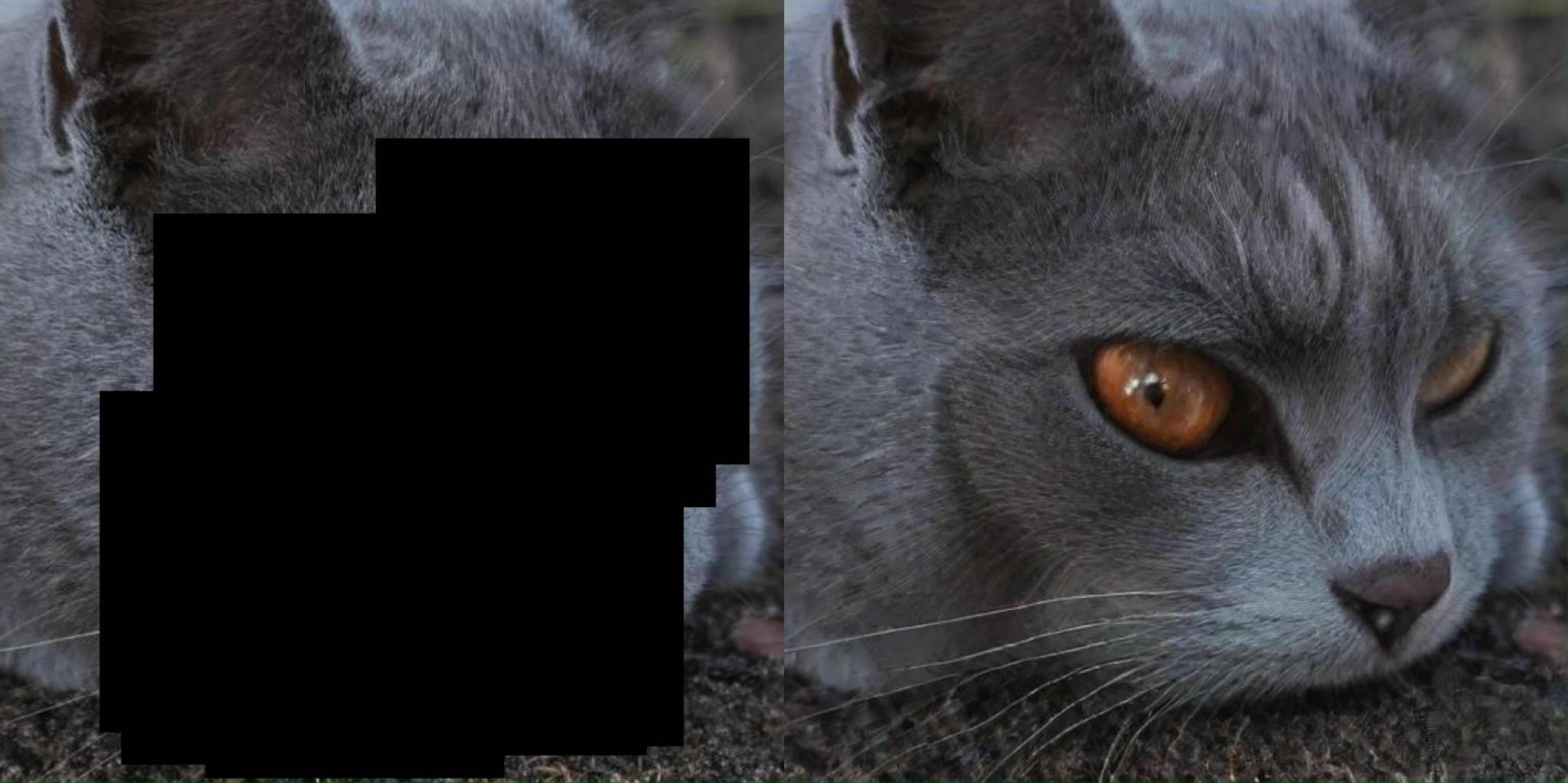}}
  \hskip 0.1em
  \subfloat{\includegraphics[width=1.72in]{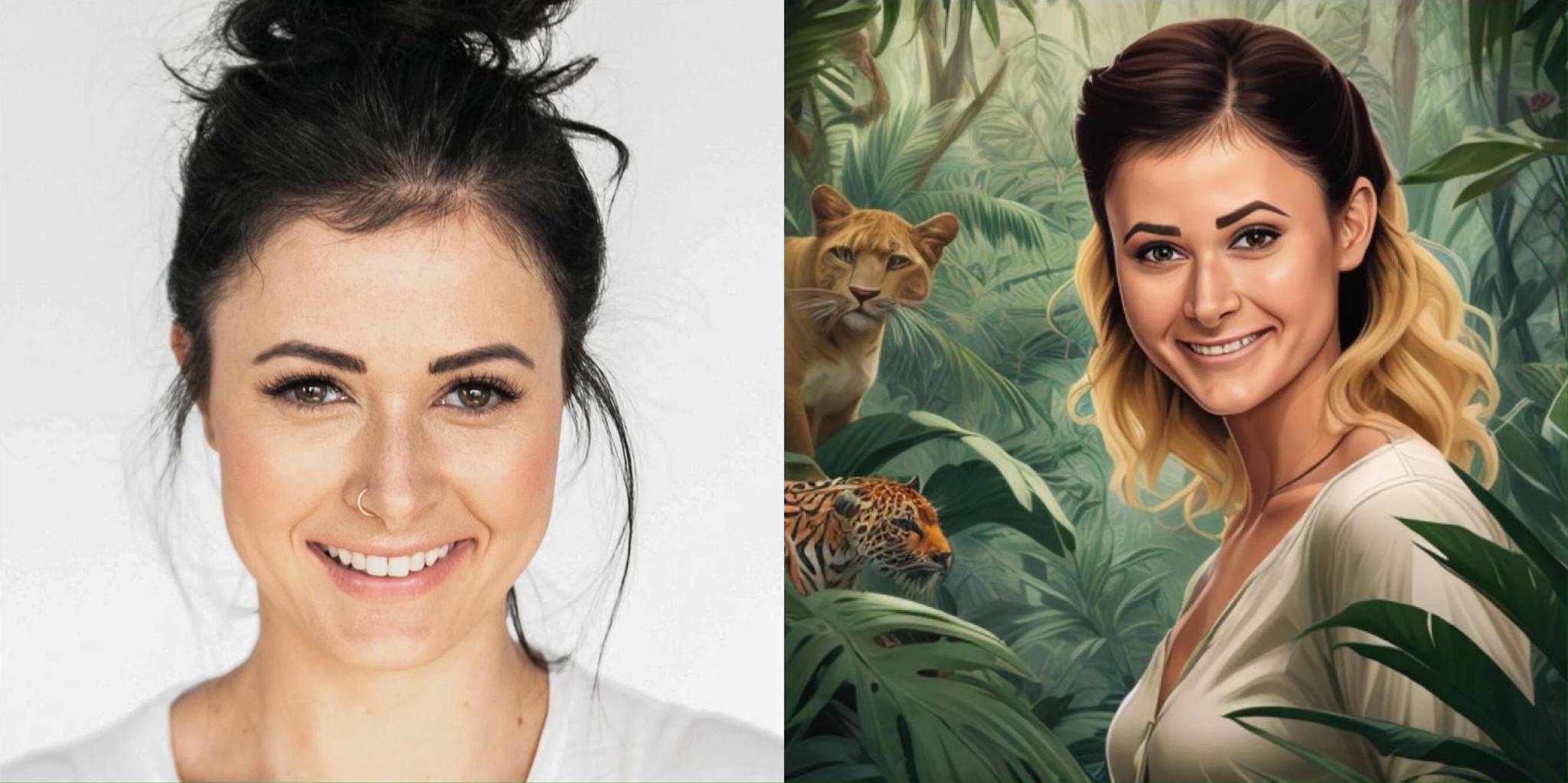}}
  \hskip 0.1em
  \subfloat{\includegraphics[width=1.72in]{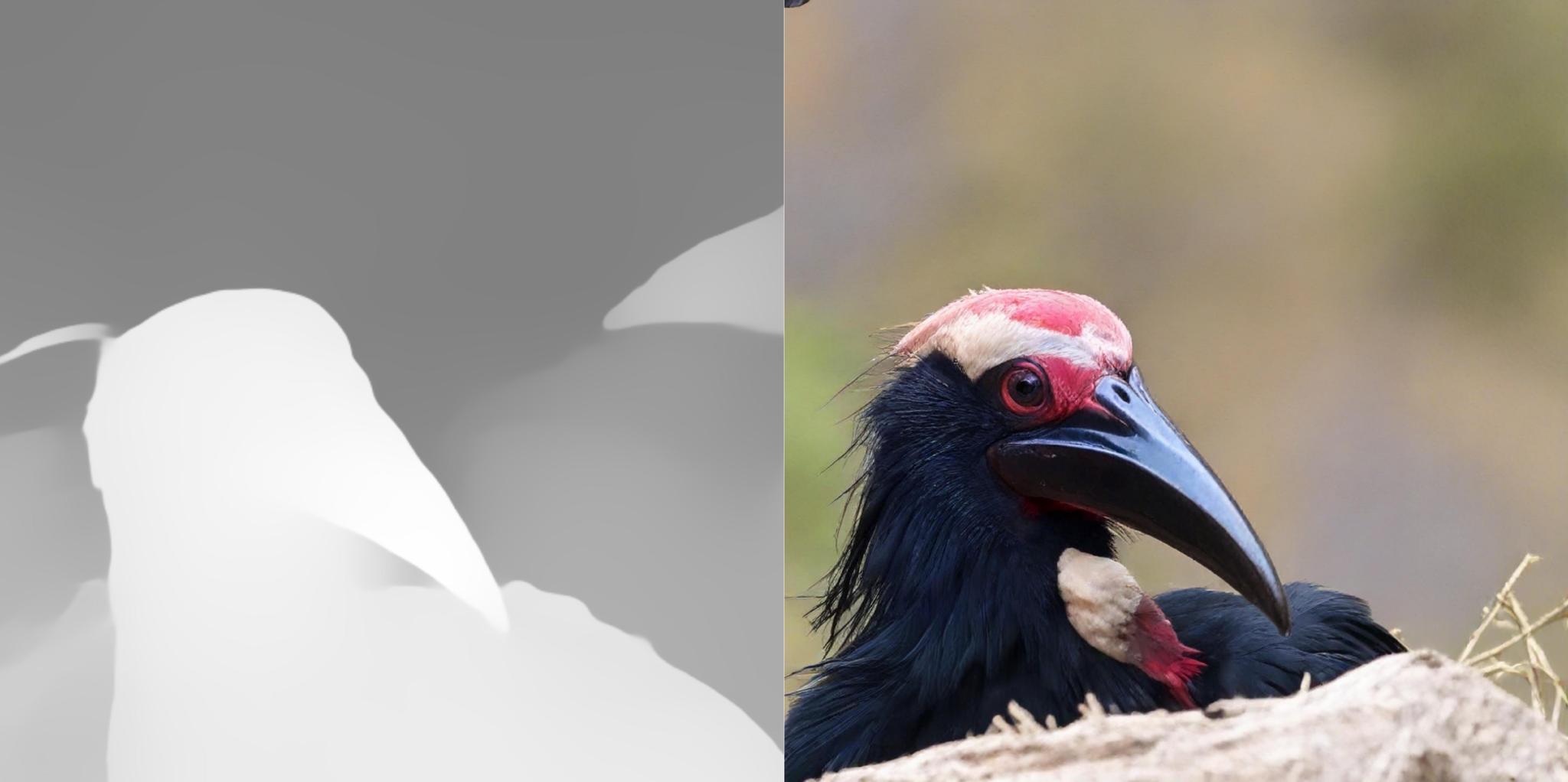}}\vspace{0.01em}
  \subfloat[Super-resolution]{\includegraphics[width=1.72in]{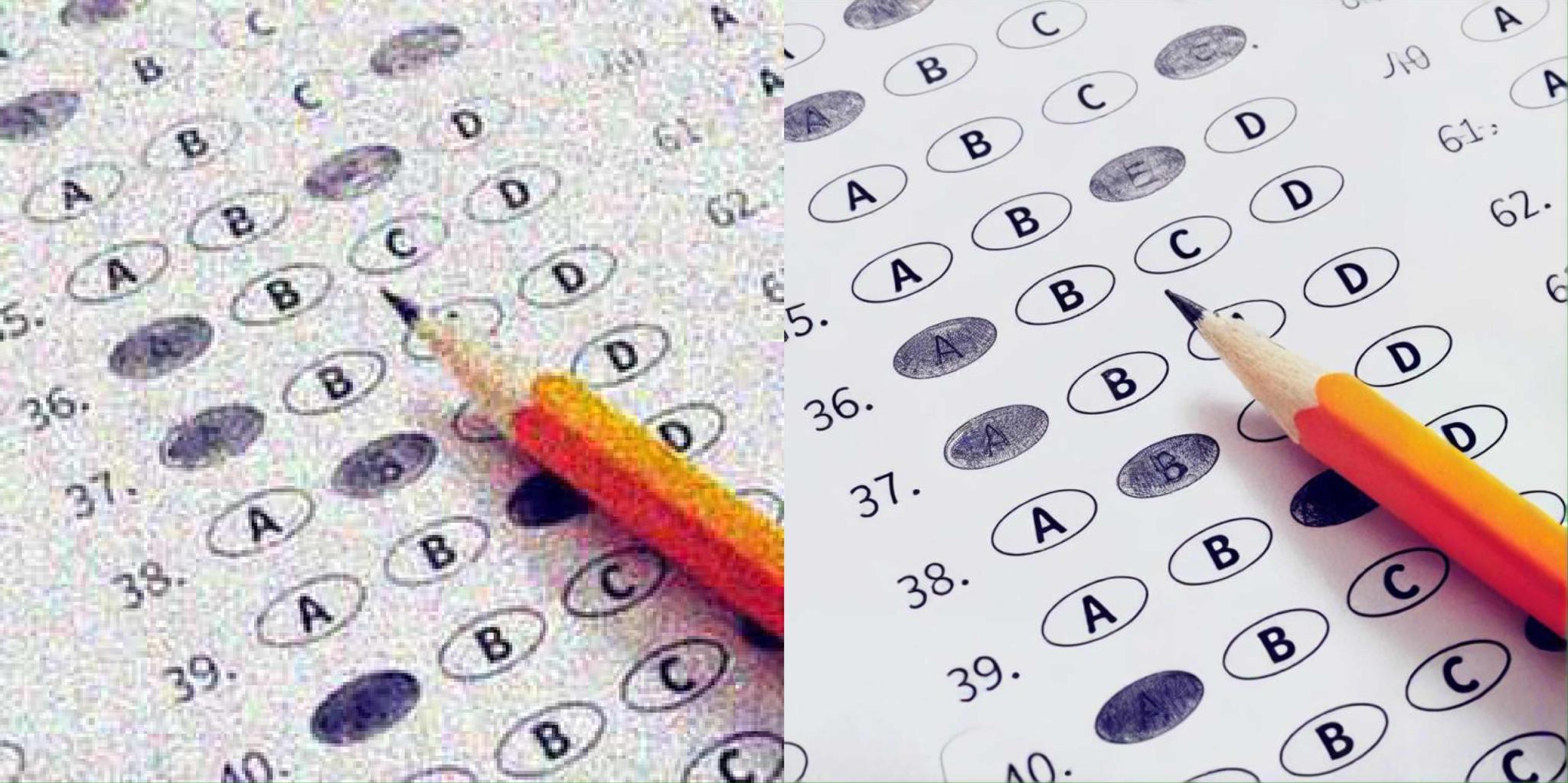}}
  \hskip 0.1em
  \subfloat[Inpainting]{\includegraphics[width=1.72in]{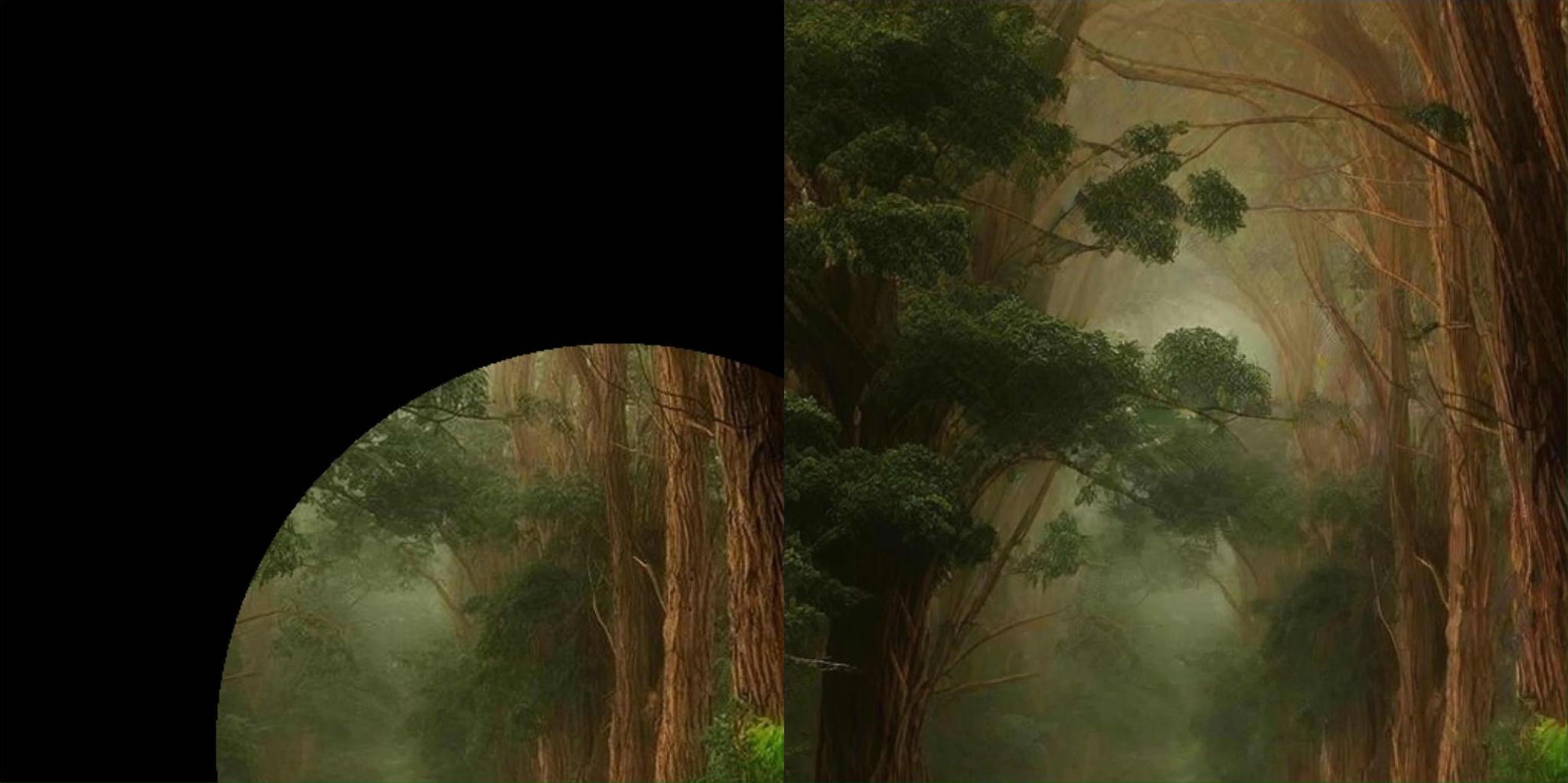}}
  \hskip 0.1em
  \subfloat[Face-Swapping]{\includegraphics[width=1.72in]{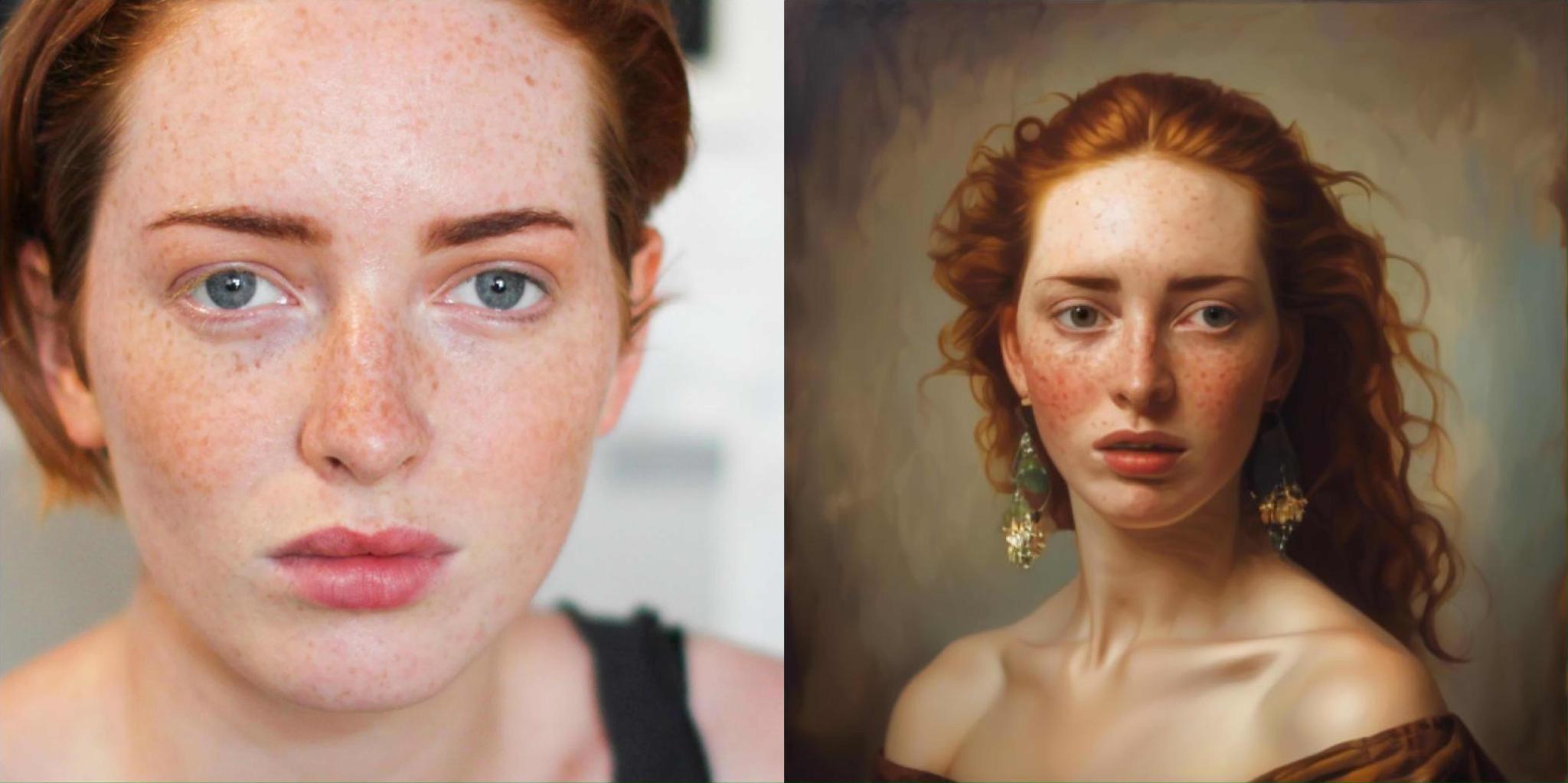}}
  \hskip 0.1em
  \subfloat[Adapters]{\includegraphics[width=1.72in]{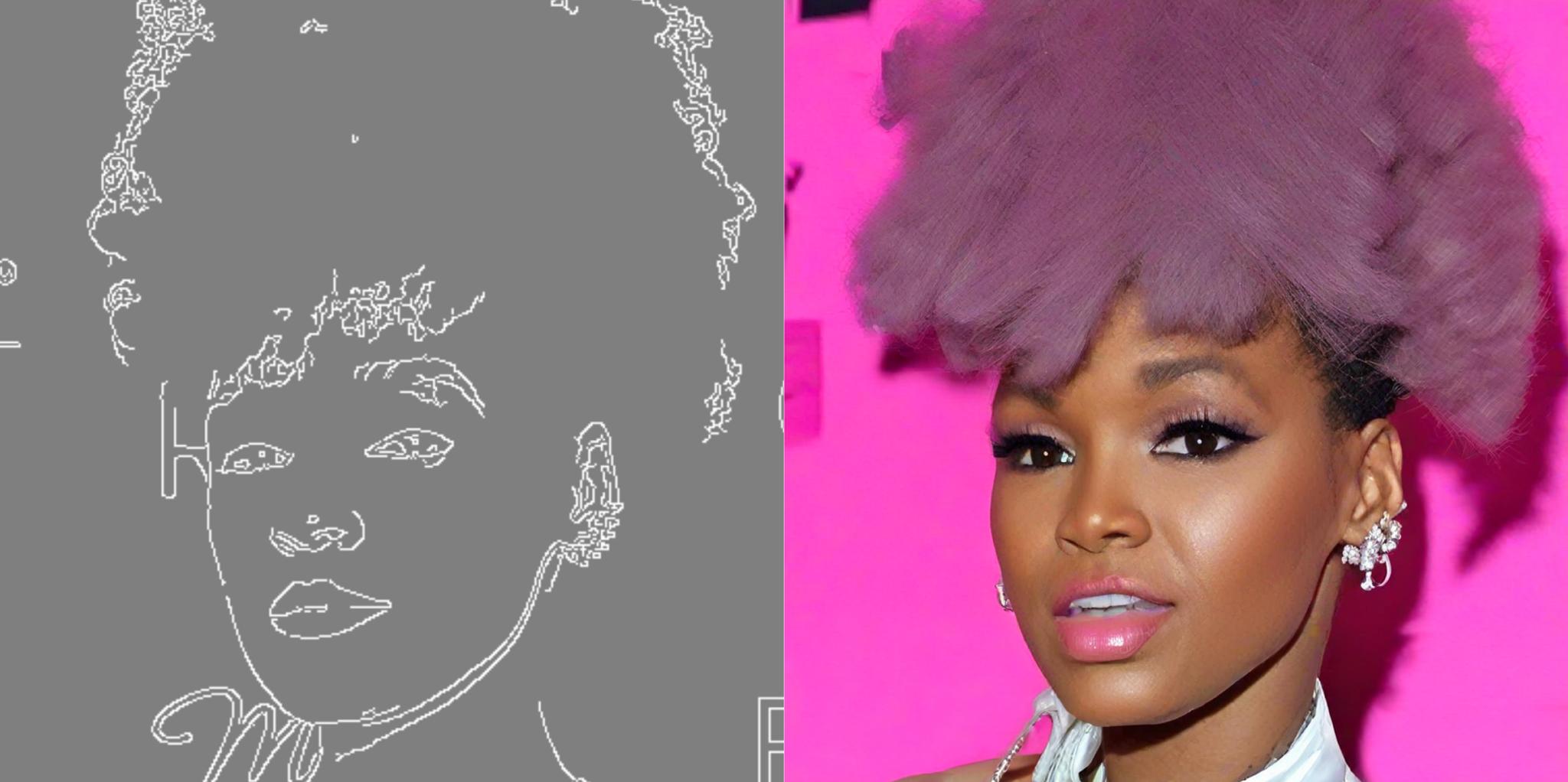}}
  \caption{Inputs (left columns) and generated samples (right columns) using the proposed method for different teacher models and tasks (\emph{super-resolution}, \emph{inpainting}, \emph{face-swapping} and adapters). Samples are generated using 4 Neural Function Evaluations (NFEs).}
  \end{figure*}

\section{Introduction}
 Diffusion Models (DM) \citep{sohl2015deep,ho2020denoising,song2020score}
 have proven to be one of the most efficient class of generative models for image synthesis \cite{dhariwal2021diffusion,ramesh2022hierarchical, rombach2022high,nichol2022glide} and have raised particular interest and enthusiasm for text-to-image applications \cite{ramesh2021zero,ramesh2022hierarchical,rombach2022high,saharia2022photorealistic,ho2022imagen,esser2024scaling,podell2023sdxl,chen2023pixart,chen2024pixart} where they outperform other approaches. However, their usability for real-time applications remains limited by the intrinsic iterative nature of their sampling mechanism. At inference time, these models aim at iteratively denoising a sample drawn from a Gaussian distribution to finally create a sample belonging to the data distribution. Nonetheless, such a denoising process requires multiple evaluations of a potentially very computationally costly neural function.

 Recently, more efficient solvers \citep{lu2022dpm,lu2022dpm+,zhang2022fast,zhao2024unipc} or diffusion distillation methods \cite{salimans2021progressive,song2023consistency,lin2024sdxl,xu2023ufogen,liu2023instaflow,ren2024hyper,luo2023latent,luo2023lcm,sauer2023adversarial,sauer2024fast,yin2023one,hsiao2024plug} aiming at reducing the number of sampling steps required to generate satisfying samples from a trained diffusion model have emerged to try to tackle this issue. Nonetheless, solvers typically require at least 10 Neural Function Evaluations (NFEs) to produce satisfying samples while distillation methods may require extensive training resources \citep{liu2023instaflow,yin2023one,meng2023distillation} or require an iterative training procedure to update the teacher model throughout training \citep{salimans2021progressive,lin2024sdxl,li2024snapfusion} limiting their applications and reach. Moreover, most of the existing distillation methods are tailored for a specific task such as text-to-image. It is still unclear how they would perform on other tasks, using different conditionings and diffusion model architectures. 

In this paper, we present \emph{Flash Diffusion}, a fast, robust, and versatile diffusion distillation method that allows to drastically reduce the number of sampling steps while maintaining a very high image generation quality. The proposed method aims at training a student model to predict in a single step a denoised multiple-step teacher prediction of a corrupted input sample. The method also drives the student distribution towards the real input sample manifold with an adversarial objective \citep{goodfellow_generative_2014} and ensures that it does not drift too much from the learned teacher distribution using distribution matching \citep{dziugaite2015training,li2015generative}. The main contributions of the paper are as follows:
\begin{itemize}
    \item We propose an efficient, \textbf{fast}, \textbf{versatile}, and LoRA compatible distillation method aiming at reducing the number of sampling steps required to generate high-quality samples from a trained diffusion model.
    \item We validate the method for text-to-image and show that it reaches SOTA results for few steps image generation on standard benchmark datasets with only two NFEs, which is equivalent to a single step with classifier-free guidance while having far fewer training parameters than competitors and requiring only a few GPU hours of training.
    \item We conduct an extensive ablation study to show the impact of the different components of the method and demonstrate its robustness and reliability.
    \item We emphasize the versatility of the method through an extensive experimental study across various tasks (\emph{text-to-image}, image \emph{inpainting}, \emph{super-resolution}, \emph{face-swapping}), diffusion model architectures (SD1.5, SDXL, Pixart-$\alpha$ and SD3) and illustrate its compatibility with adapters \citep{mou2024t2i} and existing LoRAs.
\end{itemize}

\section{Related Works}
\paragraph{Diffusion Models} Diffusion models consist in artificially corrupting input data according to a given noise schedule \citep{sohl2015deep,ho2020denoising,song2020score} such that the data distribution eventually resembles a standard Gaussian one. They are then trained to estimate the amount of noise added in order to learn a reverse diffusion process allowing them, once trained, to generate new samples from Gaussian noise. Those models can be conditioned with respect to various inputs such as images \citep{rombach2022high}, depth maps, edges, poses \citep{zhang2023adding,mou2024t2i} or text \cite{dhariwal2021diffusion,ramesh2022hierarchical,rombach2022high,nichol2022glide,esser2024scaling,ho2022imagen,podell2023sdxl} where they demonstrated very impressive results. However, the need to recourse to a large number of sampling steps (typically 50 steps) at inference time to generate high-quality samples has limited their usage for real-time applications and narrowed their usability and reach.

\paragraph{Diffusion Distillation} In order to tackle this limitation, several methods have recently emerged to reduce the number of function evaluations required at inference time. On the one hand, several papers tried to build more efficient solvers to speed up the generation process \citep{lu2022dpm,lu2022dpm+,zhang2022fast,zhao2024unipc} but these methods still require the use of several steps  (typically 10) to generate satisfying samples. On the other hand, several approaches relying on model distillation \citep{hinton2015distilling} proposed to train a student network that would learn to match the samples generated by a teacher model but in fewer steps. A simple approach would consist in building pairs of \emph{noise/teacher samples} and training a student model to match the teacher predictions in a single step \citep{luhman2021knowledge,zheng2023fast}. Nonetheless, this approach remains quite limited and struggles to match the quality of the teacher model since there is no underlying useful information to be learned by the student in full noise. Building upon this idea, several methods were proposed to first apply the \emph{forward} diffusion process to an input sample and then pass it to the student network. The student prediction is then compared to the learned distribution of the teacher model using either a regression loss  \citep{kohler2024imagine,yin2023one} an adversarial objective \citep{xu2023ufogen,sauer2023adversarial,sauer2024fast,yin2024improved} or distribution matching \citep{yin2023one,yin2024improved}.

\textbf{Progressive distillation} \citep{salimans2021progressive,meng2023distillation} is also a method that has proven to be quite promising. It consists in training a student model to predict a two-step teacher denoising of a noisy sample in a single step theoretically halving the number of required sampling steps. The teacher is then replaced by the new student and the process is repeated several times. This approach was also enriched with a GAN-based objective that allows to further reduce the number of sampling steps needed from 4-8 to a single pass \citep{lin2024sdxl}.
InstaFlow \citep{liu2023instaflow} proposed instead to rely on rectified flows \citep{liu2022flow} to ease the \emph{one-step} distillation process. However, this approach may require a significant number of training parameters and a long training procedure, making it computationally intensive.

\textbf{Consistency models} \citep{song2023consistency,song2023improved,luo2023latent,kim2023consistency} is also a promising, effective, and one of the most versatile distillation methods proposed in the literature. The main idea is to train a model to map any point lying on the \emph{Probability Flow Ordinary Differential Equation} (PF-ODE) to its origin, theoretically unlocking single-step generation. \citet{luo2023lcm} combined Latent Consistency Model (LCM) and LoRAs \citep{hu2021lora} and showed that it is possible to train a strong student with a very limited number of trainable parameters and a few GPU hours of training. Nonetheless, those models still struggle to achieve single-step generation and reach the sampling quality of peers.

In a parallel study conducted recently, the authors of \citep{yin2024improved} also introduced the combined use of a distribution matching loss and an adversarial loss, a method we also employ in our paper. Nonetheless, they do not rely on the use of a distillation loss that proved highly efficient in our experiments and do not compute the adversarial loss with respect to the same inputs. Moreover, their approach still necessitates training another denoiser to assess the score of the fake samples, significantly increasing the number of trainable parameters and, consequently, the computational burden of the method. Furthermore, the ability of their method to generalize and perform effectively across different tasks and diffusion model architectures remains unclear.

\section{Proposed Method}\label{sec:method}

In this section, we expose the proposed method that builds upon several ideas proposed in the literature. 

\subsection{Background on Diffusion Models}
Let $x_0 \in \mathcal{X}$ be a set of data such that $x_0 \sim p(x_0)$ where $p(x_0)$ is an unknown distribution. The main idea of diffusion models (DM) is to estimate the amount of noise $\varepsilon$, artificially added to an input sample $x_0$ using the \emph{forward process} $x_t = \alpha(t) \cdot x_0 +  \sigma(t)\cdot \varepsilon$ where $\varepsilon \sim \normal$. The noise schedule is controlled by two differentiable functions $\alpha(t)$, $\sigma(t)$ for any $t \in [0, T]$ such that the log signal-to-noise ratio $\log [\alpha(t)^2 / \sigma(t)^2]$ is decreasing over time. In practice, during training a diffusion model learns a parametrized function $\varepsilon_{\theta}$ conditioned on the timestep $t$ and taking as input the noisy sample $x_t$. The parameters $\theta$ are then learned via denoising score matching
\citep{vincent2011connection,song2019generative}.
\begin{equation}\label{eq:lossdiffusion}
  \mathcal{L} = \mathbb{E}_{x_0 \sim p, t \sim \pi, \varepsilon \sim \normal } \left[ \lambda(t) \left\| \varepsilon_{\theta}(x_t, t) - \varepsilon \right\|^2 \right]\,,
\end{equation}
where $\lambda(t)$ is a scaling factor, $t \in [0, 1]$ is the timestep and $\pi(t)$ is a distribution over the timesteps. We provide in the appendices an extended background on diffusion models.

\subsection{Distilling a Pretrained Diffusion Model}
For the following, we place ourselves in the context of Latent Diffusion Models \cite{rombach2022high} for image generation and refer to the teacher model as $\varepsilon_{\phi}^{\mathrm{teacher}}$, the student model as $\varepsilon_{\theta}^{\mathrm{student}}$, the training images as $x_0$ and their unknown distribution $p(x_0)$. We refer to $z_0 = \mathcal{E}(x_0)$ as the associated latent variables obtained with an encoder $\mathcal{E}$. $\pi$ is the probability density function of the timesteps $t \in [0, 1]$. 
The proposed method is mainly driven by the desire to end up with a fast, robust, and reliable approach that would be easily transposed to different use cases. The main idea of the proposed approach is quite similar to diffusion models. 

Given a noisy latent sample $z_t$ with $t \sim \pi(t)$, we propose to train a function $f_{\theta}$ to predict a denoised version $\tilde{z}_0$ of the original sample $z_0$. The main difference with a diffusion model is that instead of using $z_0$ as a target, we propose to leverage the knowledge of the teacher model and use a sample belonging to the data distribution learned by the teacher model $p_{\phi}^{\mathrm{teacher}}(z_0)$. In other words, we use the teacher model and an ODE solver $\Psi$ that is run several times to generate a denoised latent sample $\tilde{z}_0^{\mathrm{teacher}}(z_{t})$ used as a target for the student model. The main distillation loss writes as follows:
\begin{equation}\label{eq:distillation loss}
  \mathcal{L}_{\mathrm{distil}} = \mathbb{E}_{z_0, t, \varepsilon } \left[ \left\| f_{\theta}(z_t, t) - \tilde{z}_0^{\mathrm{teacher}}(z_{t}) \right\|^2 \right]\,,
\end{equation}

A similar idea was employed in \citep{sauer2024fast} but the authors generate fully synthetic samples meaning that the samples $z_t$ are pure noise, $z_t \sim \mathcal{N}\left(0, \mathbf{I} \right)$. In contrast, in our approach, we hypothesize that allowing $z_t$ to retain some information from the \emph{ground-truth} encoded sample $z_0$ could enhance the distillation process. As in \citep{luo2023latent}, when distilling a conditional DM, we also perform Classifier-Free Guidance (CFG) \citep{ho2021classifier} with the teacher to better enforce the model to respect the conditioning. This technique actually significantly improves the quality of the generated samples by the student as shown in the ablations.  Additionally, it eliminates the need for conducting CFG during inference with the student, further decreasing the method's computational cost by halving the NFEs for each step.
In practice, the guidance scale $\omega$ is uniformly sampled in $[\omega_{\min}, \omega_{\max}]$ where $0 \leq \omega_{\min} \leq \omega_{\max}$. 
  \begin{figure}
    \centering
    \captionsetup[subfigure]{position=below, labelformat = empty}
    \subfloat[\scriptsize \emph{Warm-up}]{\includegraphics[width=0.25\linewidth]{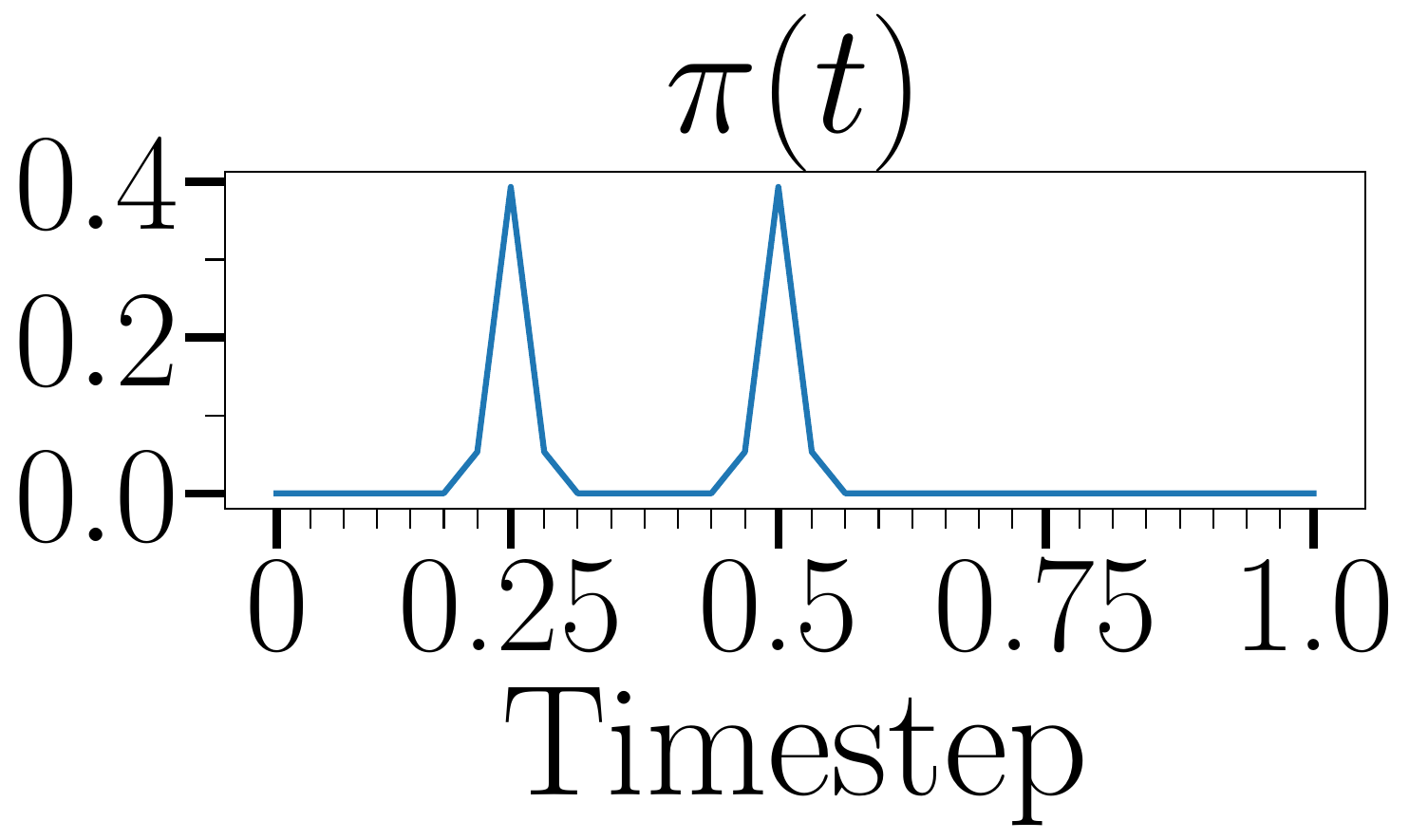}}
    \subfloat[\scriptsize Phase 1]{\includegraphics[width=0.25\linewidth]{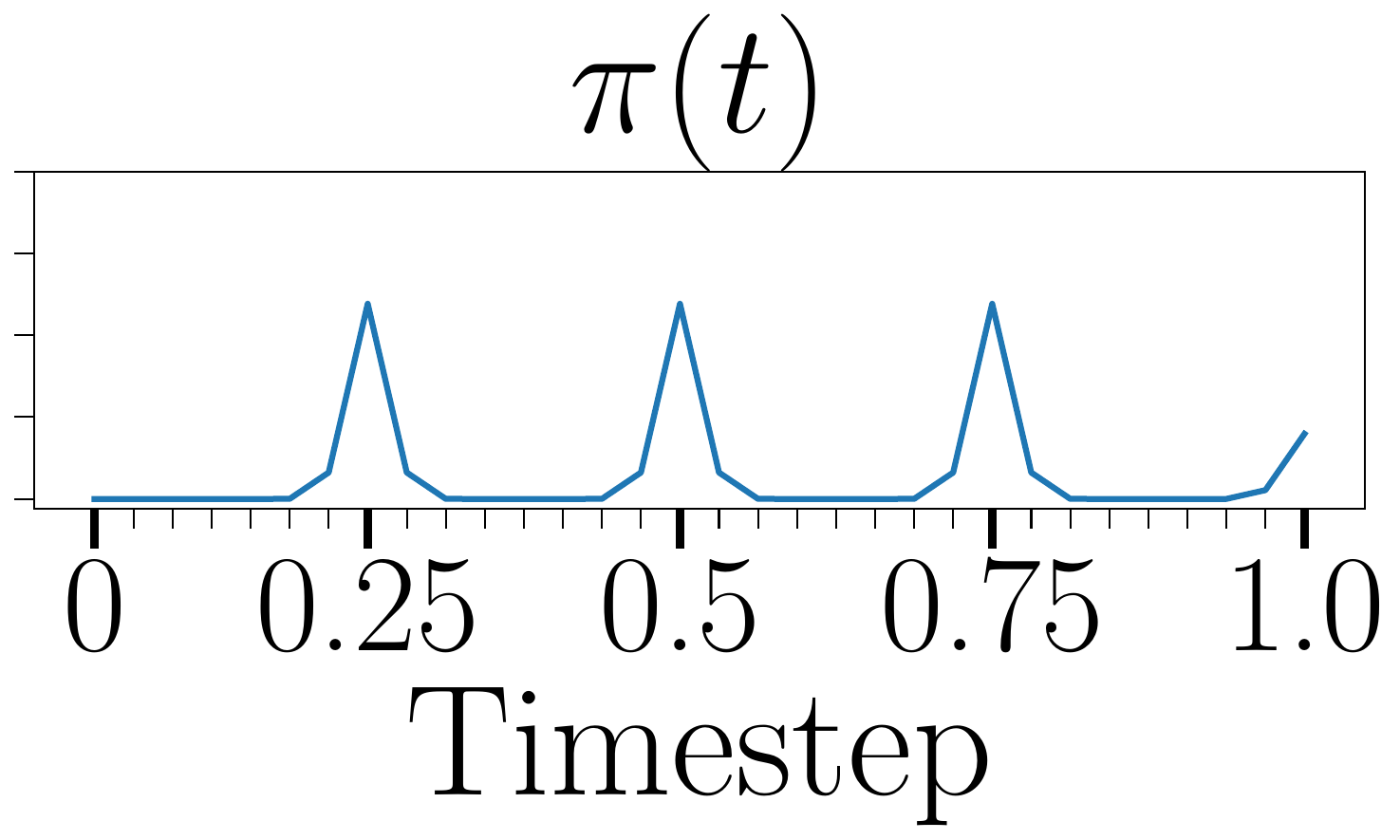}}
    \subfloat[\scriptsize Phase 2]{\includegraphics[width=0.25\linewidth]{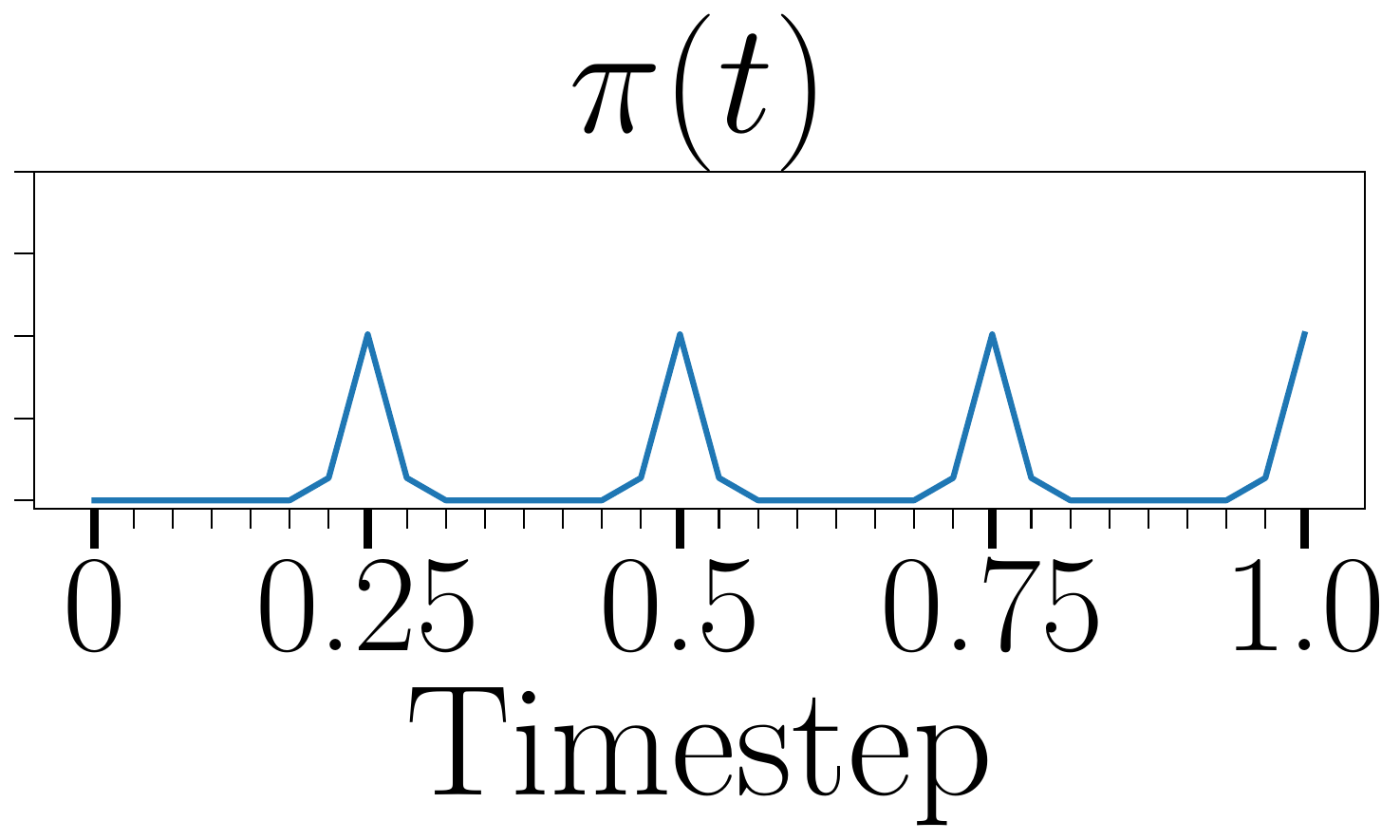}}
    \subfloat[ \scriptsize Phase 3]{\includegraphics[width=0.25\linewidth]{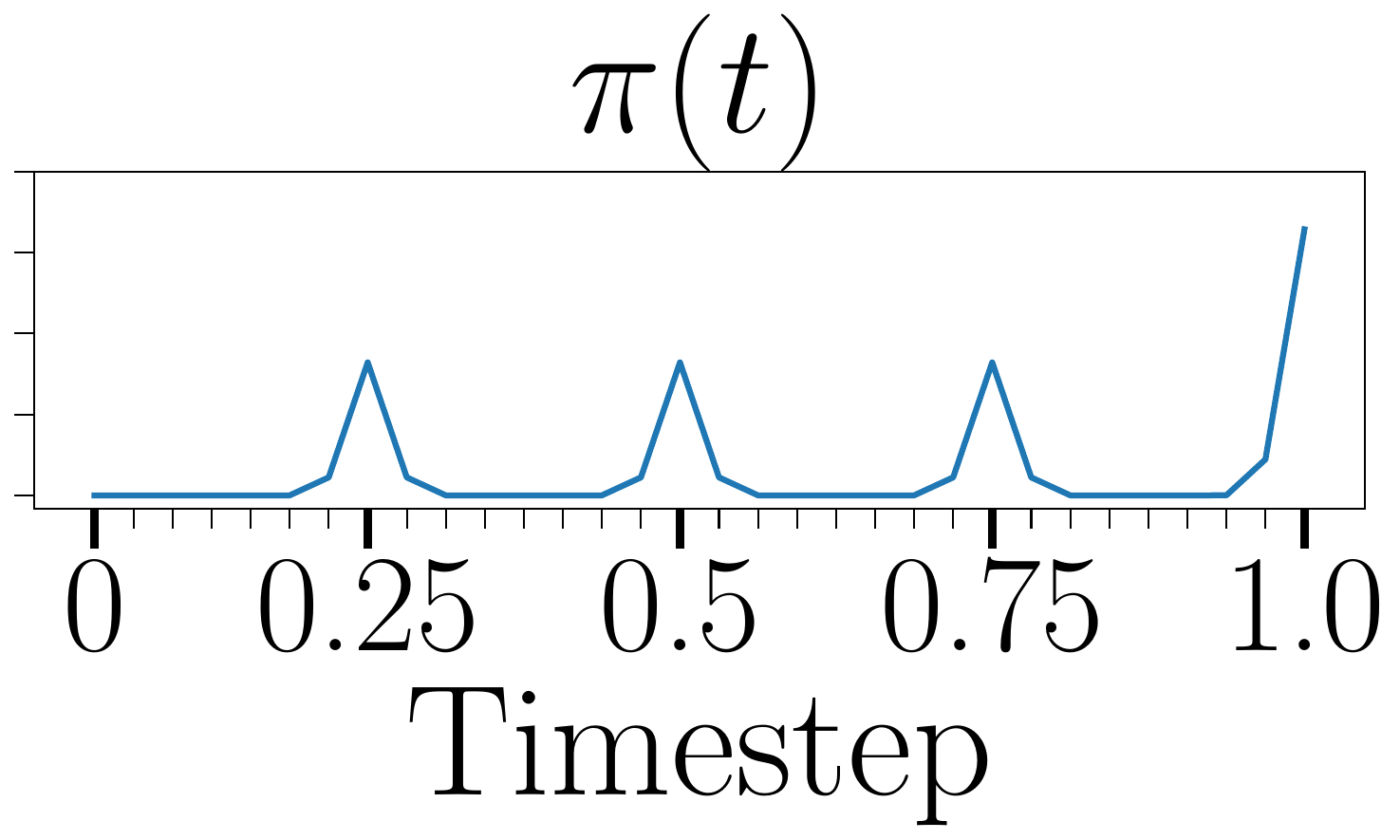}}
    \caption{Illustration of the evolution of the proposed timesteps distribution $\pi$ throughout training. $t=0$ corresponds to no noise injection while $t=1$ corresponds to the maximum noise injection (\emph{i.e.} the noisy latent sample is equivalent to a sample drawn from a standard Gaussian distribution). For each phase unless the \emph{Warm-up}, 4 timesteps are over-sampled out of the $K=32$ selected ones. As the training progresses, the probability mass is shifted towards full noise to favor single-step generation.} 
    \label{fig:timesteps}
  \end{figure}

\subsection{Timesteps Sampling}\label{sec:timesteps}

The cornerstone of our approach hinges on the selection of the timestep probability density function, denoted as $\pi(t)$. According to the continuous modeling, exposed in \citep{song2020score}, DMs are trained to remove noise from a latent sample $z_t$ for any given continuous time $t$. However, since we aim at achieving few steps data generation (typically 1-4 steps) at inference time, the learned function $\varepsilon_{\theta}$ will only be evaluated at a few discrete timesteps $\{t_i\}_{i=1}^K$. 

To tackle this issue and enforce the distillation process to focus on the most relevant timesteps, we propose to select $K$ (typically 16 or 32) uniformly spaced timesteps in $[0, 1]$ and assign a probability to each of them according to a probability mass function $\pi(t)$. We choose $\pi(t)$ as a mixture of Gaussian controlled by a series of weights $\{\beta_i\}_{i=1}^{K}$
\begin{equation}\label{eq:mass distribution}
    \pi(t) = \frac{1}{\sqrt{2\pi \sigma^2}}\sum_{i=1}^K \beta_i \exp \left(-\frac{(t - \mu_i)^2}{2\sigma^2}\right),
\end{equation}
where the mean of each Gaussian is controlled by $\{\mu_i = i/K\}_{i=1}^{K}$ and the variance is fixed to $\sigma=\sqrt{0.5/K^2}$.
This approach is such that when distilling the teacher only a small number of $K$ discrete timesteps will be sampled instead of the continuous range $[0, 1]$\footnote{In practice when training a DM, the range $[0, 1]$ is actually discretized (typically into 1000 timesteps) for computational purposes.}. Moreover, the distribution $\pi$ is defined such that out of the $K$ selected timesteps, the 4 timesteps used at inference for 1, 2 and 4 steps generation are over-sampled (typically we set $\beta_i > 0$ if $i \in [\frac{K}{4}, \frac{K}{2}, \frac{3K}{4}, K]$ and $\beta_i = 0$ otherwise). Unlike other methods \citep{sauer2023adversarial,sauer2024fast} we do not only focus on those 4 timesteps since we noticed that it can lead to a reduction of diversity in the generated samples. This is in particular emphasized in the ablation study. In practice, we notice that a warm-up phase is beneficial to the training process. Therefore, we decide to start by first imposing a higher probability to the timesteps corresponding to the least added amount of noise by setting $\beta_{K/4}=\beta_{K/2}=0.5$ and $\beta_i=0$ otherwise. We then progressively shift the probability mass towards full noise to favor single-step generation while still over-sampling the targeted 4 timesteps by setting a strictly positive value for $\beta_i$ where $i\equiv 0[K/4]$, and $\beta_i=0$ otherwise. An example for $\pi$ with $K=32$ is illustrated in Fig.~\ref{fig:timesteps}. As pictured in the figure, the $[0, 1]$ interval is split into 32 timesteps. During the \emph{warm-up} phase, the probability mass allocates a higher probability to timesteps $[0.25, 0.5]$ to ease the distillation process. As the training progresses, the probability mass function is then shifted towards full noise to favor single-step generation while always allocating a higher probability to the 4 timesteps $[0.25, 0.5, 0.75, 1]$. The impact of the timesteps distribution is further discussed in the ablations.

\begin{figure}[ht]
    \centering
    \includegraphics[width=\linewidth]{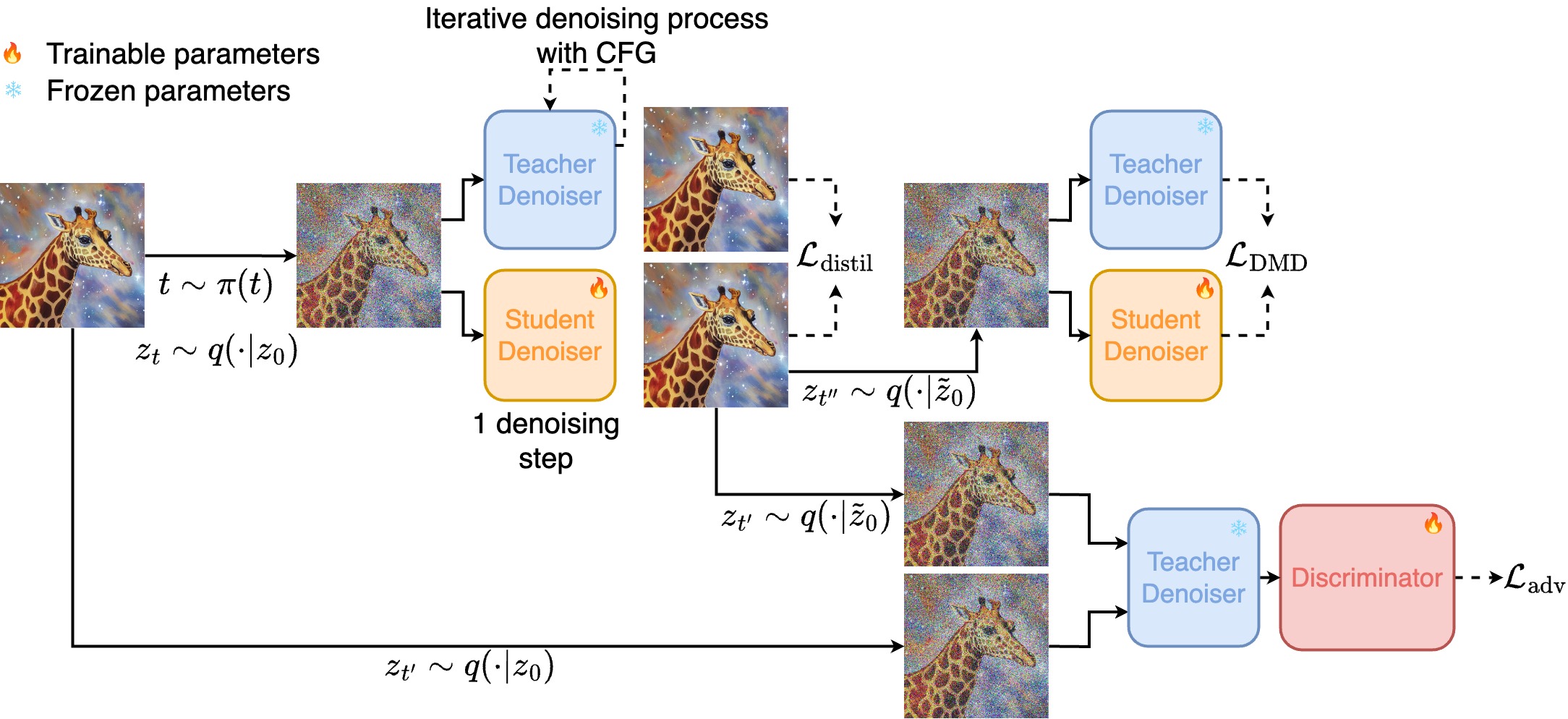}
    \caption{\emph{Flash Diffusion} training method: the student is trained with a distillation loss between multiple-step teacher and single-step student denoised samples. The student predictions are then re-noised and denoised with the teacher and student before evaluating the GAN and DMD losses.}
    \label{fig:flash_diffusion scheme}
\end{figure}

\subsection{Adversarial Objective}\label{sec:adversarial}
To further enhance the quality of the samples, we have also decided to incorporate an adversarial objective. The core idea is to train the student model to generate samples that are indistinguishable from the true data distribution $p(x_0)$. To do so, we propose to train a discriminator $D_{\nu}$ to distinguish the generated samples $\tilde{x}_0$ from the real samples $x_0 \sim p(x_0)$. As proposed in \citep{lin2024sdxl,sauer2024fast}, we also apply the discriminator directly within the latent space. This approach circumvents the necessity of decoding the samples using the VAE, a process outlined in \citep{sauer2023adversarial}, that proves to be expensive and hampers the method's scalability to high-resolution images. Drawing inspiration from \citep{lin2024sdxl,sauer2024fast}, we propose an approach where both the one-step student prediction $\tilde{z}_0$ and the input latent sample $z_0$ are re-noised following the teacher noise schedule. This process uses a timestep $t'$ uniformly chosen from the set $[0.01, 0.25, 0.5, 0.75]$ enabling the discriminator to effectively differentiate between samples based on both high and low-frequency details \citep{lin2024sdxl}.  The samples are first passed through the \emph{frozen} teacher model, followed by the trainable discriminator, to yield a real or fake prediction. When employing a UNet architecture \citep{ronneberger2015u} for the teacher model, our approach focuses on utilising only the encoder of the UNet, generating an even more compressed latent representation and further reducing the parameter count for the discriminator. The adversarial loss $\mathcal{L}_{\mathrm{adv}}$ and discriminator loss $\mathcal{L}_{\mathrm{dis}}$ write as follows:
\begin{equation}
\label{eq:adv_loss}
\begin{aligned}
  \mathcal{L}_{\mathrm{adv}} =&~\frac{1}{2} ~ \mathbb{E}_{z_0 , t', \varepsilon} \left[ \left\| D_{\nu}(f_{\theta}(z_{t'}, t')) - 1 \right\|^2\right]\,,\\
  \mathcal{L}_{\mathrm{dis.
}} =&~\frac{1}{2} ~ \mathbb{E}_{z_0, t', \varepsilon} \left[ \left\| D_{\nu}(z_0) - 1 \right\|^2 + \left\| D_{\nu}(f_{\theta}(z_{t'}, t'))\right\|^2 \right]\,,
\end{aligned}
\end{equation}
where $\nu$ denotes the discriminator parameters. We opt for these particular losses due to their reliability and stability during training, as observed in our experiments. In practical terms, the discriminator's architecture is designed as a straightforward Convolutional Neural Network (CNN) featuring a stride of 2, a kernel size of 4, SiLU activation \citep{hendrycks2016gaussian,ramachandran2017searching} and group normalization \citep{wu2018group}.

\subsection{Distribution Matching}
Inspired by the work of \citep{yin2023one}, we also propose to introduce a Distribution Matching Distillation (DMD) loss to ensure that the generated samples closely mirror the data distribution learned by the teacher. Specifically, this involves minimizing the Kullback–Leibler (KL) divergence between the student distribution  $p^{\mathrm{student}}_{\theta}$ and $p^{\mathrm{teacher}}_{\phi}$, the data distribution learned by the teacher \citep{wang2024prolificdreamer}:
\begin{equation}\label{eq:dmd}
  \begin{aligned}
    \mathcal{L}_{\mathrm{DMD}} &=~D_{KL}(p^{\mathrm{student}}_{\theta}||p^{\mathrm{teacher}}_{\phi})\,.
  \end{aligned}
\end{equation}
Taking the gradient of the KL divergence with respect to the student model parameters $\theta$ leads to the following update:
\[
  \nabla_{\theta} \mathcal{L}_{\mathrm{DMD}} = \mathbb{E} \bigg[\bigg( s^{\mathrm{student}}(y) -s^{\mathrm{teacher}}(y)) \big)\bigg) \nabla f_{\theta}(z_{t}, t)\bigg]\,,
\]
where $s^{\mathrm{teacher}}$ and $s^{\mathrm{student}}$ are the score functions of the teacher and student distributions respectively and $y=f_{\theta}(z_{t}, t)$ is the student prediction. Inspired by \citep{yin2023one}, the one-step student prediction $\tilde{z}_0$ is re-noised using a uniformly sampled timestep $t'' \sim \mathcal{U}([0,1])$ and the teacher noise schedule. The new noisy sample is passed through the \emph{frozen} teacher model to get the score function for the teacher distribution $ s^{\mathrm{teacher}}(f_{\theta}(z_{t''}, t'')) = -(\varepsilon_{\phi}^{\mathrm{teacher}}(x_{t''}, t'')/\sigma(t''))$.
In our approach, we utilize the student model for the score function of the student distribution, instead of a dedicated diffusion model. This choice significantly reduces the number of trainable parameters and computational costs.

\begin{figure}[t]
  \centering
  \captionsetup[subfigure]{position=above, labelformat = empty}

  \subfloat[\scriptsize \centering 1 NFE]{\includegraphics[width=1.06in]{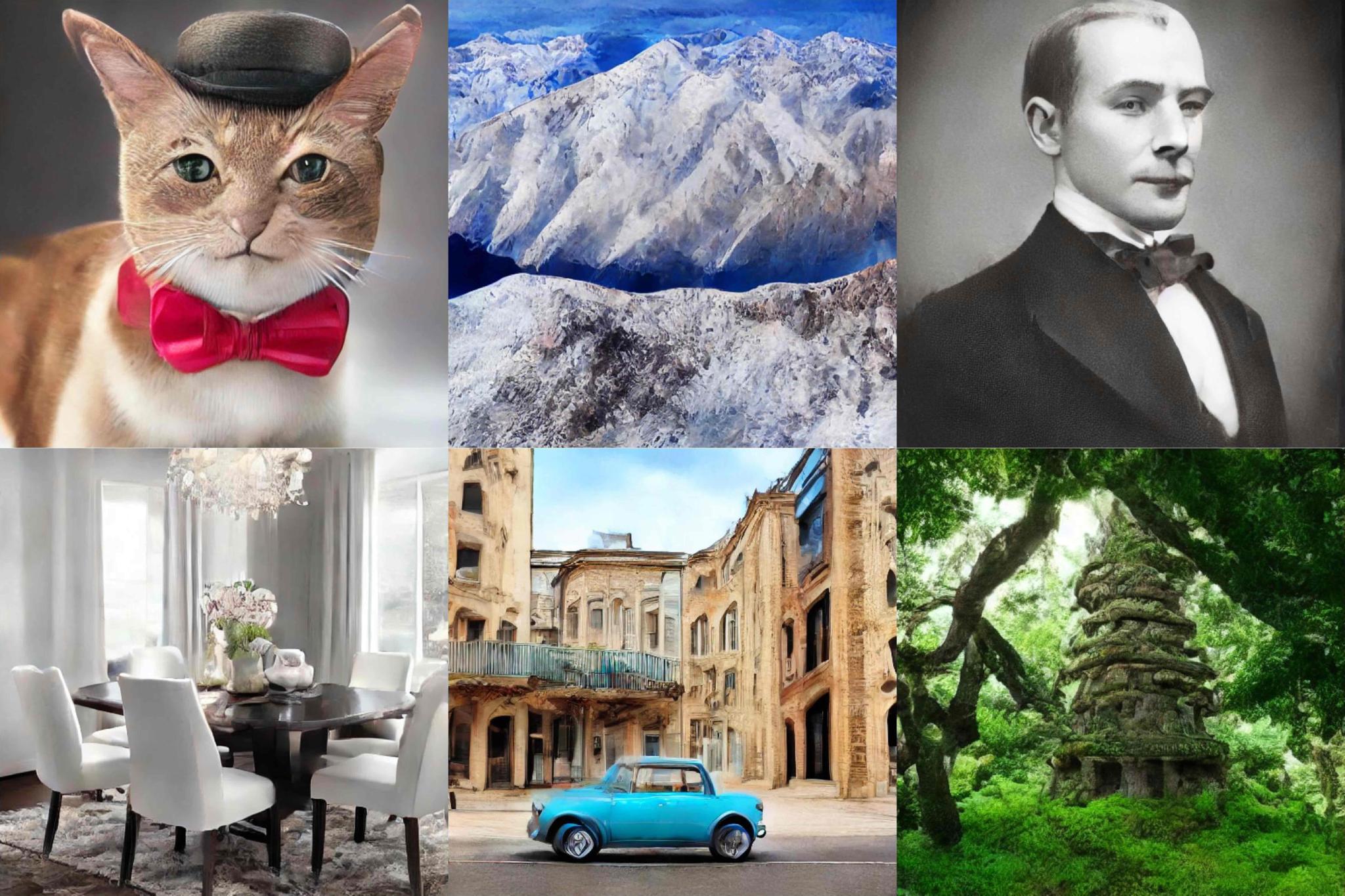}}\hspace{0.05em}
  \subfloat[\scriptsize \centering 2 NFEs]{\includegraphics[width=1.06in]{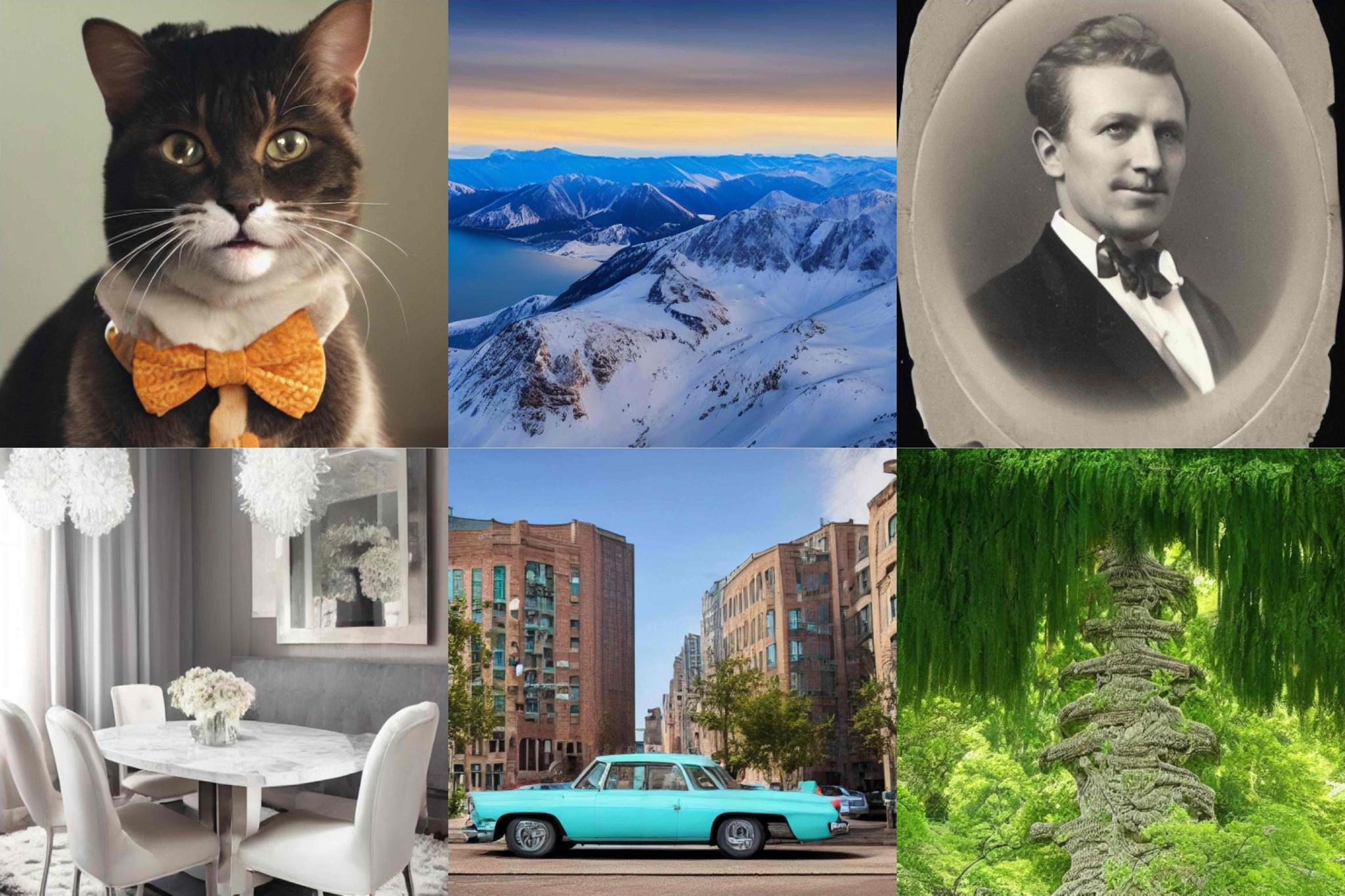}}\hspace{0.05em}
  \subfloat[\scriptsize \centering 4 NFEs]{\includegraphics[width=1.06in]{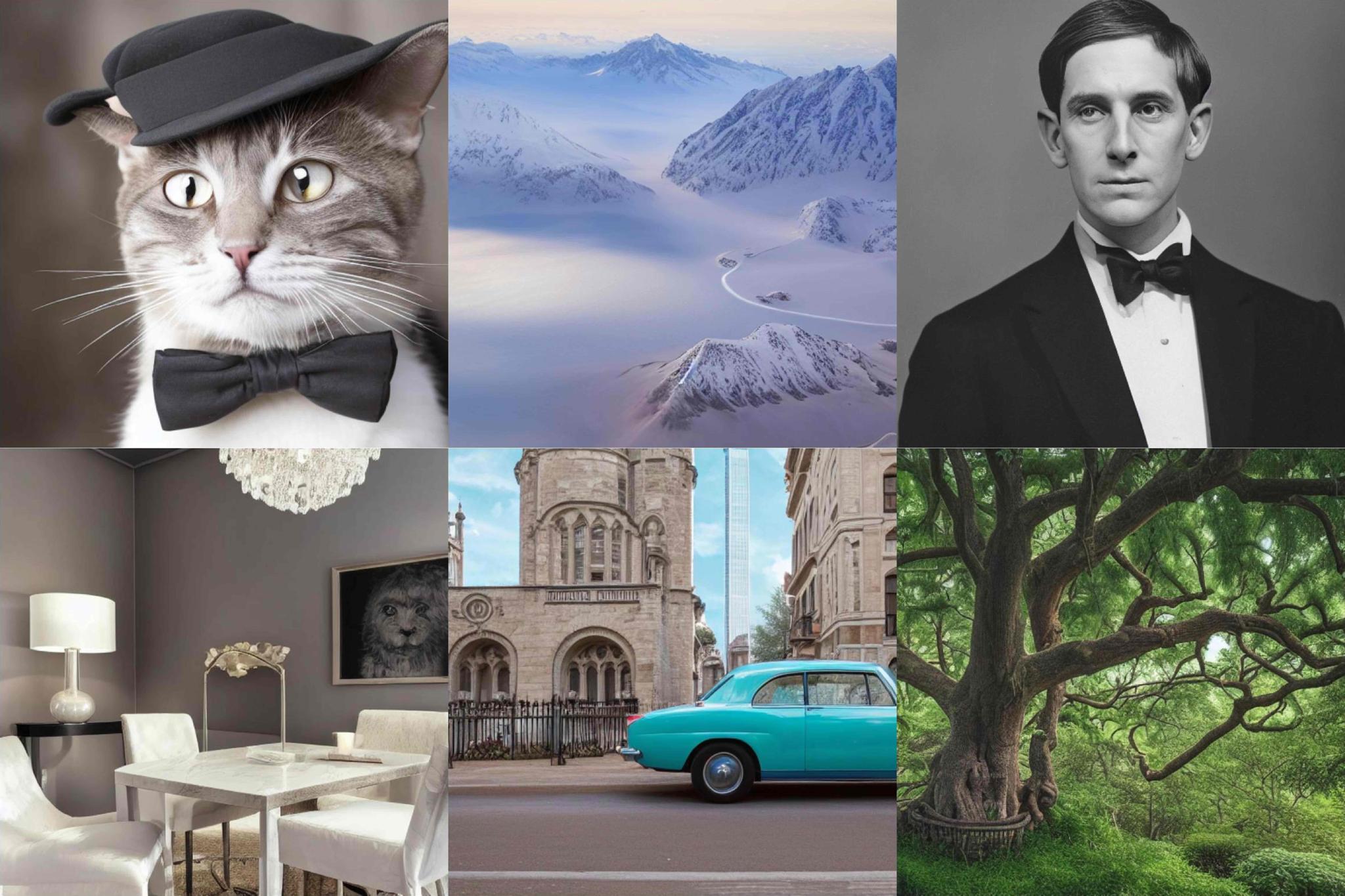}}
  \caption{Qualitative evaluation of the sample quality as the number of NFEs increases for the proposed method applied to SD1.5 model. Best viewed zoomed in.}
  \label{fig:steps}
  \end{figure}

\subsection{Model Training}
While striving for robustness and versatility, we also aimed to design a model with a minimal number of trainable parameters, since it involves the loading of computationally intensive functions (teacher and student).
To do so, we propose to rely on the parameter-efficient method LoRA \citep{hu2021lora} and apply it to our student model. This way, we drastically reduce the number of parameters and speed up the training process.

In a nutshell, our student model is trained to minimize a weighted combination of the distillation Eq. \eqref{eq:distillation loss}, the adversarial Eq. \eqref{eq:adv_loss}, and the distribution matching Eq. \eqref{eq:dmd} losses:
\begin{equation}\label{eq:loss}
  \mathcal{L} = \mathcal{L}_{\mathrm{distil}} + \lambda_{\mathrm{adv}} \mathcal{L}_{\mathrm{adv}} + \lambda_{\mathrm{DMD}}\mathcal{L}_{\mathrm{DMD}}\,.
\end{equation} 
The training process is illustrated in Fig.~\ref{fig:flash_diffusion scheme} and detailed in the appendices.


\section{Experiments}\label{sec:experiments}
In this section, we assess the effectiveness of our proposed method across various tasks and datasets. First, as it is common in the literature, we quantitatively compare the method with several approaches in the context of text-to-image generation.
Then, we conduct an extensive ablation study to assess the importance and impact of each component proposed in the method. Finally, we highlight the versatility of our method across several tasks, conditioning, and denoiser architectures.

\subsection{Text-to-Image Quantitative Evaluation}\label{sec:text-to-image-quantitative}
First, we apply our distillation approach to the publicly available SD1.5 model \citep{rombach2022high} and report both FID \citep{heusel2017gans} and CLIP score \citep{radford2021learning} on the COCO2014 and COCO2017 datasets \citep{lin2014microsoft}. The model is trained on the LAION dataset \citep{schuhmann2022laion} with aesthetic scores above 6. For COCO2017, we rely on the evaluation approach proposed in \citep{meng2023distillation} and we pick 5,000 prompts from the validation set to generate synthetic images. For COCO2014, we follow \citep{kang2023scaling} and pick 30,000 prompts from the validation set. We then compute the FID against the real images in the respective validation sets. We report the results in Tables (a) and (b) in Fig.~\ref{tab:ablations}. Our method achieves a FID of 22.6 and 12.27 on COCO2017 and COCO2014 respectively with only 2 NFEs corresponding to SOTA results for few steps image generation. On COCO2017, our approach also achieves a CLIP score of 30.6 and 31.1 for 2 and 4 NFEs respectively. Importantly, our method only requires the training of 26.4M parameters (out-of the 900M teacher parameters) and merely 26 H100 GPUs hours of training time. This is in stark contrast with many competitors who depend on training the entire UNet architecture of the student. See the appendices for more details on the training procedure.


\subsection{Ablation Study}\label{sec:ablation}
In this section, we conduct a comprehensive ablation study to assess the influence of the main parameters and choices made in the proposed method. For all the ablations, we train the model for 20k iterations with SD1.5 model as a teacher. All the results are reported on the COCO2017 using 2 NFEs.


\begin{figure*}[p]
  \scriptsize
  \captionsetup[subfigure]{position=below}
  \begin{minipage}[b]{0.65\linewidth}
  \centering
  \subfloat[]{
  \begin{tabular}[b]{lcccc}
\toprule                 
\multirow{2}{*}{Method (\# NFE)} & \# Train. &\multirow{2}{*}{FID $\downarrow$} & \multirow{2}{*}{CLIP $\uparrow$} \\
& Param.&&&\\
\midrule
SD1.5 (50)&  \multirow{2}{*}{N/A} &  20.1 & 31.8  \\
SD1.5 (16) &&  31.7 & 32.0\\
\midrule
Prog. Distil. (2) & \multirow{3}{*}{900M} & 37.3 & 27.0\\
Prog. Distil. (4)&   & 26.0 & 30.0 \\
Prog. Distil. (8)&  & 26.9 & 30.0\\
\midrule
InstaFlow (1)& 900M & 23.4 & 30.4 \\
\midrule
    CFG Dist. (16) &  850M &  24.2 & 30.0 \\
\midrule
Ours (2)  &  \multirow{2}{*}{26.4 M} &  22.6 & 30.6 \\
Ours (4)&   &  \textbf{22.5} & \textbf{31.1} \\
\bottomrule
\end{tabular}}\quad
\subfloat[]{\begin{tabular}[b]{lcccc}
\toprule                 
\multirow{2}{*}{Method (\# NFE)} & \# Train.  &\multirow{2}{*}{FID $\downarrow$}\\
& Param.&&\\
\midrule
DPM++$^{\dagger}$ (8)& N/A & 22.44\\
UniPC$^{\dagger}$  (8)& N/A & 23.30\\
\midrule
UFOGen (1)& 1,700M &  12.78 \\
InstaFlow (1) & 900M & 13.10 \\
DMD$^{\dagger}$ (1) & 1,700M &  14.93\\
\midrule
LCM-LoRA$^{\dagger}$ (1)&\multirow{3}{*}{67.5M} & 77.90 \\
LCM-LoRA$^{\dagger}$ (2)& &  24.28 \\
LCM-LoRA$^{\dagger}$ (4)& &  23.62 \\
\midrule
Ours (2)  & \multirow{2}{*}{26.4M} & \textbf{12.27}\\
Ours (4) &  & 12.41\\
\bottomrule
\end{tabular}}
\end{minipage}
\begin{minipage}[b]{0.35\linewidth}
  \subfloat[]{\includegraphics[width=0.9\linewidth]{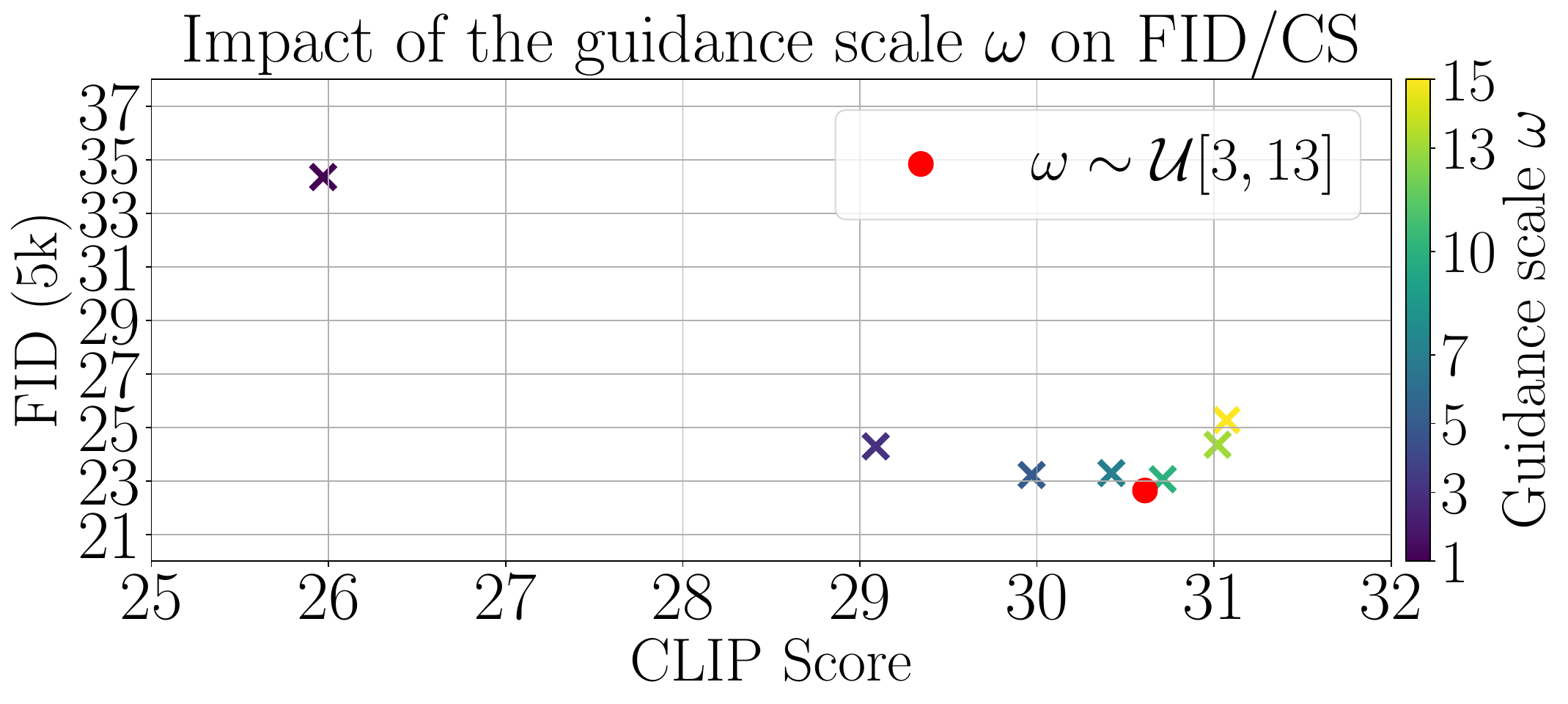}}\quad
  \subfloat[]{
  \begin{tabular}[b]{lccccc}
\toprule                 
Loss & FID $\downarrow $& CLIP $ \uparrow$\\
\midrule
\scriptsize
$\mathcal{L}_{\mathrm{distil.}} $& 27.12 & 29.85\\
$\mathcal{L}_{\mathrm{distil.}} + \mathcal{L}_{\mathrm{DMD}}$ &  26.88 & 30.45 \\
$\mathcal{L}_{\mathrm{distil.}} + \mathcal{L}_{\mathrm{adv}}$  &  23.41 & 30.14\\
$\mathcal{L}_{\mathrm{distil.}} + \mathcal{L}_{\mathrm{DMD}} + \mathcal{L}_{\mathrm{adv}}$ & \textbf{22.64} & \textbf{30.61} \\
\bottomrule
\end{tabular}}

  \end{minipage}
  \begin{minipage}[b]{\linewidth}
  \centering
\subfloat[]{
    \begin{tabular}[b]{lccccc}
      \toprule                 
      $\pi(t)$ & FID $\downarrow $& CLIP $ \uparrow$\\
      \midrule
      $\pi^{\mathrm{uniform}}(t)$ & 24.25 & 30.11 \\
      $\pi^{\mathrm{gaussian}}(t)$ & 35.89 & 28.15 \\
      $\pi^{\mathrm{sharp}}(t)$  & 23.35 &  30.58\\
      $\pi^{\mathrm{ours}}(t)$ & \textbf{22.64} & \textbf{30.61}\\
      \bottomrule
      \end{tabular}}\quad
  \subfloat[]{
  \begin{tabular}[b]{lccccc}
  \toprule                 
  $\mathcal{L}_{\mathrm{distil.}}$ & FID $\downarrow $& CLIP $ \uparrow$\\
  \midrule
  LPIPS & 24.89 & 30.56\\
  MSE & \textbf{22.64} & \textbf{30.61} \\
  \bottomrule
  \end{tabular}}\quad
              \subfloat[]{
    \begin{tabular}[b]{lccccc}
    \toprule                 
    $\mathcal{L}_{\mathrm{adv.}}$ & FID $\downarrow $& CLIP $ \uparrow$\\
    \midrule
    Hinge & 25.02 & 30.17 \\
    WGAN  & 24.58 & 30.36 \\
    LSGAN & \textbf{22.64} & \textbf{30.61}\\
    \bottomrule
    \end{tabular}}\quad
    \subfloat[]{
  \begin{tabular}[b]{lccccc}
  \toprule                 
  $K$ & FID $\downarrow $& CLIP $ \uparrow$\\
  \midrule
  16 & 23.35 & 30.11 \\
  32 & \textbf{22.64} & \textbf{30.61}\\
  64 & 22.87 & 30.58\\
  \bottomrule
  \end{tabular}}

\end{minipage}

\caption{\emph{From left to right and top to bottom:} a) FID-5k and CLIP score on COCO2017 validation set for SD1.5 as teacher. b) FID-30k on MS COCO2014 validation set for SD1.5 as teacher ($^{\dagger}$ results from \citep{yin2023one}). c) Influence of the guidance scale used to generate with the teacher, d) the loss terms e) the timestep sampling $\pi(t)$, f) the distillation loss, g) the GAN loss and h) the value of $K$ in Eq.~\eqref{eq:mass distribution}.}
\label{tab:ablations}
\end{figure*}

\paragraph{Influence of the loss terms}
We first train the model using different loss combinations and report the results in Table (d) in Figure \ref{tab:ablations}. As highlighted in the table, both $\mathcal{L}_{\mathrm{adv}}$ and $\mathcal{L}_{\mathrm{DMD}}$ have a noticeable impact on the final performance since $\mathcal{L}_{\mathrm{adv}}$ seems to allow reaching a better image quality, as indicated by lower FID, while $\mathcal{L}_{\mathrm{DMD}}$ improves prompt adherence, reflected in higher CLIP scores.
Experiments conducted using only $\mathcal{L}_{\mathrm{adv}}$ and $\mathcal{L}_{\mathrm{DMD}}$ revealed notable inconsistencies and even divergence in outcomes, emphasizing the crucial contribution of the distillation loss to the method's stability and reliability. In Tables (f) and (g), we also report results for different $\mathcal{L}_{\mathrm{distil.}}$ (LPIPS \citep{zhang2018unreasonable} and MSE) and $\mathcal{L}_{\mathrm{adv}}$ (Hinge \citep{lim2017geometric}, WGAN \citep{arjovsky2017wasserstein} and LSGAN \citep{mao2017least}). For $\mathcal{L}_{\mathrm{distil.}}$, MSE allows to achieve better results in terms of FID and CLIP score than LPIPS. For the GAN loss, the use of LSGAN seems the best-suited choice and we also noticed that it leads to stabler trainings.

\paragraph{Influence of the timestep sampling}\label{sec:ablation_timesteps}
In this section, we stress the influence of $\pi(t)$, the timesteps distribution. We compare the proposed timestep distribution to a uniform distribution across $K=32$ timesteps, a normal distribution $\pi^{\mathrm{gaussian}}(t)$ centered on $t=0.5$ and $\pi^{\mathrm{sharp}}$, a \emph{sharp} version of our proposed distribution that only allows sampling 4 distinct timesteps. Results are shown in Table (e) of Fig.~\ref{tab:ablations}. The proposed distribution significantly improves the performance compared to $\pi^{\mathrm{uniform}}$ and $\pi^{\mathrm{gaussian}}$. Moreover, allowing to sample more than 4 distinct timesteps seems to be beneficial to the final performance since a noticeable decrease in the FID score is observed. This can be explained by the fact that the student model can distil more useful information from the teacher model by sampling a wider range of timesteps and not over-fit the 4 selected ones.



\paragraph{Influence of the guidance scale during training}
For this ablation, unlike in the previous sections, we generate samples from the teacher model using a \textbf{fixed guidance scale $\omega$} set to either $1, 3, 5, 7, 10, 13$ or $15$. We report the evolution of the FID and CLIP score accordingly in graph (c) in Fig \ref{tab:ablations}. In line with the behavior observed with the teacher, the choice of the guidance scale has a strong impact on the final performance. While the CLIP score measuring prompt adherence tends to increase with the guidance scale, there exists a trade-off with the FID score that eventually increases with the guidance scale resulting in a potential loss of image quality. We represent by the red dot the setting that we propose which consists in uniformly sampling a guidance scale within a given range. 


\subsection{On the Method's Versatility}
To highlight the versatility of the proposed method, we apply the same approach to diffusion models trained with different conditionings, backbones, or adapters \citep{mou2024t2i}. 


\subsubsection{Backbones' Study}\label{sec: backbones}

\begin{figure*}[p]
  \centering
  \captionsetup[subfigure]{position=above, labelformat=empty}
  \subfloat[\scriptsize SDXL\\(40 NFEs)]{\includegraphics[width=.69in]{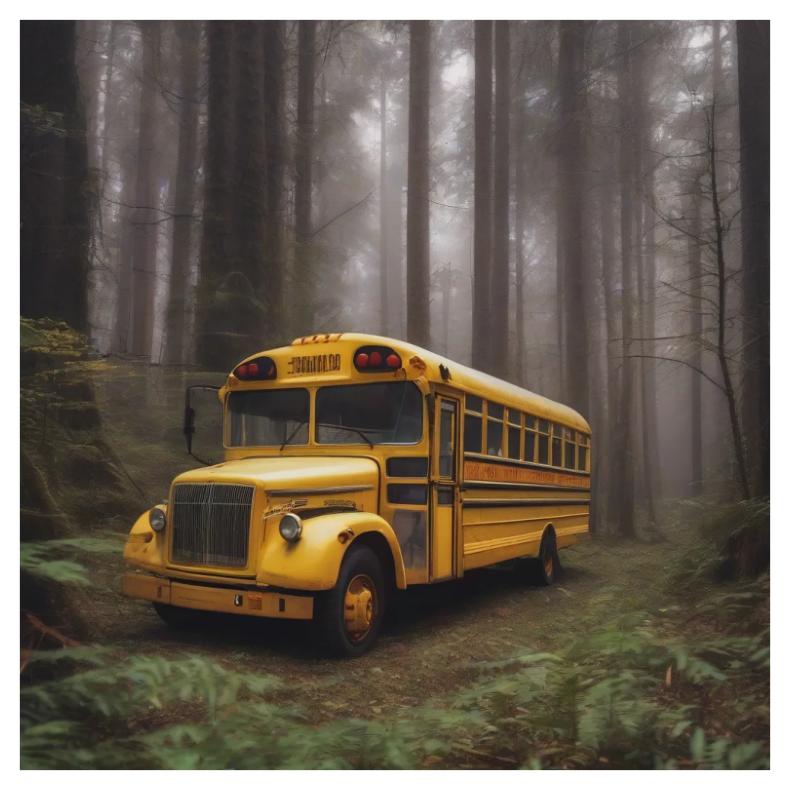}}\hspace{-0.5em}
  \subfloat[\scriptsize LCM\\(4 NFEs)]{\includegraphics[width=.69in]{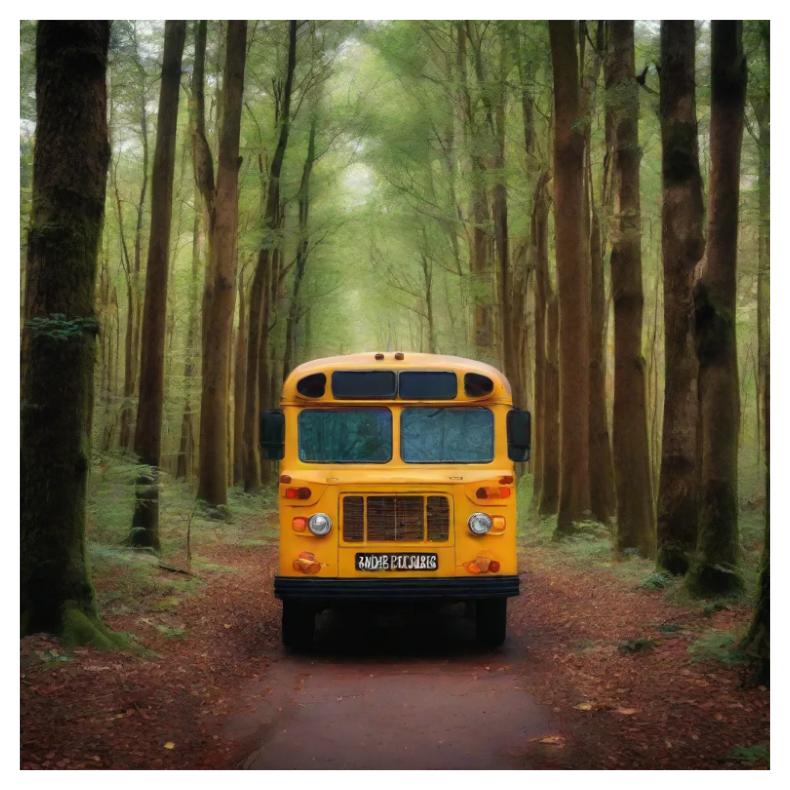}}\hspace{-0.5em}
  \subfloat[\scriptsize Lightning\\(4 NFEs)]{\includegraphics[width=.69in]{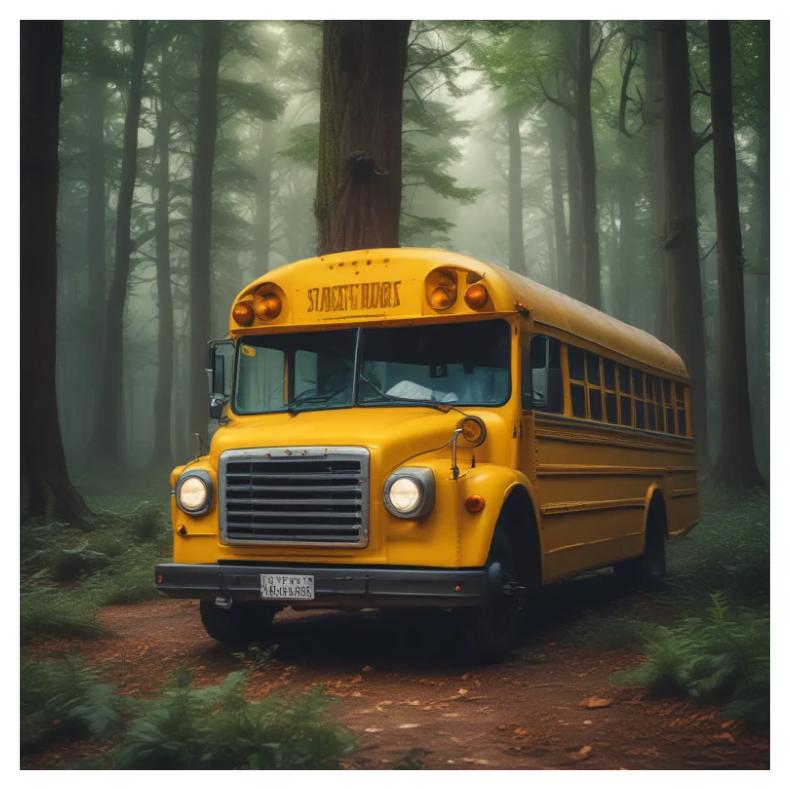}}\hspace{-0.5em}
  \subfloat[\scriptsize HyperSD\\(4 NFEs)]{\includegraphics[width=.69in]{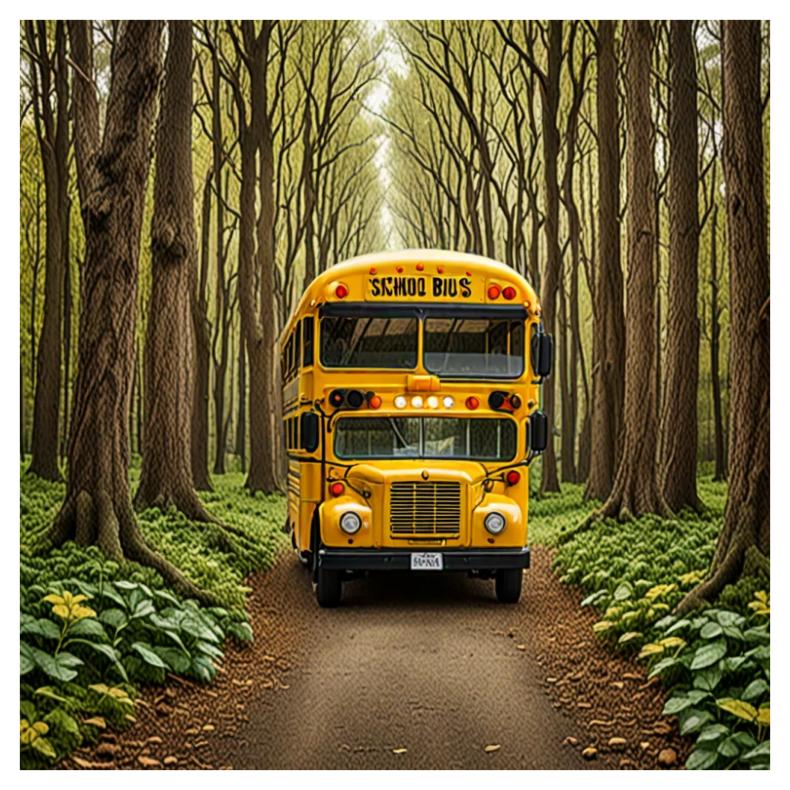}}\hspace{-0.5em}
  \subfloat[\scriptsize Ours\\(4 NFEs)]{\includegraphics[width=.69in]{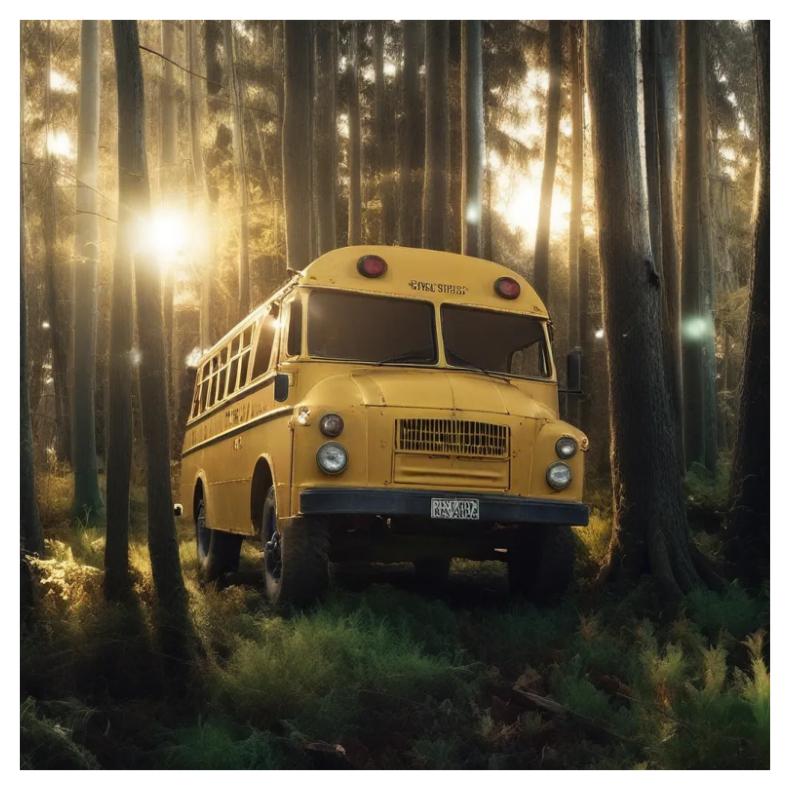}}\hspace{0.2em}
  \subfloat[\scriptsize Pixart-$\alpha$\\(40 NFEs)]{\includegraphics[width=.68in]{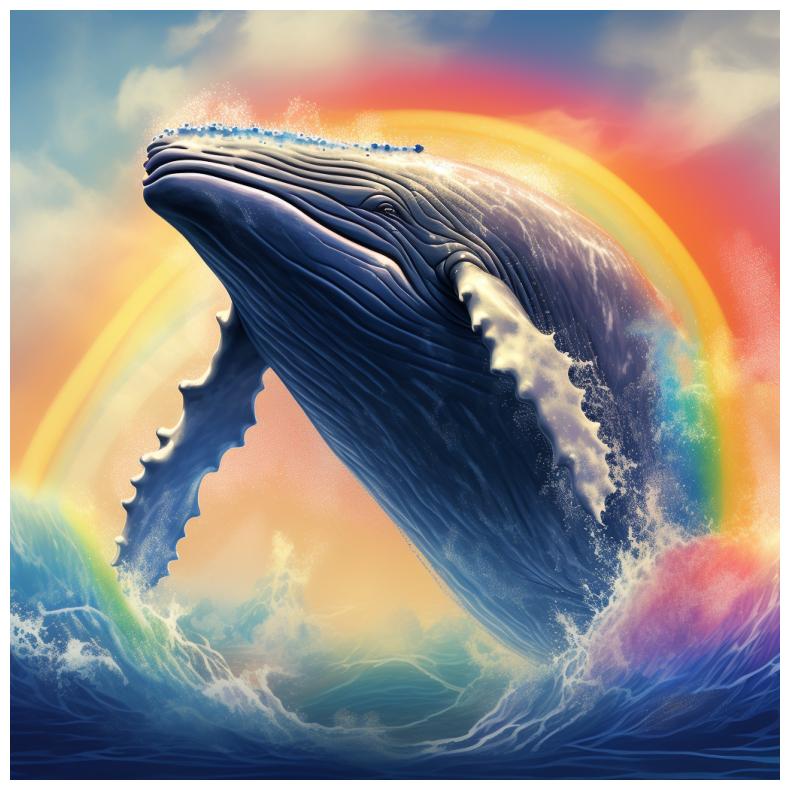}}
  \subfloat[\scriptsize LCM\\(4 NFEs)]{\includegraphics[width=.68in]{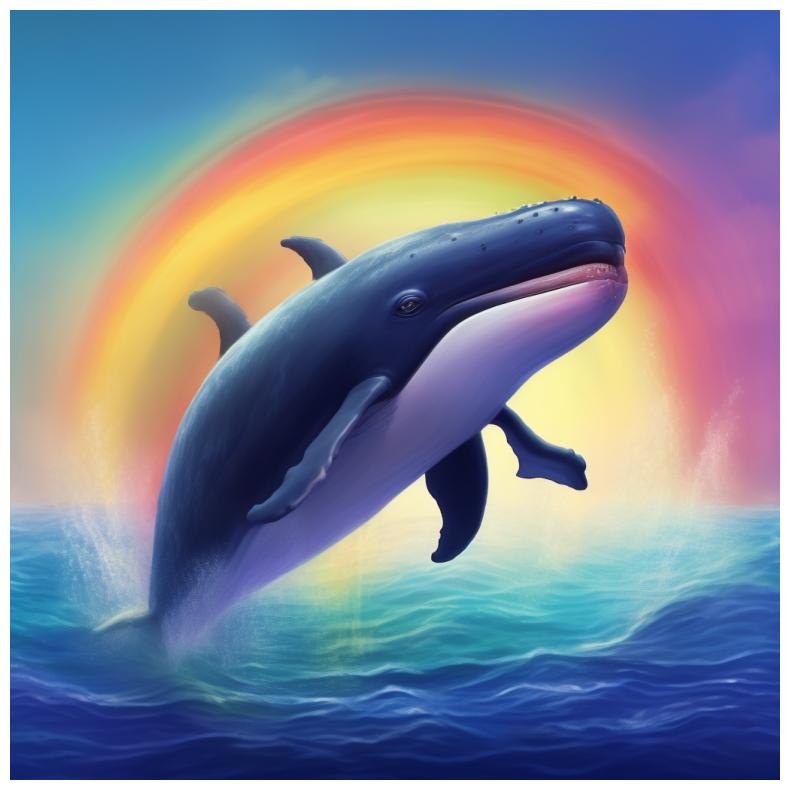}}
  \subfloat[\scriptsize Ours\\(4 NFEs)]{\includegraphics[width=.68in]{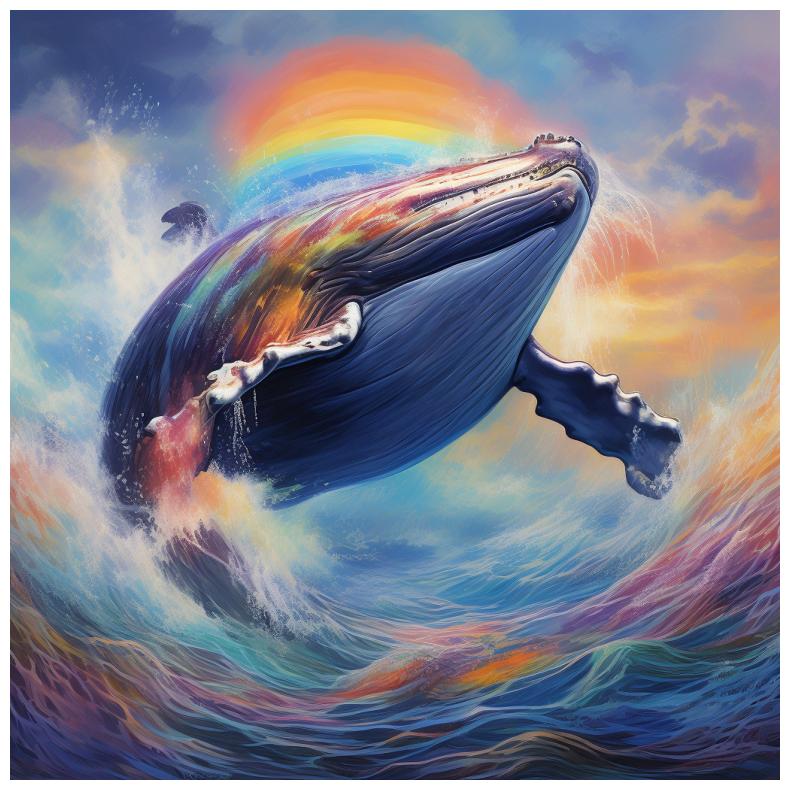}}\hspace{0.2em}
\subfloat[\scriptsize SD 3\\(40 NFEs)]{\includegraphics[width=.68in]{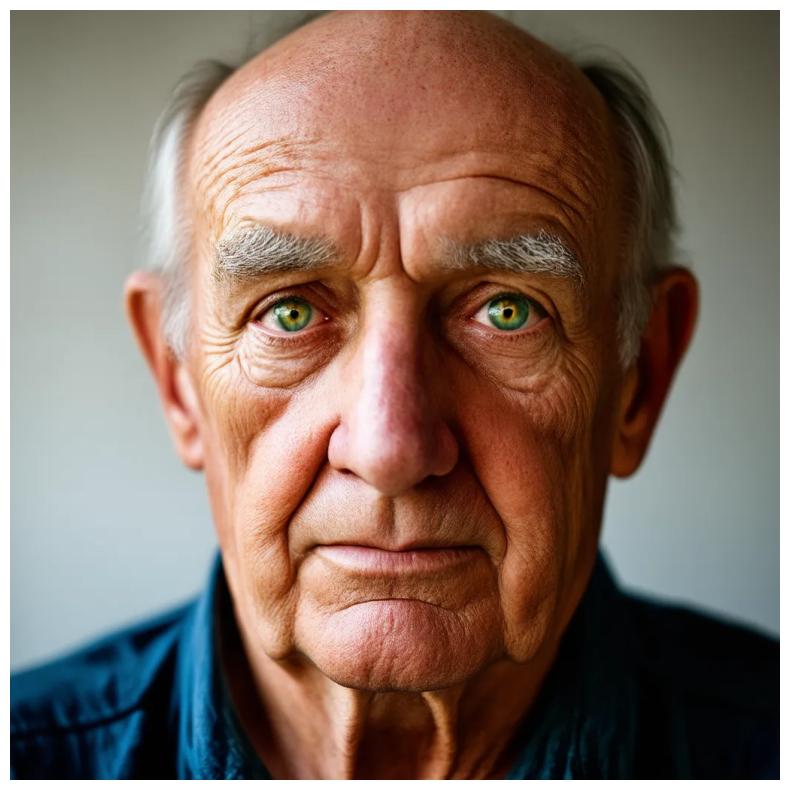}}
  \subfloat[\scriptsize Ours\\(4 NFEs)]{\includegraphics[width=.68in]{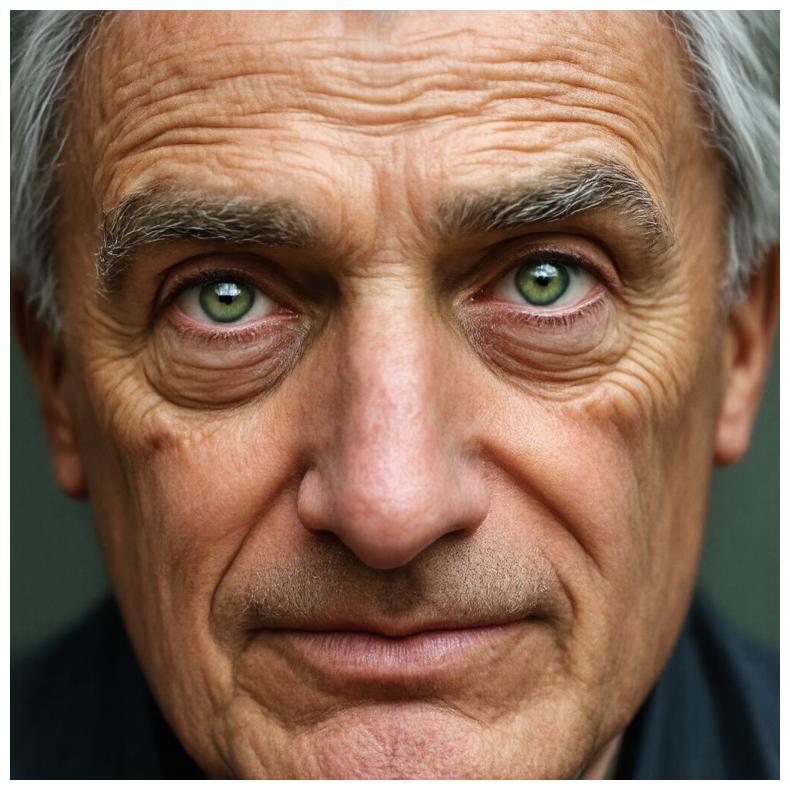}}
    \\[-1.2em]
  \subfloat{\includegraphics[width=.69in]{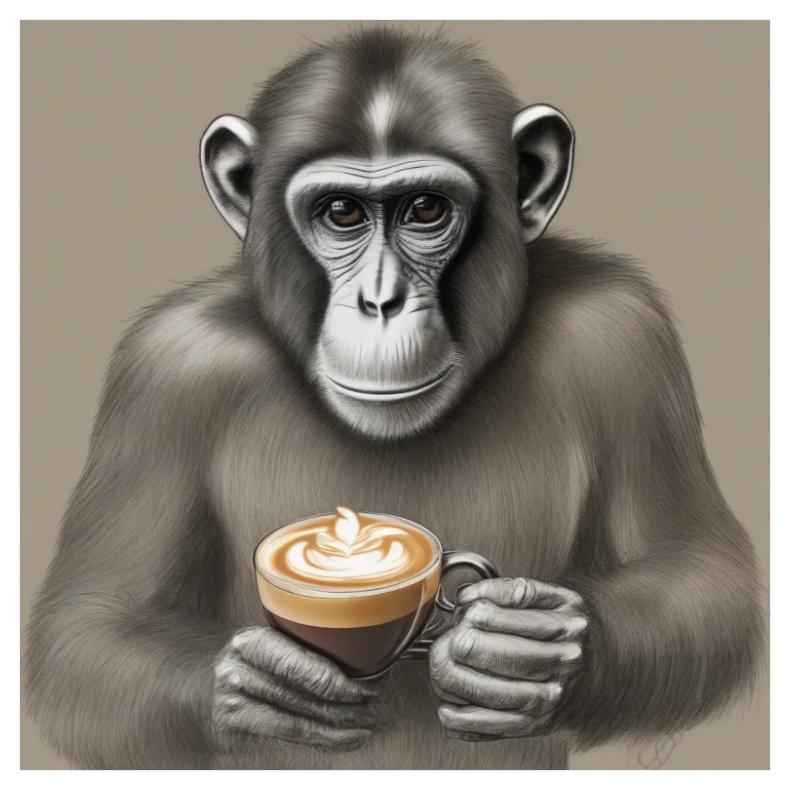}}\hspace{-0.5em}
  \subfloat{\includegraphics[width=.69in]{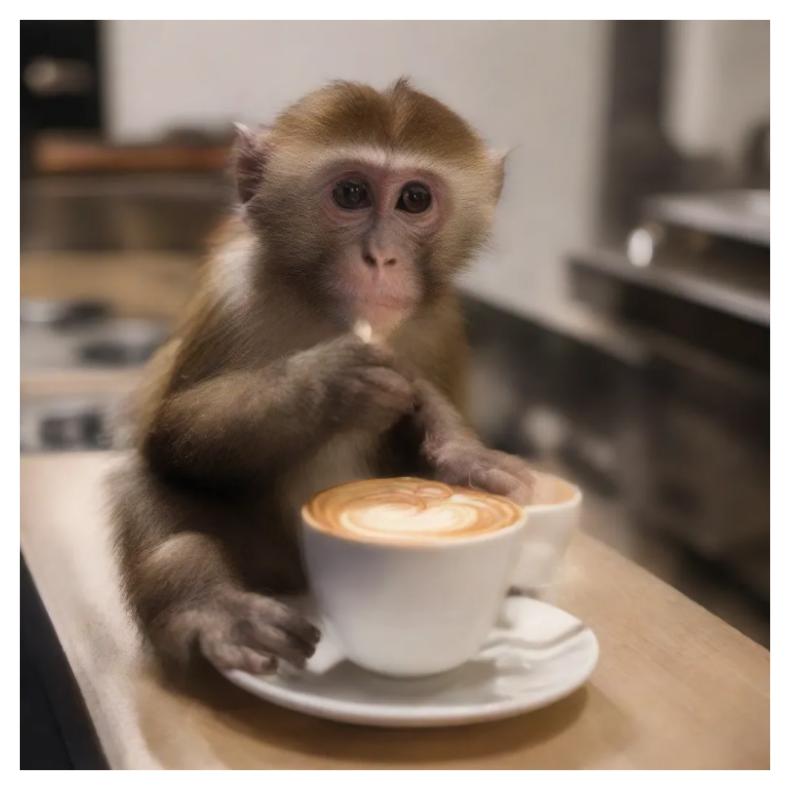}}\hspace{-0.5em}
  \subfloat{\includegraphics[width=.69in]{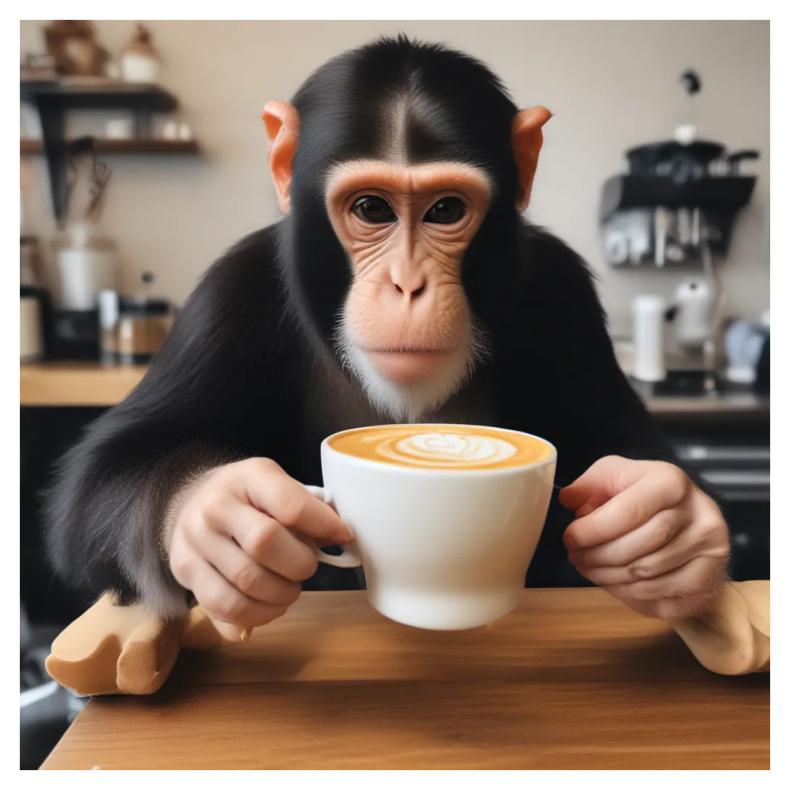}}\hspace{-0.5em}
  \subfloat{\includegraphics[width=.69in]{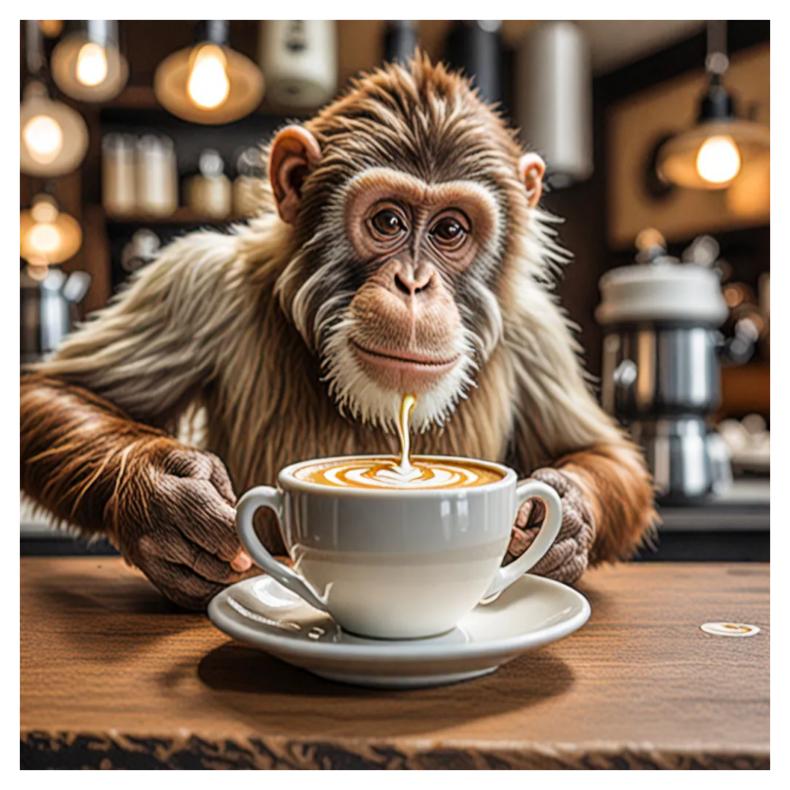}}\hspace{-0.5em}
  \subfloat{\includegraphics[width=.69in]{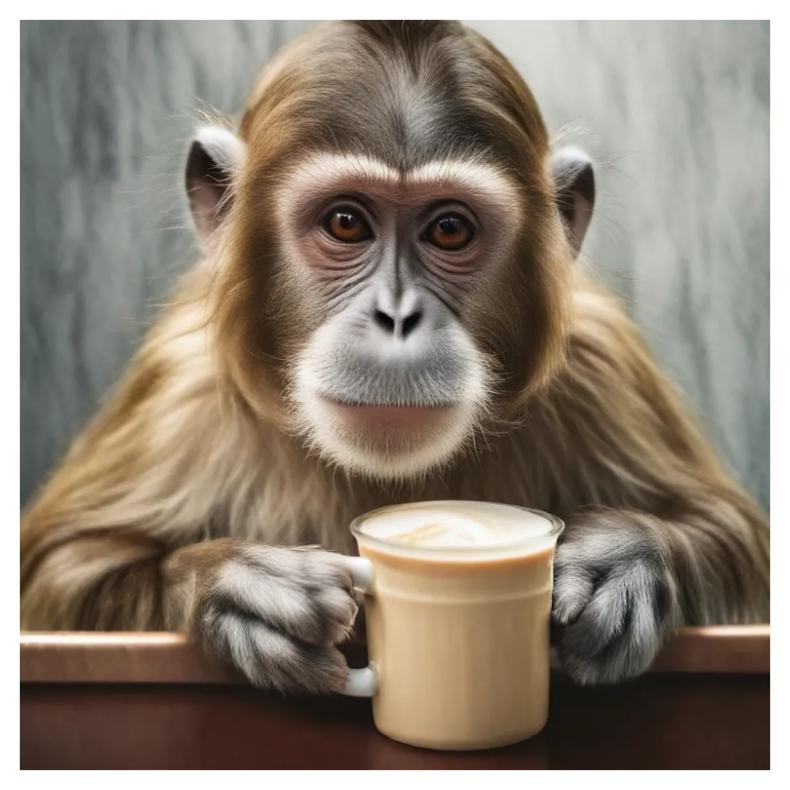}}\hspace{0.2em}
   \subfloat{\includegraphics[width=.68in]{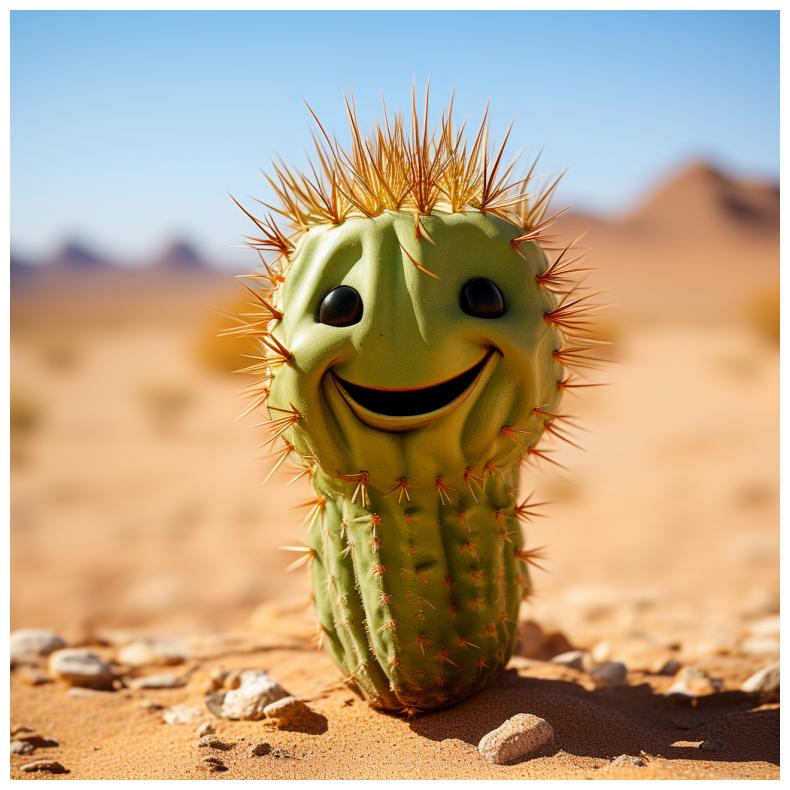}}
  \subfloat{\includegraphics[width=.68in]{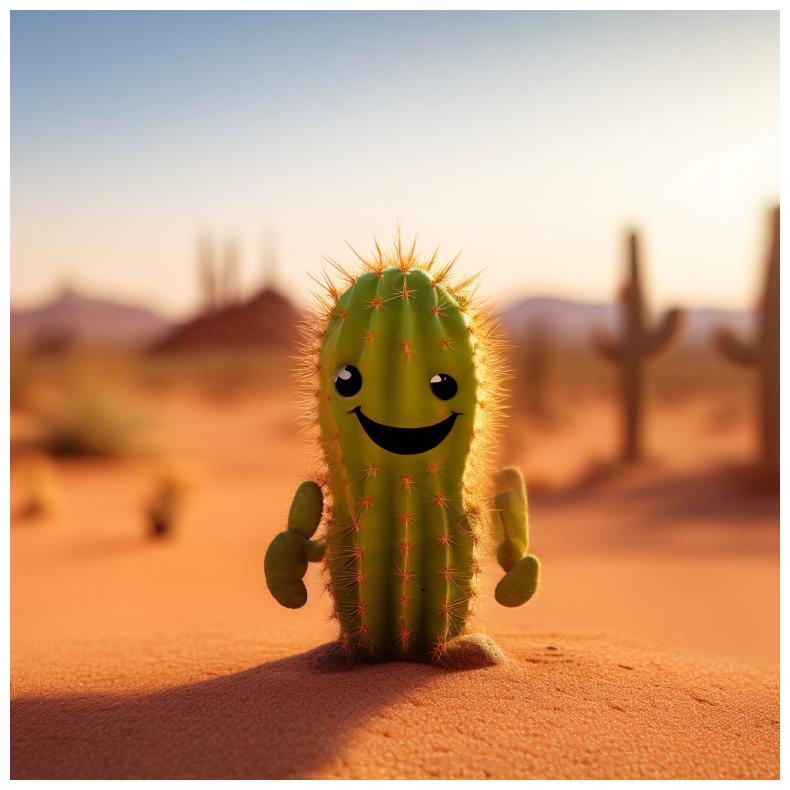}}
  \subfloat{\includegraphics[width=.68in]{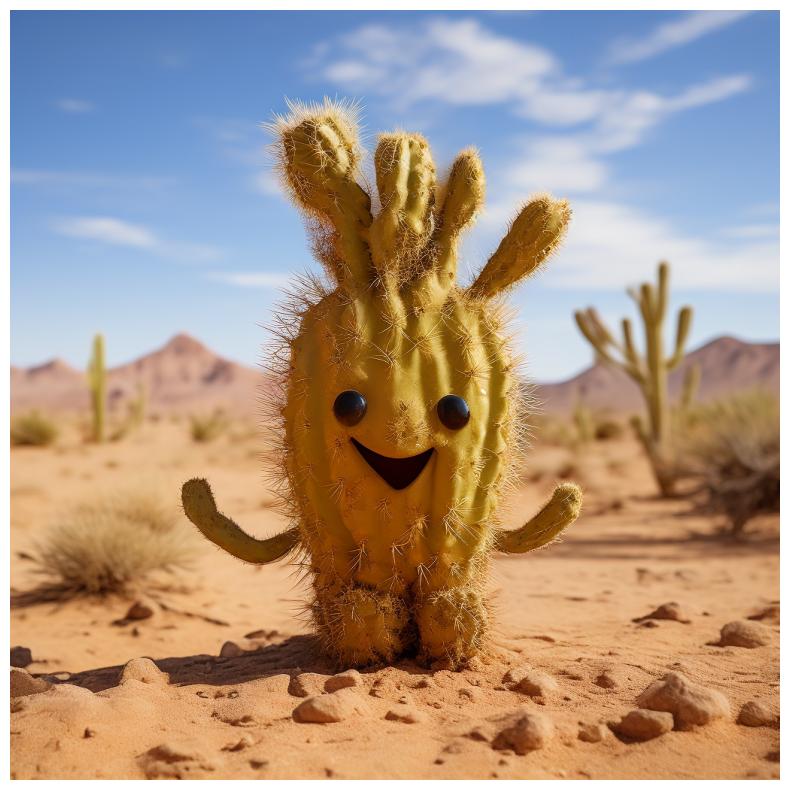}}\hspace{0.2em}
   \subfloat{\includegraphics[width=.68in]{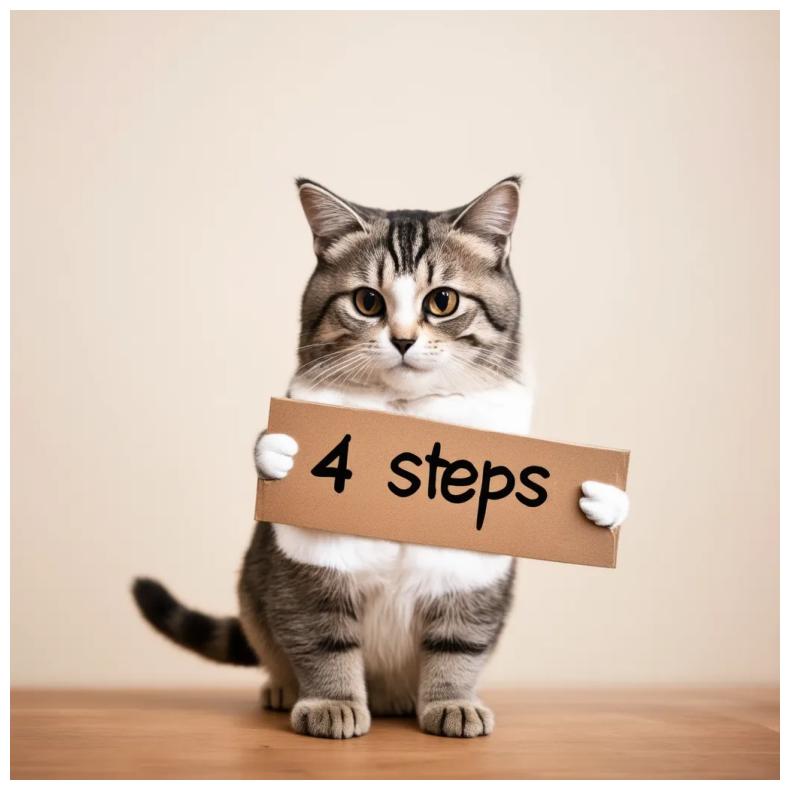}}
  \subfloat{\includegraphics[width=.68in]{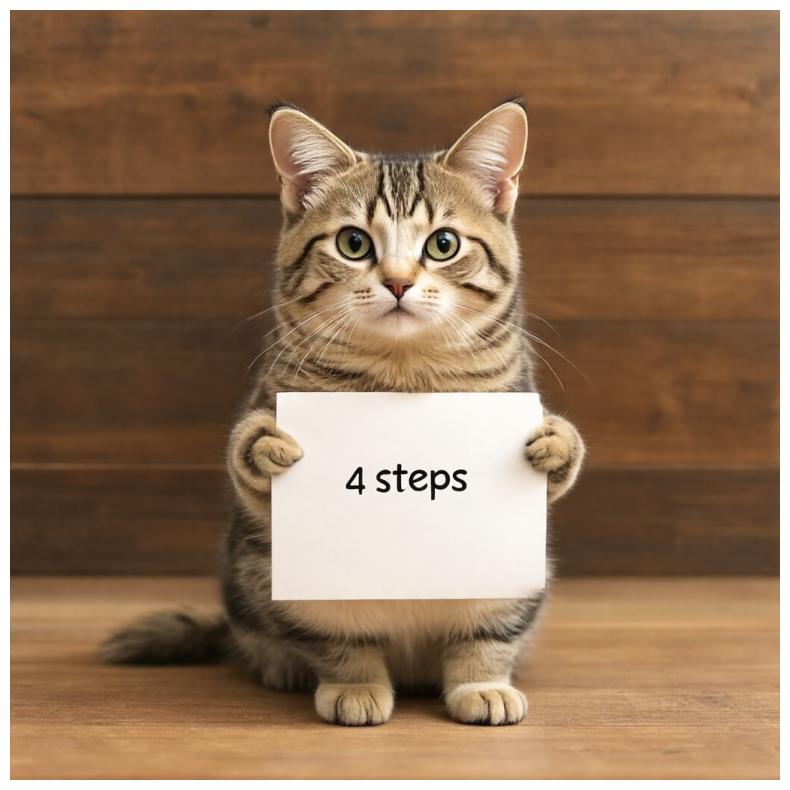}}
      \\[-1.2em]
    \subfloat{\includegraphics[width=.69in]{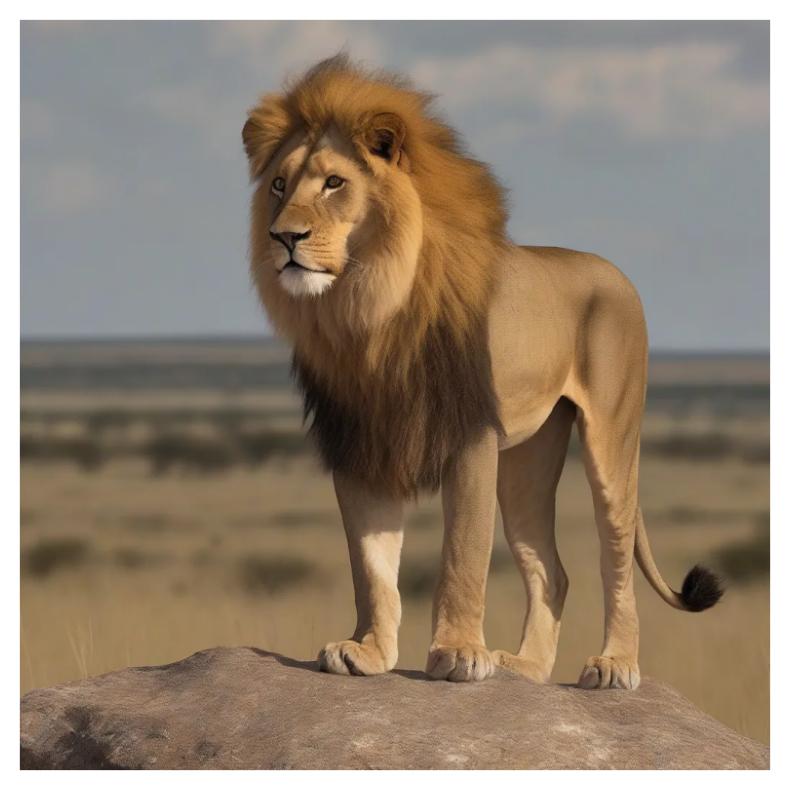}}\hspace{-0.5em}
  \subfloat{\includegraphics[width=.69in]{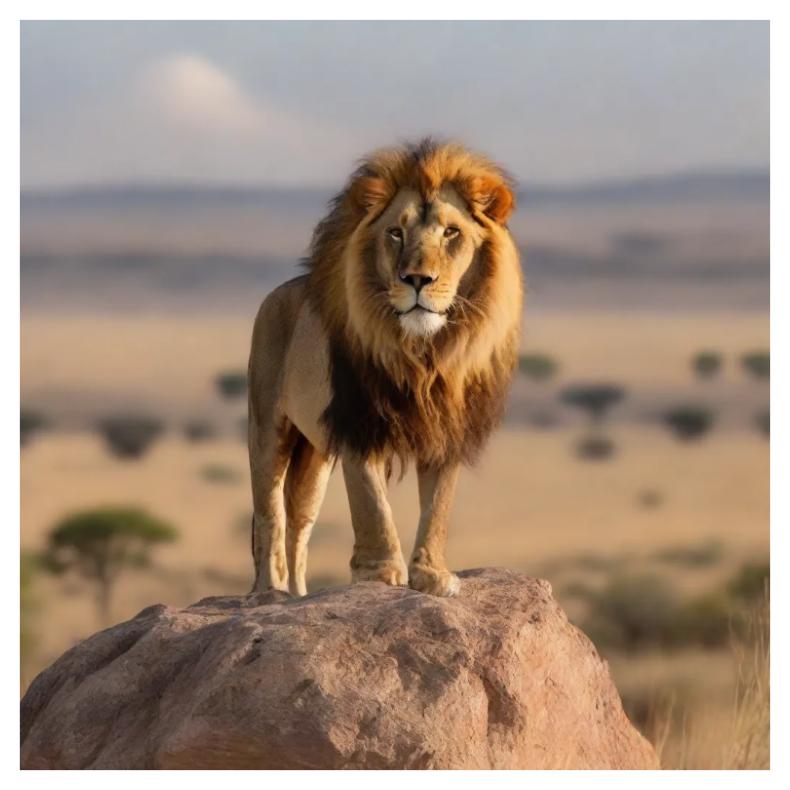}}\hspace{-0.5em}
  \subfloat{\includegraphics[width=.69in]{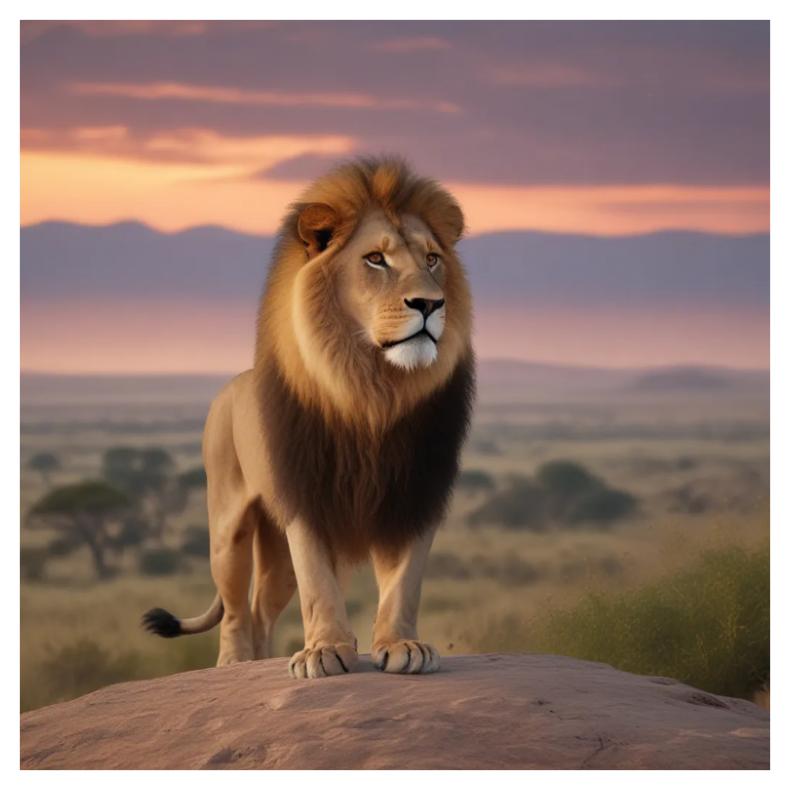}}\hspace{-0.5em}
  \subfloat{\includegraphics[width=.69in]{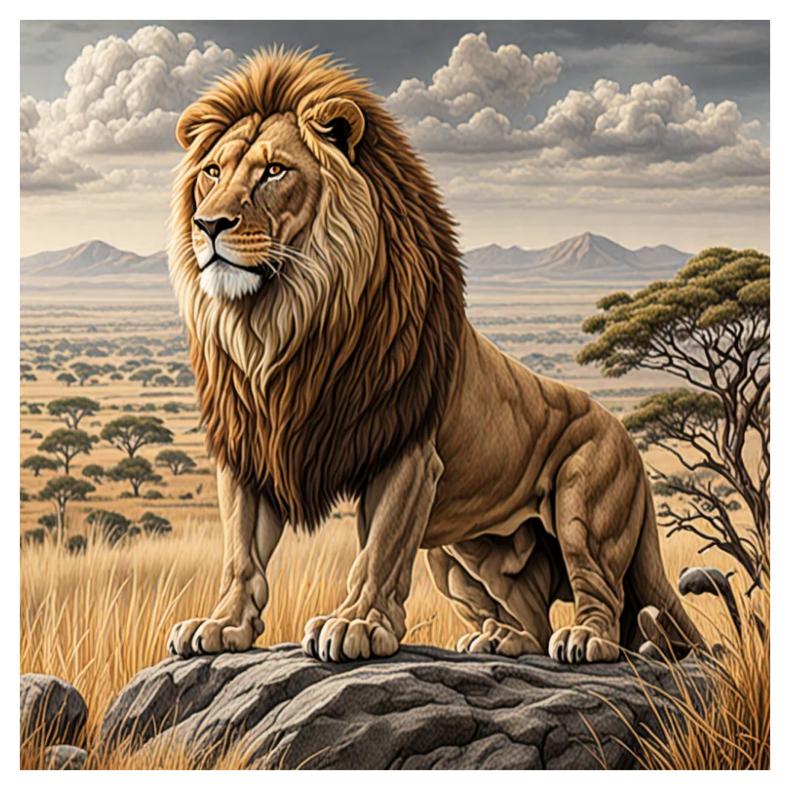}}\hspace{-0.5em}
  \subfloat{\includegraphics[width=.69in]{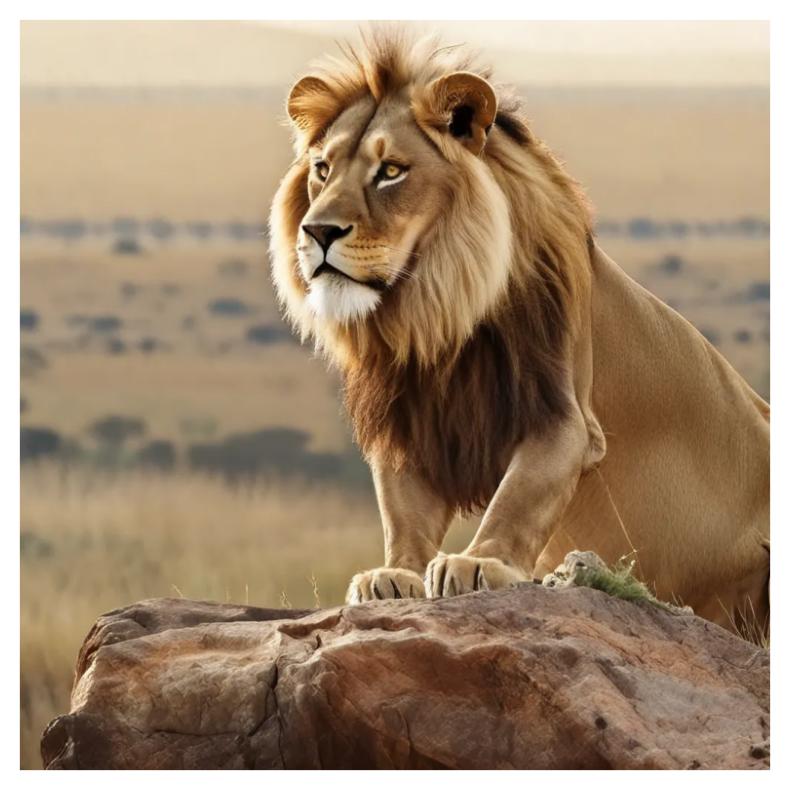}}\hspace{0.2em}
   \subfloat{\includegraphics[width=.68in]{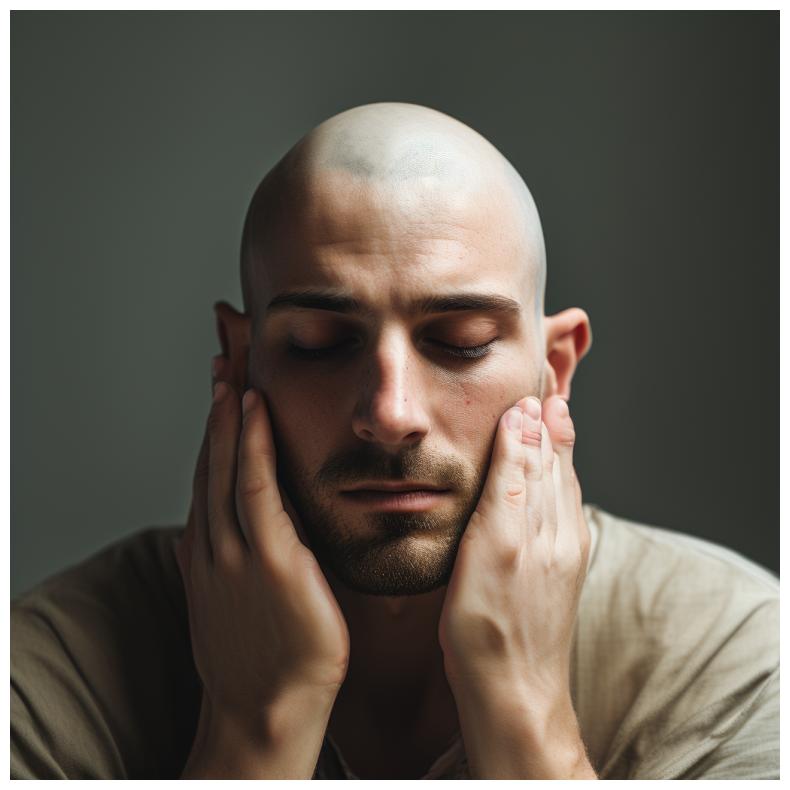}}
  \subfloat{\includegraphics[width=.68in]{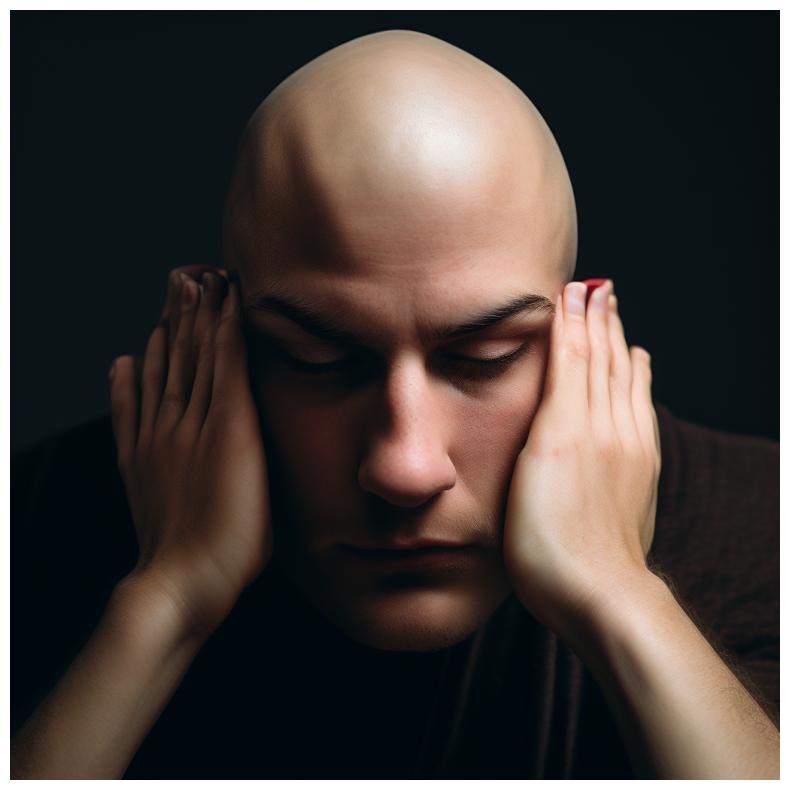}}
  \subfloat{\includegraphics[width=.68in]{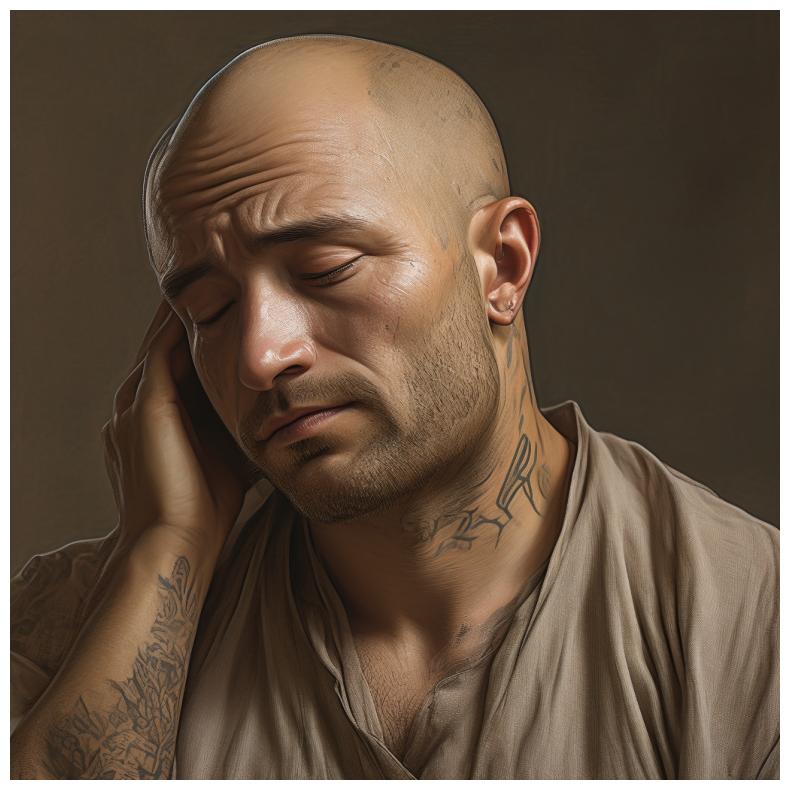}}\hspace{0.2em}
   \subfloat{\includegraphics[width=.68in]{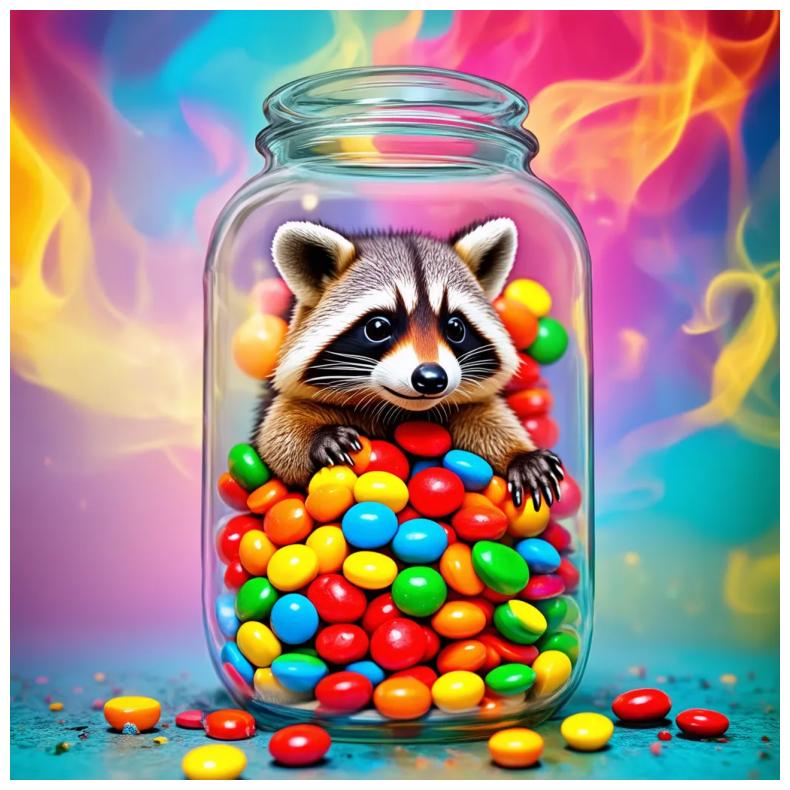}}
  \subfloat{\includegraphics[width=.68in]{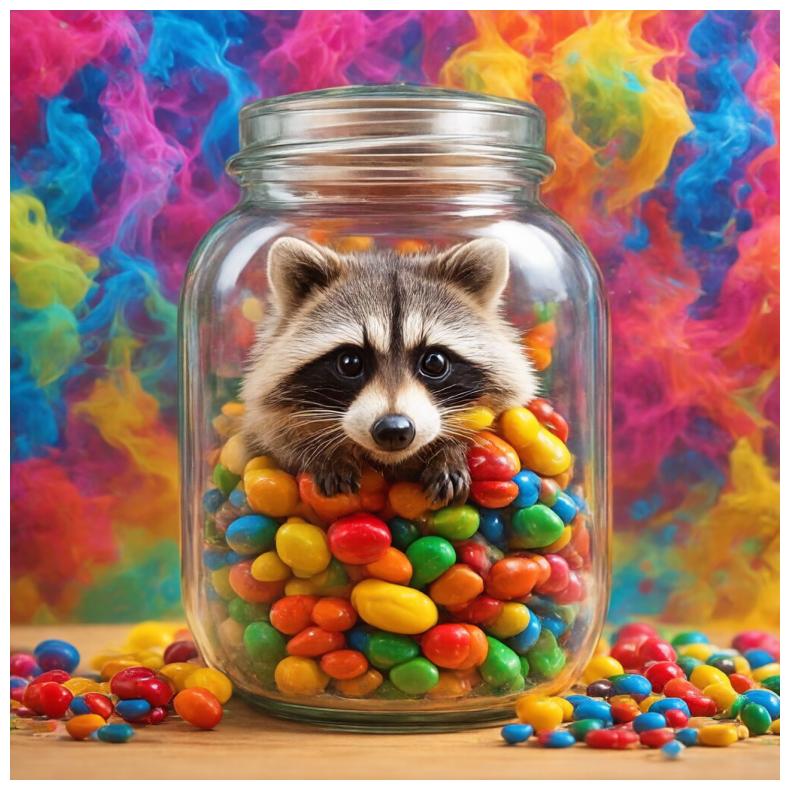}}

    \caption{\emph{From left to right:} Application of \emph{Flash Diffusion} to SDXL (UNet), Pixart-$\alpha$ (DiT) and Stable Diffusion 3 (MMDiT) teachers. Teacher samples are generated with a guidance scale of 5, 3, and 5 respectively. The proposed approach is compared to LoRA based competitors and appears to be able to generate samples that are visually closer to the learned teacher distribution. Best viewed zoomed in. Additional samples are provided in the appendices.}
        \label{fig:qualitative}


\end{figure*}

\paragraph{Flash SDXL}\label{sec:flash_sdxl}
In this section, we illustrate the ability of the method to adapt to a SDXL \citep{podell2023sdxl} teacher model. We provide in Table \ref{tab:COCO2014_dit_sdxl}, the FID and CLIP score computed on the 10k first prompts of COCO2014 validation set. We compare the proposed approach to several distillation methods proposed in the literature using publicly available checkpoints. Our method can outperform peers in terms of FID while maintaining quite good prompt alignment capabilities. In addition, we also provide a visual overview of the generated samples in Fig. \ref{fig:qualitative} for the teacher, the trained student model and LoRA-compatible approaches proposed in the literature (LCM \citep{luo2023latent}, SDXL-lightning \citep{lin2024sdxl} and Hyper-SD \citep{ren2024hyper}). Teacher samples are generated with a guidance scale of 5. For a fair comparison with competitors, we include prompts used in \citep{lin2024sdxl} for this qualitative evaluation. The proposed approach appears to be able to generate samples that are visually closer to the learned teacher distribution. In particular, HyperSD and lightning seem to struggle to generate samples that are realistic despite creating sharp samples. See the appendices for the comprehensive experimental setup and additional comparisons. Additionally, since our student share the same architecture as the teacher, we notice that our approach can be combined with existing LoRAs in a \emph{training-free} manner. We show at the bottom right of Fig.~\ref{fig:inpainting_upscaler}, 4 steps generations for 6 existing SDXL LoRAs directly plugged to our trained Flash SDXL model. We provide additional samples in the appendices.  

\begin{table}[ht]
  \scriptsize
  \begin{minipage}[t]{.45\linewidth}
\begin{tabular}[t]{lcccc}
  \toprule                 
  Model (\# NFE)&  FID $\downarrow$ & CLIP $\uparrow$\\
  \midrule
  SDXL (40) & 18.4 & 33.9\\
  \midrule
  LCM (8) & 21.7 & 32.7\\
  Turbo (4) & 23.7  & \textbf{33.7}\\
  Lightning (4) & 24.6 & 32.9\\
  \midrule
  Lightning$^{\dagger}$ (4) & 25.1 & 32.8\\
  HyperSD$^{\dagger}$ (4)  &27.8 & 33.3 \\
  \midrule
  Ours$^{\dagger}$ (4)  & \textbf{21.6} & 32.7\\
  \bottomrule
\end{tabular}
$^{\dagger}$ LoRAs
\end{minipage}
\hspace{1.2em}
\begin{minipage}[t]{.45\linewidth}
\scriptsize
\begin{tabular}[t]{lcccc}
\toprule                 
Model (\# NFE) & FID $\downarrow$ & CLIP $\uparrow$\\
\midrule
Pixart (40) &28.1 & 31.6 \\
\midrule
Ours$^{\dagger}$ (4) & 29.3 & 30.3\\
\bottomrule
\vspace{0.42em}
\end{tabular}
\begin{tabular}[b]{lcccc}
\toprule                 
Model (\# NFE) & FID $\downarrow$ & CLIP $\uparrow$\\
\midrule
SD3 (40) & 24.4 & 33.5 \\
\midrule
Ours$^{\dagger}$ (4) & 27.5 & 32.8\\
\bottomrule
\end{tabular}
\end{minipage}
  \caption{FID and CLIP score on 10k samples of COCO2014 validation set for SDXL, Pixart-$\alpha$ and SD3 teacher.}
  \label{tab:COCO2014_dit_sdxl}
\end{table}

\paragraph{Flash Pixart (DiT)}\label{sec:flash_dit}
In this section, we propose to apply the proposed method to a DiT denoiser backbone \citep{peebles2023scalable} using Pixart-$\alpha$ \citep{chen2023pixart} as teacher. We compare the student generations using 4 NFEs to the teacher generations using 40 NFEs (20 steps) as well as Pixart-LCM \citep{luo2023lcm} in Fig. \ref{fig:qualitative} and provide metrics in Table \ref{tab:COCO2014_dit_sdxl}. The proposed method can generate high-quality samples that sometimes seem even more visually appealing than the teacher. Moreover, driven by the adversarial approach the student model trained with our method generates images with more vivid colors and sharper details than LCM. It is noteworthy that the student model does not lose the capability of the teacher to generate samples that are coherent with the prompt. In addition, we provide in Table \ref{tab:COCO2014_dit_sdxl} FID and CLIP scores computed on the 10k first prompts of COCO2014 validation set for our model and the teacher. See the appendices for the comprehensive experimental setup and additional samples as well as discussion on the variability of the output samples with respect to the prompt.


\paragraph{Flash SD3 (MMDiT)}
Finally, we also show the compatibility of our approach with the recently propose MMDiT architecture of Stable Diffusion 3 \cite{esser2024scaling}. The method is again able to successfully distil the teacher model and generate samples in only 4 NFEs. We train a 90.4M parameter LoRA model with a batch size of 2 and a learning rate of $1e^{-5}$ together with Adam optimizer \citep{kingma2014adam} for both the student and the discriminator. We provide in Fig. \ref{fig:qualitative} samples generated with the teacher model and our method and quantitative results in Table.~\ref{tab:COCO2014_dit_sdxl}. 

\subsubsection{Conditionings' Study}

\begin{figure}[ht]
    \centering    
    \captionsetup[subfigure]{position=above, labelformat = empty}
    \subfloat[\scriptsize Original]{\includegraphics[width=.64in]{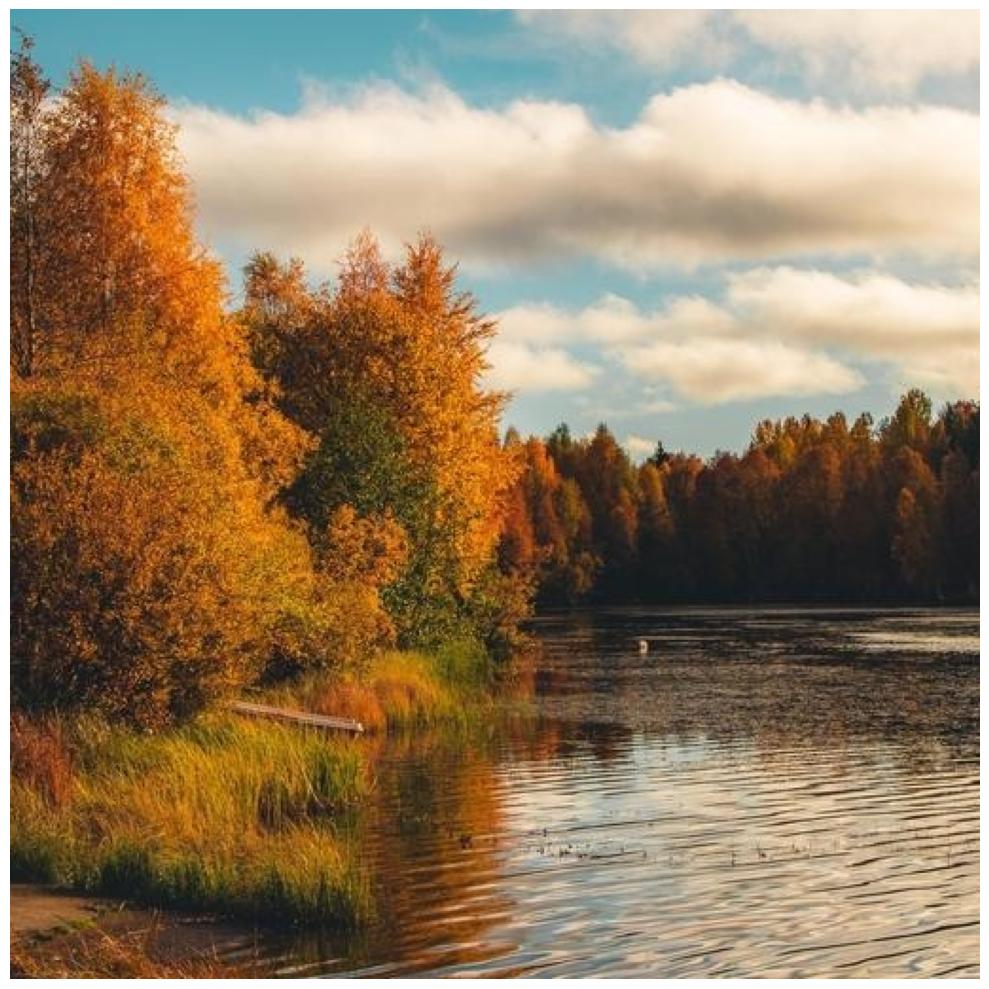}}
    \subfloat[\scriptsize Masked Image]{\includegraphics[width=0.64in]{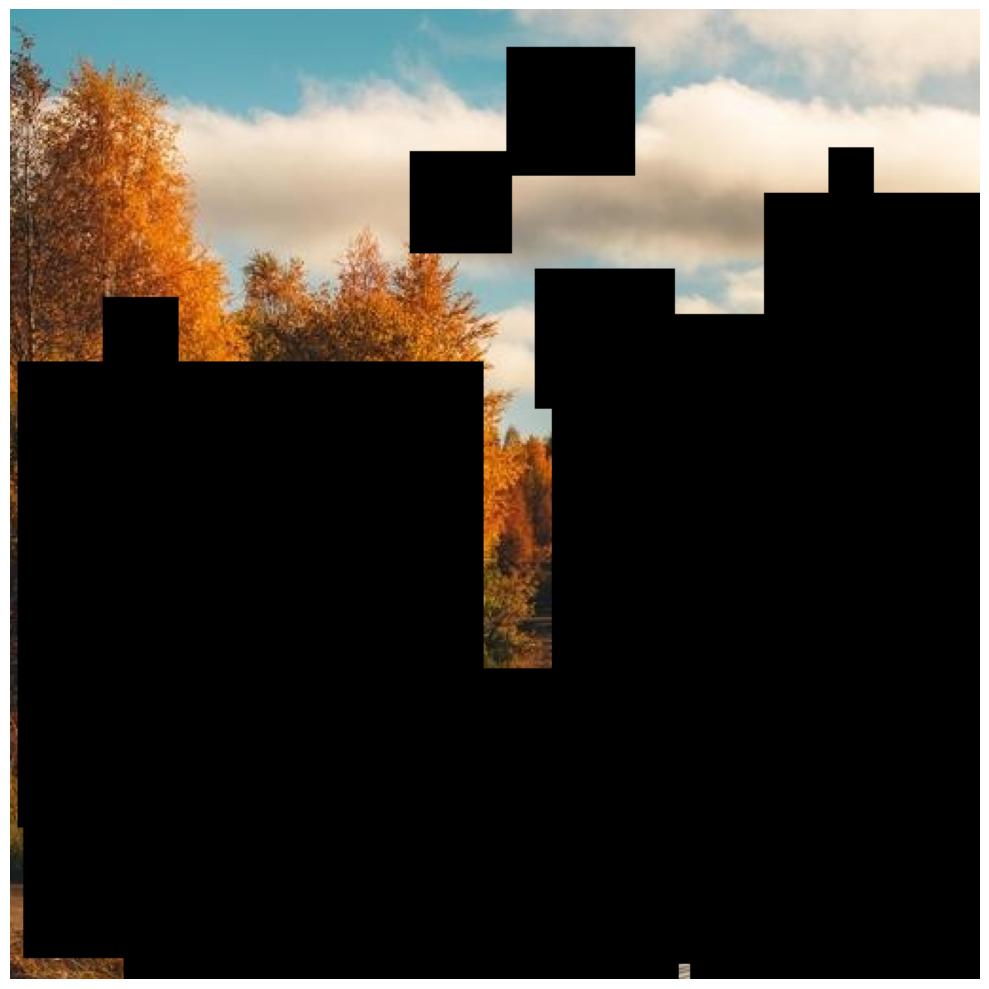}}
    \subfloat[\scriptsize Ref. (8 NFE)]{\includegraphics[width=0.64in]{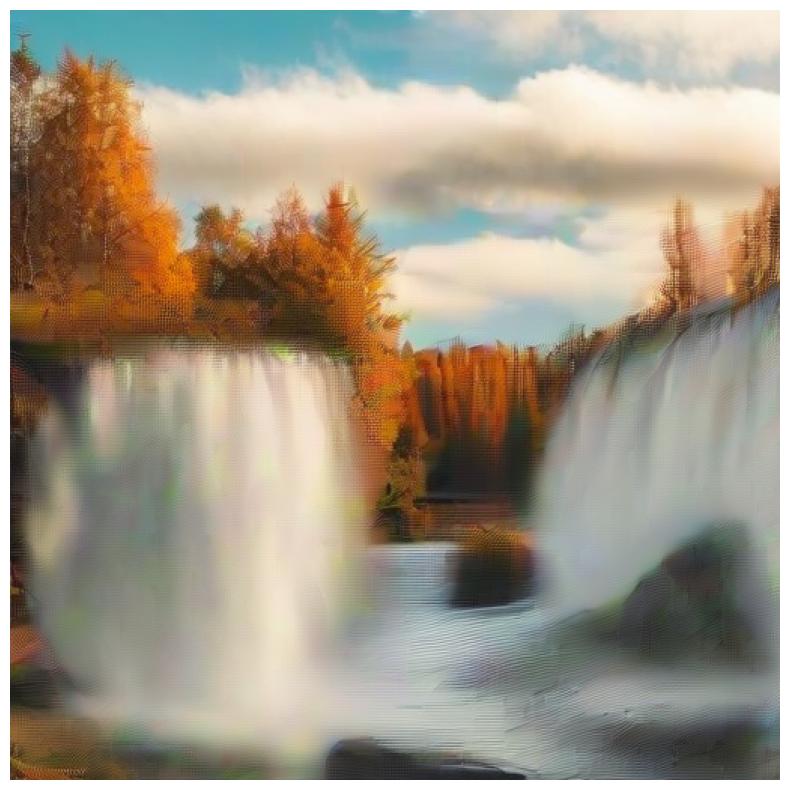}}
    \subfloat[\scriptsize Ref. (40 NFE)]{\includegraphics[width=0.64in]{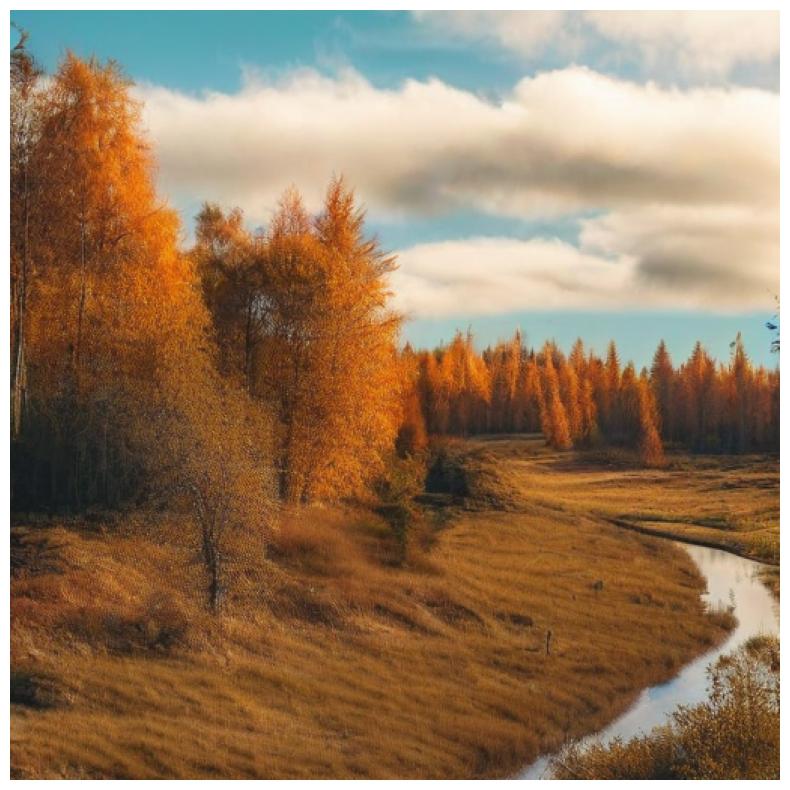}}
    \subfloat[\scriptsize Ours (4 NFE)]{\includegraphics[width=0.64in]{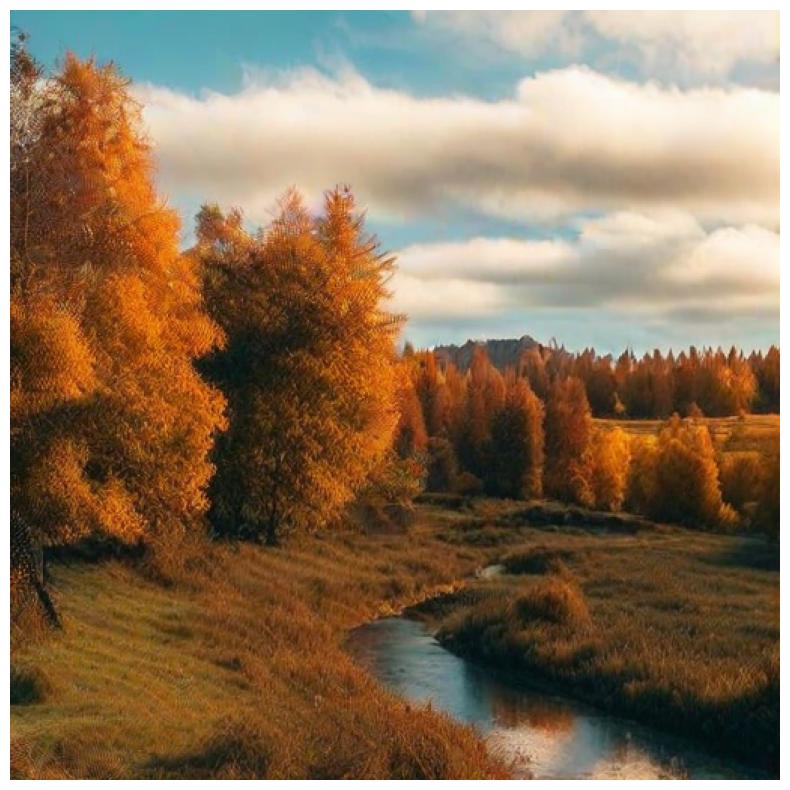}}\\[-1.1em]
    \subfloat{\includegraphics[width=.64in]{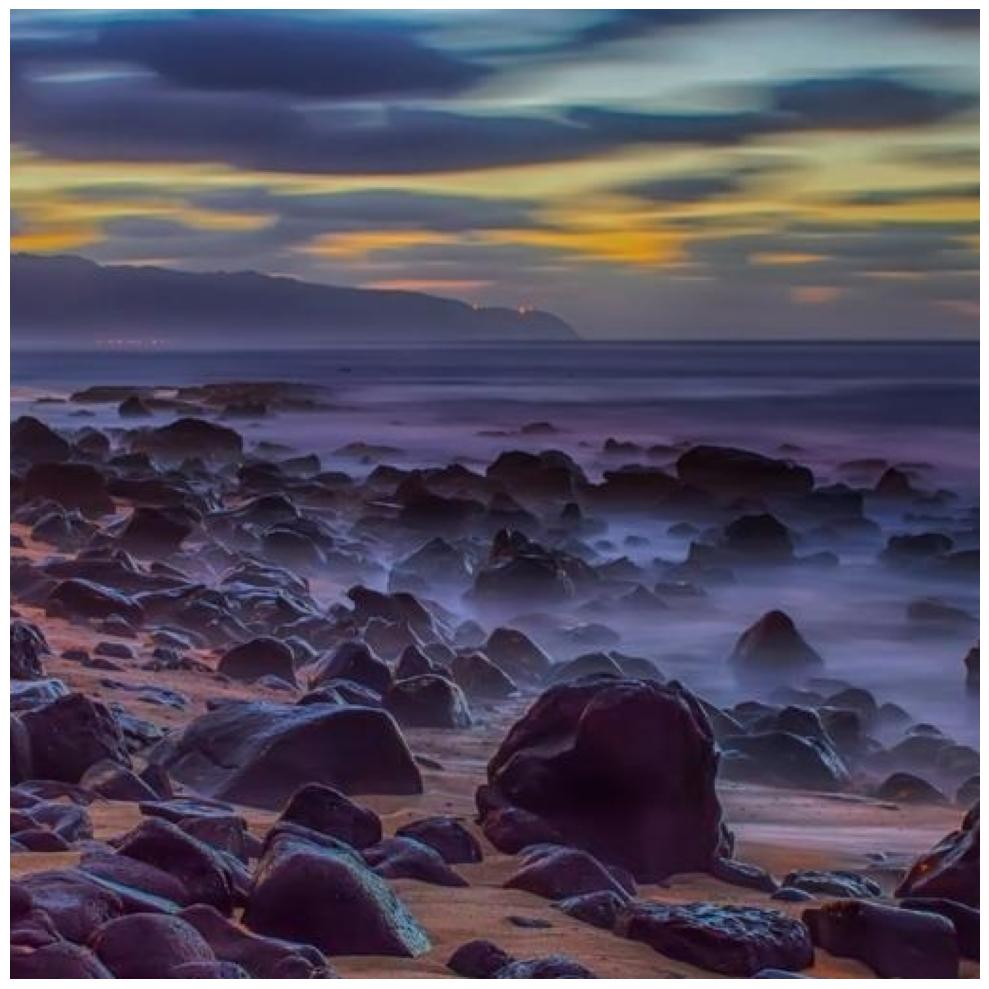}}
    \subfloat{\includegraphics[width=0.64in]{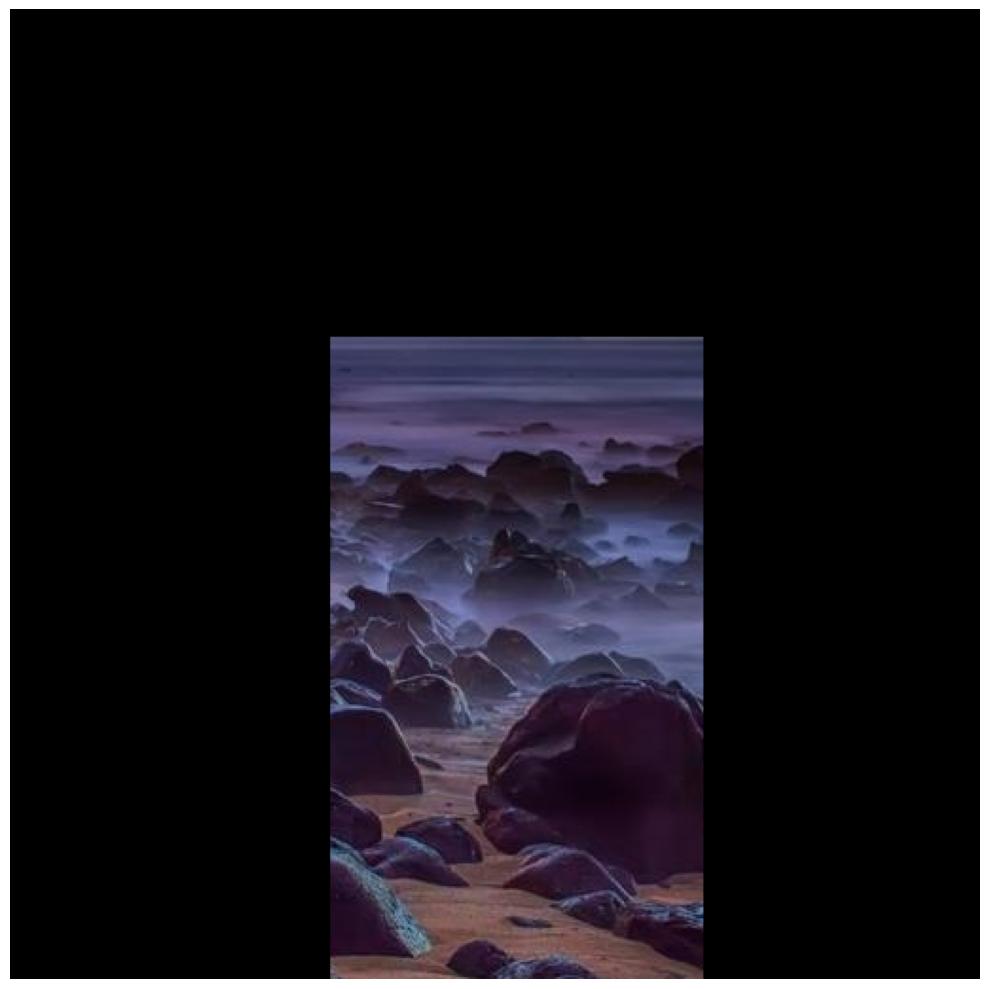}}
    \subfloat{\includegraphics[width=0.64in]{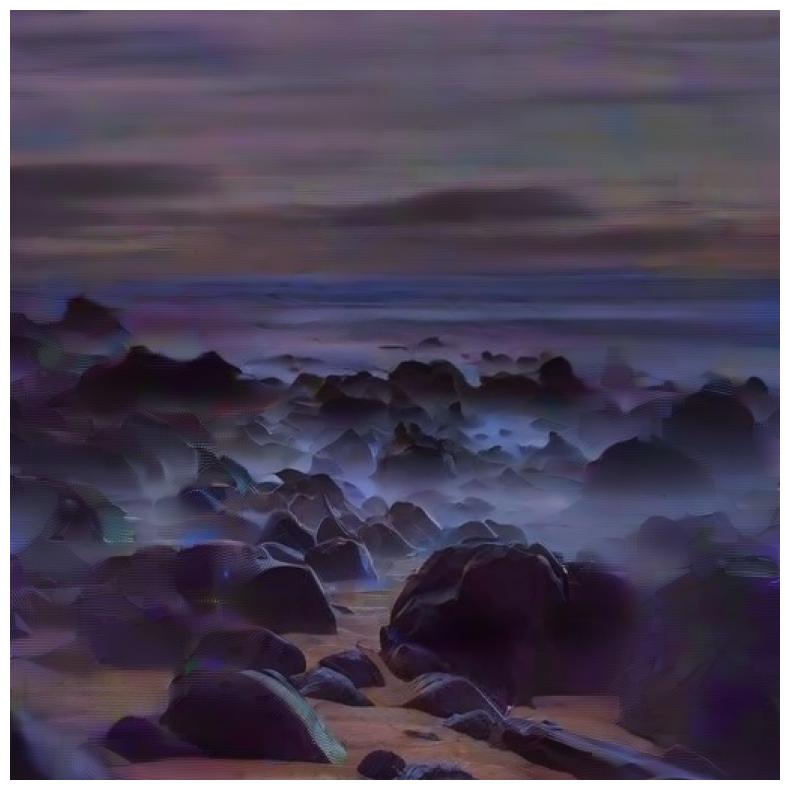}}
    \subfloat{\includegraphics[width=0.64in]{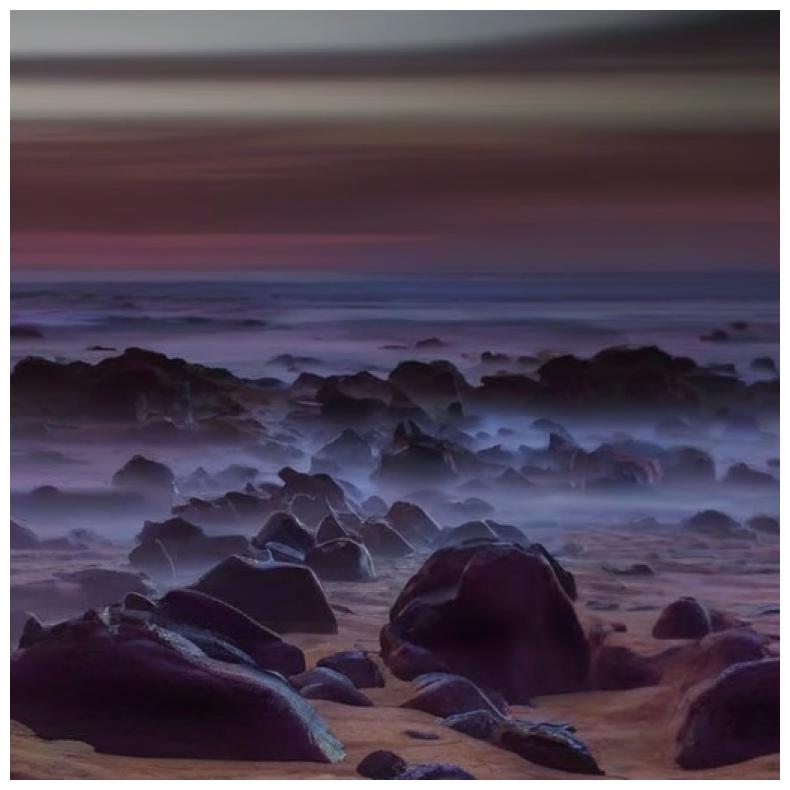}}
    \subfloat{\includegraphics[width=0.64in]{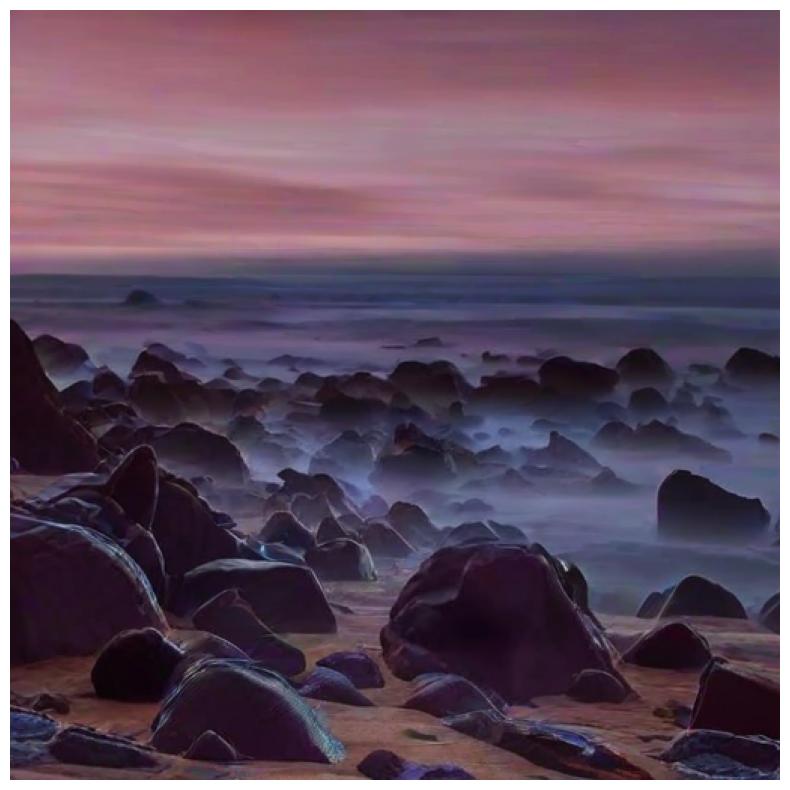}}\\[-1.1em]
    \subfloat{\includegraphics[width=.64in]{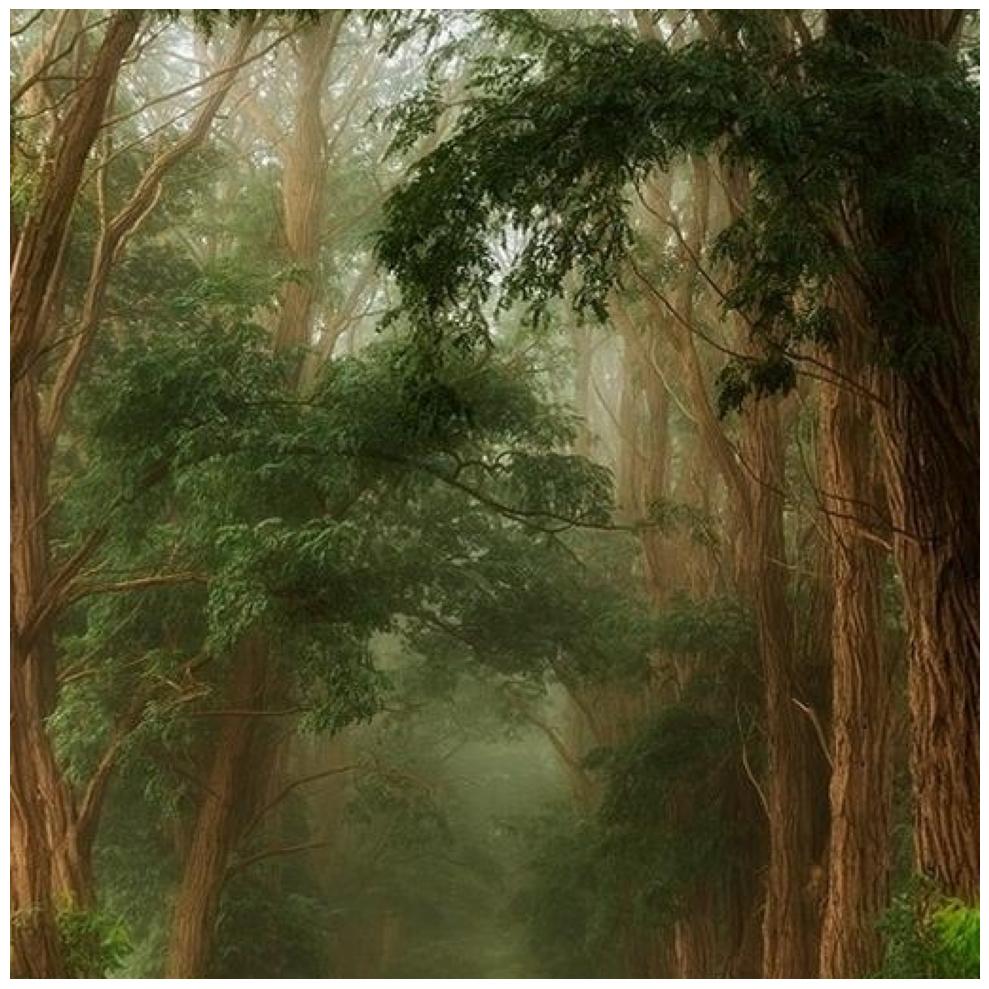}}
    \subfloat{\includegraphics[width=0.64in]{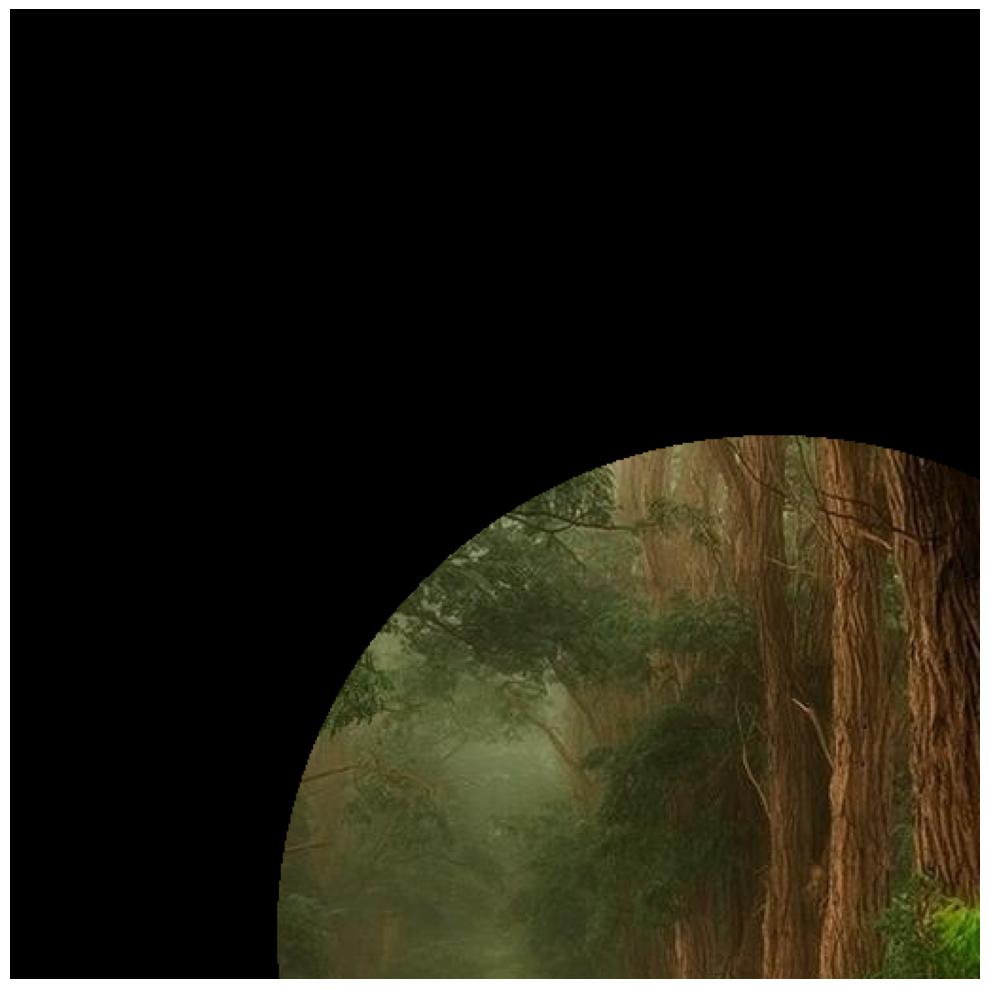}}
    \subfloat{\includegraphics[width=0.64in]{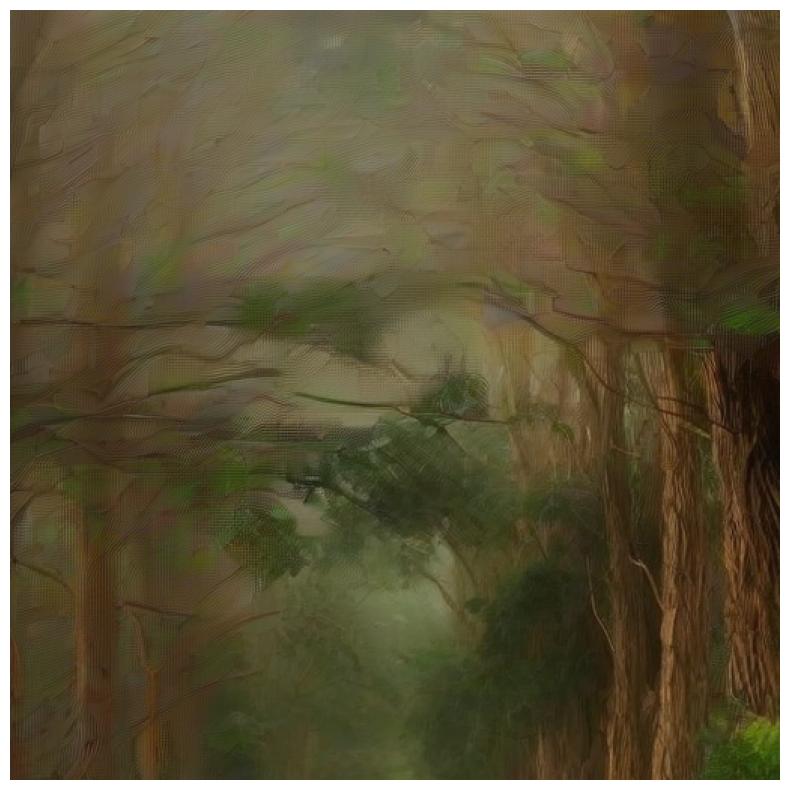}}
    \subfloat{\includegraphics[width=0.64in]{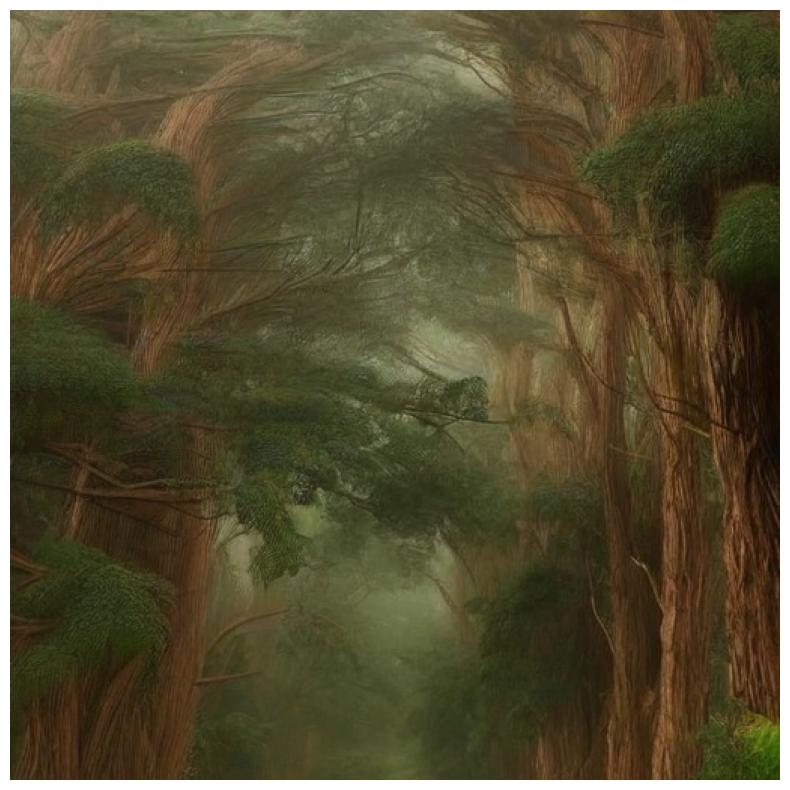}}
    \subfloat{\includegraphics[width=0.64in]{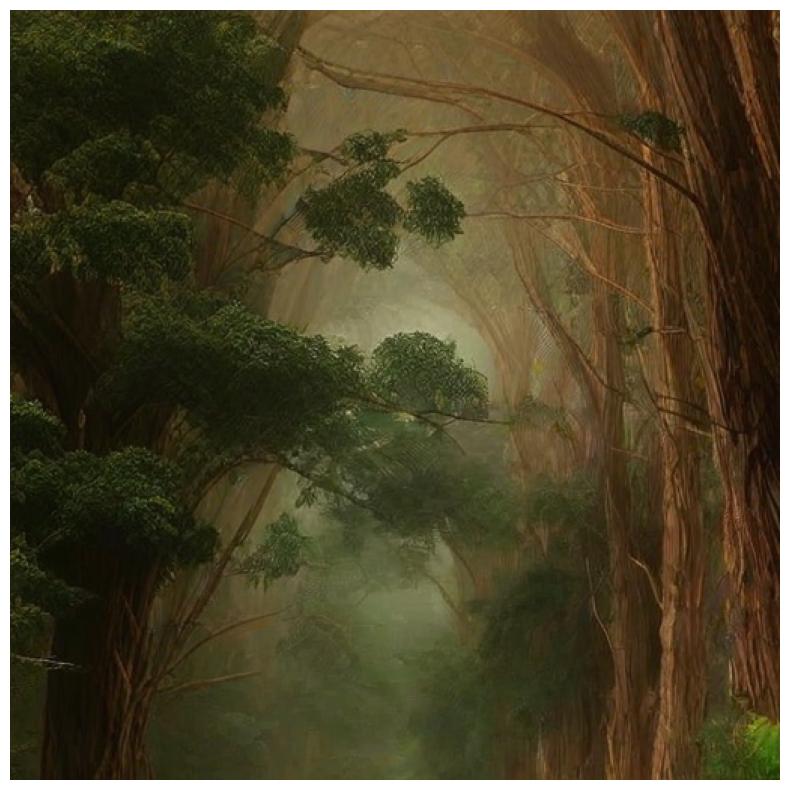}}\\[-0.3em]
        \subfloat[\scriptsize Source image]{\includegraphics[width=.64in]{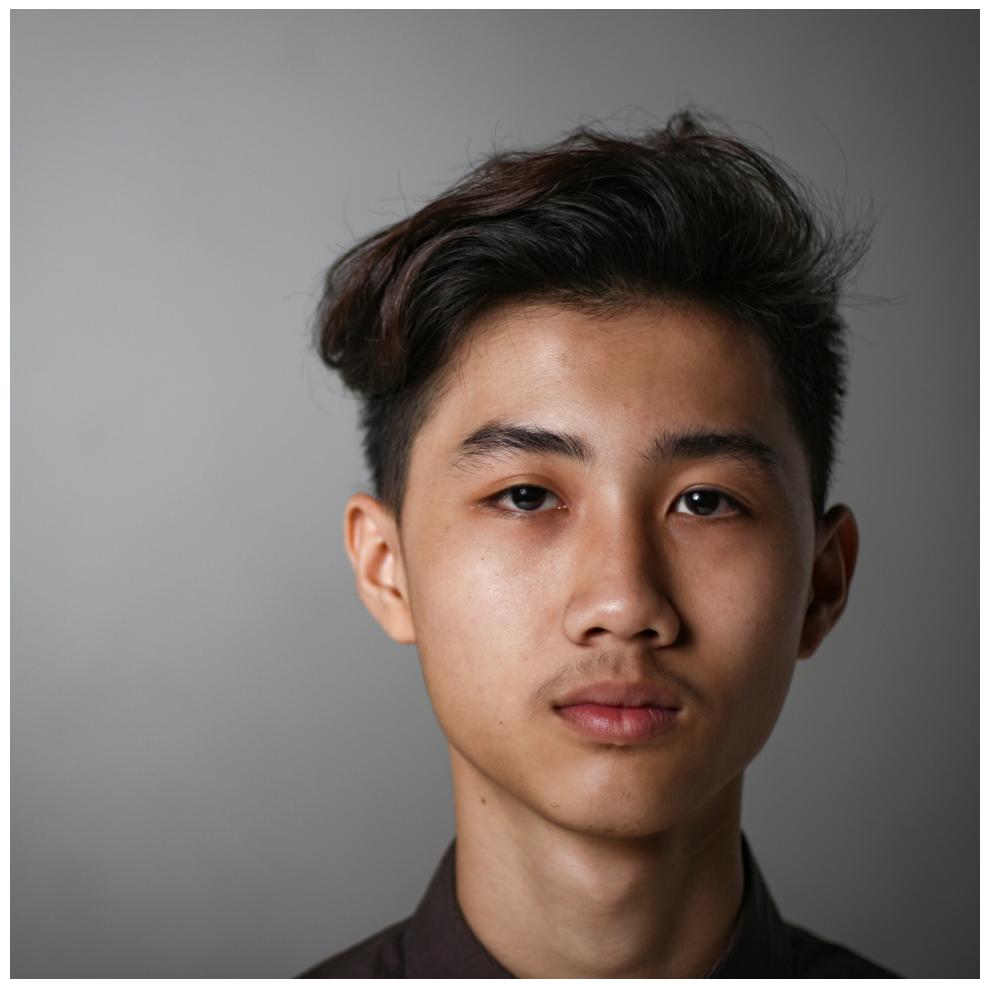}}
      \subfloat[\scriptsize Target image]{\includegraphics[width=0.64in]{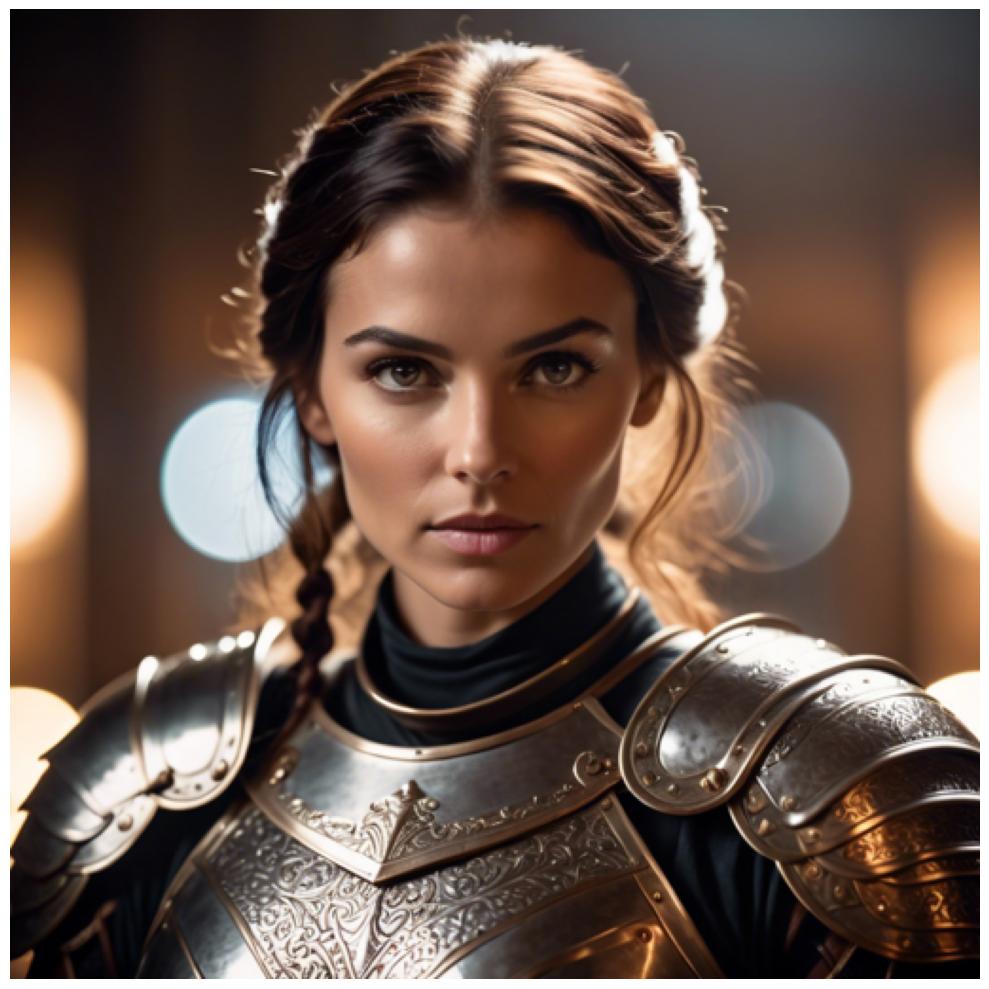}}
       \subfloat[\scriptsize Ref. (8 NFE)]{\includegraphics[width=0.64in]{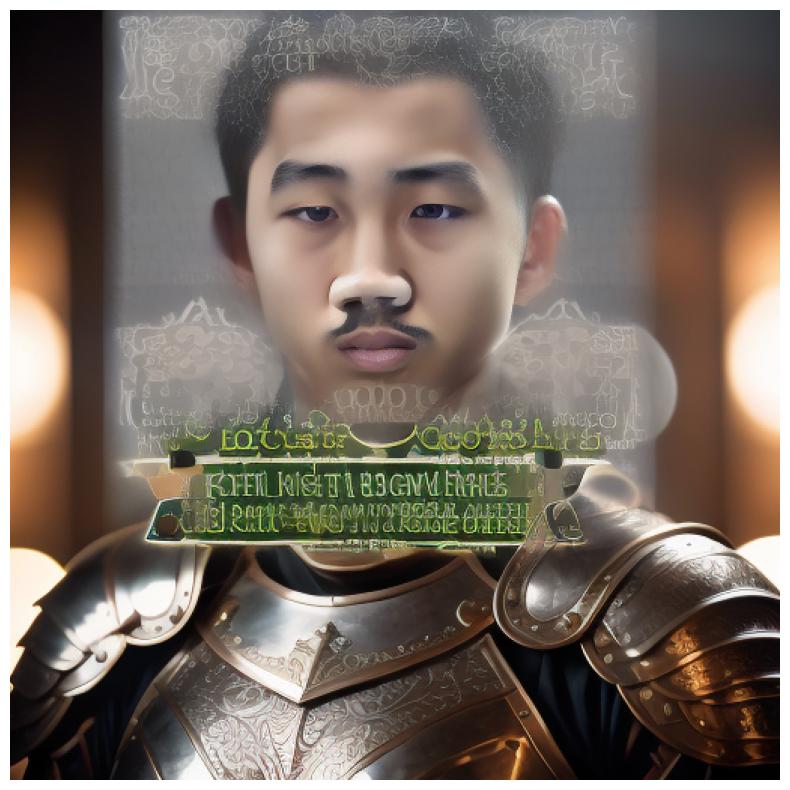}}
      \subfloat[\scriptsize Ref. (40 NFE)]{\includegraphics[width=0.64in]{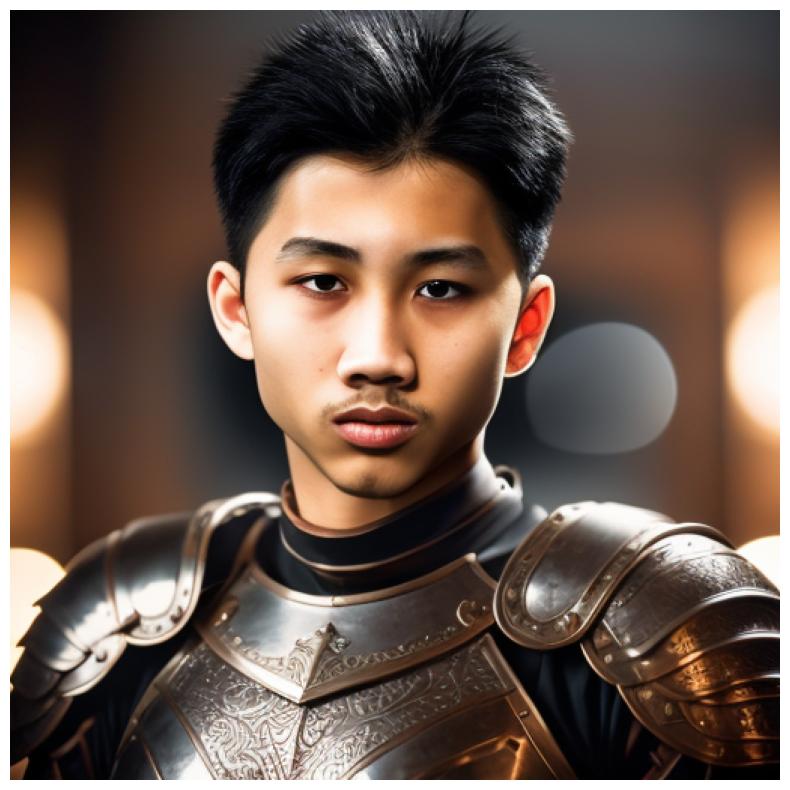}}
      \subfloat[\scriptsize Ours (4 NFEs)]{\includegraphics[width=0.64in]{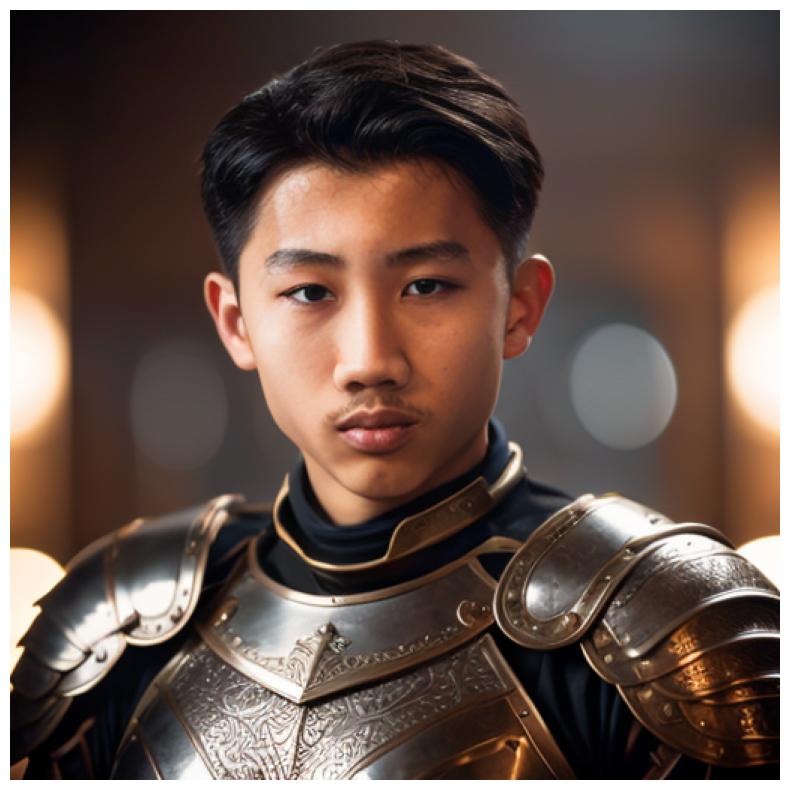}}\\[-1.1em]
      \subfloat{\includegraphics[width=.64in]{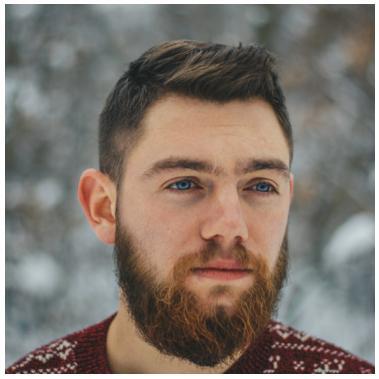}}
      \subfloat{\includegraphics[width=.64in]{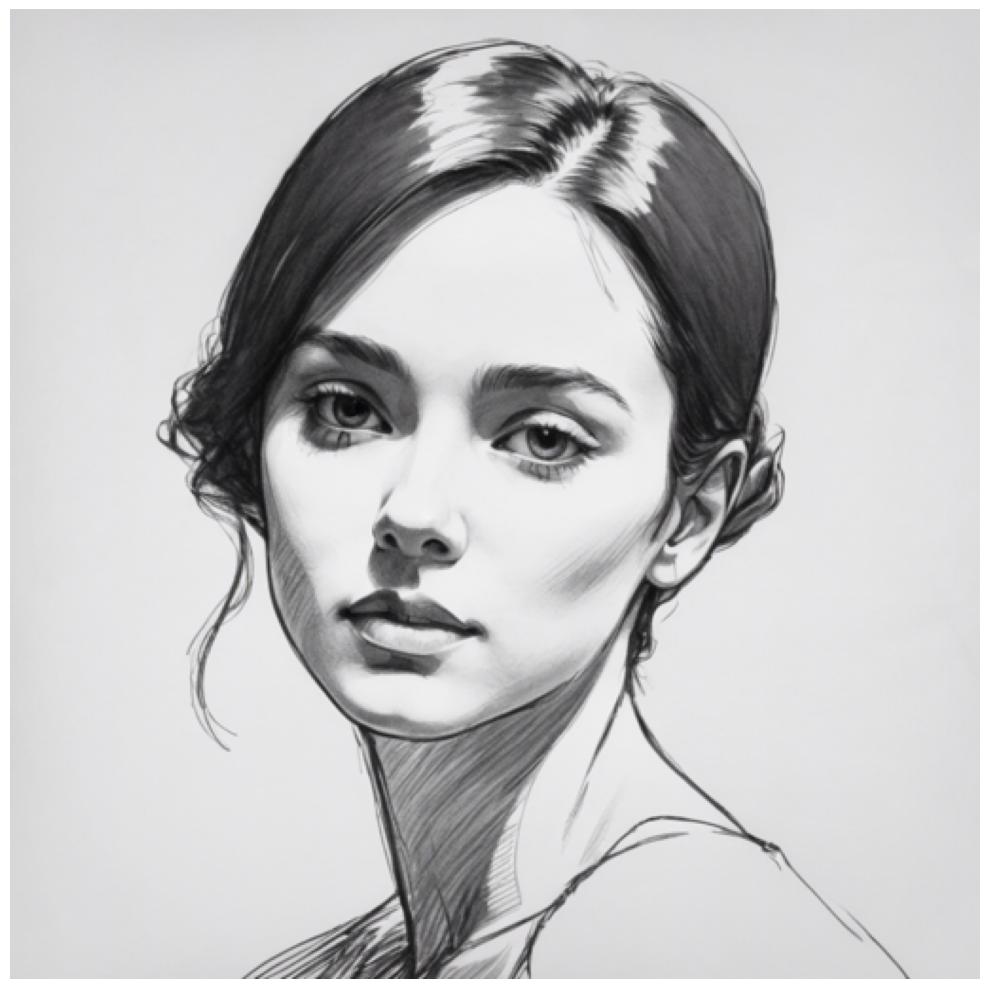}}
      \subfloat{\includegraphics[width=.64in]{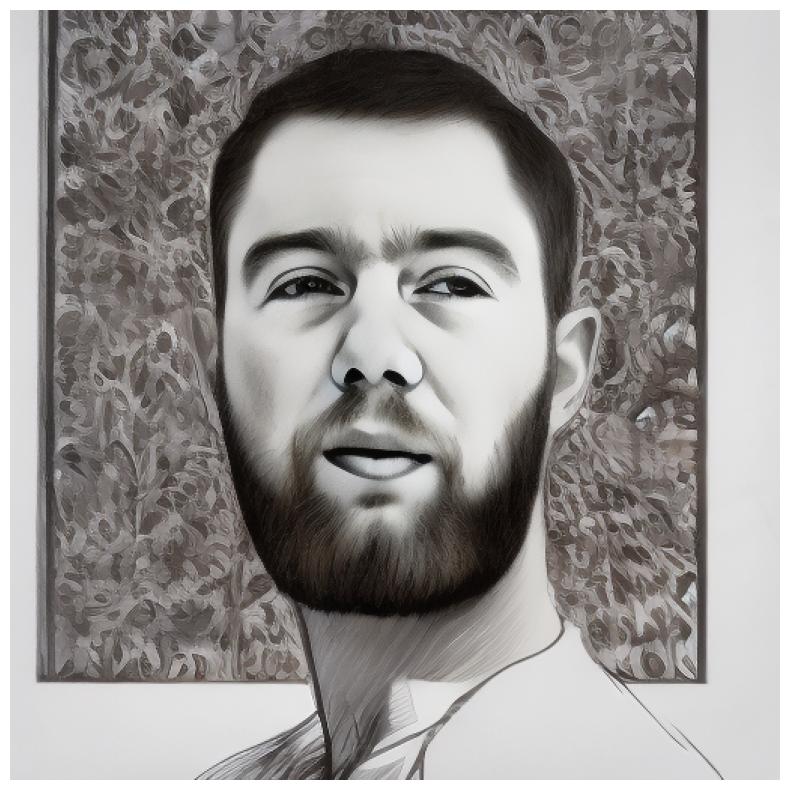}}
      \subfloat{\includegraphics[width=.64in]{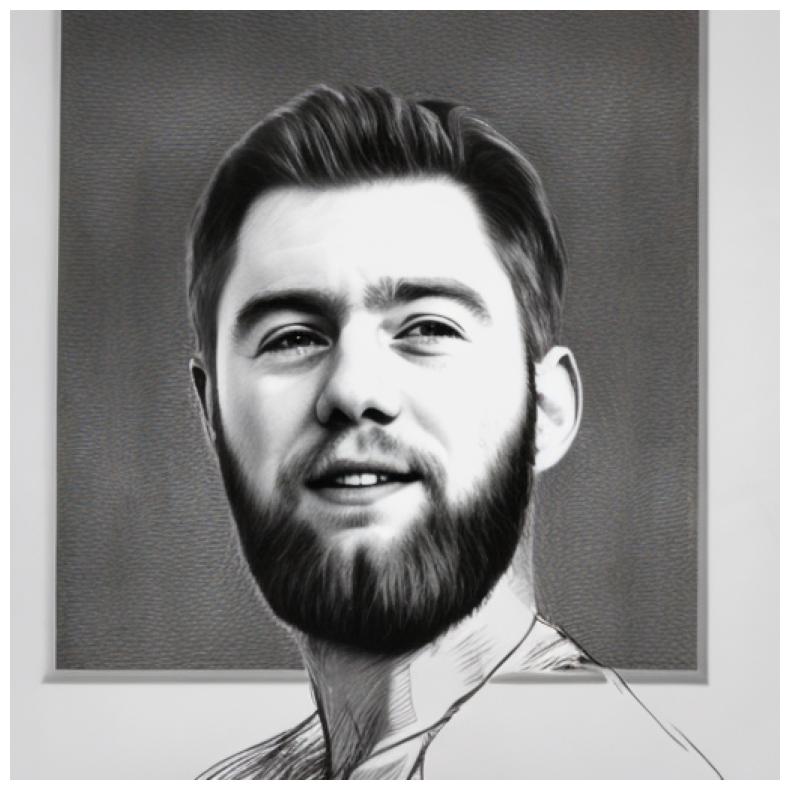}}
      \subfloat{\includegraphics[width=.64in]{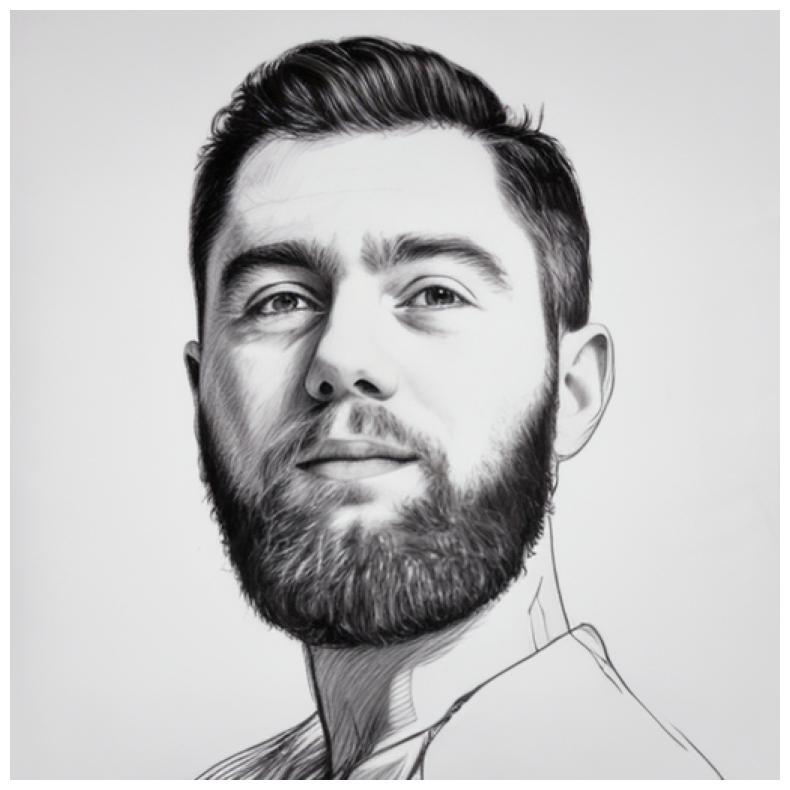}}\\[-1.1em]
      \subfloat{\includegraphics[width=.64in]{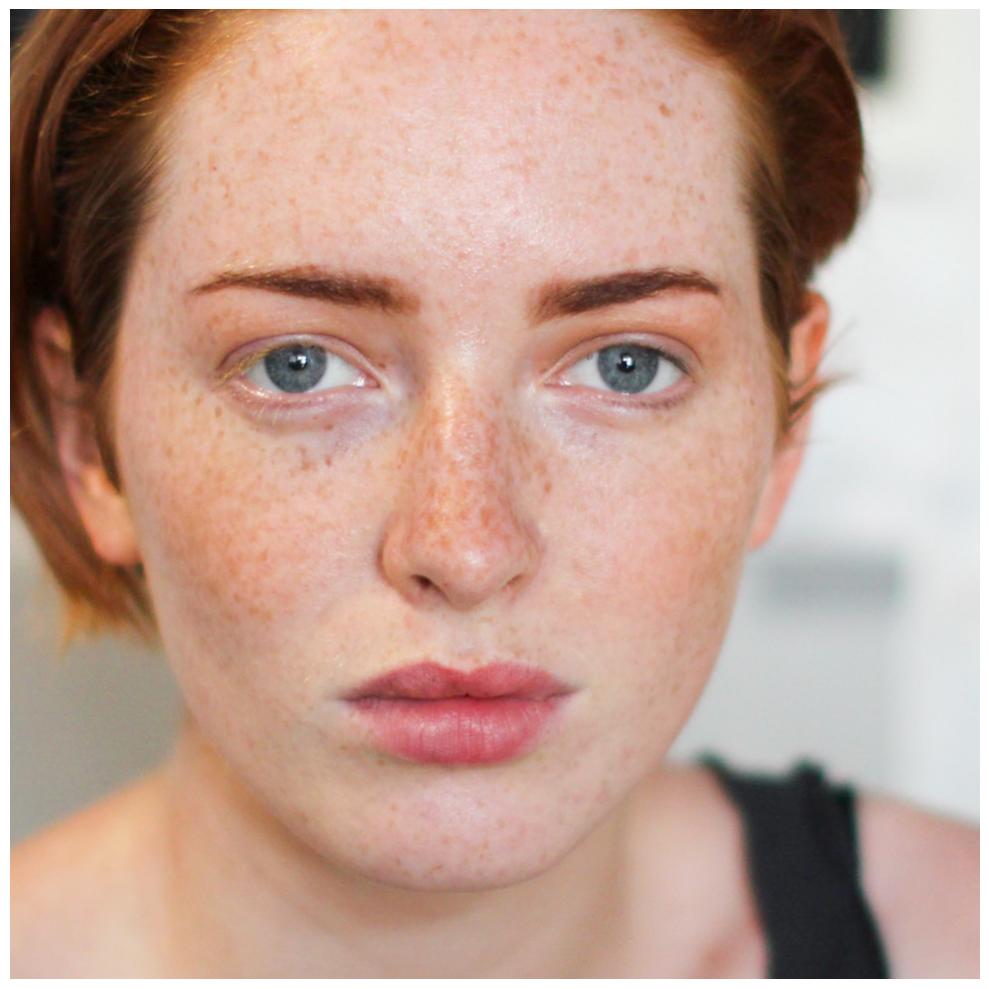}}
      \subfloat{\includegraphics[width=.64in]{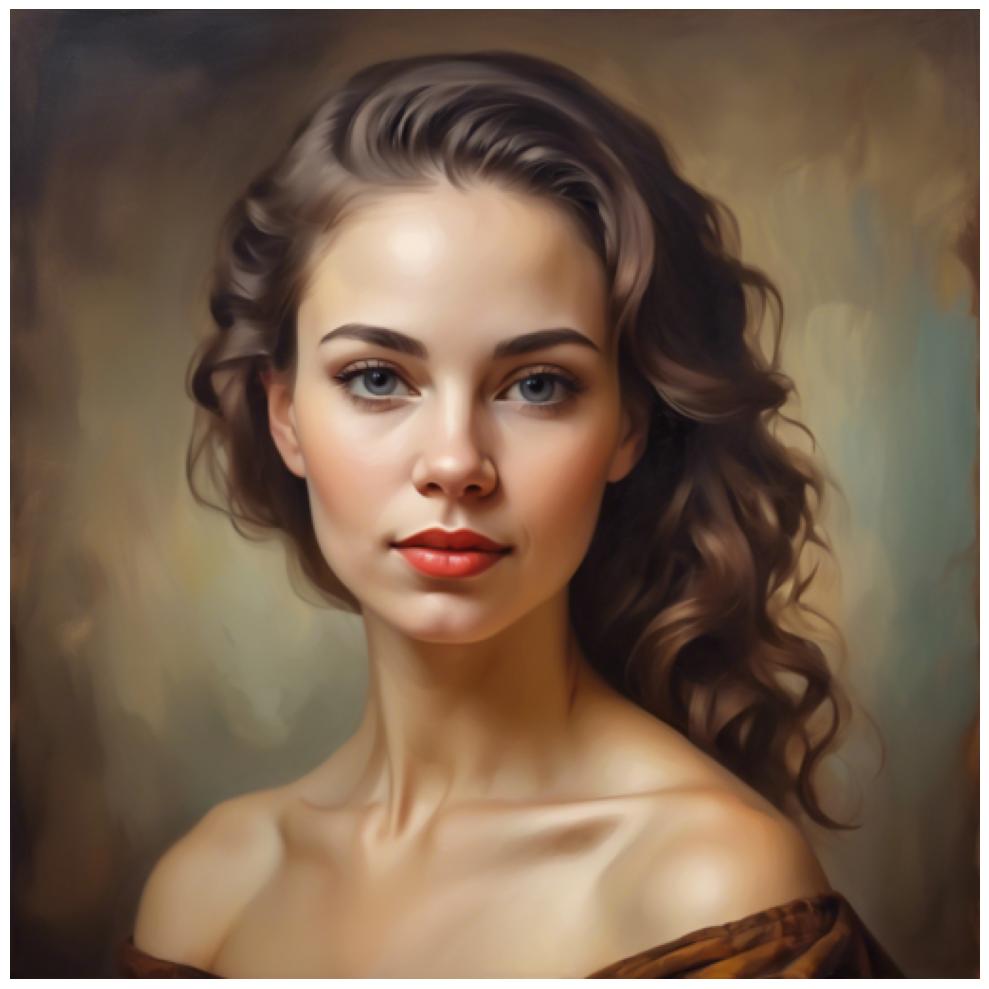}}
      \subfloat{\includegraphics[width=.64in]{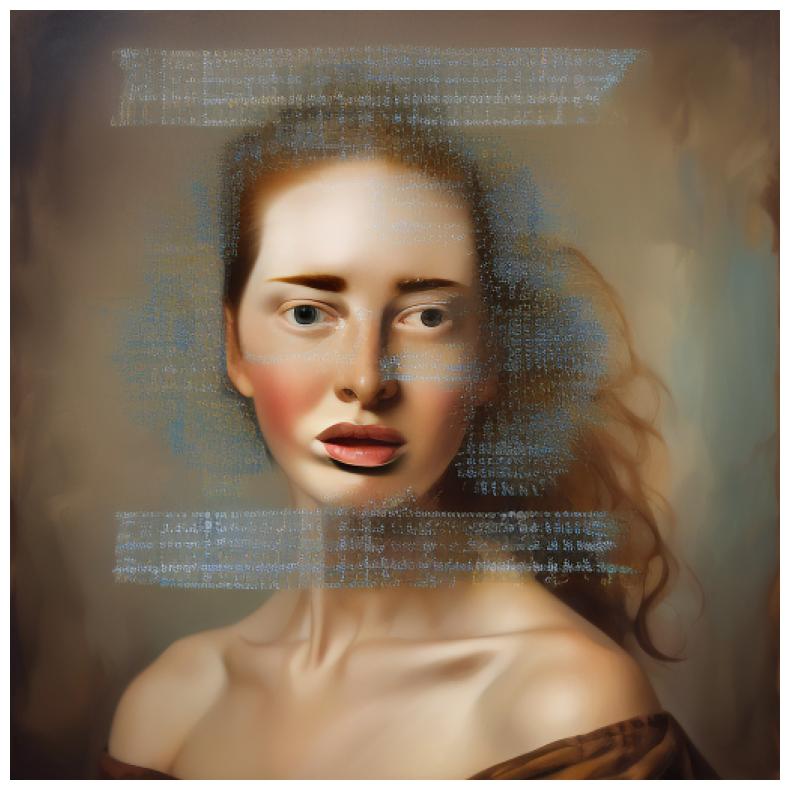}}
      \subfloat{\includegraphics[width=.64in]{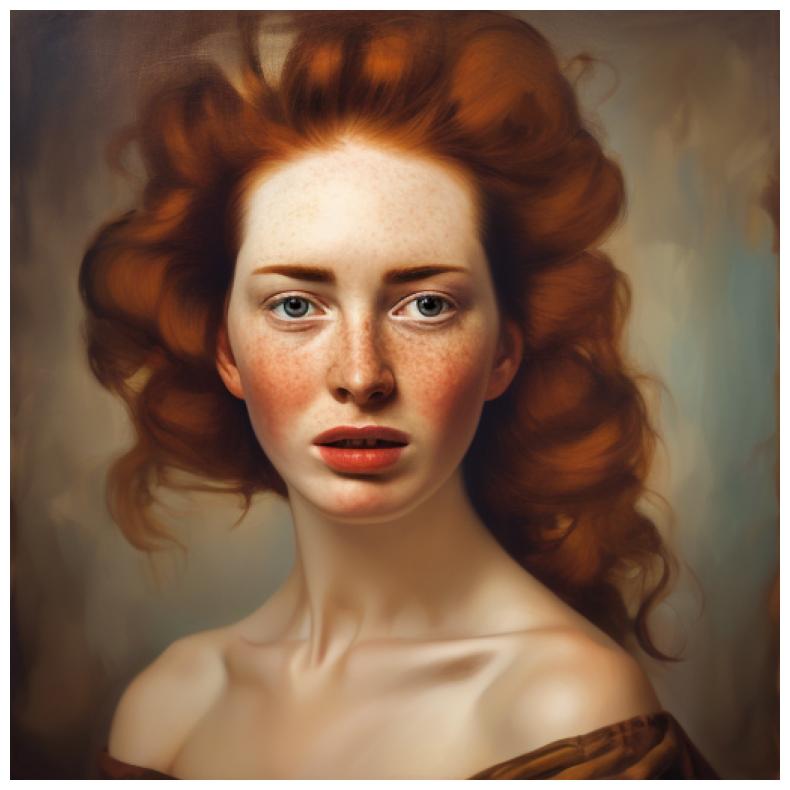}}
      \subfloat{\includegraphics[width=.64in]{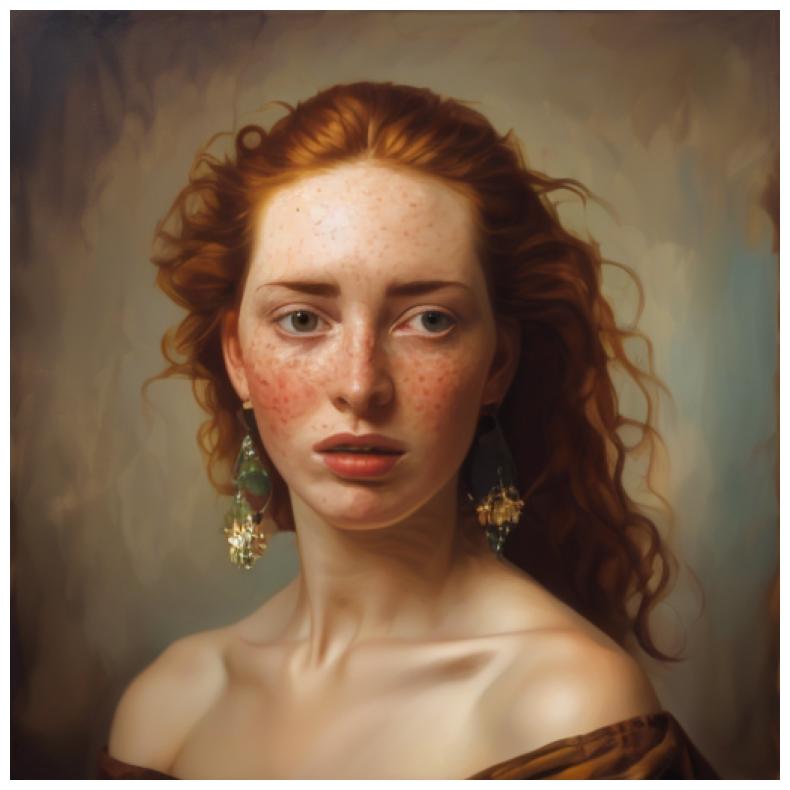}}\\[-0.3em]
 \subfloat[\scriptsize LR image]{\includegraphics[width=0.8in]{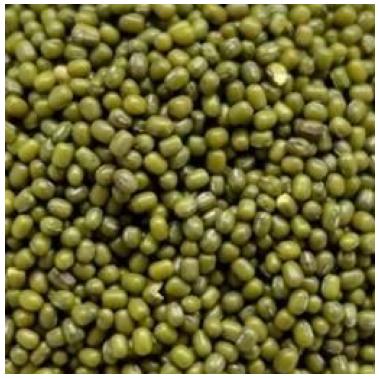}}
    \subfloat[\scriptsize Ref. (8 NFEs)]{\includegraphics[width=.8in]{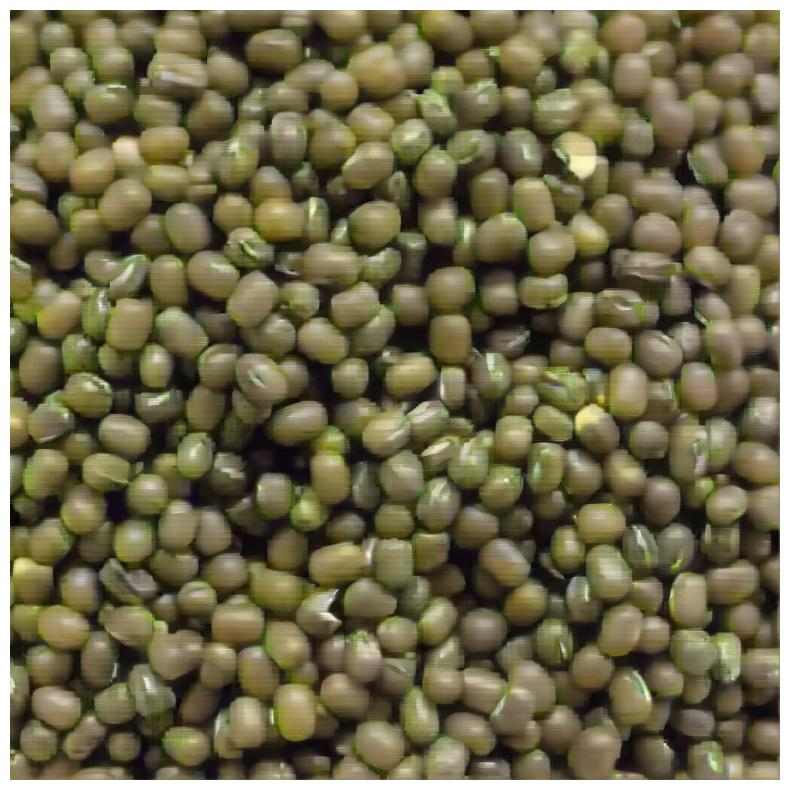}}
   \subfloat[\scriptsize Ref. (40 NFEs)]{\includegraphics[width=.8in]{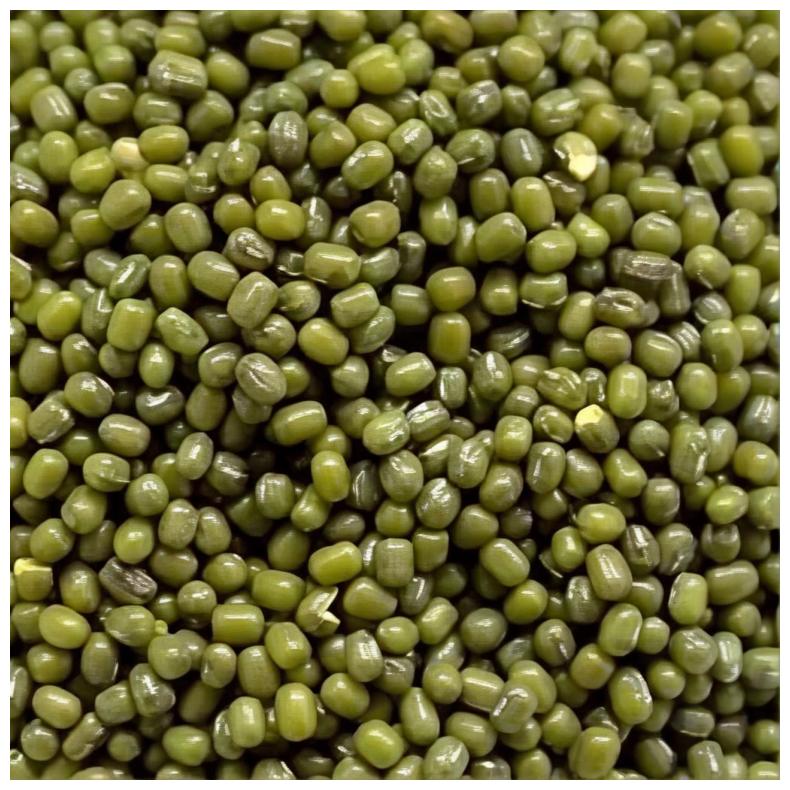}}
   \subfloat[\scriptsize Ours (4 NFEs)]{\includegraphics[width=.8in]
{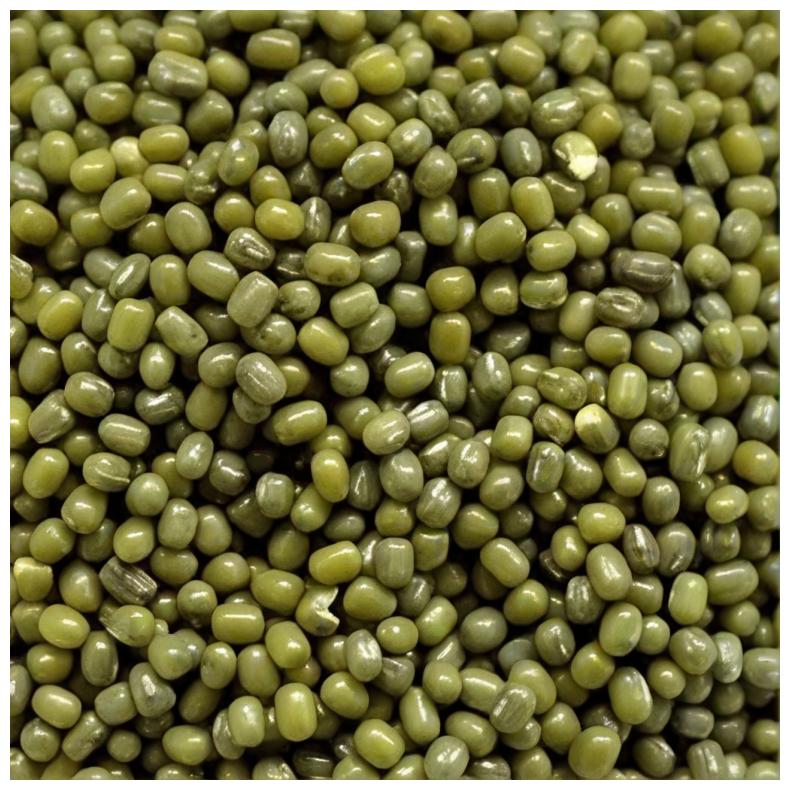}}\hspace{0.6em}\\[-1.1em]
\subfloat{\includegraphics[width=0.8in]{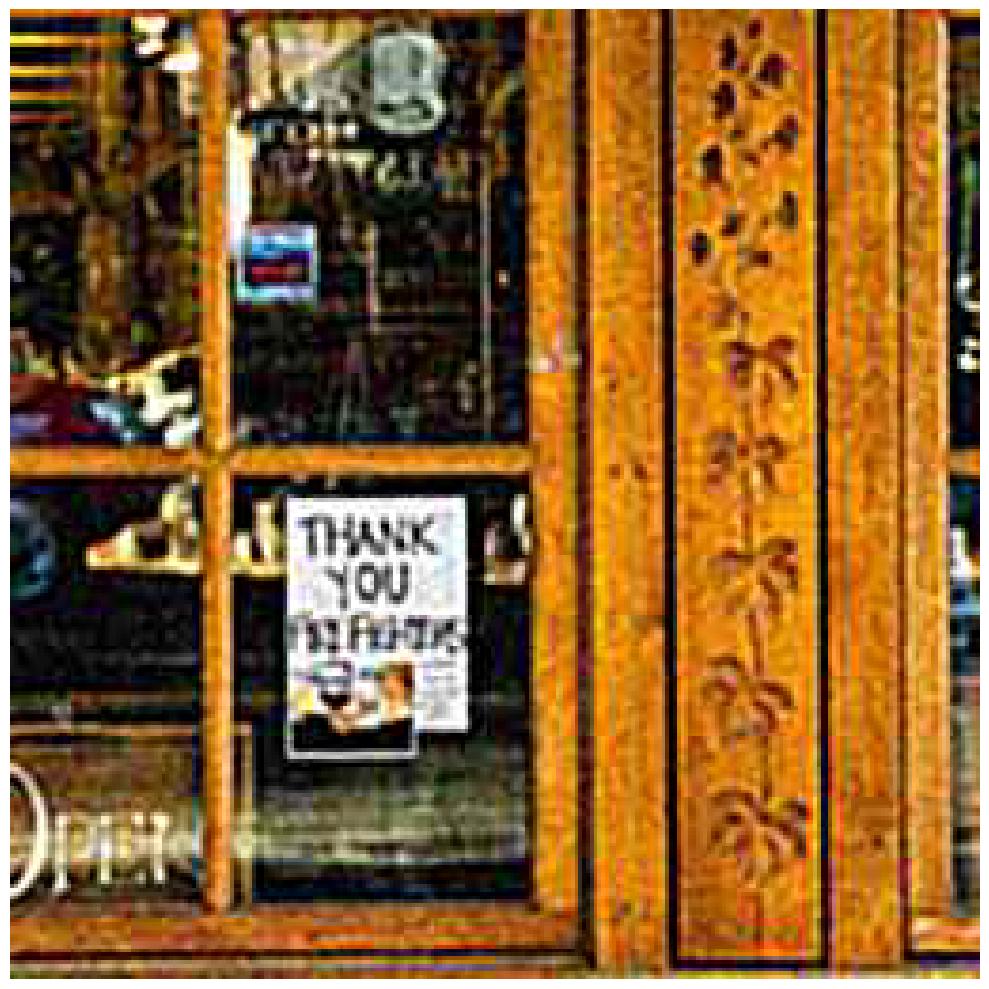}}
\subfloat{\includegraphics[width=.8in]{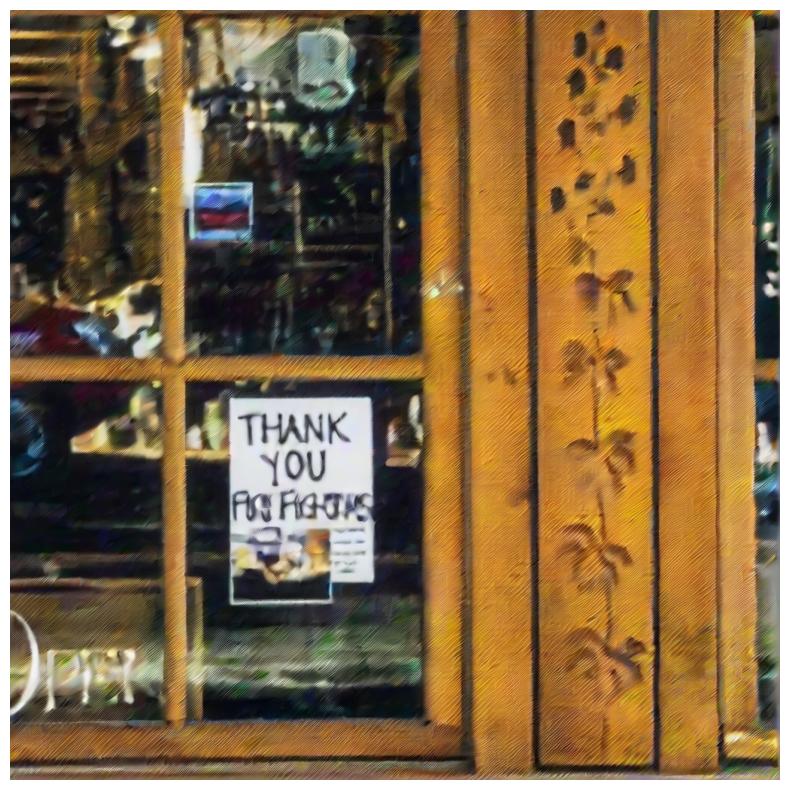}}
\subfloat{\includegraphics[width=.8in]{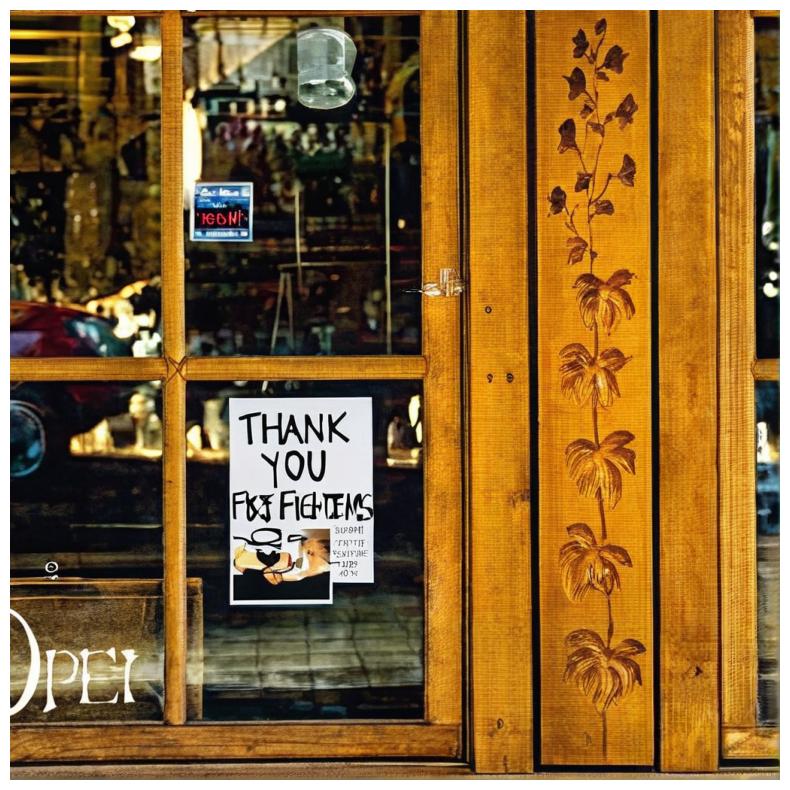}}
\subfloat{\includegraphics[width=.8in]
{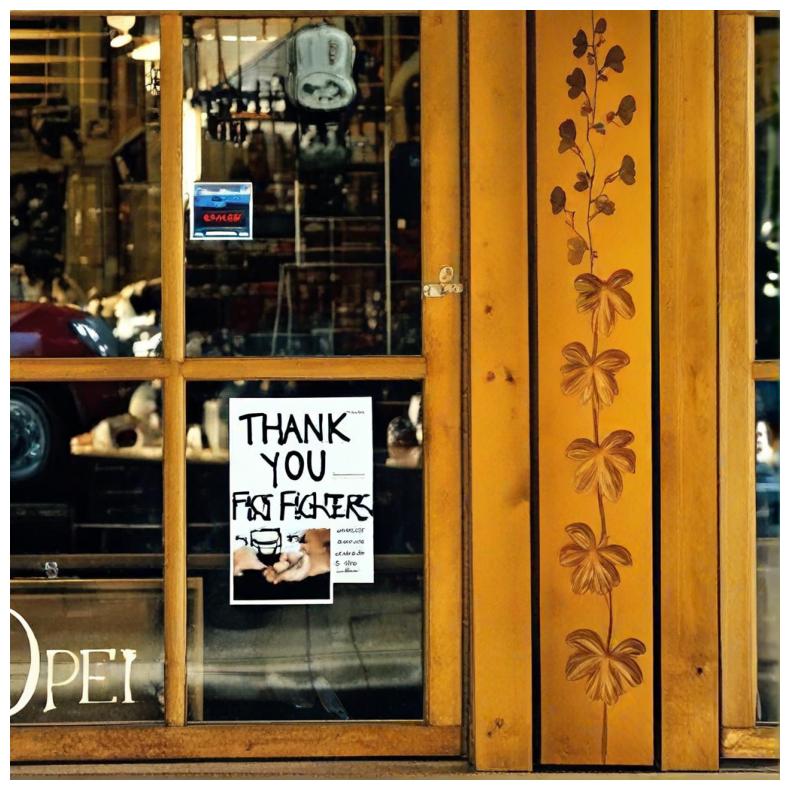}}
\hspace{0.6em}\\[.5em]
    \subfloat[\scriptsize T2I Adapters]{\includegraphics[width=1.in]{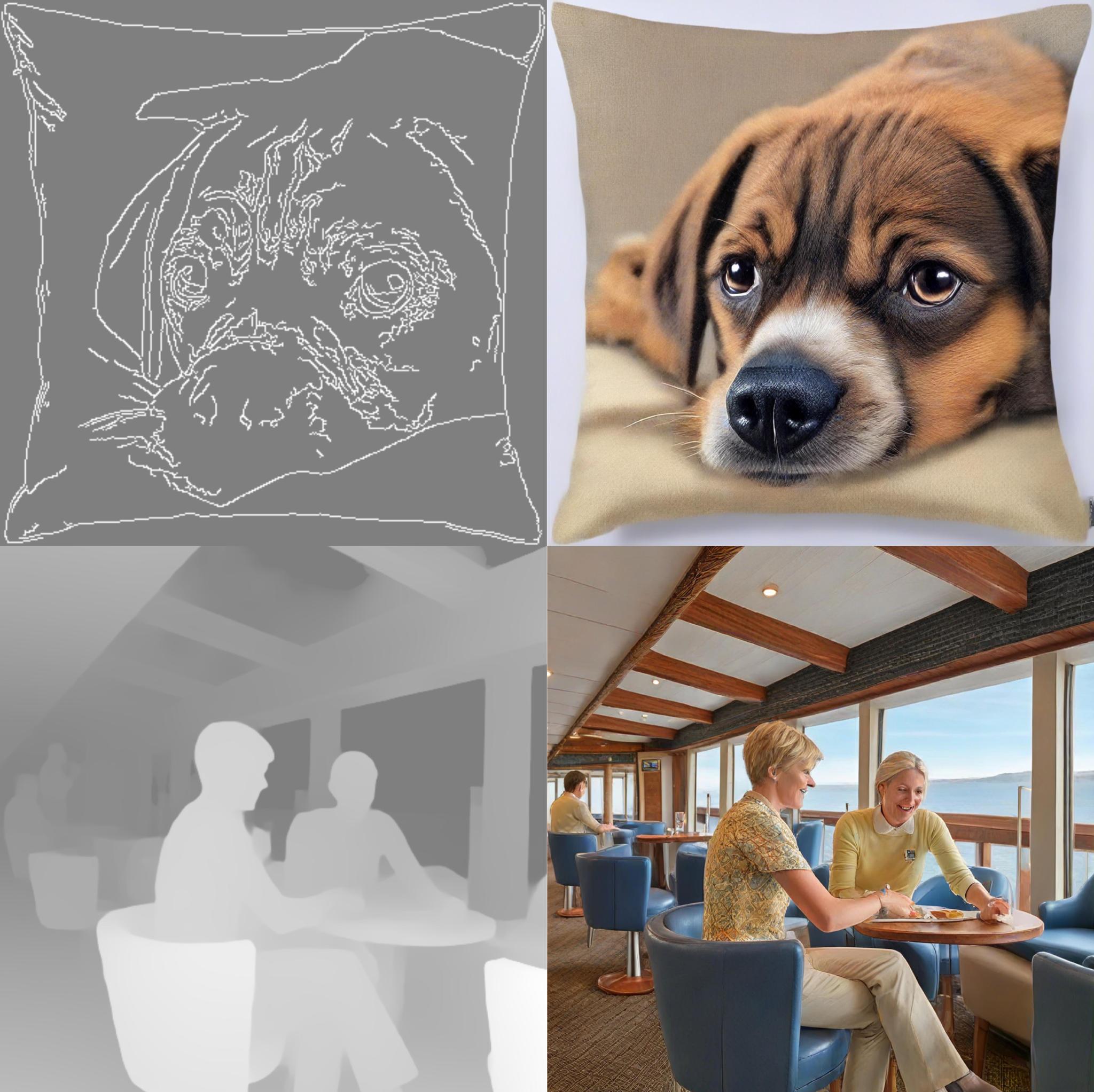}}
        \hspace{0.6em}
        \subfloat[\scriptsize \emph{Training-free} LoRAs compatibility]{\includegraphics[width=1.5in]{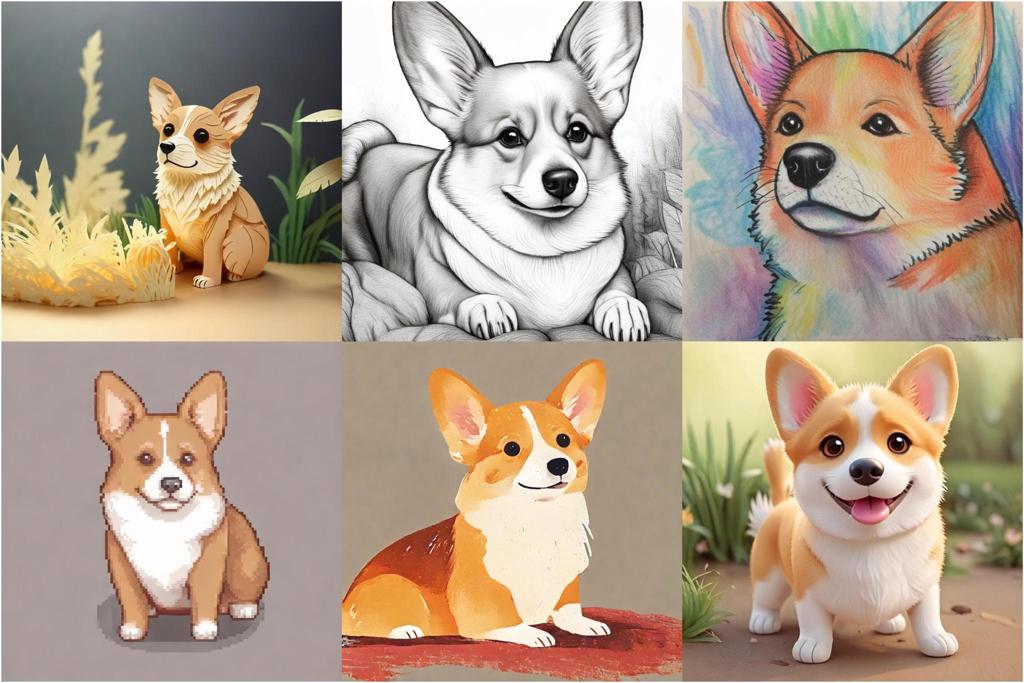}}
    \caption{\emph{From top to bottom:} \emph{Flash Diffusion} applied to 1) an \emph{inpainting} model, 2) a \emph{face-swapping} model and 3) a \emph{super-resolution} model as well as T2I adapters. At the bottom right, we show the 4 steps generations from 6 different LoRAs directly applied on top of Flash SDXL (no training needed).}
    \label{fig:inpainting_upscaler}
    \end{figure}

\paragraph{Inpainting, Super-Resolution and Face-Swapping} In this section, we consider 1) an \emph{in-house} \emph{inpainting} diffusion model conditioned on both a masked image, a mask, and a prompt, 2) a \emph{super-resolution} model trained to upscale input images by a factor of 4 and 3) a \emph{face-swapping} model conditioned on a source image and trained to replace the face of the person in the target image with the one in the source image. We show some samples in Fig.~\ref{fig:inpainting_upscaler} using either our student model using 4 NFEs or the teacher generations using 4 steps (\emph{i.e.} 8 NFEs) and 20 steps. As highlighted in the figure, the proposed method is able to generate samples that are visually close to the teacher generations while using far fewer NFEs demonstrating the ability of the method to adapt to different conditionings and tasks. See the appendices for the comprehensive experimental setup and additional samples.

\paragraph{Adapters}
We show the compatibility of the proposed approach with T2I adapters \citep{mou2024t2i}. In this case, the student model is trained to output samples conditioned on both a prompt and an additional conditioning given either with edges or a depth map. Samples are shown in Fig. \ref{fig:inpainting_upscaler}.

    \section{Conclusion}
    In this paper, we proposed a new versatile, fast, and efficient distillation method for diffusion models. The proposed method relies on the training of a student model to generate samples that are close to the data distribution learned by a teacher model using a combination of a distillation loss, an adversarial loss, and a distribution matching loss. We also proposed to rely on the LoRA method to reduce the number of training parameters and speed up the training process. We evaluated the proposed method on a \emph{text-to-image} task and showed that it can achieve SOTA results on COCO2014 and COCO2017 datasets. We also stressed and illustrated the versatility of the method by applying it to several tasks (\emph{inpainting}, \emph{super-resolution}, \emph{face-swapping}), different denoiser architectures (UNet, DiT, MMDiT), and adapters where the trained student model was able to produce high-quality samples using only a few number of NFEs. Future work would consists in trying to reduce even more the number of NFEs or trying to enhance the quality of the samples by applying \emph{Direct Preference Optimization} \citep{rafailov2024direct,wallace2023diffusion} directly to the student model. 

\clearpage

\bibliography{aaai25}

\clearpage

\section{Extended Background}\label{app:backgroud}

\subsection{Diffusion Models}
Let $x_0 \in \mathcal{X}$ be a set of input data such that $x_0 \sim p(x_0)$ where $p(x_0)$ is an unknown distribution. Diffusion models (DM) are a class of generative models that define a Markovian process $(x_t)_{t \in [0,T]}$ consisting in creating a noisy version $x_t$ of $x_0$ by iteratively injecting Gaussian noise to the data $x_0$. This process is such that as $t$ increases the distribution of the noisy samples $x_t$ eventually becomes equivalent to an isotropic Gaussian distribution. The noise schedule is controlled by two differentiable functions $\alpha(t)$, $\sigma(t)$ for any $t \in [0, T]$ such that the log signal-to-noise ratio $\log [\alpha(t)^2 / \sigma(t)^2]$ is decreasing over time. Given any $t \in [0, T]$, the distribution of the noisy samples given the input $q(x_t|x_0)$ is called the \emph{forward process} and is defined by $q(x_t|x_0) = \mathcal{N} \left(x_t; \alpha(t) \cdot x_0, \sigma(t)^2 \cdot \mathbf{I} \right)$  from which we can sample as follows:
\begin{equation}\label{eq:repametrization}
  x_t = \alpha(t) \cdot x_0 +  \sigma(t)\cdot \varepsilon\hspace{1em}\text{with}\hspace{1em}\varepsilon \sim \normal\,.
\end{equation}
The main idea of diffusion models is to learn to denoise a noisy sample $x_t \sim q(x_t|x_0)$ in order to learn the \emph{reverse process} allowing to ultimately create samples $\tilde{x}_0$ directly from pure noise. In practice, during training a diffusion model consists in learning a parametrized function $x_{\theta}$ conditioned on the timestep $t$ and taking as input the noisy sample $x_t$ such that it predicts a denoised version of the original sample $x_0$. The parameters $\theta$ are then learned via denoising score matching
\citep{vincent2011connection,song2019generative}.
\begin{equation}\label{eq:lossdiffusion_app}
  \mathcal{L} = \mathbb{E}_{x_0 \sim p(x_0), t \sim \pi(t), \varepsilon \sim \normal } \left[ \lambda(t) \left\| x_{\theta}(x_t, t) - x_0 \right\|^2 \right]\,,
\end{equation}
where $\lambda(t)$ is a scaling factor that depends on the timestep $t \in [0, 1]$ and $\pi(t)$ is a distribution over the timesteps. Note that Eq.~\eqref{eq:lossdiffusion_app} is actually equivalent to learning a function $\varepsilon_{\theta}$ estimating the amount of noise $\varepsilon$ added to the original sample using the repametrization $\varepsilon_{\theta}(x_t, t) = \big(x_t - \alpha(t) \cdot x_{\theta}(x_t, t)\big) / \sigma(t)$. \citet{song2020score} showed that $\varepsilon_{\theta}$ can be used to generate new data points from Gaussian noise by solving the following PF-ODE \citep{song2020score,salimans2021progressive,kingma2021variational,lu2022dpm}:
\begin{equation}\label{eq:pfode}
  \mathrm{d}x_t = \left[f(x_t, t) -\frac{1}{2}g^2(t)\nabla \log p_{\theta}(x_t)\right]\mathrm{d}t\,,
\end{equation}
where $f(x_t, t)$ and $g(t)$ are respectively the \emph{drift} and \emph{diffusion} functions of the PF-ODE defined as follows:
\begin{equation*}
\begin{aligned}
  f(x_t, t) &= \diff{\log \alpha(t)}{t}x_t\,,\\ g^2(t) &= \diff{\sigma(t)^2}{t} - 2 \diff{\log \alpha(t)}{t}\sigma^2(t)\,.
  \end{aligned}
\end{equation*}
$\nabla \log  p_{\theta}(x_t) = -\frac{\varepsilon_{\theta}(x_t, t)}{\sigma(t)}$ is called the \emph{score function} of $p_{\theta}(x_t)$.  The PF-ODE can be solved using a neural ODE integrator \citep{chen2018neural} consisting in iteratively applying the learned function $\varepsilon_{\theta}$ according to given update rules such as the Euler \citep{song2020score} or the Heun solver \citep{karras2022elucidating}.

A conditional diffusion model can be trained to generate samples from a conditional distribution $p(x_0|c)$ by learning conditional denoising functions $\varepsilon_{\theta}(x_t, t, c)$ or $x_{\theta}(x_t, t, c)$ \citep{ramesh2021zero,ramesh2022hierarchical,rombach2022high,saharia2022photorealistic,ho2022imagen,esser2024scaling,podell2023sdxl,chen2023pixart,chen2024pixart}. In that particular setting, Classifier-Free Guidance (CFG) \citep{ho2021classifier} has proven to be a very efficient way to better enforce the model to respect the conditioning and so improve the sampling quality. CFG is a technique that consists in dropping the conditioning $c$ with a certain probability during training and replacing the conditional noise estimate $\varepsilon_{\theta}(x_t, t, c)$ with a linear combination at inference time as follows:
\begin{equation}\label{eq:cfg}
  \varepsilon_{\theta}(x_t, t, c) = \omega \cdot \varepsilon_{\theta}(x_t, t, c) + (1 - \omega) \cdot \varepsilon_{\theta}(x_t, t, \varnothing)\,,
\end{equation}
where $\omega > 0$ is called the \emph{guidance scale}.

\subsection{Consistency Models}\label{sec:consistency}
Since our approach is inspired by the idea exposed in consistency models \citep{song2023consistency,luo2023latent}, we recall some elements of those models. Consistency Models (CM) are a new class of generative models designed primarily to learn a consistency function $f_{\theta}$ that maps any  sample $x_t$ lying on a trajectory of the PF-ODE given in Eq.~\eqref{eq:pfode} directly to the original sample $x_0$ while ensuring the \emph{self-consistency} property for any $t \in [\varepsilon, T]$, $\varepsilon>0$ \citep{song2023consistency,luo2023latent,song2023improved}:
\begin{equation}\label{eq:consistency}
  f_{\theta}(x_t, t) = f_{\theta}(x_{t'}, t'),\hspace{1em}\forall (t, t') \in [\varepsilon, T]^2\,.
\end{equation}

In order to ensure the self-consistency property, the authors of \citep{song2023consistency} proposed to parametrized $f_{\theta}$ as follows:
\begin{equation*}
  f_{\theta}(x_t, t) = c_{\mathrm{skip}}(t) \cdot x_t + c_{\mathrm{out}}(t) \cdot F_{\theta}(x_t, t)\,,
\end{equation*}
where $F_{\theta}$ is parametrized using a neural network and $c_{\mathrm{skip}}$ and $c_{\mathrm{out}}$ are differentiable functions \citep{song2023consistency,luo2023latent}. A consistency model can be trained either from scratch (\emph{Consistency Training}) or can be used to distil an existing DM (\emph{Consistency Distillation}) \citep{song2023consistency,luo2023latent}. In both cases, the objective of the model is to learn $f_{\theta}$ such that it matches the output of a target function $f_{\theta^-}$ the weights of which are updated using Exponential Moving Average (EMA), for any given points $(x_t, x_{t'})$ lying on a trajectory of the PF-ODE: 
\begin{equation*}
  \mathcal{L} = \mathbb{E}_{x_0, t \sim \pi(t), \varepsilon \sim \normal } \left[ \left\| f_{\theta}(x_t, t) - f_{\theta^-}(x_{t'}, t') \right\|^2 \right]\,.
\end{equation*}
In other words, given a noisy sample $x_t$ obtained with Eq.~\eqref{eq:repametrization}, the idea is to enforce that $f_{\theta}(x_t, t) = f_{\theta^-}(x_{t'}, t')$ where $x_{t'}$ is obtained using either Eq.~\eqref{eq:repametrization} with the same noise $\varepsilon$ and input $x_0$ for \emph{Consistency Training} \citep{song2023consistency,song2023improved} or using a trained diffusion model $\varepsilon_{\phi}^{\mathrm{teacher}}$ and an ODE solver $\Psi$ for \emph{Consistency Distillation} \citep{song2023consistency,song2023improved}. Once the model is trained, one may theoretically generate a sample $\tilde{x}_0$ in a single step by first drawing a noisy sample $x_T \sim \mathcal{N}\left(0, \mathbf{I} \right)$ and then applying the learned function $f_{\theta}$ to it. In practice, several iterations are required to generate a satisfying sample and so the estimated sample $\tilde{x}_0$ is iteratively re-noised 
 and denoised several times using the learned function $f_{\theta}$.

\section{Training Process}

The training process is detailed in Alg.~\ref{alg:flash_diffusion} and illustrated in Fig.~\ref{alg:flash_diffusion} of the main manuscript. In more detail, we first pick a random sample $x_0 \sim p(x_0)$ belonging to the unknown data distribution. This sample is then encoded with an encoder $\mathcal{E}$ to get the corresponding latent sample $z_0$. A timestep $t$ is drawn according to the timesteps probability mass function $\pi$ detailed in Sec. \ref{sec:timesteps} to create a noisy sample $z_t$ using Eq.~\eqref{eq:repametrization}. The teacher model $\varepsilon_{\phi}^{\mathrm{teacher}}$ and the ODE solver $\Psi$ are then used to solve the PF-ODE and so generate a synthetic sample $\tilde{z}_0^{\mathrm{teacher}}$ belonging to the distribution learned by the teacher model. At the same time, the student model $f_{\theta}^{\mathrm{student}}$ is used to generate a denoised sample $\tilde{z}_0^{\mathrm{student}} = f_{\theta}^{\mathrm{student}}(z_t, t)$ in a single step. The distillation loss is then computed according to Eq.~\eqref{eq:distillation loss}. Then, we re-noise the one-step student prediction $\tilde{z}_0^{\mathrm{student}}$ as well as the input latent sample $z_0$ and compute the adversarial loss as explained in Sec. \ref{sec:adversarial}. Finally, for distribution matching, we take again the one-step student prediction $\tilde{z}_0^{\mathrm{student}}$ and re-noise it using a uniformly sampled timestep $t \sim \mathcal{U}([0,1])$. The new noisy sample is passed through the teacher model to get the teacher score $s^{\mathrm{teacher}}$ function while we use the student model (and not a dedicated diffusion model as in \citep{yin2023one}) to get the student score function $s^{\mathrm{student}}$. The distribution matching loss is then computed as explained in Sec. \ref{sec:method}.

Overall, our proposed method relies on the training of only a few number of parameters. This is achieved through applying LoRA to the student model, utilizing a frozen teacher model for the adversarial approach, and employing the student denoiser directly rather than introducing a new diffusion model to calculate the fake scores for the distribution matching loss. This approach not only drastically cuts down on the number of parameters but also accelerates the training process.
\begin{algorithm*}[t]
\caption{Flash Diffusion}
\label{alg:flash_diffusion}
\begin{algorithmic}[1]
\STATE \textbf{Input:} A trained teacher DM $\varepsilon_{\phi}^{\mathrm{teacher}}$, a trainable student DM $f_{\theta}^{\mathrm{student}}$, an ODE solver $\Psi$, the number of sampling teacher  steps $K$, a timesteps distribution $\pi(t)$, the guidance scale range $[\omega_{\mathrm{min}}, \omega_{\mathrm{max}}]$, $\lambda_{\mathrm{adv}}$, $\lambda_{\mathrm{dmd}}$ the losses weights
\STATE{\bfseries Initialisation: $\theta \leftarrow \phi$}\; \algorithmiccomment{Initialise the student with teacher's weights}
\WHILE{not converged}
    \STATE{$(z, c) \sim \mathcal{Z}\times\mathcal{C}$, $\omega \sim \mathcal{U}\big([\omega_{\mathrm{min}}, \omega_{\mathrm{max}}]\big)$} \;\algorithmiccomment{Draw a sample and guidance scale}
    \STATE{$ t_i \sim \pi(t)$, $\varepsilon \sim \normal$ } \; \algorithmiccomment{Sample a timestep and noise}
    \STATE{$\tilde{z}_{t_i} \leftarrow \alpha(t_i) \cdot z_0 + \sigma(t_i)\cdot \varepsilon$} \;
    \FOR{$j = i-1 \rightarrow 0$}
      \STATE{$\tilde{\varepsilon} =  \omega \cdot \varepsilon_{\phi}^{\mathrm{teacher}}(\tilde{z}_{t_{j+1}}, t_{j+1}, c) + (1 - \omega) \cdot \varepsilon_{\phi}^{\mathrm{teacher}}(\tilde{z}_{t_{j+1}}, t_{j+1}, \varnothing) $} \;\algorithmiccomment{CFG}
        \STATE{$\tilde{z}_{t_j}\leftarrow \Psi(\tilde{\varepsilon}, t_{j+1}, \tilde{z}_{t_{j+1}})$} \;\algorithmiccomment{ODE solver update}
    \ENDFOR
    \STATE{$\tilde{z}_0^{\mathrm{teacher}} \leftarrow \tilde{z}_{t_0}$}\;
    \STATE{$\tilde{z}_0^{\mathrm{student}} \leftarrow f_{\theta}^{\mathrm{student}}(\tilde{z}_{t_i}, t_i)$}\;
    \STATE{$\mathcal{L} \leftarrow \mathcal{L}_{\mathrm{distil}}(\tilde{z}_0^{\mathrm{student}}, \tilde{z}_0^{\mathrm{teacher}})$ + $\lambda_{\mathrm{adv}} \cdot \mathcal{L}_{\mathrm{adv}}(\tilde{z}_0^{\mathrm{student}}, z_0)$ + $\lambda_{\mathrm{dmd}} \cdot \mathcal{L}_{\mathrm{DMD}}(\tilde{z}_0^{\mathrm{student}})$}\;
\ENDWHILE
\end{algorithmic}
\end{algorithm*}

\section{Experimental Details}

\subsection{Experimental Setup for Text-to-Image}\label{sec:app_t2i}

To compute the FID, we rely on the \emph{clean-fid} library \citep{parmar2022aliased} while we use an OpenCLIP-G backbone \citep{ilharco2021} to compute the CLIP scores. The models are trained on the LAION dataset \citep{schuhmann2022laion} where we select samples with aesthetic scores above 6 and re-caption the samples using CogVLM \citep{wang2023cogvlm}.

\paragraph{Flash SD1.5}
In this section, we provide the detailed experimental setup used to perform the quantitative evaluation of the method. For this experiment, we use SD1.5 model as teacher and initialize the student with SD1.5's weights. The student model is trained for 20k iterations on 2 H100-80Gb GPUs (amounting to 26 H100 hours of training) with a batch size of 4 and a learning rate of $10^{-5}$ for both the student and the discriminator. We use the timestep distribution $\pi(t)$ detailed in the main paper with $K=32$ and shift modes every 5000 iterations. We also start with both $\lambda_{\mathrm{adv}} = 0$ and $\lambda_{\mathrm{DMD}} = 0$ and progressively increase each time we change the timestep distribution so they reach final values set to $0.3$ and $0.7$ respectively. The schedule is $[0, 0.1, 0.2, 0.3]$ for $\lambda_{\mathrm{adv}}$ and $[0, 0.3, 0.5, 0.7]$ for $\lambda_{\mathrm{DMD}}$. The guidance scale $\omega$ used to denoise using the teacher model is uniformly sampled from $[3, 13]$. The distillation loss is set to the MSE loss and the GAN loss is set to the LSGAN loss.

\begin{figure}[ht]
  \centering
  \captionsetup[subfigure]{position=below, labelformat = empty}
  \subfloat[\scriptsize \centering (a) $\pi^{\mathrm{uniform}}$]{\includegraphics[width=0.25\linewidth]{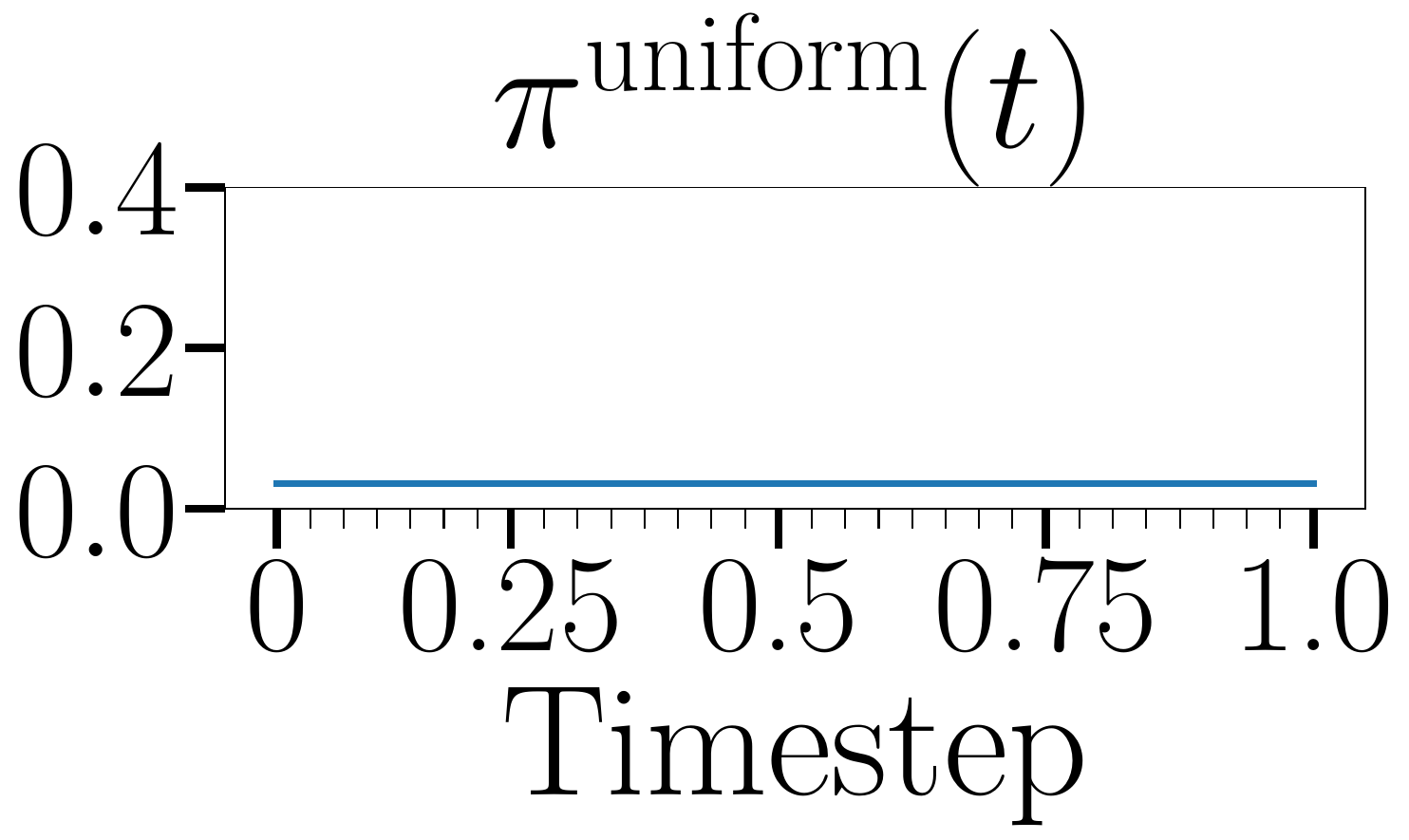}}
  \hspace{2em}
  \subfloat[\scriptsize \centering (b) $\pi^{\mathrm{gaussian}}$]{\includegraphics[width=0.25\linewidth]{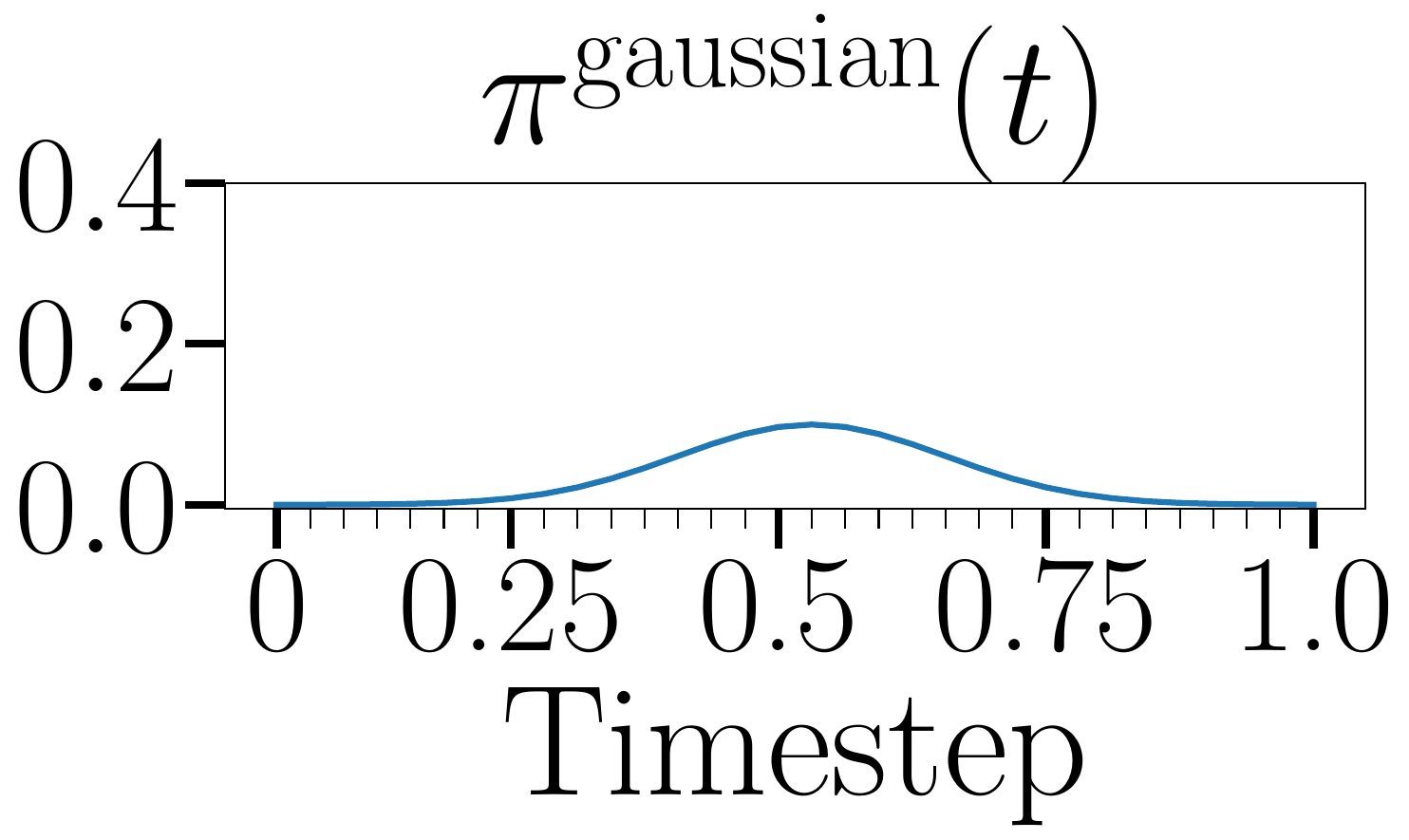}}\\
  \subfloat[\scriptsize \centering (c) $\pi^{\mathrm{sharp}}$]{
  \subfloat[\scriptsize \emph{Warm-up}]{\includegraphics[width=0.25\linewidth]{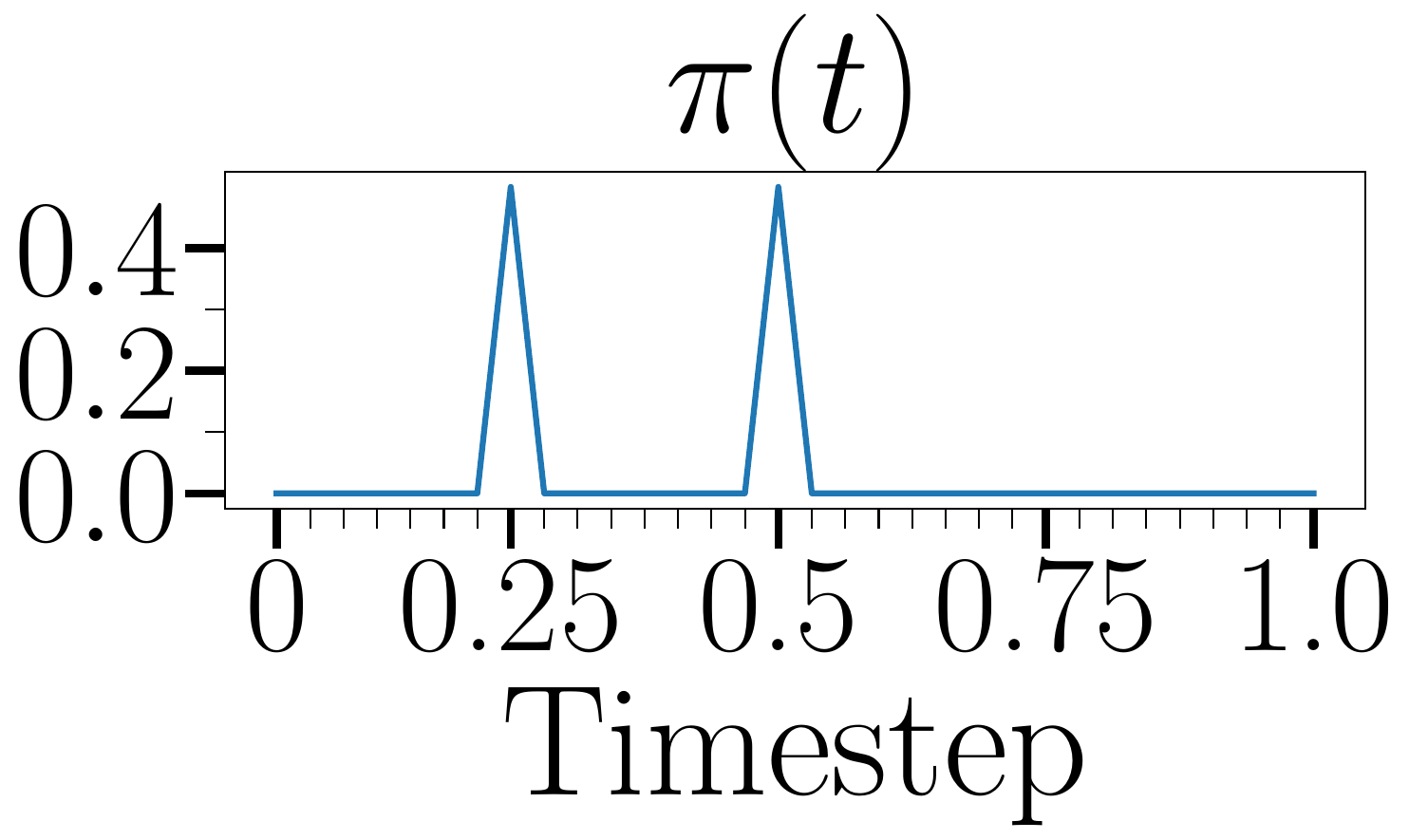}}
  \subfloat[\scriptsize Phase 1]{\includegraphics[width=0.25\linewidth]{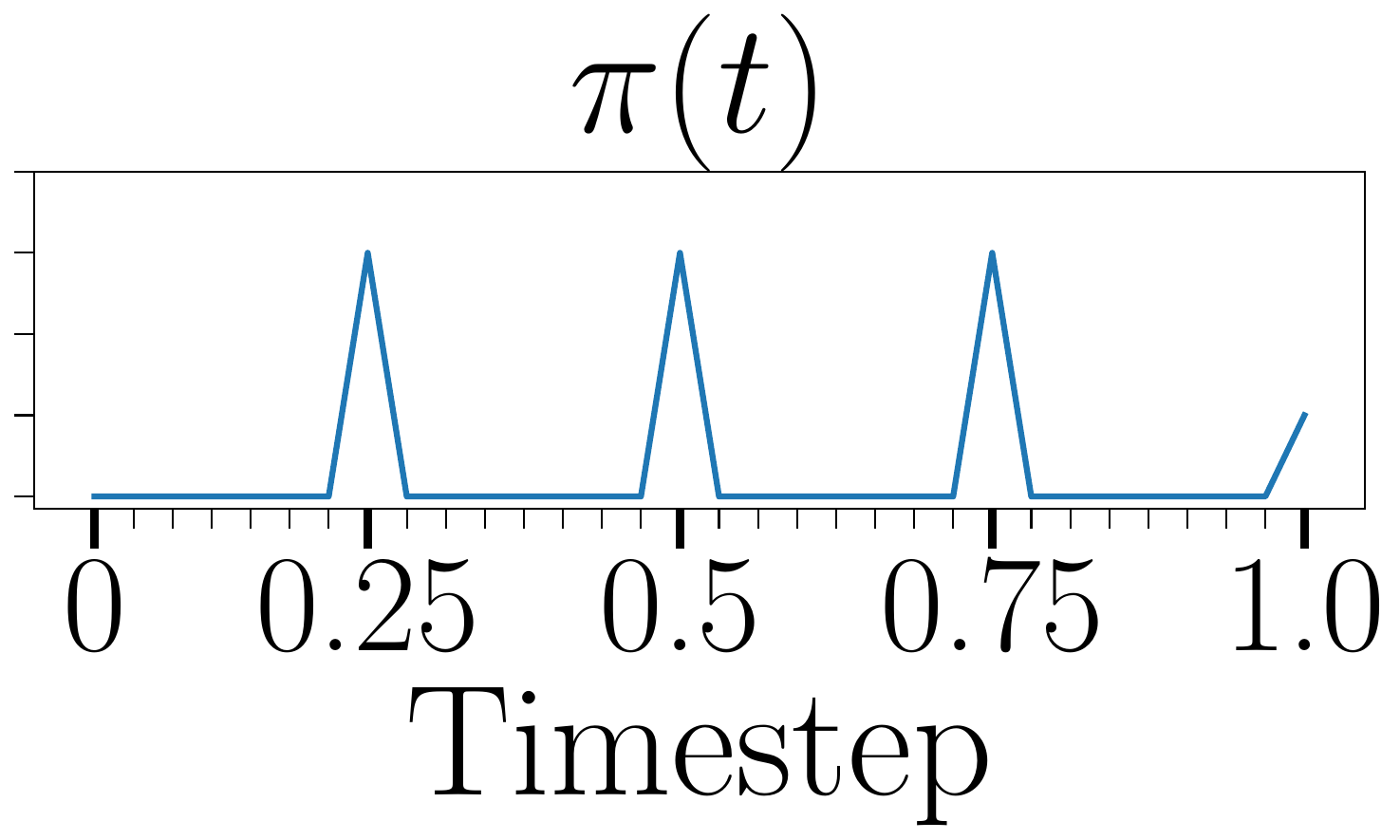}}
  \subfloat[\scriptsize Phase 2]{\includegraphics[width=0.25\linewidth]{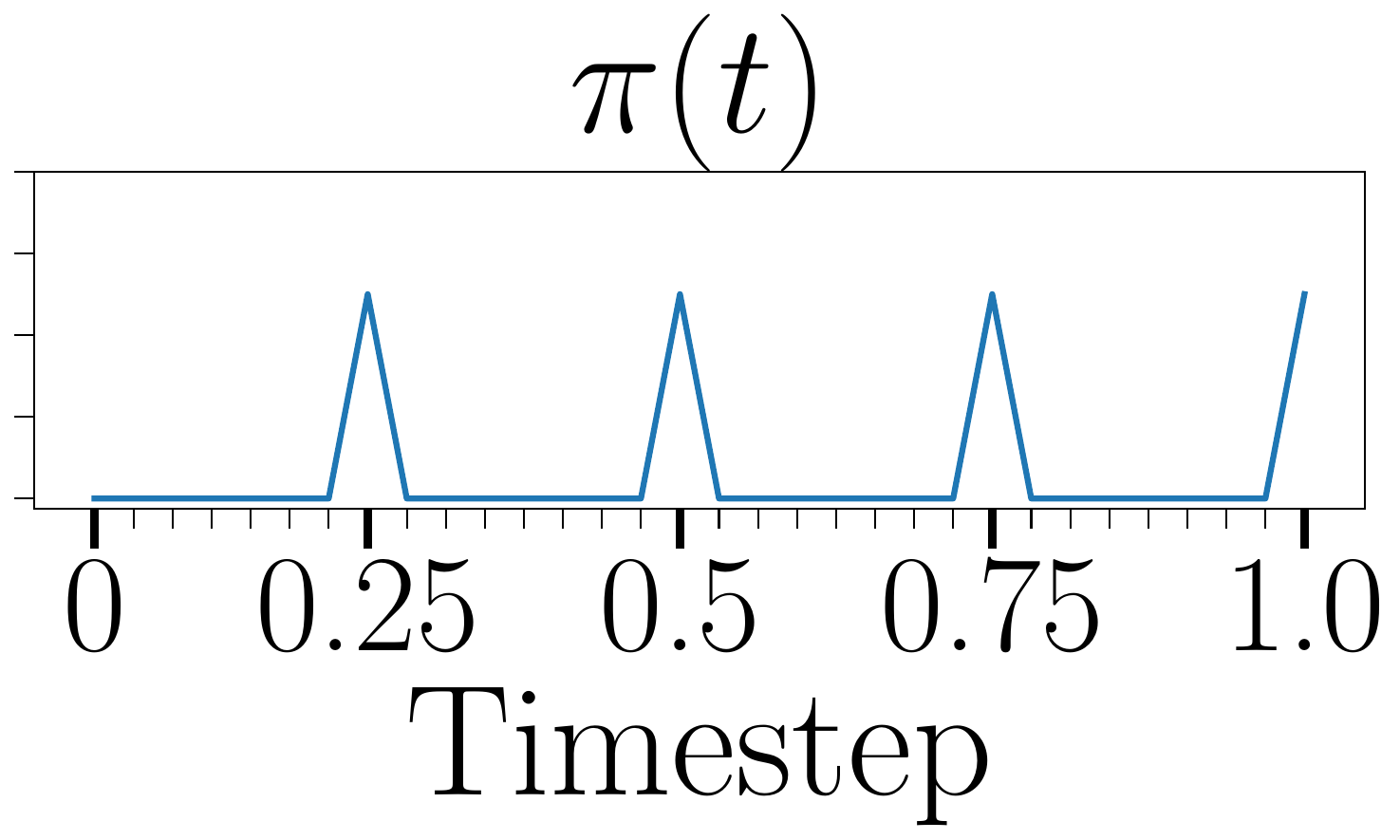}}
  \subfloat[ \scriptsize Phase 3]{\includegraphics[width=0.25\linewidth]{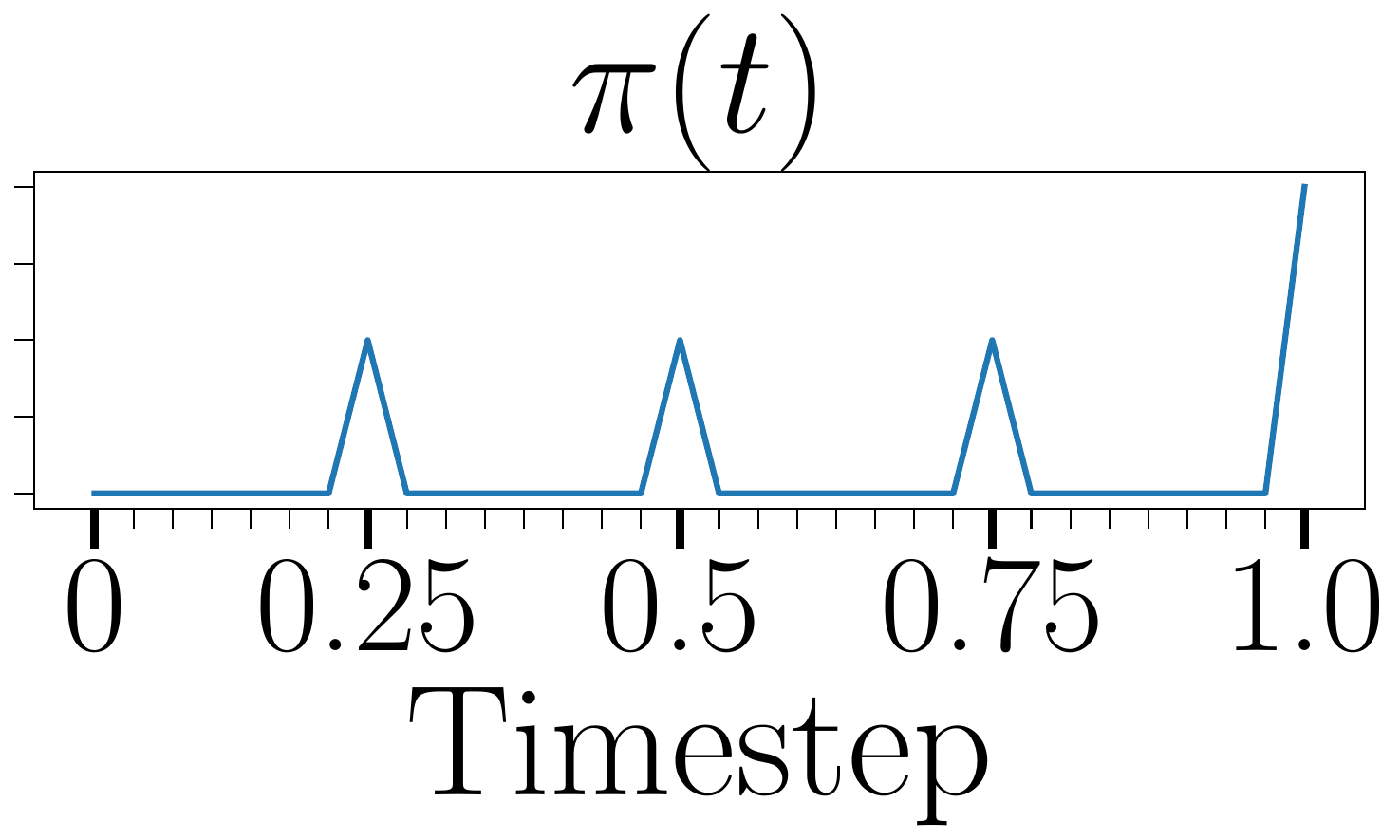}}}\\
  \subfloat[\scriptsize \centering (d) $\pi^{\mathrm{ours}}$]{\subfloat[\scriptsize \emph{Warm-up}]{\includegraphics[width=0.25\linewidth]{plots/timesteps/mixture_00_00_05_05.pdf}}
  \subfloat[\scriptsize Phase 1]{\includegraphics[width=0.25\linewidth]{plots/timesteps/mixture_01_03_03_03.pdf}}
  \subfloat[\scriptsize Phase 2]{\includegraphics[width=0.25\linewidth]{plots/timesteps/mixture_025_025_025_025.pdf}}
  \subfloat[ \scriptsize Phase 3]{\includegraphics[width=0.25\linewidth]{plots/timesteps/mixture_04_02_02_02.pdf}}}
  \caption{Illustration of the timestep distributions used in the ablation study.}
  \label{fig:app_timesteps}
\end{figure}

When ablating the timesteps distribution, we use the following distributions: $\pi^{\mathrm{uniform}}(t)$, $\pi^{\mathrm{gaussian}}(t)$, $\pi^{\mathrm{sharp}}(t)$ and $\pi^{\mathrm{ours}}(t)$ that are represented in Fig.~\ref{fig:app_timesteps}.

\paragraph{Flash SDXL}
In this section, we train a LoRA student model (108M trainable parameters) sharing the same UNet architecture as SDXL. The model is trained for 20k iterations on 4 H100-80Gb GPUs (amounting to a total of 176 H100 hours of training) with a batch size of 2 and a learning rate of $10^{-5}$ for both the student and the discriminator. The student weights are initialized with the teacher's one. The timestep distribution $\pi(t)$ is detailed in the main paper and chosen such that $K=32$. We also shift modes every 5000 iterations. As for SD1.5, we set $\lambda_{\mathrm{adv}} = 0$ and $\lambda_{\mathrm{DMD}} = 0$ and progressively increase each time we change the timestep distribution so they reach final values set to $0.3$ and $0.7$ respectively. The schedule is $[0, 0.1, 0.2, 0.3]$ for $\lambda_{\mathrm{adv}}$ and $[0, 0.3, 0.5, 0.7]$ for $\lambda_{\mathrm{DMD}}$. We use a guidance scale $\omega$ uniformly sampled from $[3, 13]$ with a distillation loss chosen as LPIPS and the GAN loss is set to the LSGAN loss.

\paragraph{Flash Pixart (DiT)}
We train a LoRA student model (66.5M trainable parameters) sharing the same architecture as the teacher for 40k iterations on 4 H100-80Gb GPUs (amounting to a total of 188 H100 hours of training) with a batch size of 2 and a learning rate of $1e^{-5}$ together with Adam optimizer \citep{kingma2014adam} for both the student and the discriminator. The weights of the student model are initialized using the teacher's. We use the timestep distribution $\pi(t)$ such that $K=16$ and shift modes every 10000 iterations. We also start with both $\lambda_{\mathrm{adv}} = 0$ and $\lambda_{\mathrm{DMD}} = 0$ and progressively increase each time we change the timestep distribution so they reach final values set to $0.3$ and $0.7$ respectively. The schedule is $[0, 0.05, 0.1, 0.2]$ for $\lambda_{\mathrm{adv}}$ and $[0, 0.3, 0.5, 0.7]$ for $\lambda_{\mathrm{DMD}}$. The guidance scale $\omega$ used to denoise using the teacher model is uniformly sampled from $[2, 9]$. The distillation loss is LPIPS loss and the GAN loss is set as the LSGAN loss.

\subsection{Experimental Setup for \emph{Inpainting}}
For the \emph{inpainting} experiment, we use an \emph{in-house} diffusion-based model whose backbone architecture is similar to the one of SDXL \citep{podell2023sdxl} and weights are initialized using the teacher. The student model is trained on 512x512 input image resolution for 20k iterations on 2 H100-80Gb GPUs with a batch size of 4 and a learning rate of $10^{-5}$ for both the student and the discriminator. The timestep distribution $\pi(t)$ is chosen with $K=16$. Modes are shifted every 5000 iterations. We again start with both $\lambda_{\mathrm{adv}} = 0$ and $\lambda_{\mathrm{DMD}} = 0$ and progressively increase each time we change the timestep distribution so they reach final values set to $0.3$ and $0.7$ respectively. The schedule is $[0, 0.1, 0.2, 0.3]$ for $\lambda_{\mathrm{adv}}$ and $[0, 0.3, 0.5, 0.7]$ for $\lambda_{\mathrm{DMD}}$. The guidance scale $\omega$ is uniformly sampled from $[3, 13]$. The distillation loss is set as the MSE loss and the GAN loss is set as the LSGAN loss.

\subsection{Experimental Setup for \emph{Super-Resolution}}
For the \emph{super-resolution} experiment, we use an \emph{in-house} diffusion-based model whose backbone architecture is similar to the one of SDXL \citep{podell2023sdxl}. The student model is trained with 256x256 low-resolution images used as conditioning and outputs 1024x1024 images. The student model is initialized using the teacher's weights and is trained for 20k iterations on 2 H100-80Gb GPUs with a batch size of 4 and a learning rate of $10^{-5}$ for both the student and the discriminator. We set $K=16$ for $\pi(t)$ and shift modes every 5000 iterations. We start with $\lambda_{\mathrm{adv}} = 0$ and $\lambda_{\mathrm{DMD}} = 0$ and progressively increase each time we change the timestep distribution so they reach final values set to $0.3$ and $0.7$ respectively. The schedule is $[0, 0.1, 0.2, 0.3]$ for $\lambda_{\mathrm{adv}}$ and $[0, 0.3, 0.5, 0.7]$ for $\lambda_{\mathrm{DMD}}$. The guidance scale $\omega$ used to denoise using the teacher model is uniformly sampled from $[1.2, 1.8]$. The distillation loss is set as the MSE loss and the GAN loss is chosen as the LSGAN loss.

\subsection{Experimental Setup for \emph{Face-Swapping}}
For the \emph{face-swapping} experiment, we use an \emph{in-house} diffusion-based model whose backbone architecture is similar to the one of SD2.2 \citep{rombach2022high}. The student model is trained on 512x512 input images and target images. We use a face detector to extract the face from the source image and use it as conditioning. The student model is then initialized using the teacher's weights and is trained for 15k iterations on 2 H100-80Gb GPUs with a batch size of 8 and a learning rate of $10^{-5}$ for both the student and the discriminator. We use the timestep distribution $\pi(t)$ with $K=16$ and shift modes every 5000 iterations. We also start with both $\lambda_{\mathrm{adv}} = 0$ and $\lambda_{\mathrm{DMD}} = 0$ and progressively increase each time we change the timestep distribution so they reach final values set to $0.3$ and $0.7$ respectively. The schedule is $[0, 0.1, 0.2, 0.3]$ for $\lambda_{\mathrm{adv}}$ and $[0, 0.3, 0.5, 0.7]$ for $\lambda_{\mathrm{DMD}}$. The guidance scale $\omega$ used to denoise using the teacher model is uniformly sampled from $[2.0, 7.0]$. The distillation loss is set as the MSE loss and the GAN loss is chosen as the LSGAN loss.

\subsection{Experimental Setup for Adapters}
In this study, the student model is trained using the proposed method and unchanged hyper-parameters unless the guidance that was sampled in $\mathcal{U}([3.0, 7.0])$ and $K$ is set to 16 to speed up the training. For both adapters, we use a conditioning scale of 0.8 to generate the samples with the student model.

\section{Additional Sampling Results}
In this section, we provide additional samples for each task considered in the main paper. The prompts for Fig.~\ref{fig:qualitative} of the main manuscript are from top to bottom \emph{A photograph of a school bus in a magic forest}, \emph{A monkey making latte art} and \emph{A majestic lion stands proudly on a rock, overlooking the vast African savannah} (SDXL), \emph{A whale with a big mouth and a rainbow on its back jumping out of the water}, \emph{A small cactus with a happy face in the Sahara desert},  \emph{A close-up of a person with a shaved head, gazing downwards, with a hand resting on their forehead} (Pixart-$\alpha$) and \emph{A cat holding a sign that says "4 steps"}, \emph{A close up of an old elderly man with green eyes looking straight at the camera} and \emph{A raccoon trapped inside a glass jar full of colorful candies, the background is steamy with vivid colors} (SD3).

\subsection{Flash SDXL}\label{sec:app_flash_sdxl}
\begin{figure*}[p]
  \centering
  \captionsetup[subfigure]{position=above, labelformat = empty}
  \subfloat[\scriptsize Teacher\\(40 NFEs)]{\includegraphics[width=1.0in]{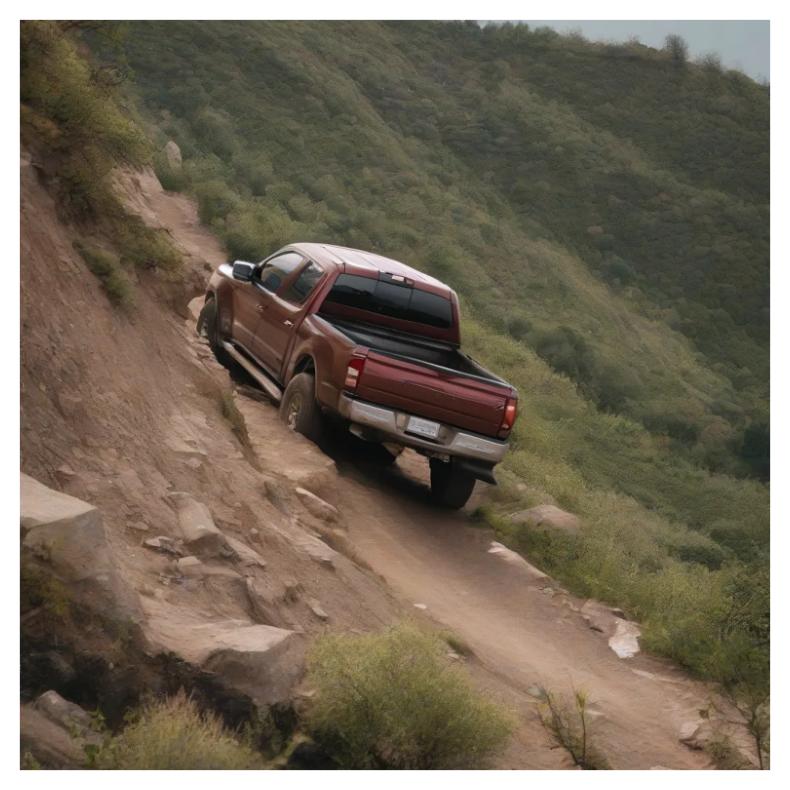}}
  \subfloat[\scriptsize LCM\\(4 NFEs)]{\includegraphics[width=1.0in]{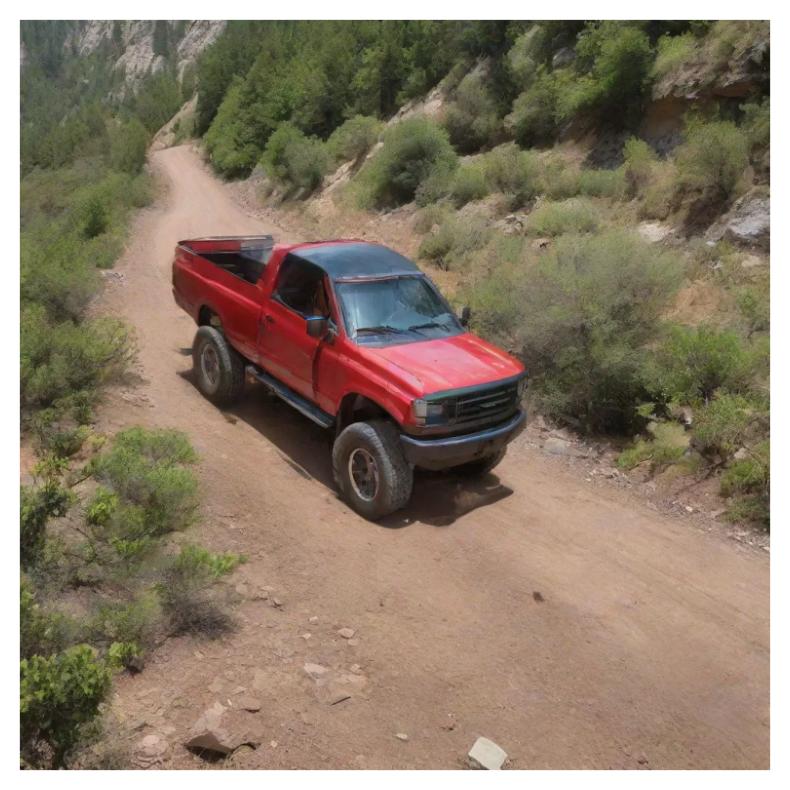}}
  \subfloat[\scriptsize Lightning\\(4 NFEs)]{\includegraphics[width=1.0in]{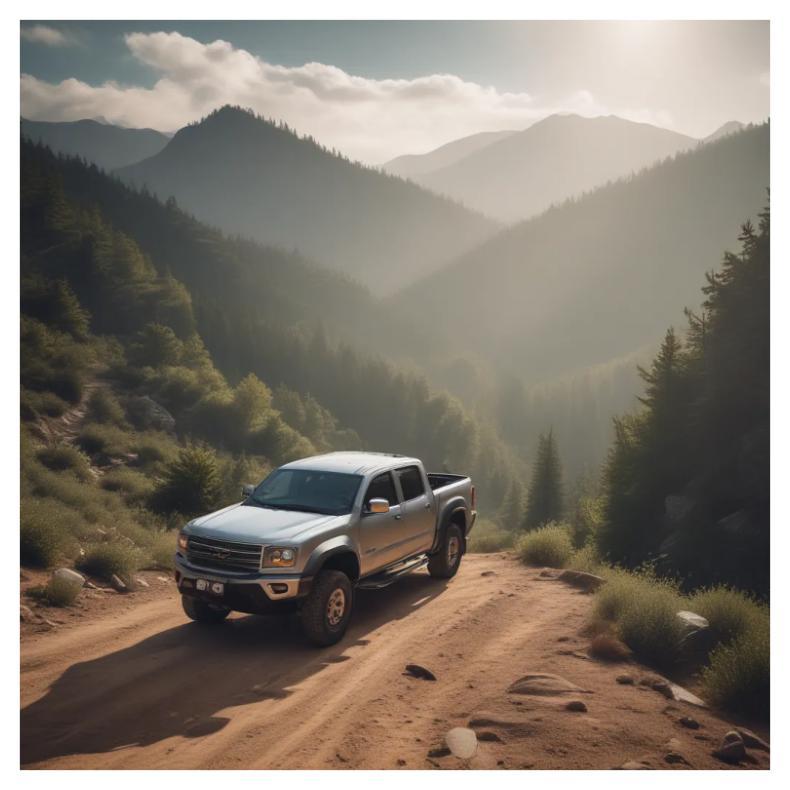}}
  \subfloat[\scriptsize HyperSD\\(4 NFEs)]{\includegraphics[width=1.0in]{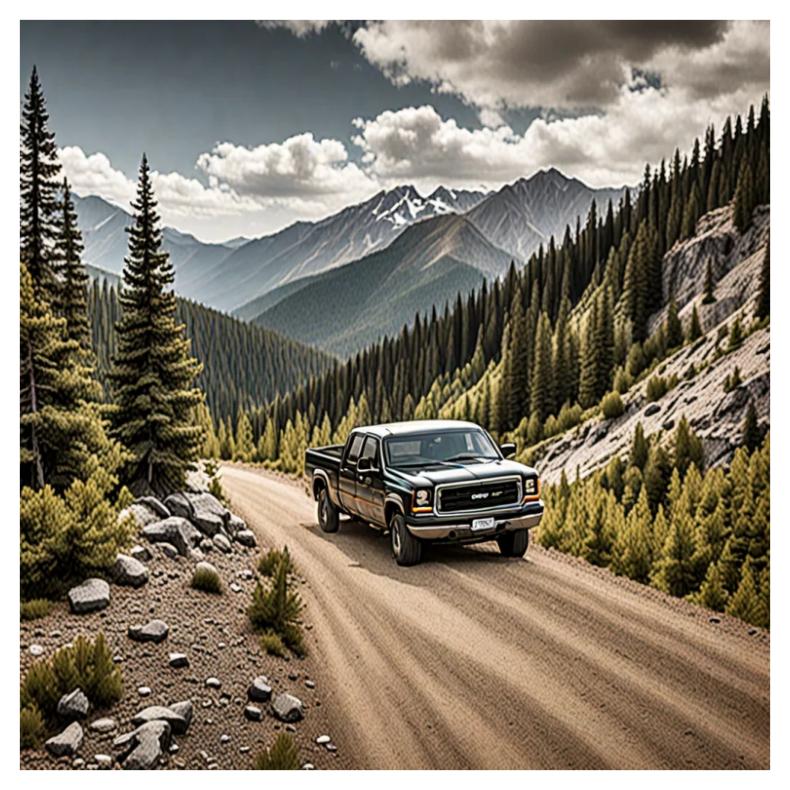}}
   \subfloat[\scriptsize Ours\\(4 NFEs)]{\includegraphics[width=1.0in]{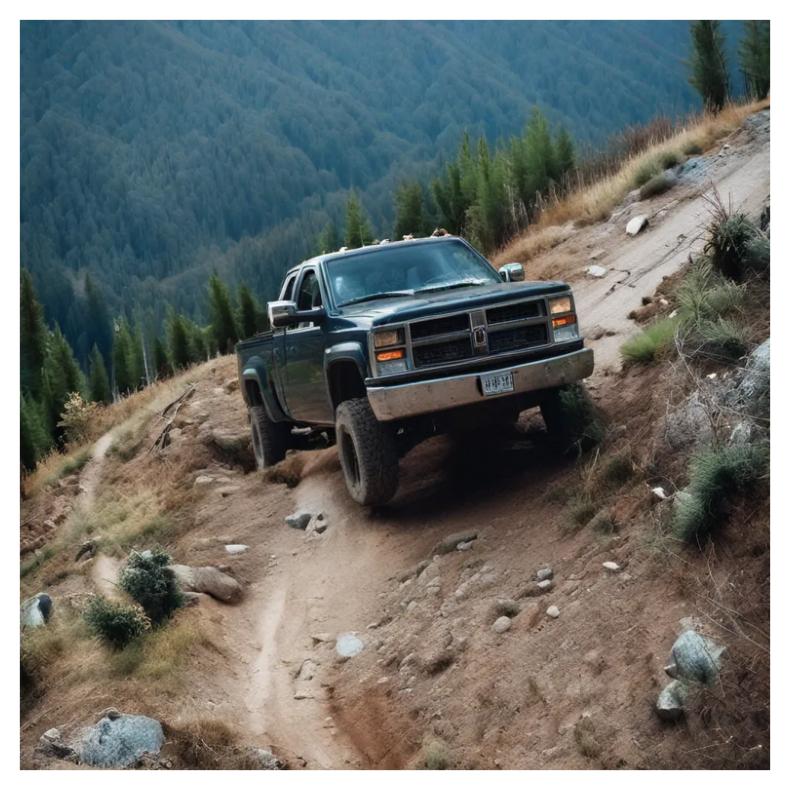}}\\\vspace{-1.em}
   \subfloat{\scriptsize \emph{A pickup truck going up a mountain switchback}}\\\vspace{-1.em}
  \subfloat{\includegraphics[width=1.0in]{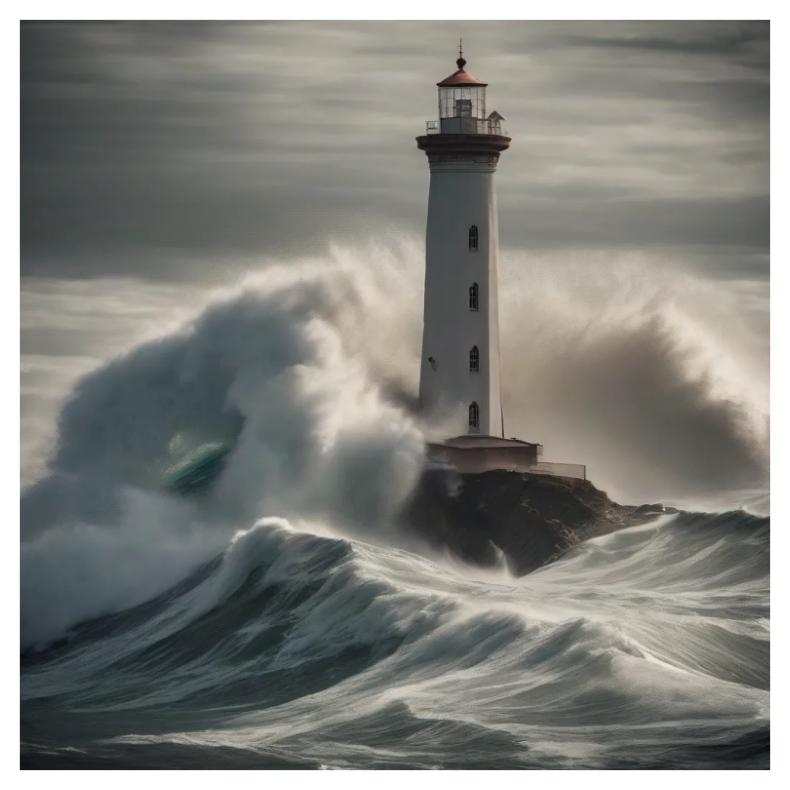}}
   \subfloat{\includegraphics[width=1.0in]{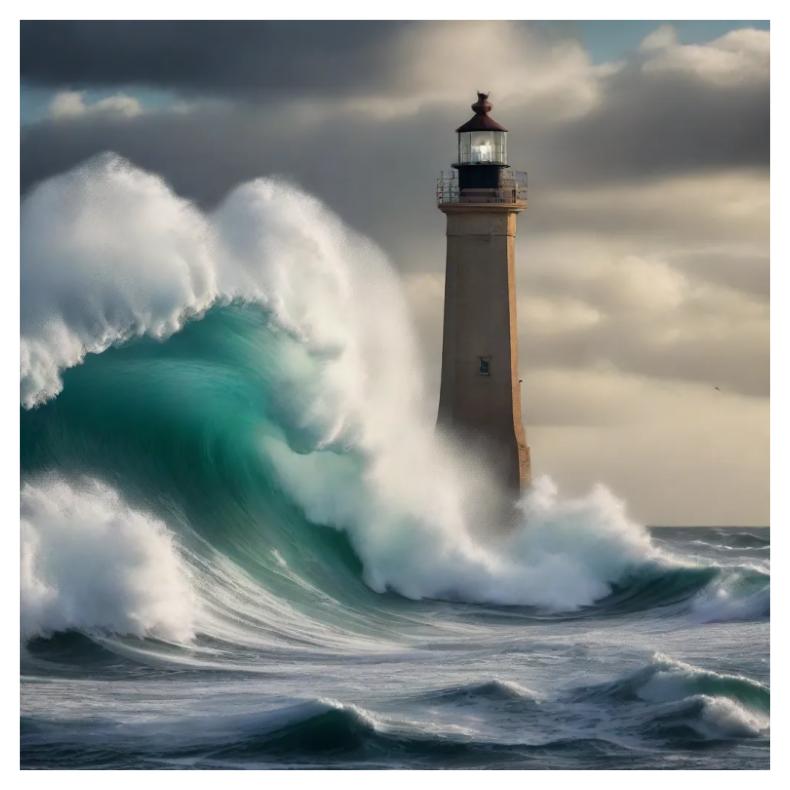}}
  \subfloat{\includegraphics[width=1.0in]{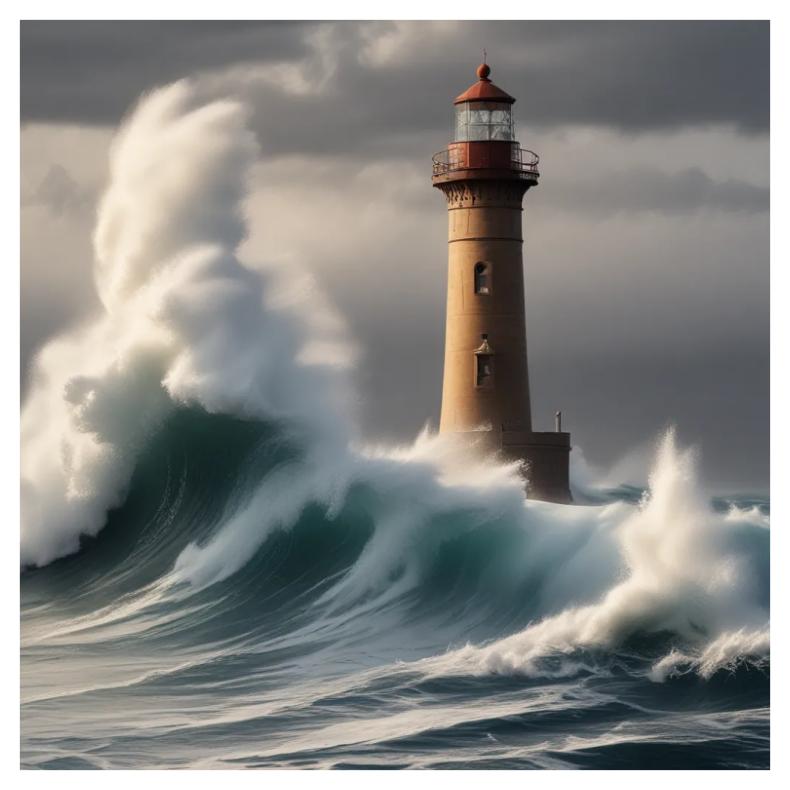}}
  \subfloat{\includegraphics[width=1.0in]{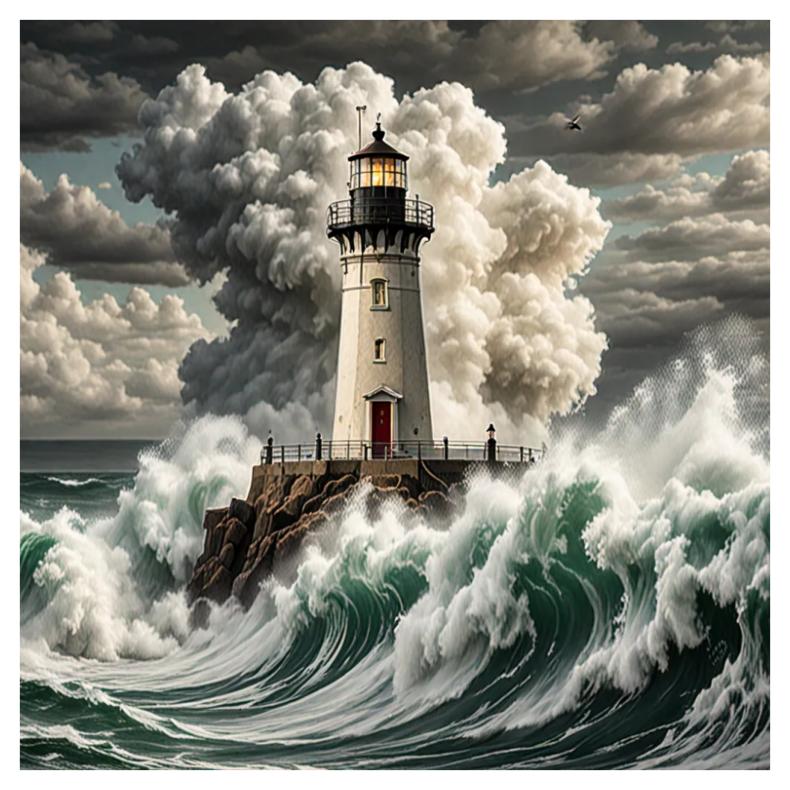}}
  \subfloat{\includegraphics[width=1.0in]{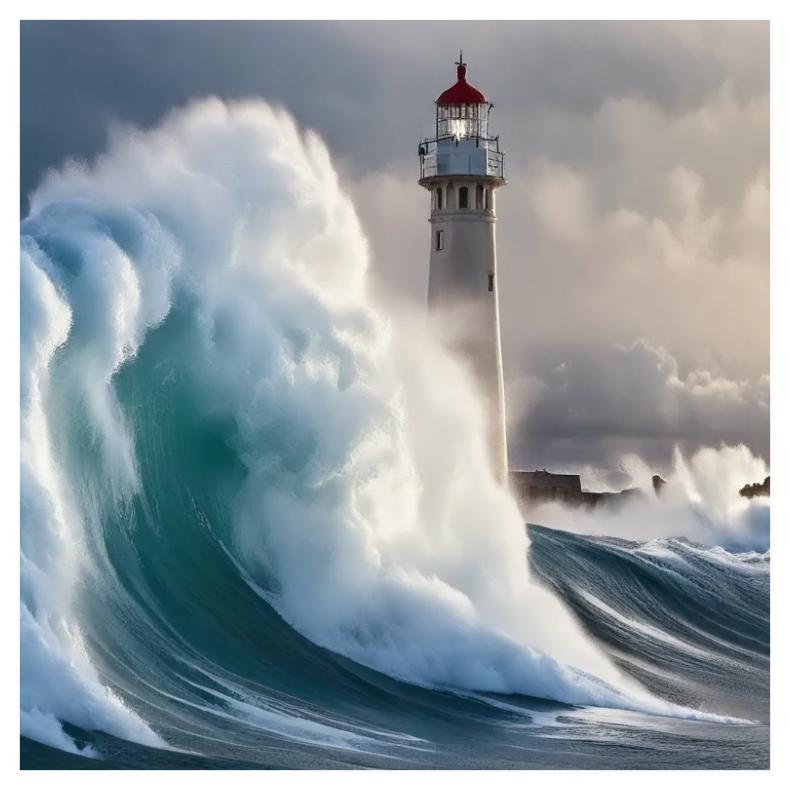}}\\\vspace{-1.em}
  \subfloat{\scriptsize \emph{A giant wave breaking on a majestic lighthouse}}\\\vspace{-1.em}
  \subfloat{\includegraphics[width=1.0in]{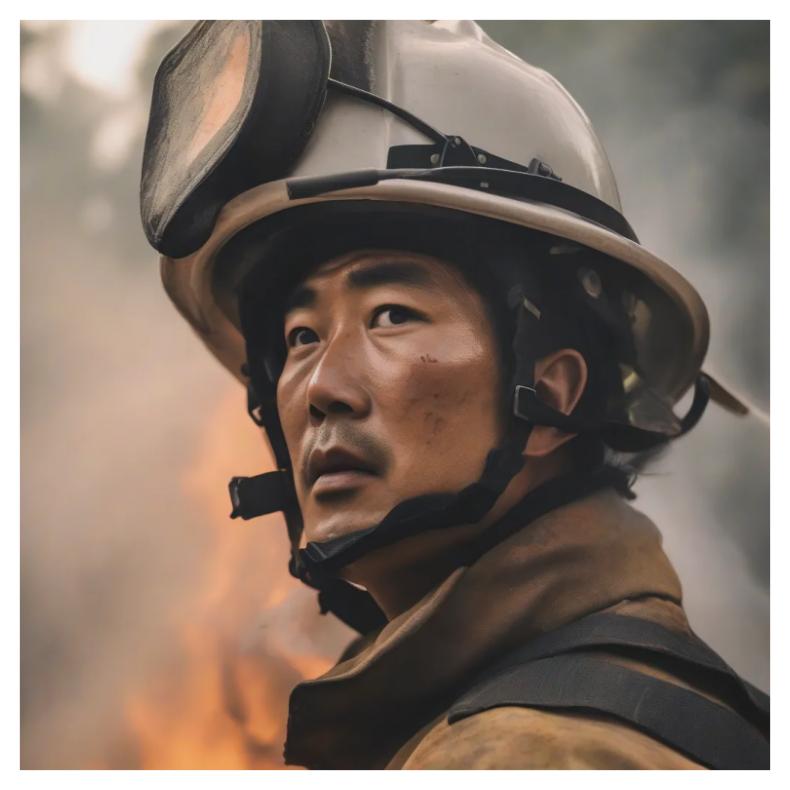}}
  \subfloat{\includegraphics[width=1.0in]{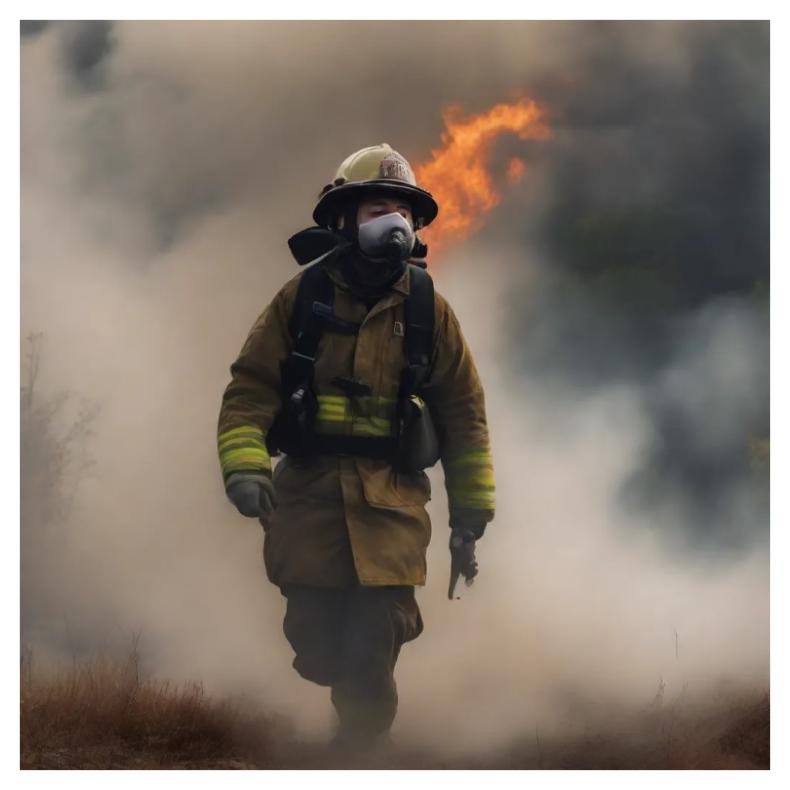}}
  \subfloat{\includegraphics[width=1.0in]{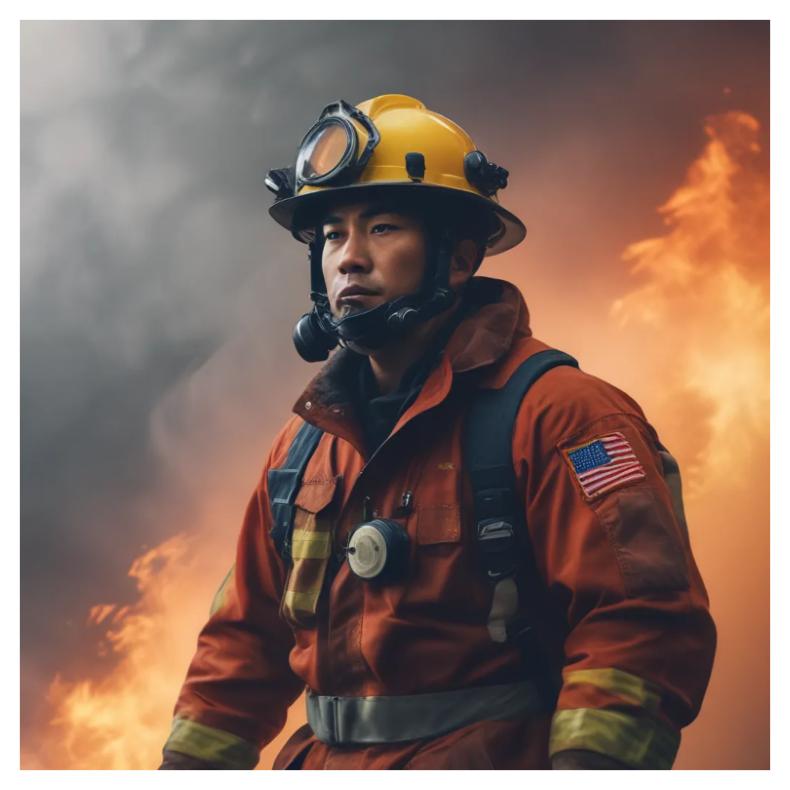}}
   \subfloat{\includegraphics[width=1.0in]{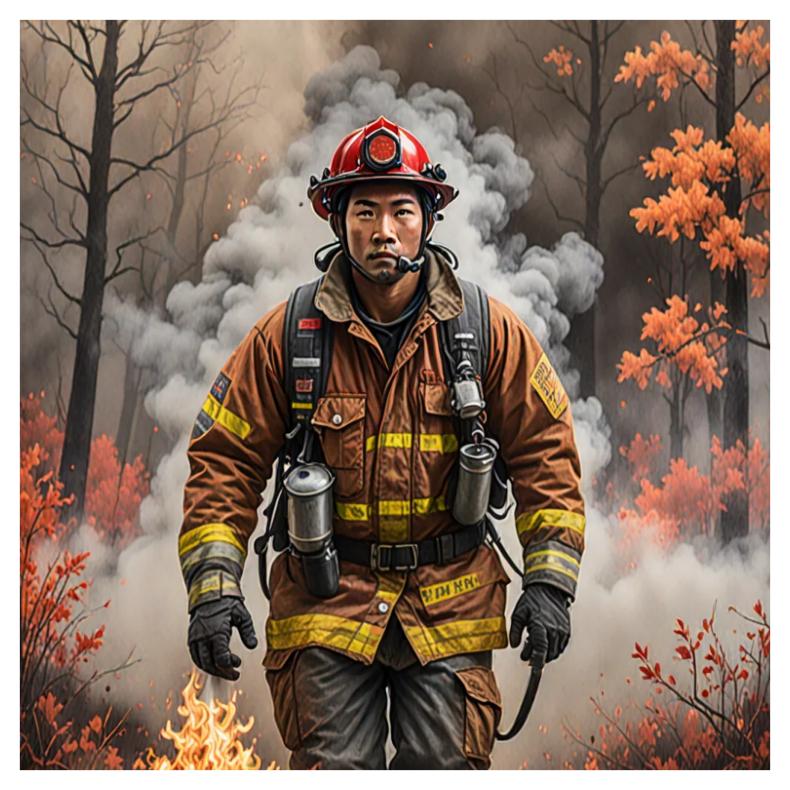}}
  \subfloat{\includegraphics[width=1.0in]{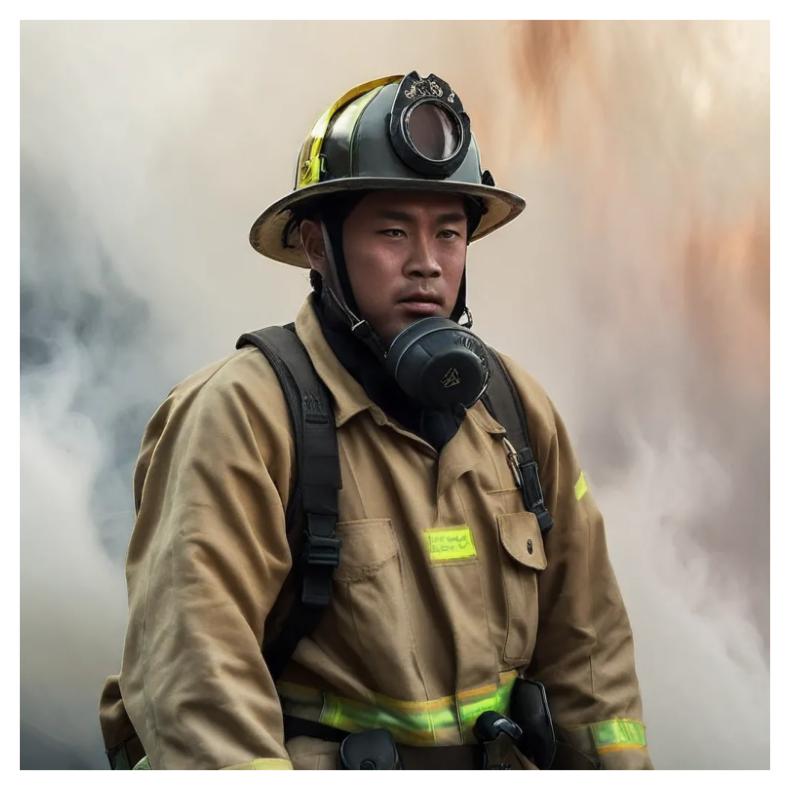}}\\\vspace{-1.em}
  \subfloat{\scriptsize \emph{An Asian firefighter with a rugged jawline rushes through the billowing smoke of an autumn blaze}}\\\vspace{-1.em}
  \subfloat{\includegraphics[width=1.0in]{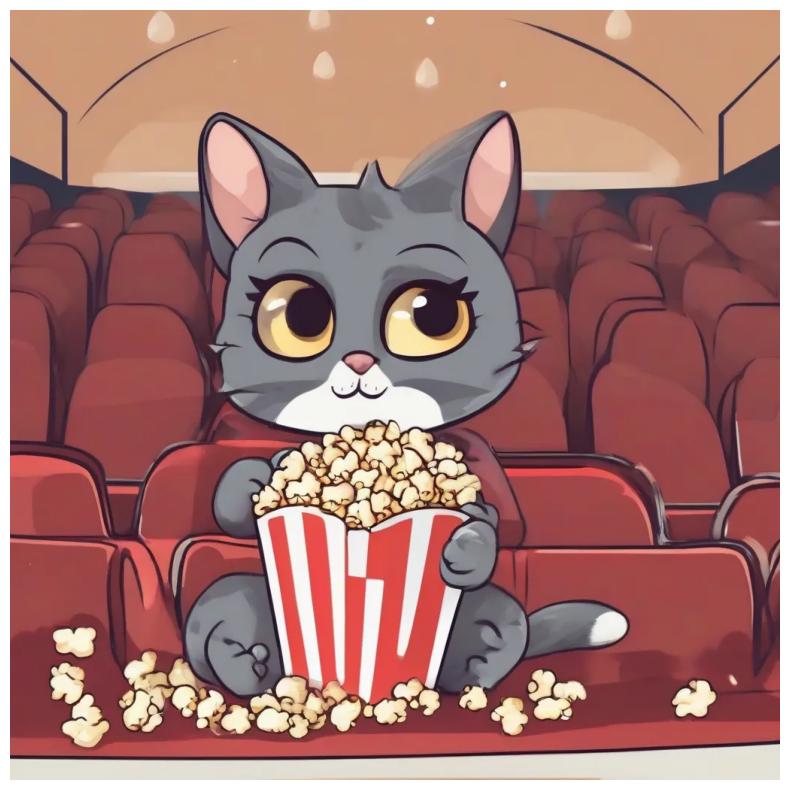}}
  \subfloat{\includegraphics[width=1.0in]{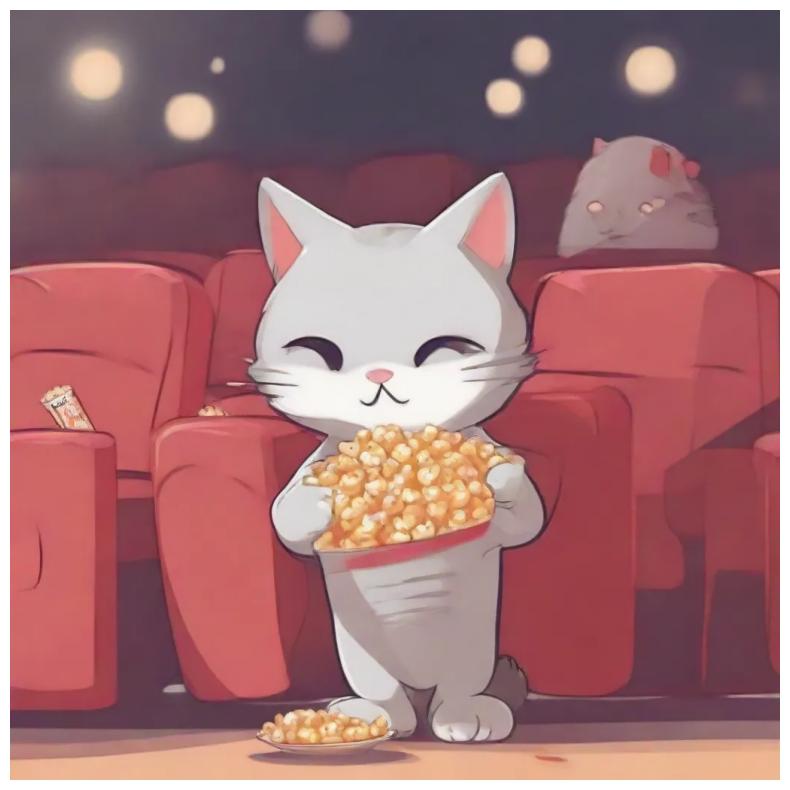}}
   \subfloat{\includegraphics[width=1.0in]{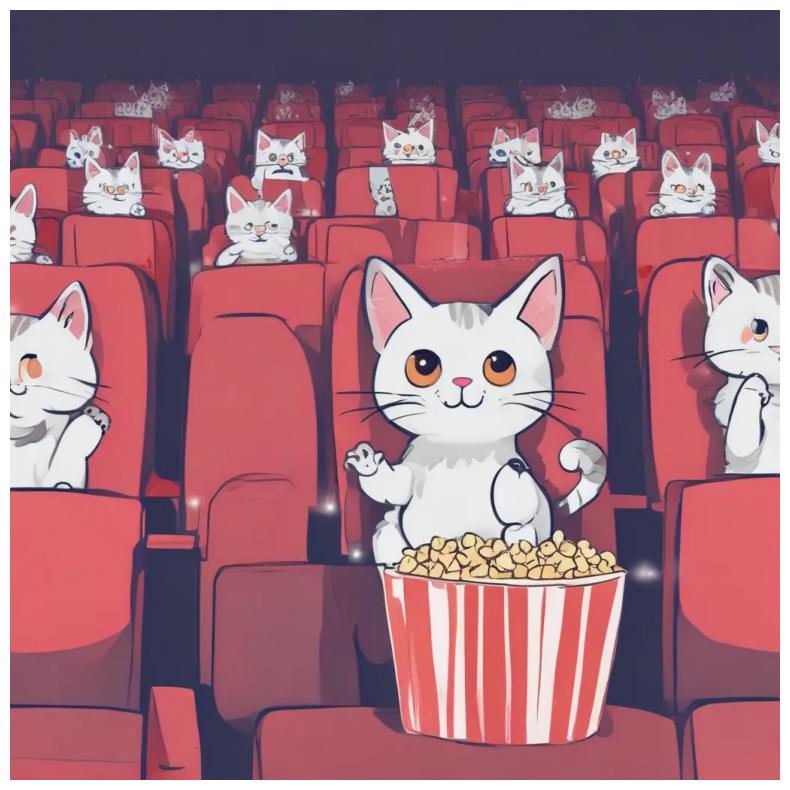}}
  \subfloat{\includegraphics[width=1.0in]{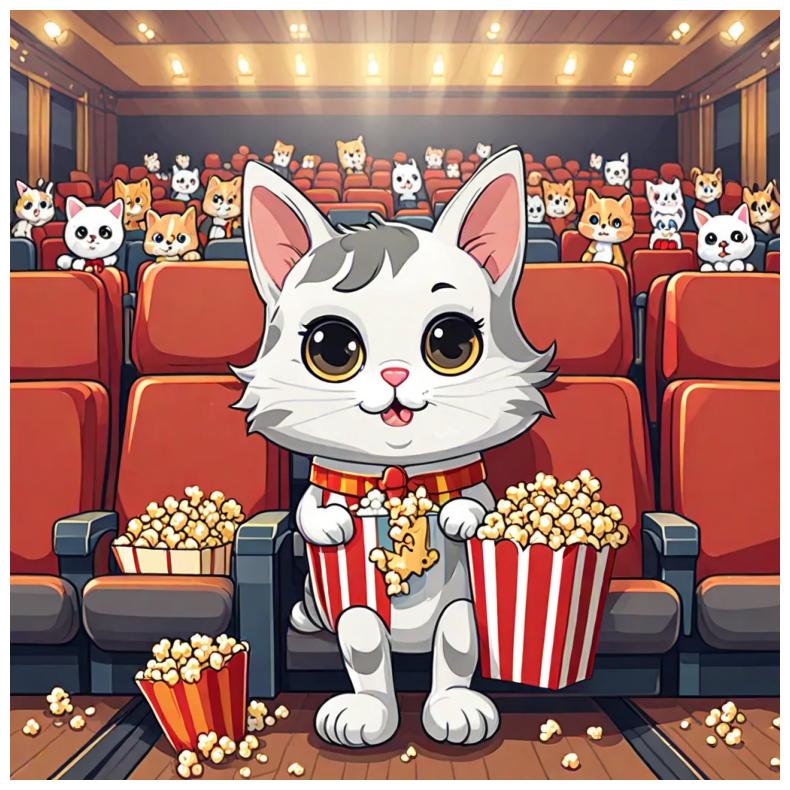}}
  \subfloat{\includegraphics[width=1.0in]{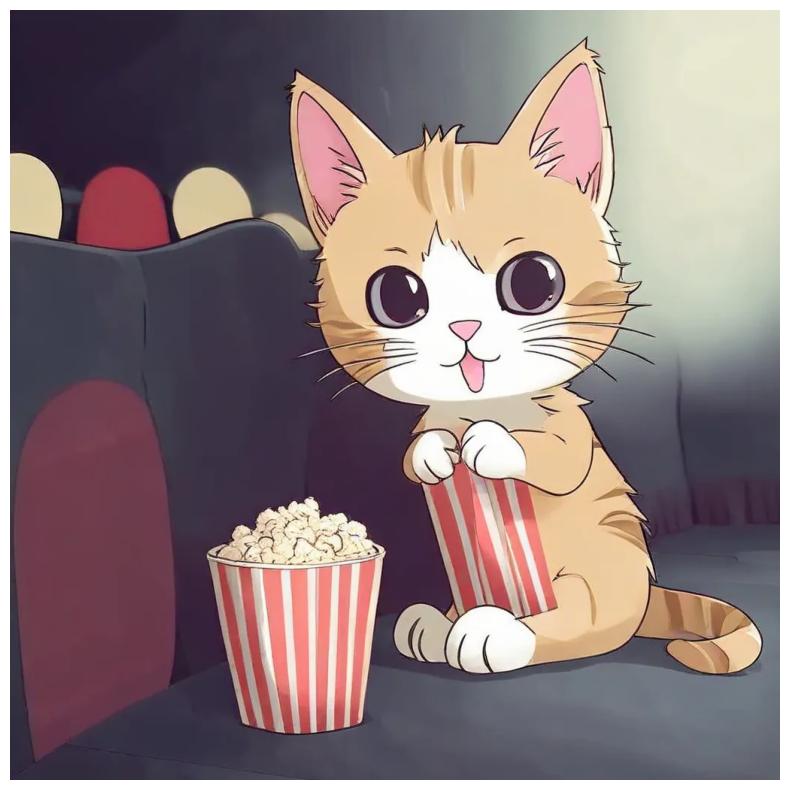}}
  \\\vspace{-1.em}
  \subfloat{\scriptsize \emph{Cute cartoon small cat sitting in a movie theater eating popcorn, watching a movie}}\\\vspace{-1.em}
  \subfloat{\includegraphics[width=1.0in]{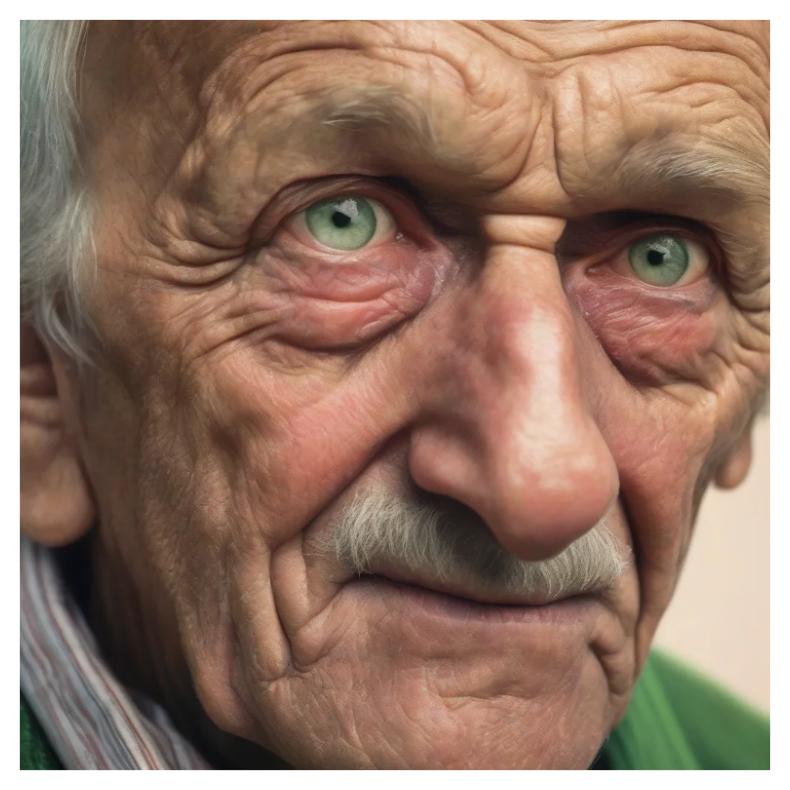}}
  \subfloat{\includegraphics[width=1.0in]{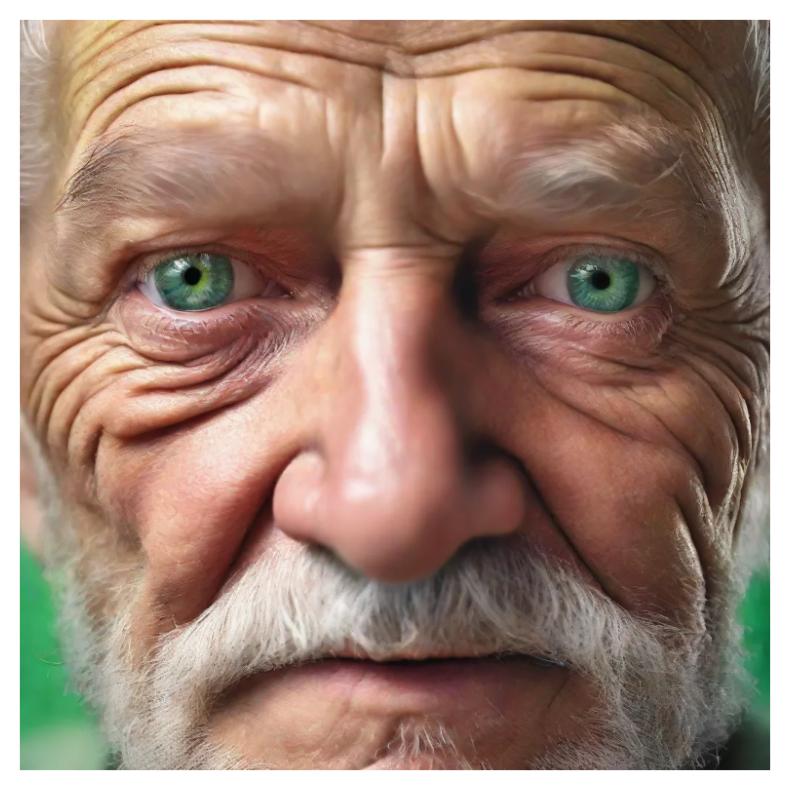}}
  \subfloat{\includegraphics[width=1.0in]{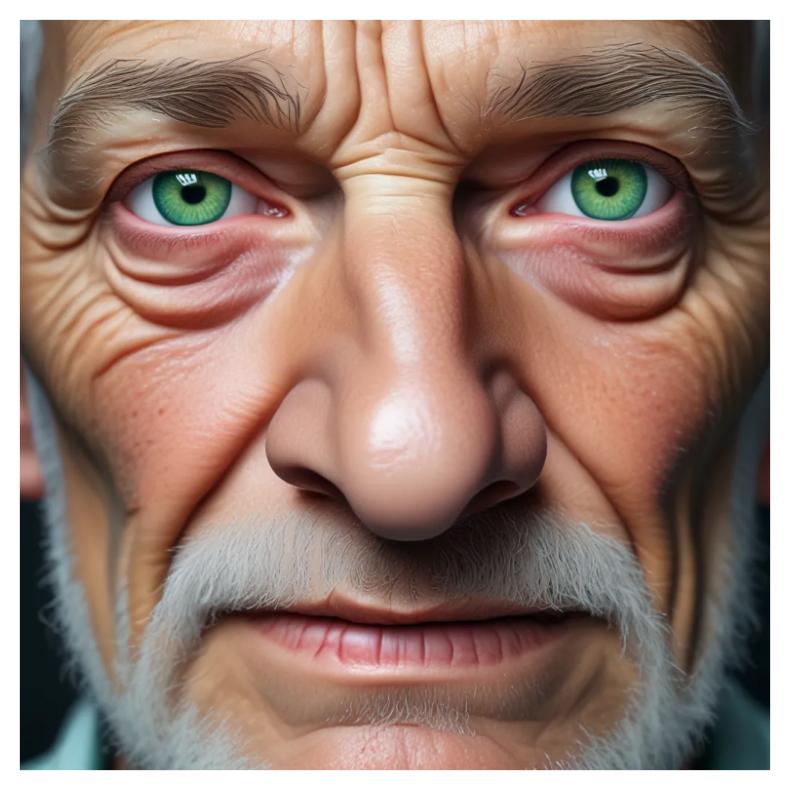}}
  \subfloat{\includegraphics[width=1.0in]{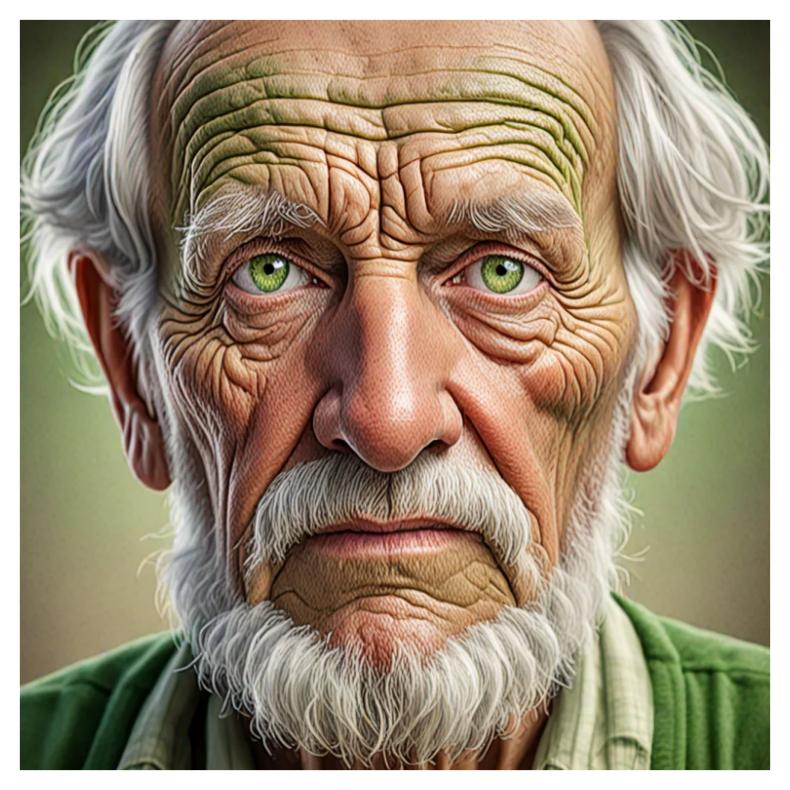}}
  \subfloat{\includegraphics[width=1.0in]{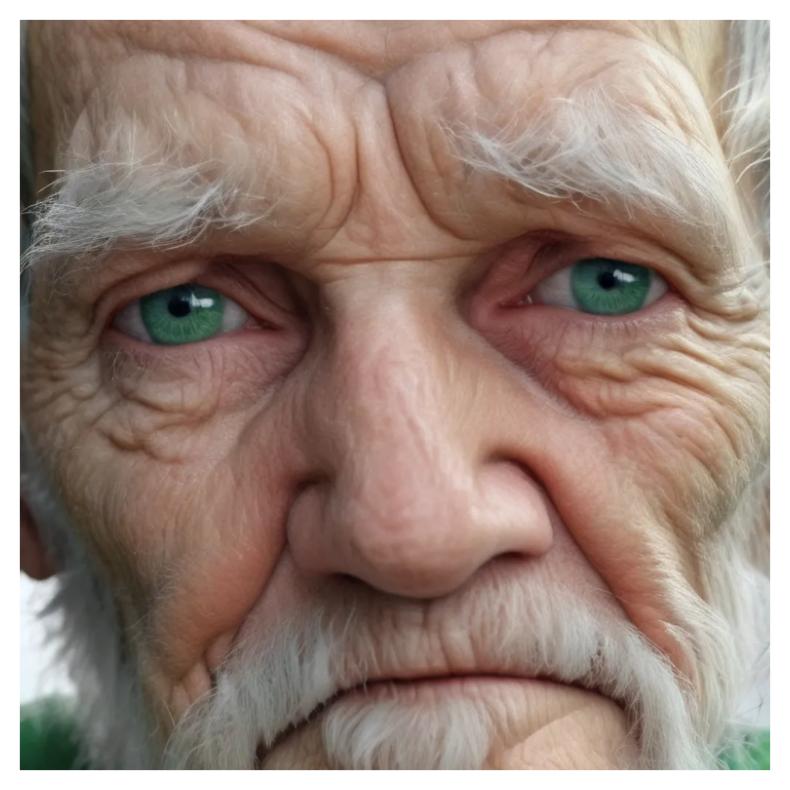}}\\\vspace{-1.em}
  \subfloat{\scriptsize \emph{A very realistic close up of an old elderly man with green eyes looking straight at the camera, vivid colors}}\\\vspace{-1.em}
  \subfloat{\includegraphics[width=1.0in]{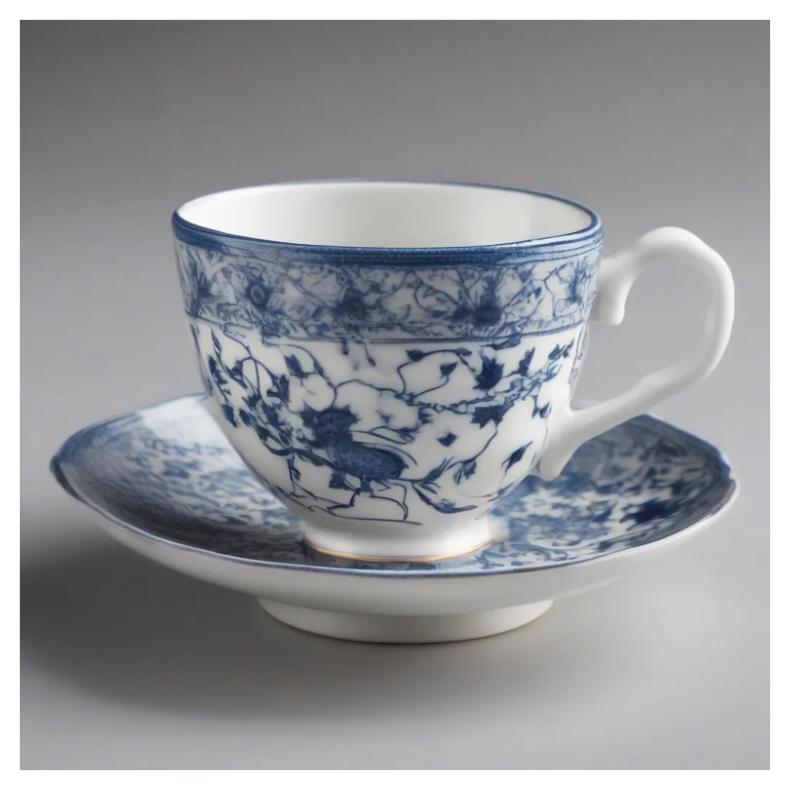}}
  \subfloat{\includegraphics[width=1.0in]{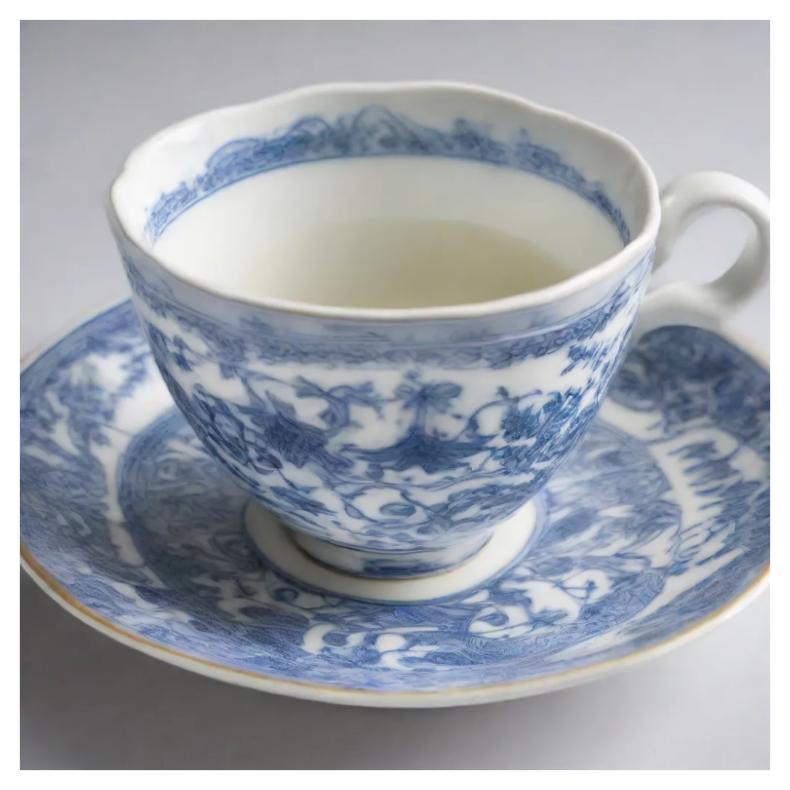}}
   \subfloat{\includegraphics[width=1.0in]{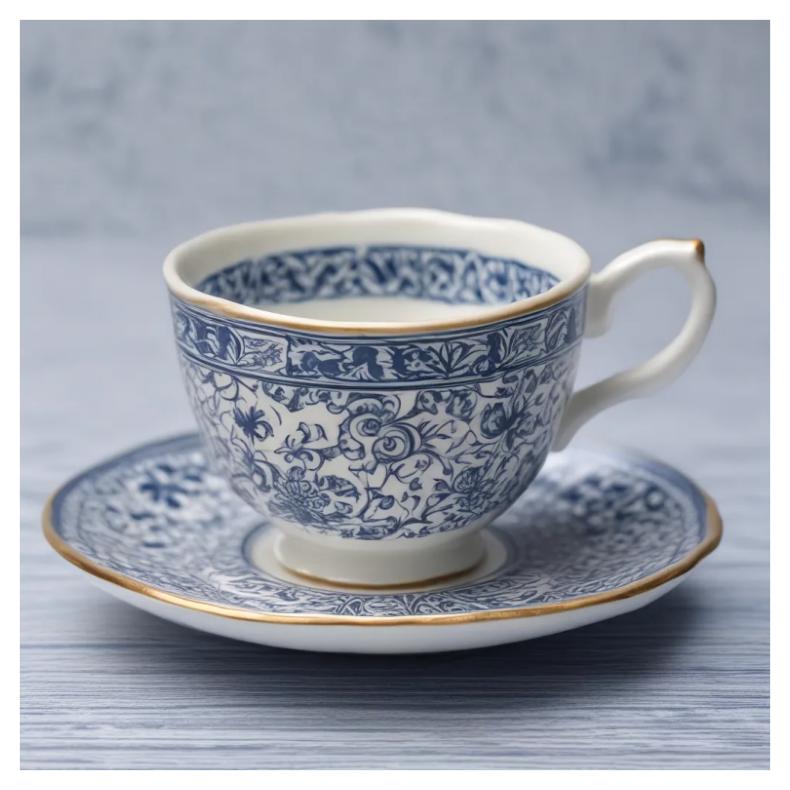}}
  \subfloat{\includegraphics[width=1.0in]{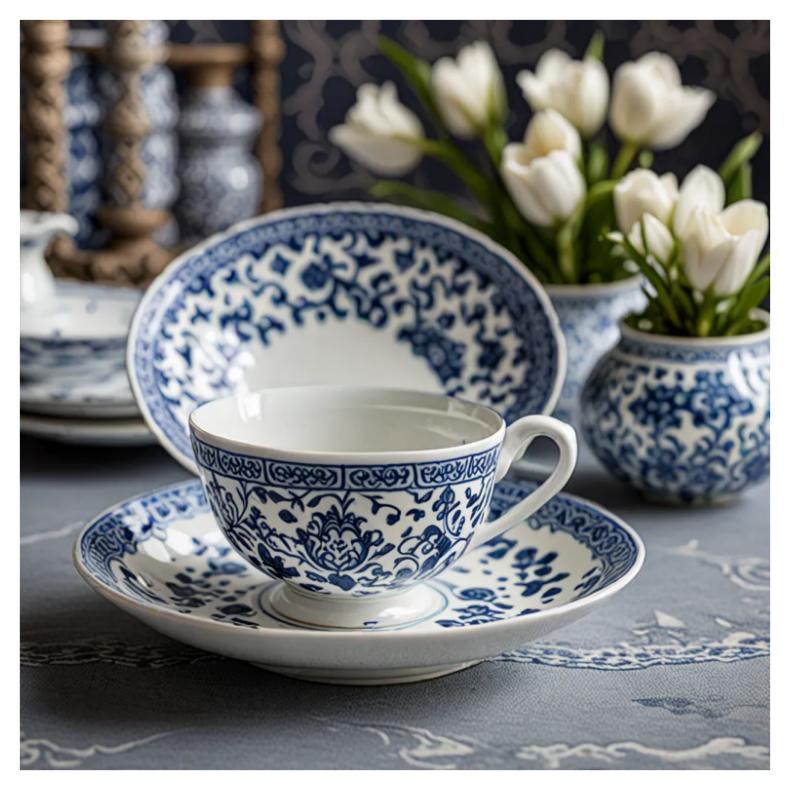}}
  \subfloat{\includegraphics[width=1.0in]{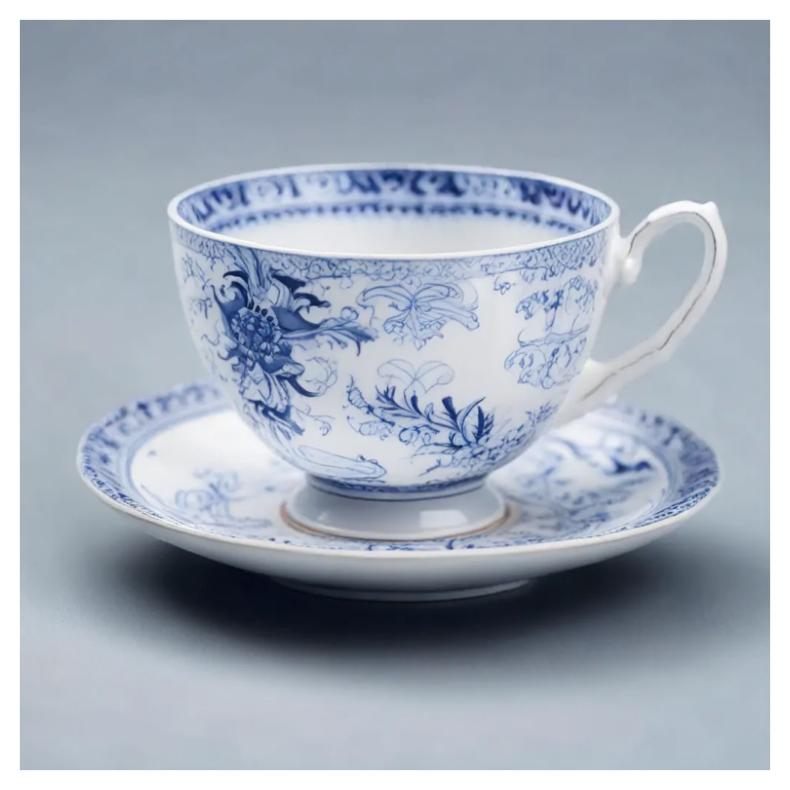}}\\\vspace{-1.em}
  \subfloat{\scriptsize \emph{A delicate porcelain teacup sits on a saucer, its surface adorned with intricate blue patterns}}
  \caption{Application of \emph{Flash Diffusion} to a SDXL teacher model. The proposed method 4 NFEs generations are compared to the teacher generations using 40 NFEs as well as 
  LoRA approaches proposed in the literature (LCM \citep{luo2023latent}, SDXL-lightning \citep{lin2024sdxl} and Hyper-SD \citep{ren2024hyper}). Teacher samples are generated with a guidance scale of 5. Best viewed zoomed in.}
  \label{fig:app_sdxl}
  \end{figure*}

\begin{figure*}[p]
      \centering
      \includegraphics[width=0.5\linewidth]{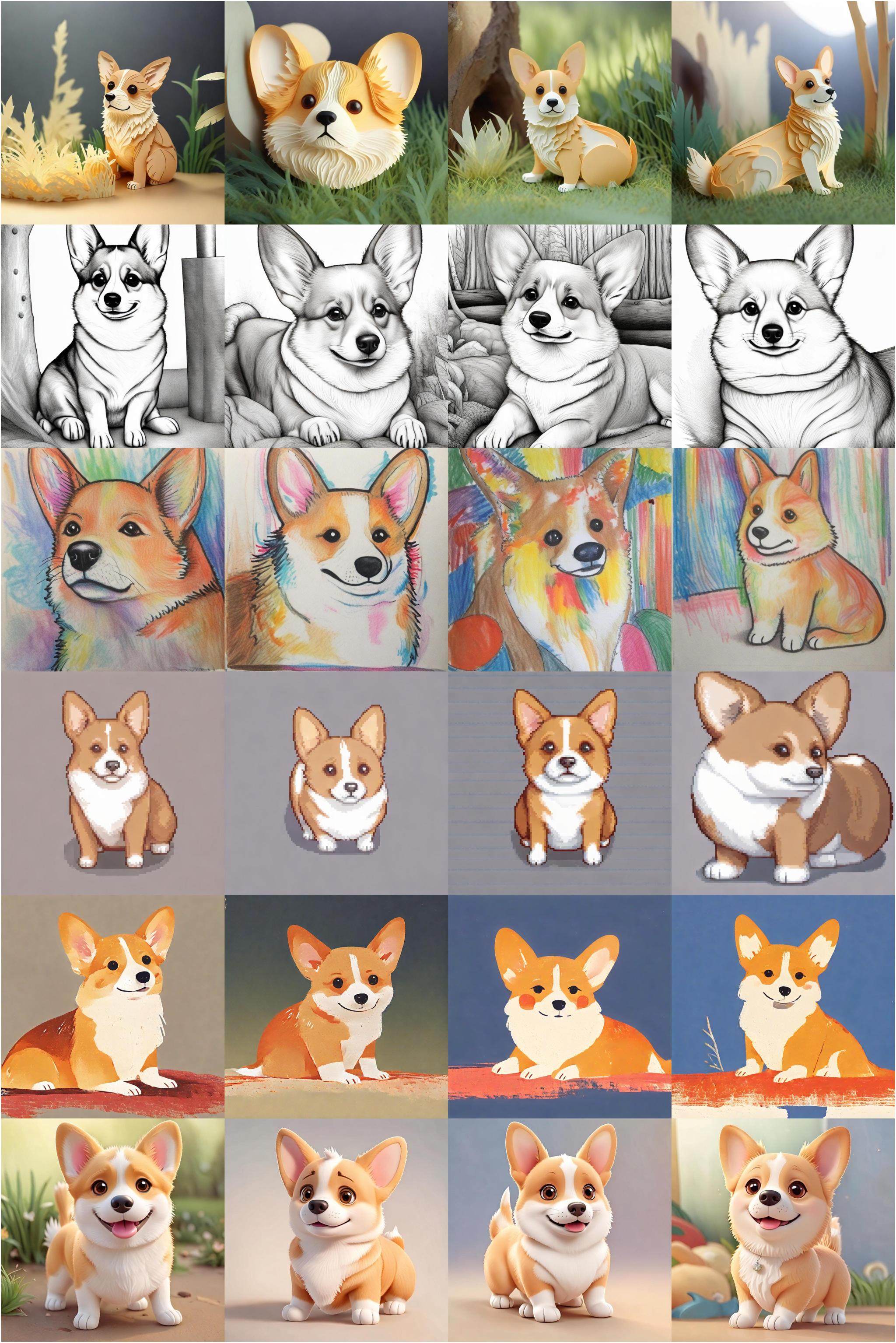}
      \caption{Application of 6 SDXL LoRAs on top of Flash SDXL in a \emph{training-free} manner. We show samples using 4 NFEs for each LoRA. }
      \label{fig:loras}
  \end{figure*}

In Fig.~\ref{fig:app_sdxl}, we provide addition samples enriching the qualitative comparision performed in the main manuscript. Again, to be fair to the competitors, we use some prompts from \citep{lin2024sdxl} to generate the samples. As mentioned in the paper, the proposed approach appears to be able to generate samples that are visually closer to the learned teacher distribution. We also provide additional samples of 6 LoRAs directly plugged on top of Flash SDXL in a \emph{training-free} manner in Fig.~\ref{fig:loras}.


\subsection{Flash Pixart (DiT)}\label{sec:app_flash_dit}
\begin{figure*}[p]
  \centering
  \captionsetup[subfigure]{position=above, labelformat = empty}
  \subfloat[\scriptsize Teacher\\(8 NFEs)]{\includegraphics[width=1.0in]{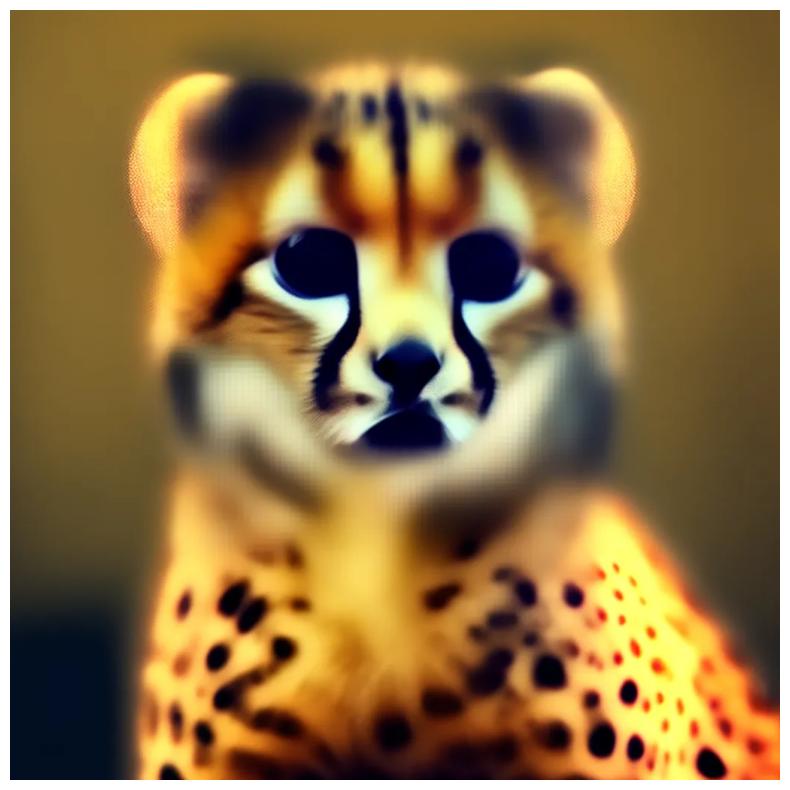}}
  \subfloat[\scriptsize Teacher\\(40 NFEs)]{\includegraphics[width=1.0in]{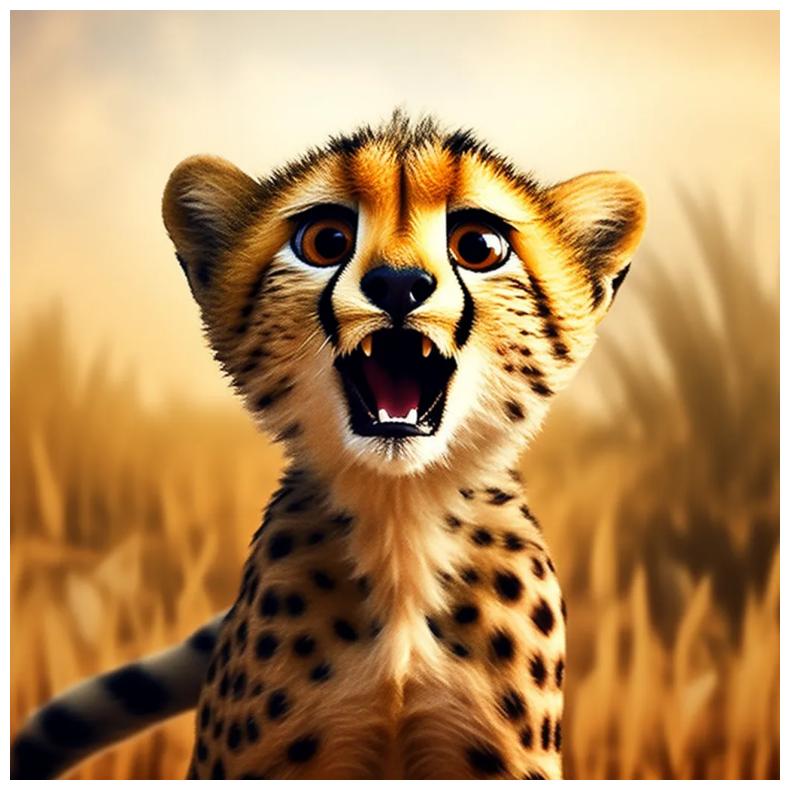}}
  \subfloat[\scriptsize LCM\\(4 NFEs)]{\includegraphics[width=1.0in]{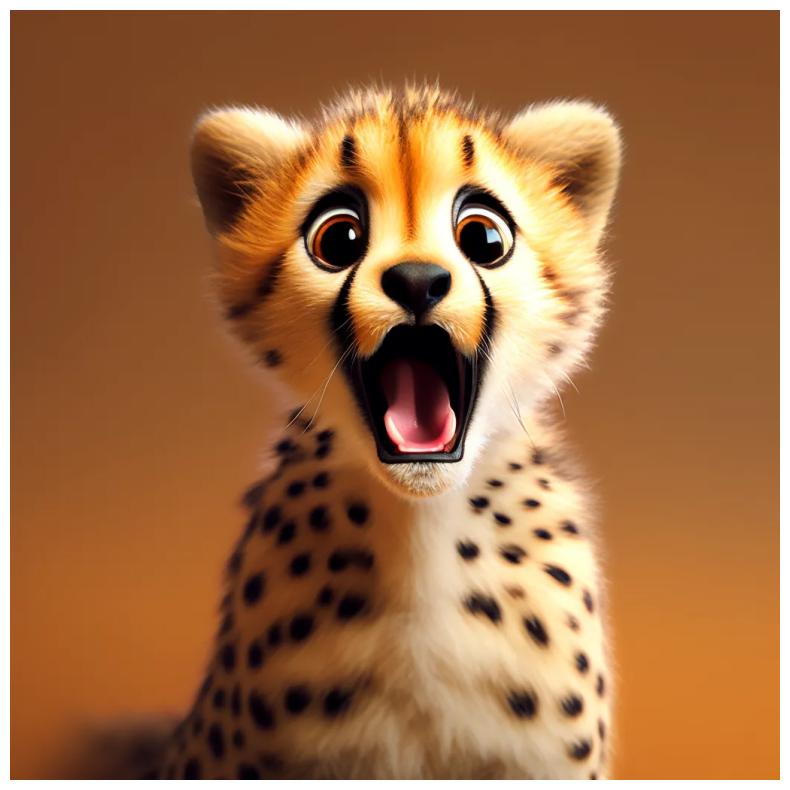}}
  \subfloat[\scriptsize Ours\\(4 NFEs)]{\includegraphics[width=1.0in]{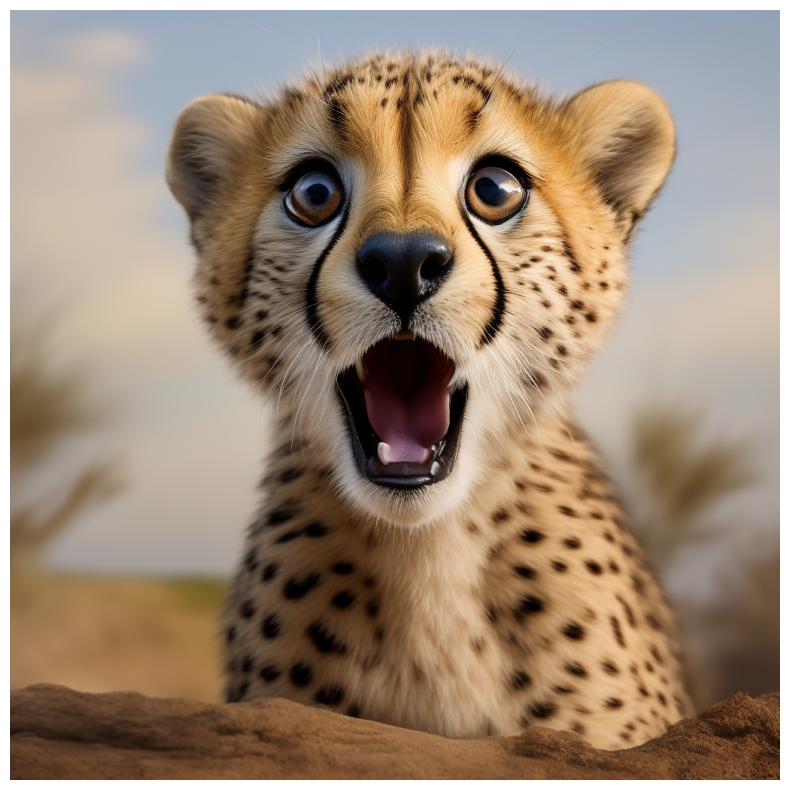}}\\\vspace{-1.em}
  \subfloat{\scriptsize \emph{A cute cheetah looking amazed and surprised}}\\\vspace{-1.em}
  \subfloat{\includegraphics[width=1.0in]{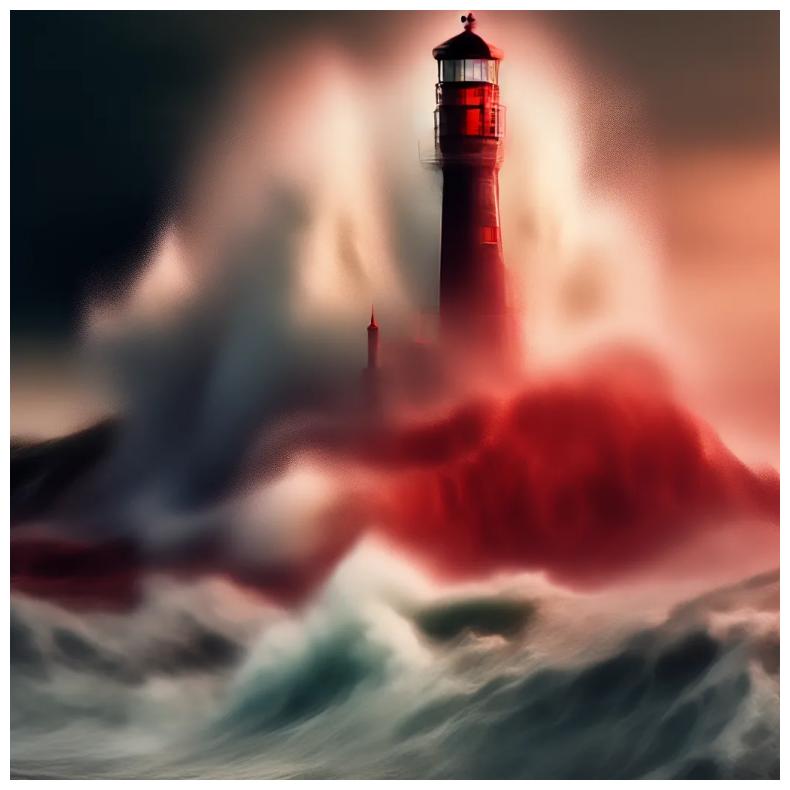}}
   \subfloat{\includegraphics[width=1.0in]{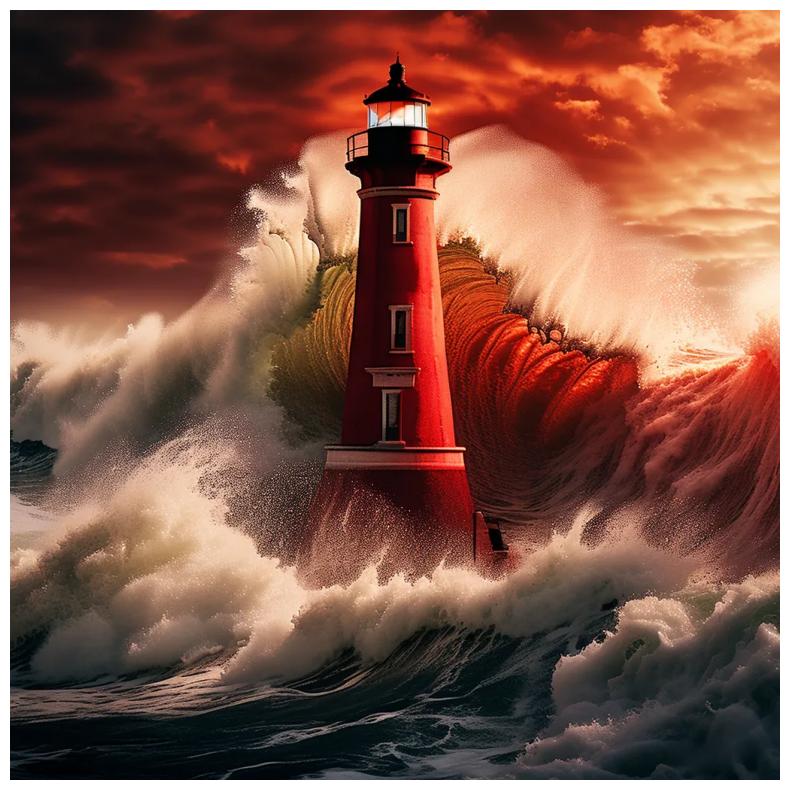}}
  \subfloat{\includegraphics[width=1.0in]{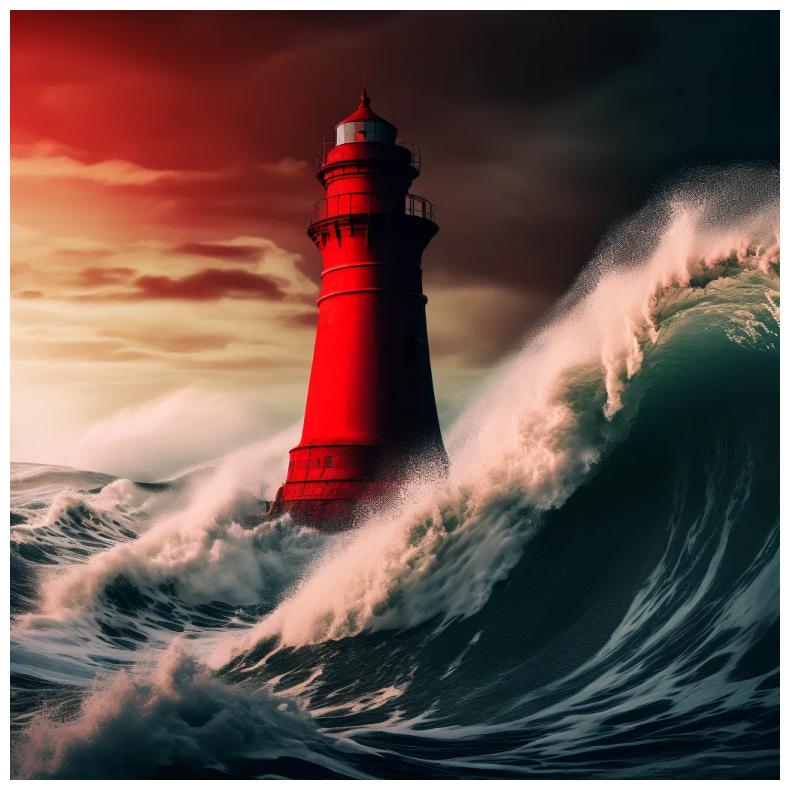}}
  \subfloat{\includegraphics[width=1.0in]{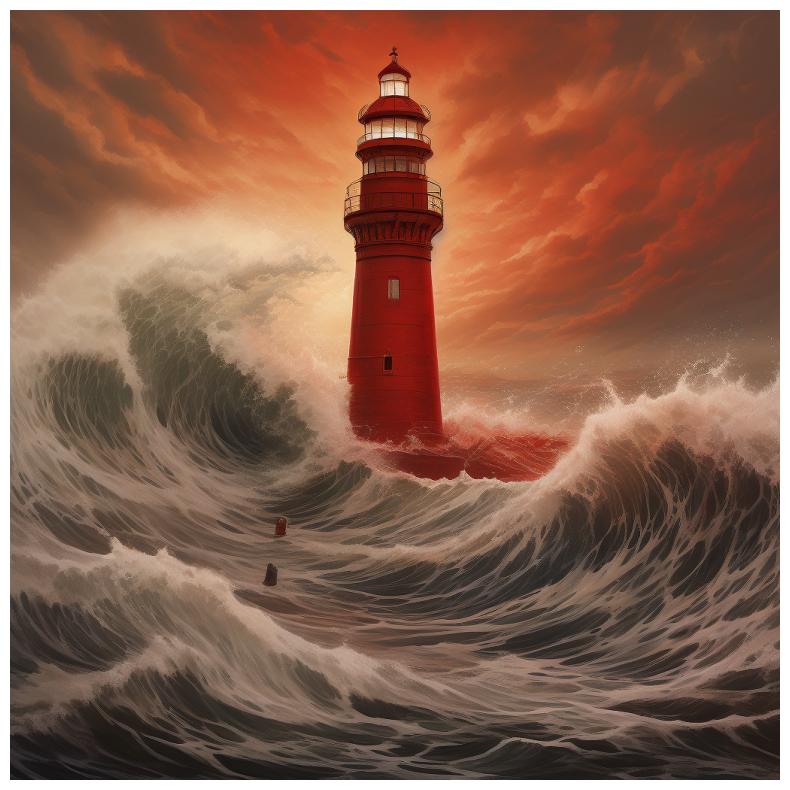}}
  \\\vspace{-1.em}
  \subfloat{\scriptsize \emph{A giant wave shoring on big red lighthouse}}\\\vspace{-1.em}
  \subfloat{\includegraphics[width=1.0in]{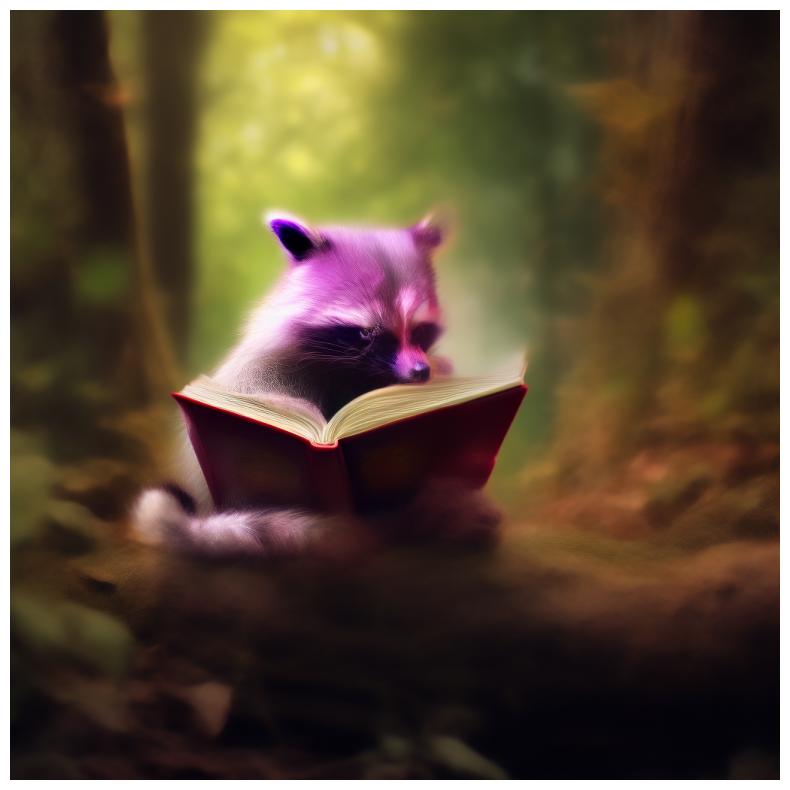}}
   \subfloat{\includegraphics[width=1.0in]{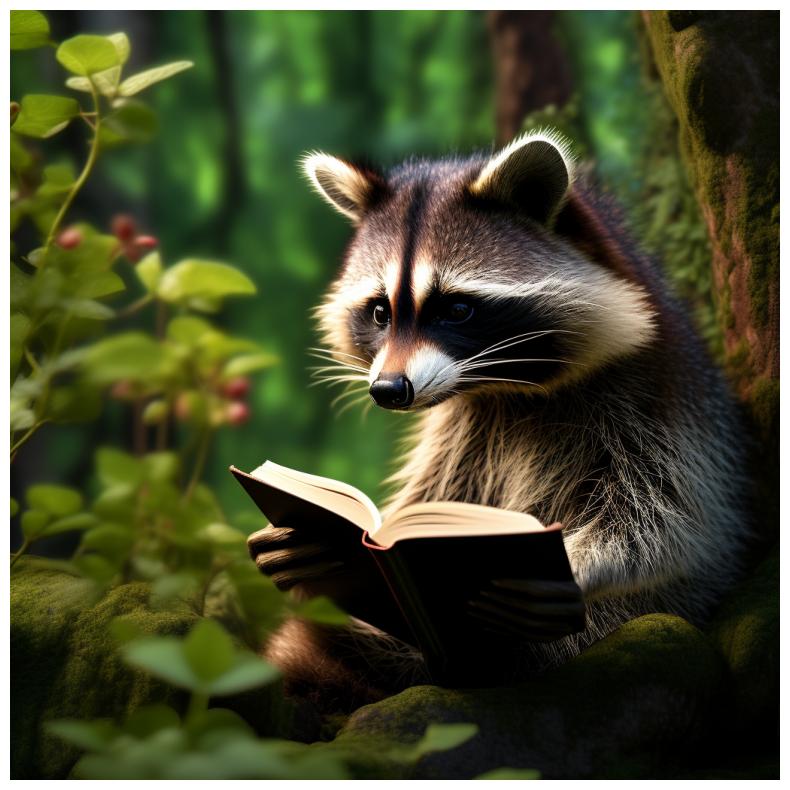}}
  \subfloat{\includegraphics[width=1.0in]{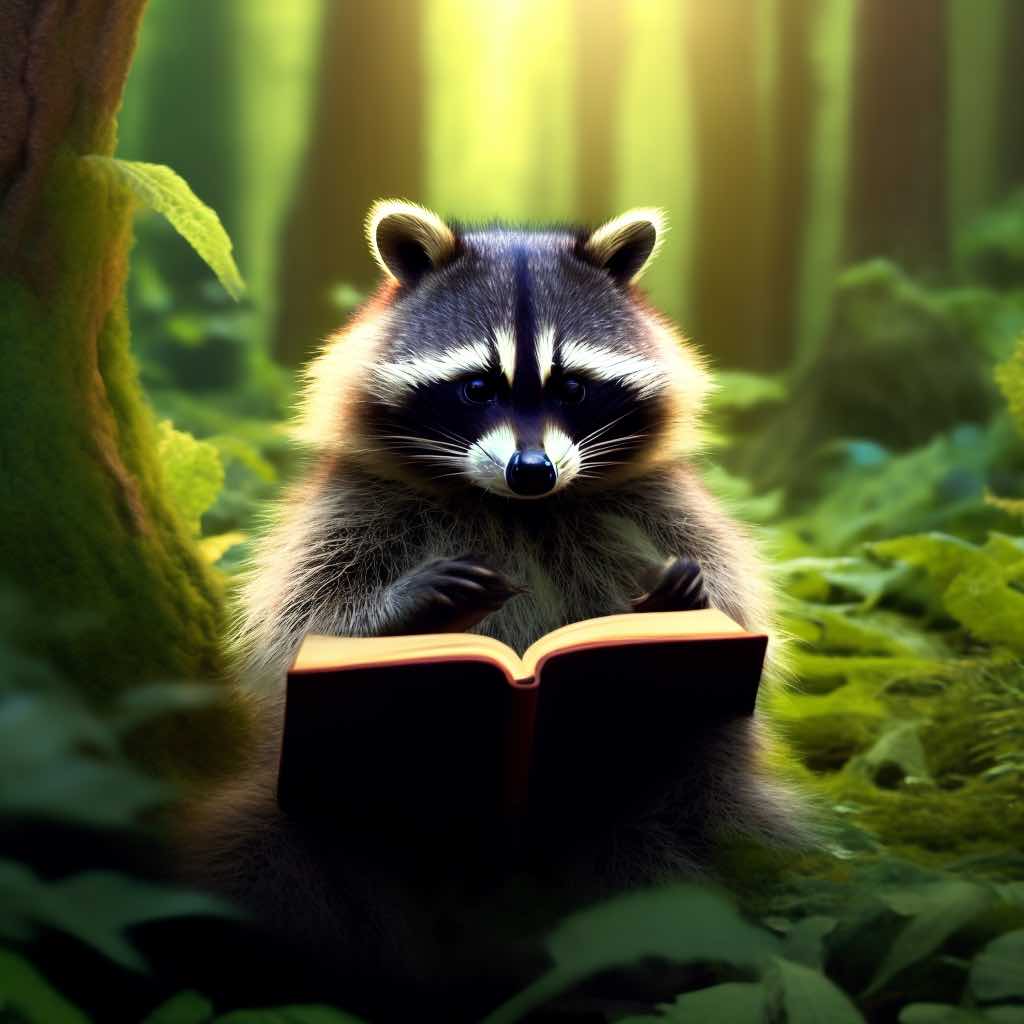}}
  \subfloat{\includegraphics[width=1.0in]{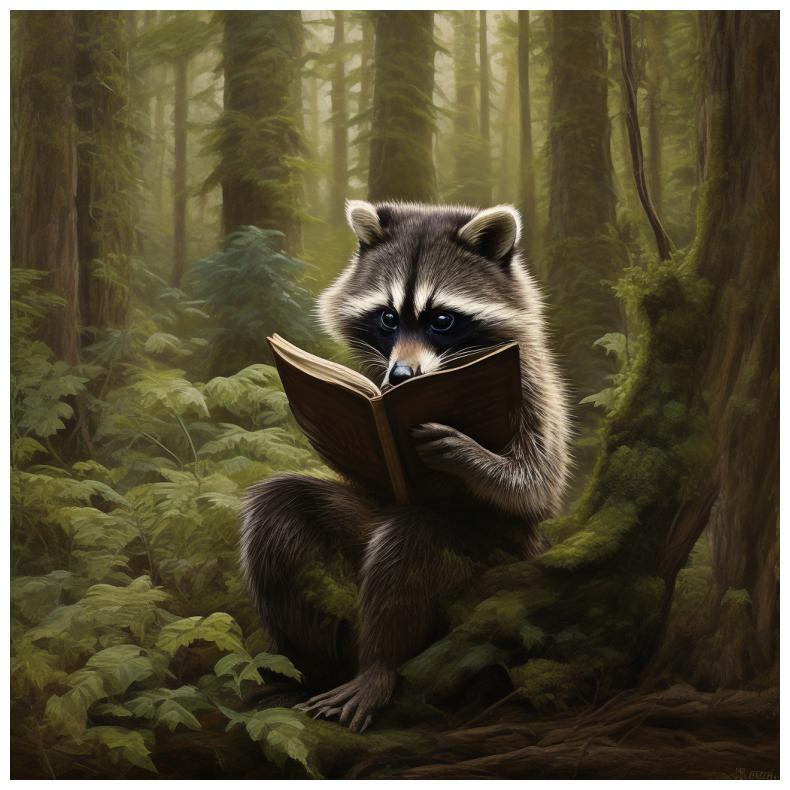}}
  \\\vspace{-1.em}
  \subfloat{\scriptsize \emph{A raccoon reading a book in a lush forest}}\\\vspace{-1.em}
  \subfloat{\includegraphics[width=1.0in]{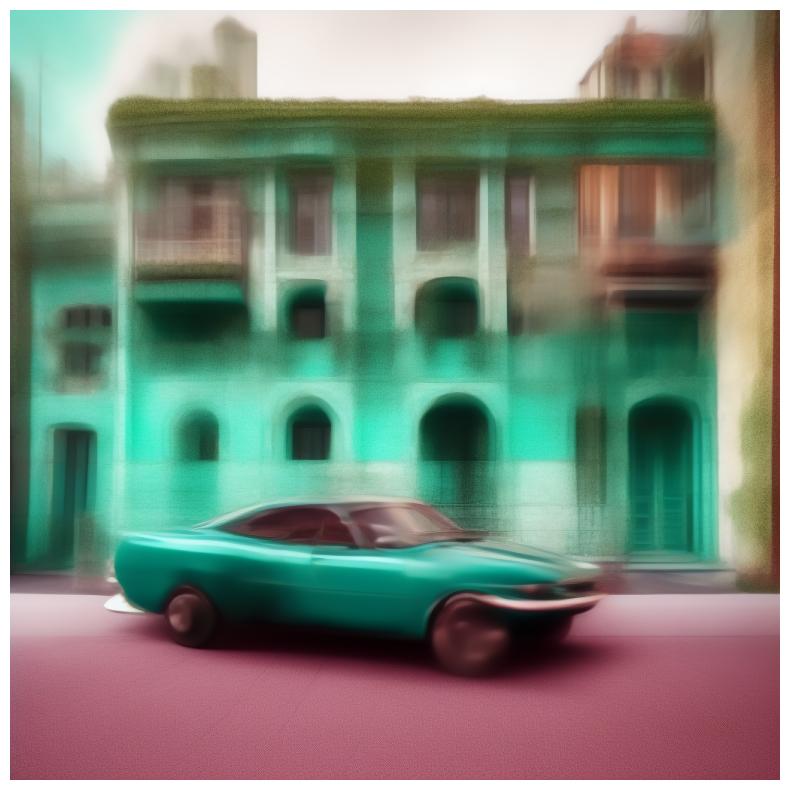}}
   \subfloat{\includegraphics[width=1.0in]{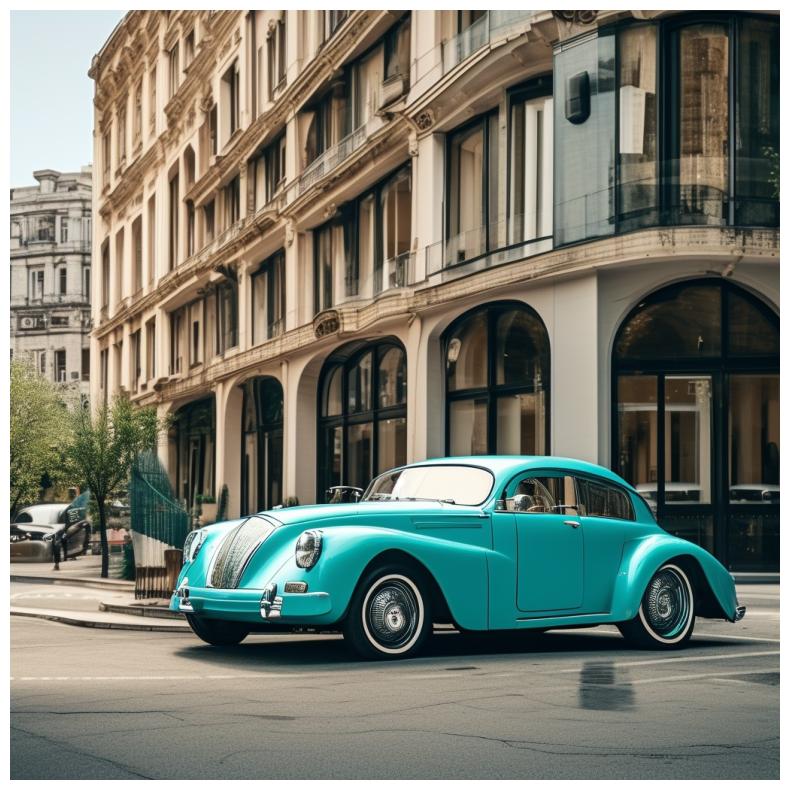}}
  \subfloat{\includegraphics[width=1.0in]{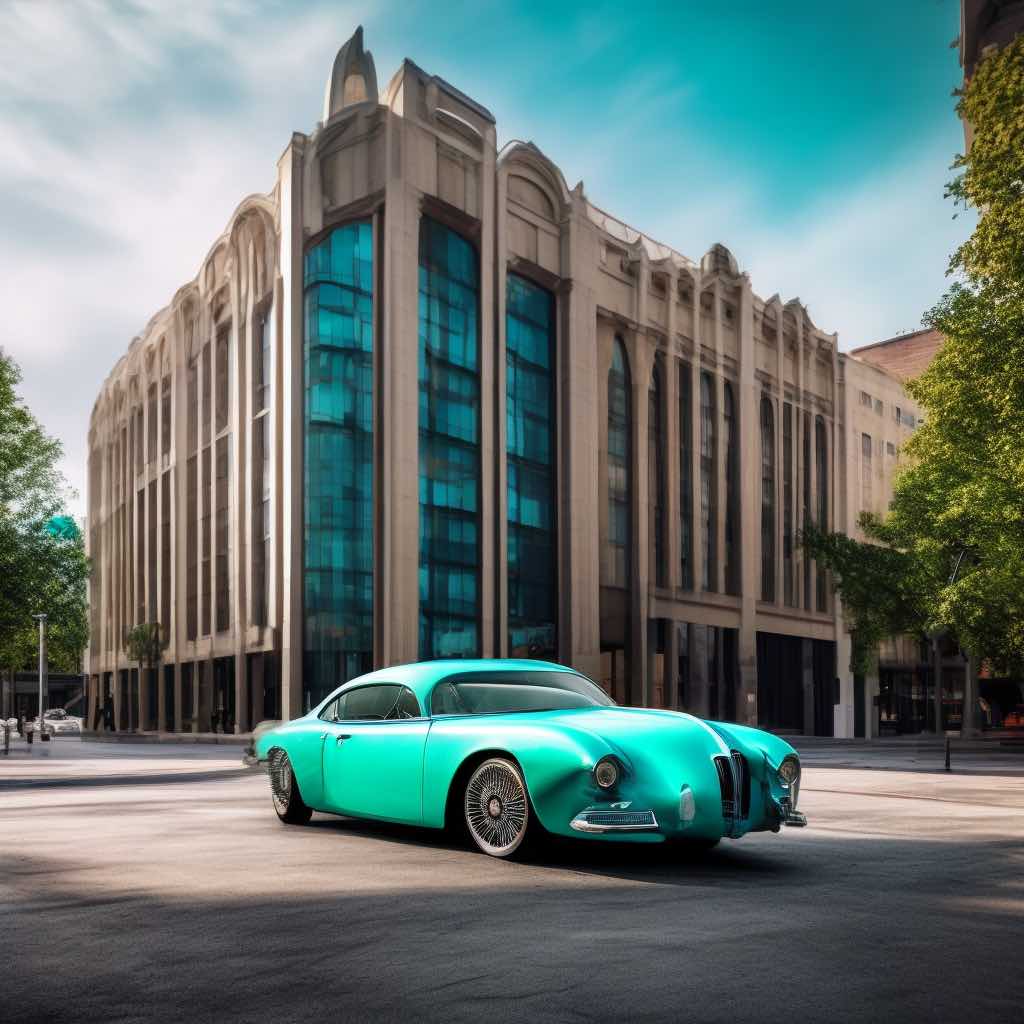}}
  \subfloat{\includegraphics[width=1.0in]{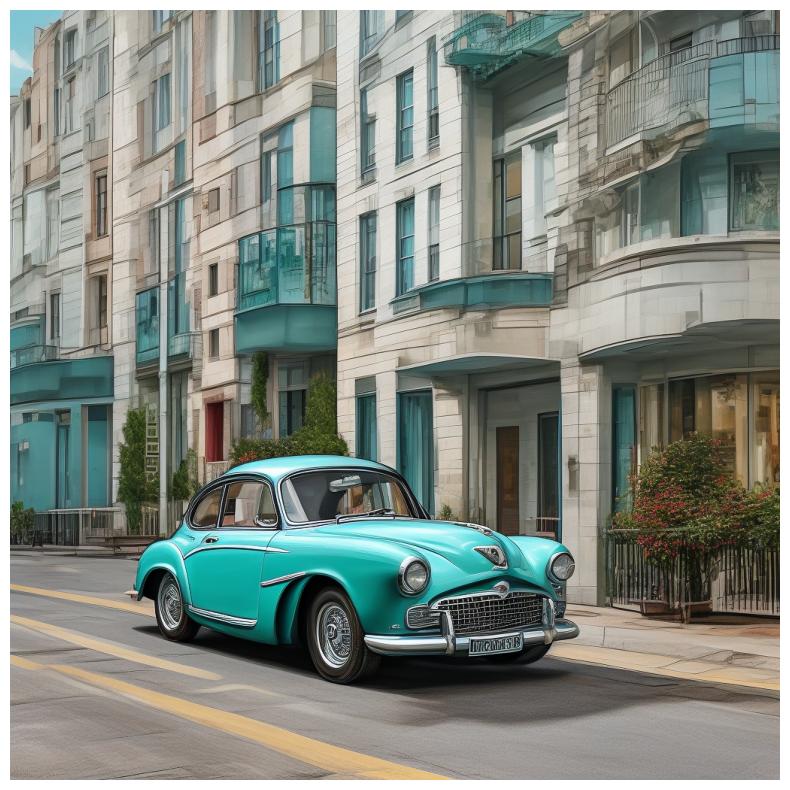}}
  \\\vspace{-1.em}
  \subfloat{\tiny \emph{A classic turquoise car is parked outside a modern building with curved balconies}}\\\vspace{-1.em}
  \subfloat{\includegraphics[width=1.0in]{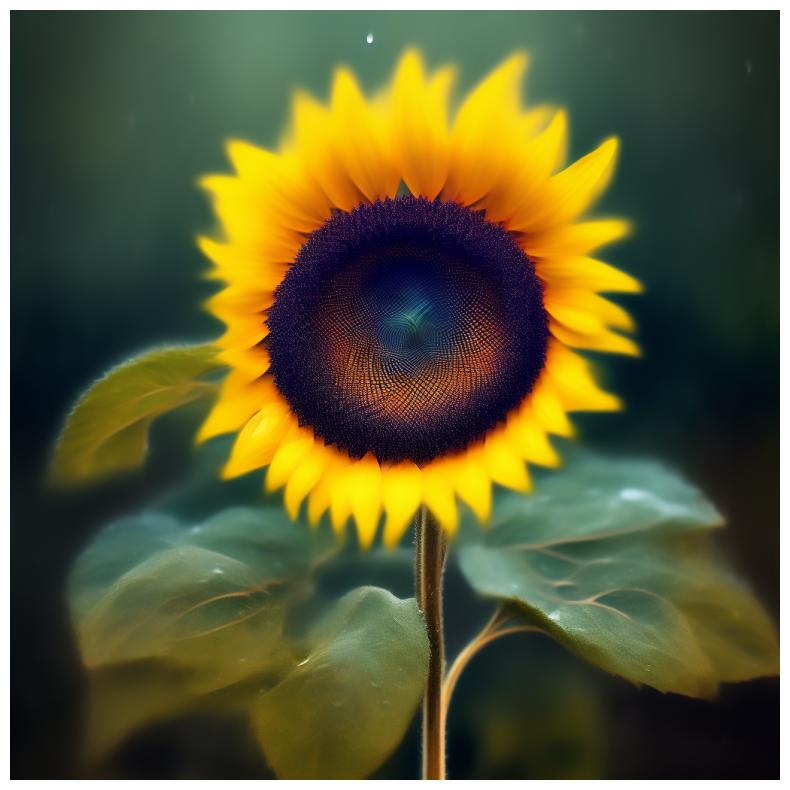}}
   \subfloat{\includegraphics[width=1.0in]{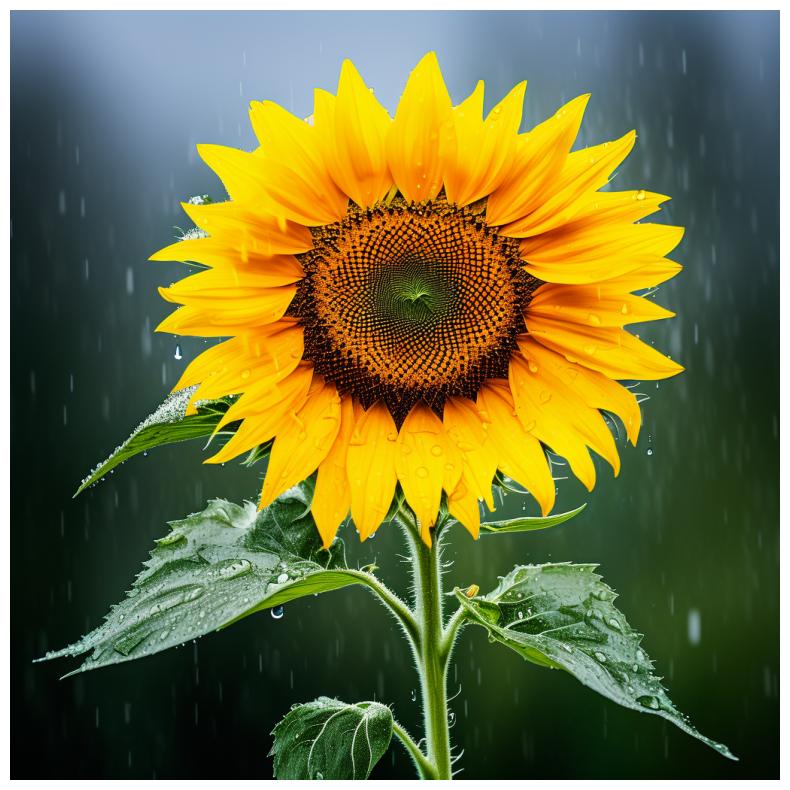}}
  \subfloat{\includegraphics[width=1.0in]{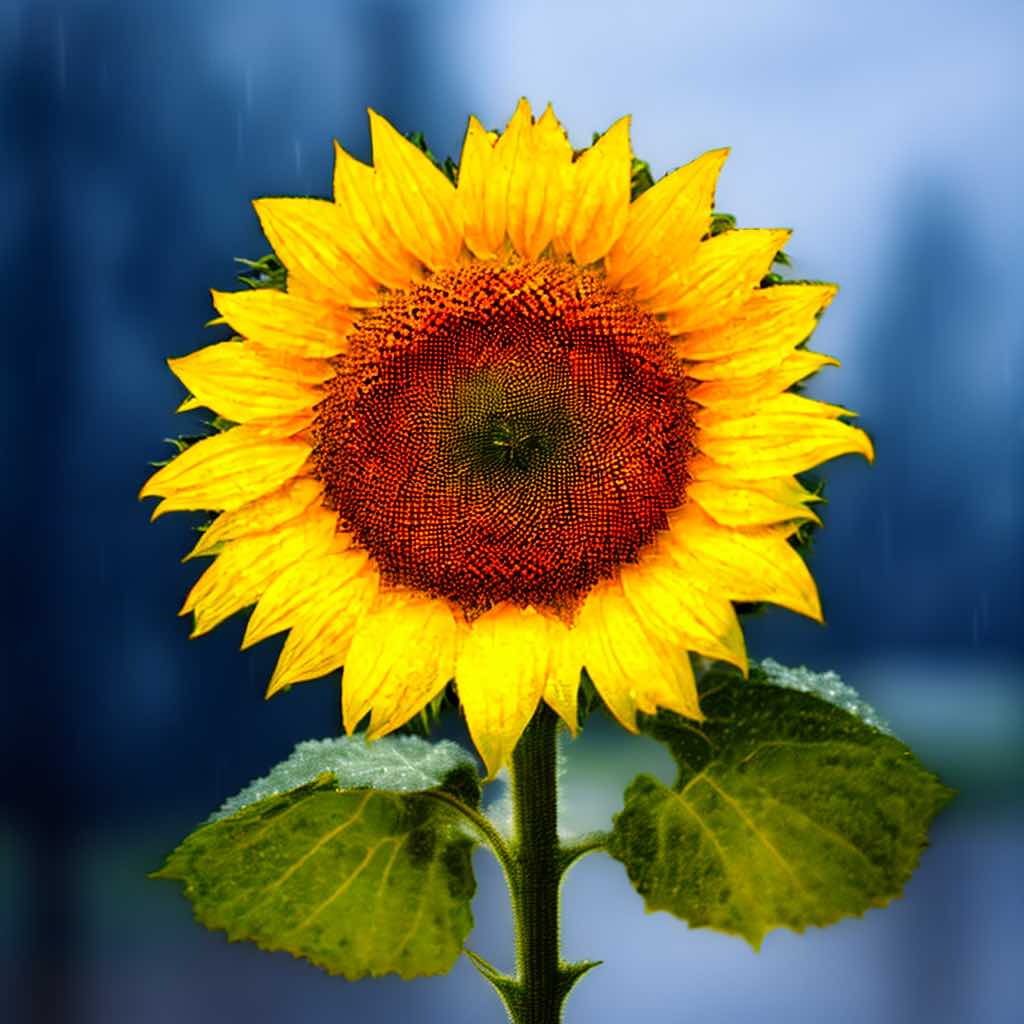}}
  \subfloat{\includegraphics[width=1.0in]{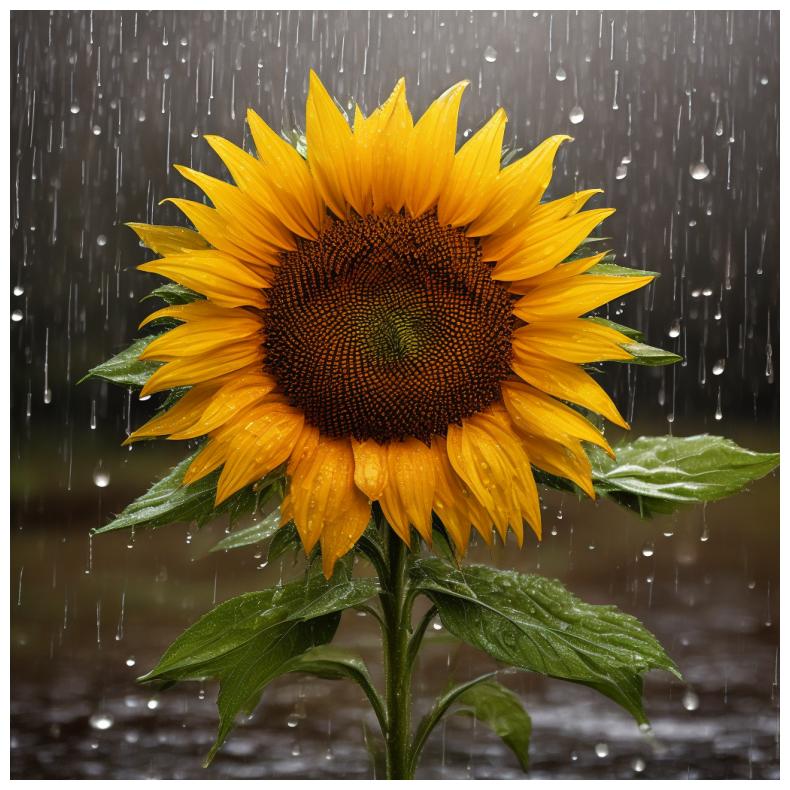}}\\
  \vspace{-1.em}
  \subfloat{\scriptsize \emph{A beautiful sunflower in rainy day}}\\\vspace{-1.em}
  \subfloat{\includegraphics[width=1.0in]{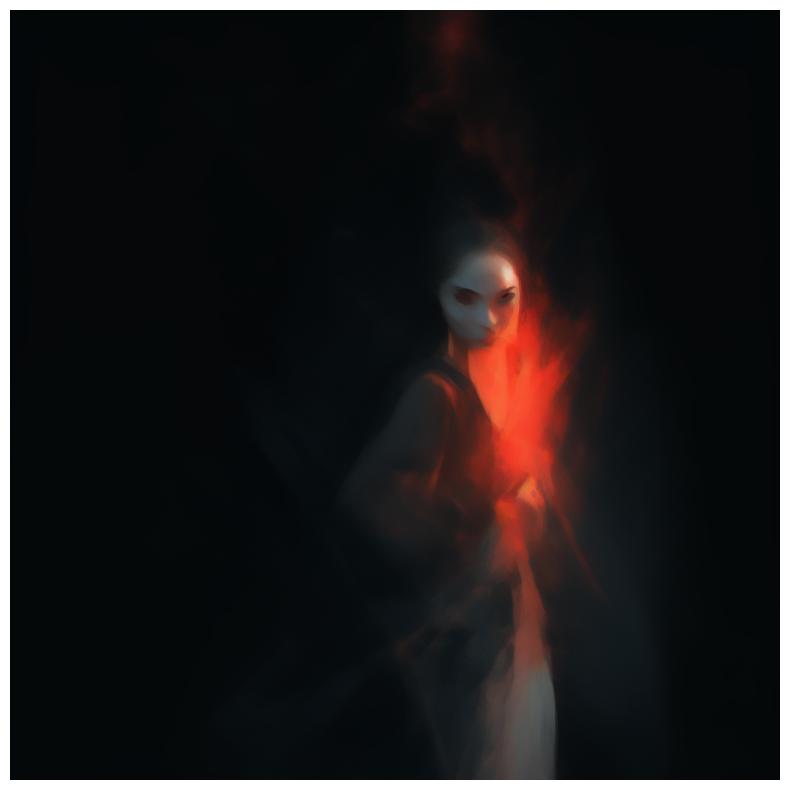}}
   \subfloat{\includegraphics[width=1.0in]{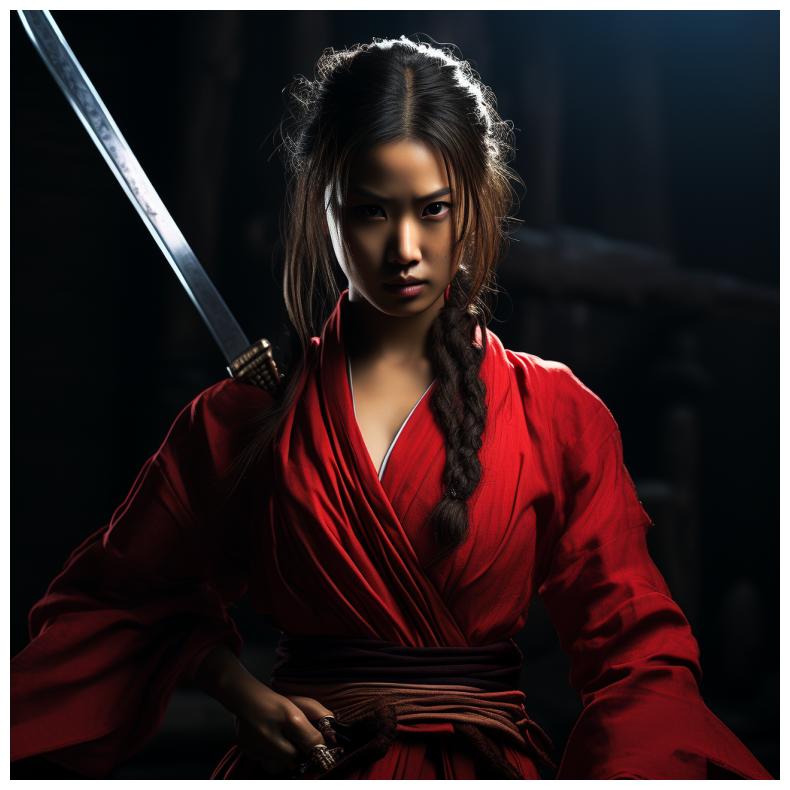}}
  \subfloat{\includegraphics[width=1.0in]{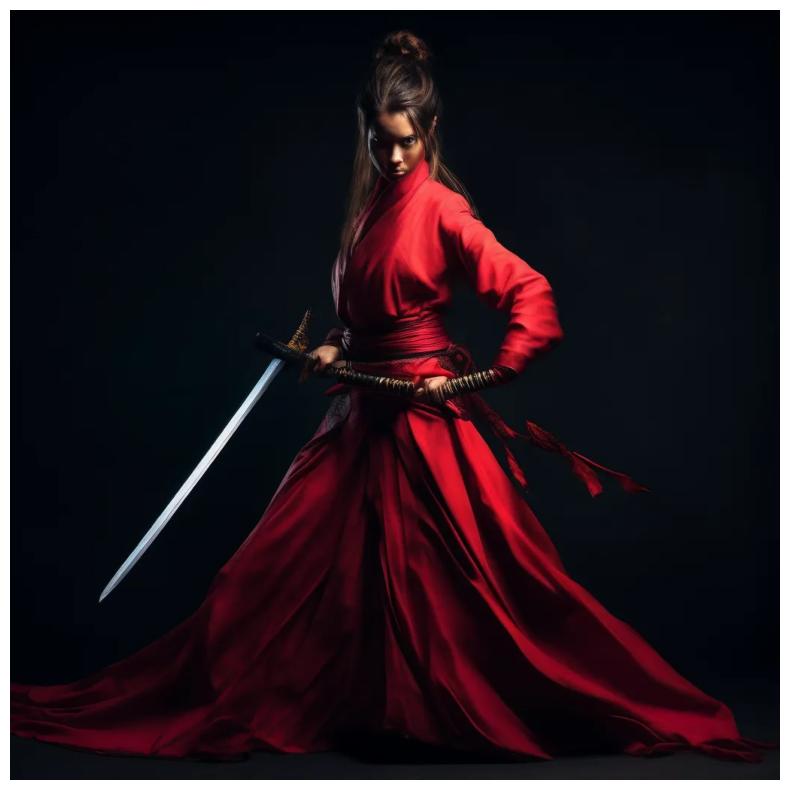}}
  \subfloat{\includegraphics[width=1.0in]{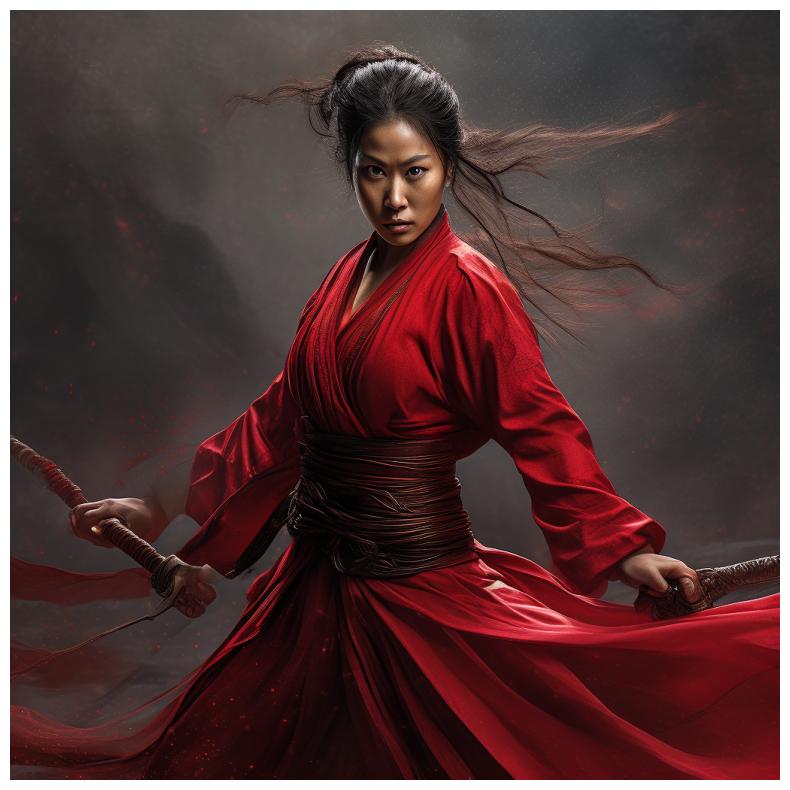}}
  \\\vspace{-1.em}
  \subfloat{\scriptsize \emph{A woman in a red traditional outfit wields a sword, poised in an intense stance against a dark background}}
  \caption{Application of \emph{Flash Diffusion} to a DiT-based Diffusion model, namely Pixart-$\alpha$. The proposed method 4 NFEs generations are compared to the teacher generations using 8 NFEs and 40 NFEs as well as Pixart-LCM \citep{luo2023lcm} with 4 steps. Teacher samples are generated with a guidance scale of 3.}
  \label{fig:app_dit}
  \end{figure*}

\begin{figure*}[p]
  \centering
  \captionsetup[subfigure]{position=below, labelformat = empty}
  \subfloat[\scriptsize \emph{A famous professor giraffe in a classroom standing in front of the blackboard teaching}]{\includegraphics[width=2.5in]{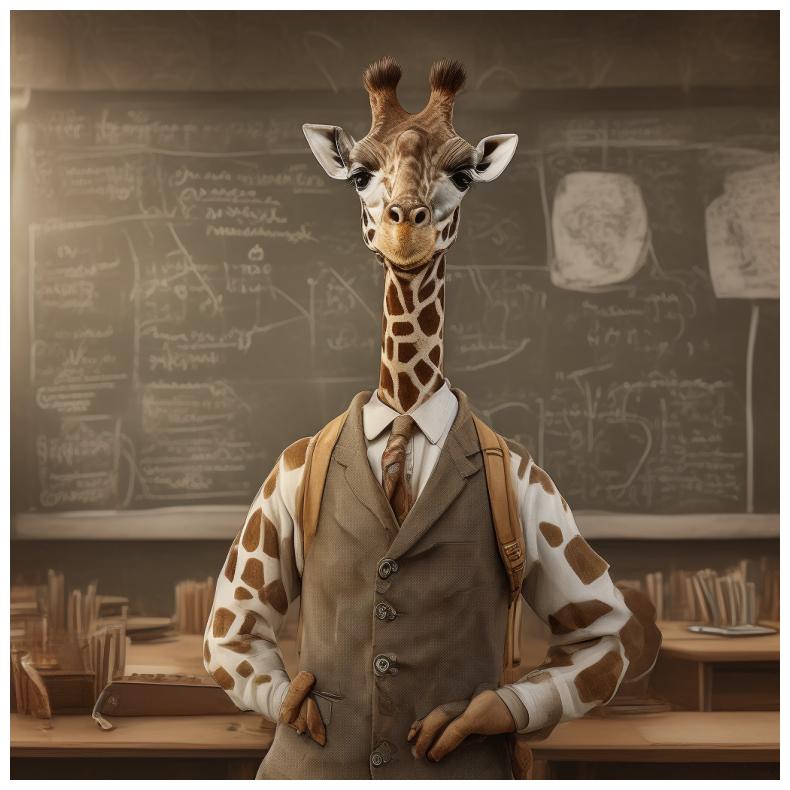}}\hspace{1em}
  \subfloat[\scriptsize \emph{A close up of an old elderly man with green eyes looking straight at the camera}]{\includegraphics[width=2.5in]{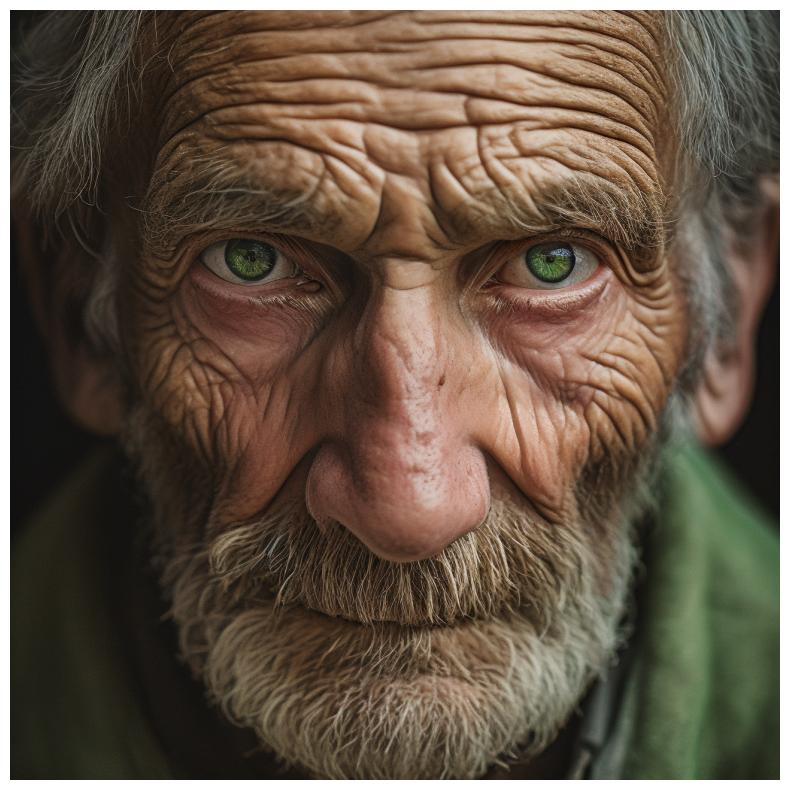}}\hspace{1em}
  \subfloat[\scriptsize \emph{A cute fluffy rabbit pilot walking on a military aircraft carrier, 8k, cinematic}]{\includegraphics[width=2.5in]{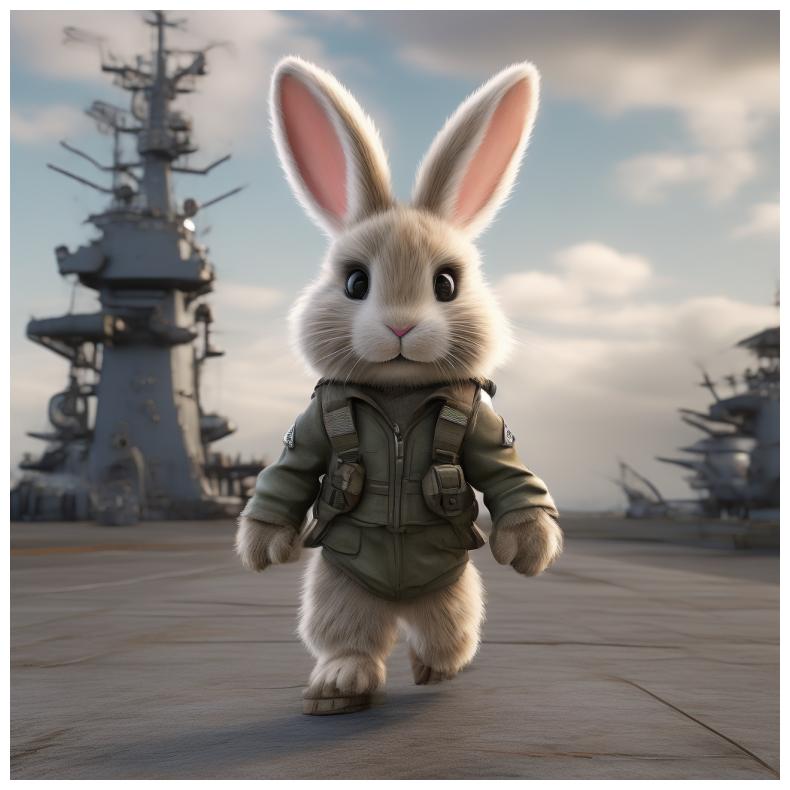}}\hspace{1em}
  \subfloat[\scriptsize \emph{Pirate ship sailing on a sea with the milky way galaxy in the sky and purple glow lights}]{\includegraphics[width=2.5in]{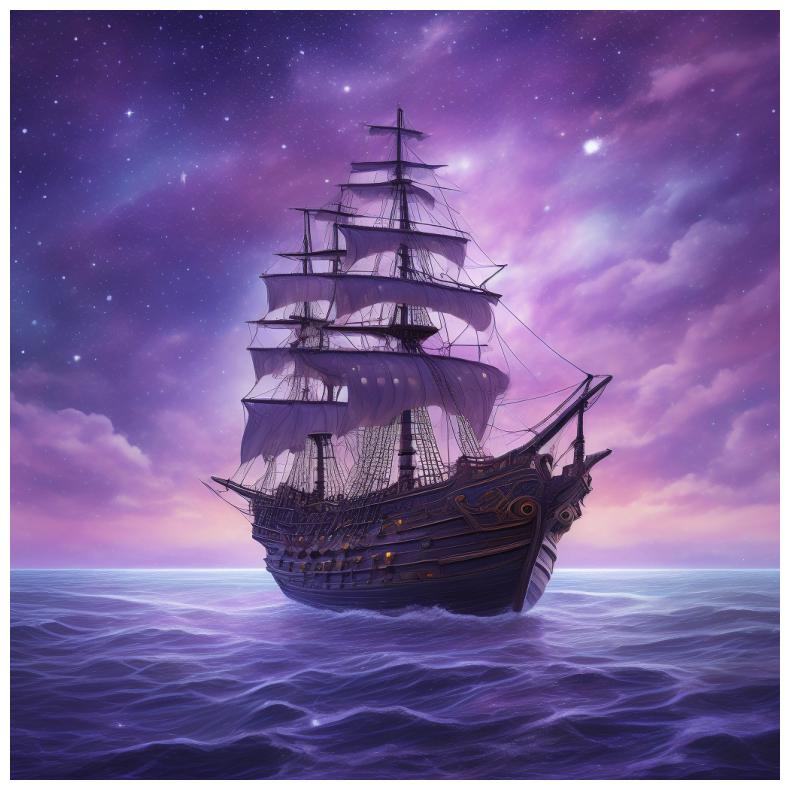}}
  \caption{Application of \emph{Flash Diffusion} to a DiT-based Diffusion model Pixart-$\alpha$.}
  \label{fig:app_dit_1}
\end{figure*}

\begin{figure*}[p]
  \centering
  \captionsetup[subfigure]{position=below, labelformat = empty}
  \subfloat[\scriptsize \emph{A photograph of a woman with headphone coding on a computer, photograph, cinematic, high details, 4k}]{\includegraphics[width=2.5in]{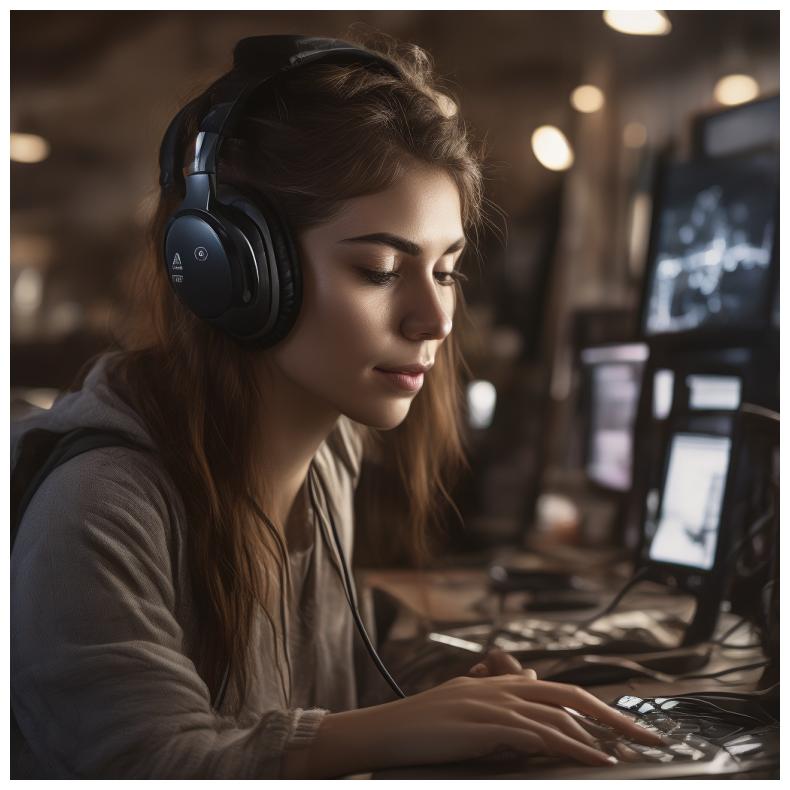}}\hspace{1em}
  \subfloat[\scriptsize \emph{A super realistic kungfu master panda Japanese style}]{\includegraphics[width=2.5in]{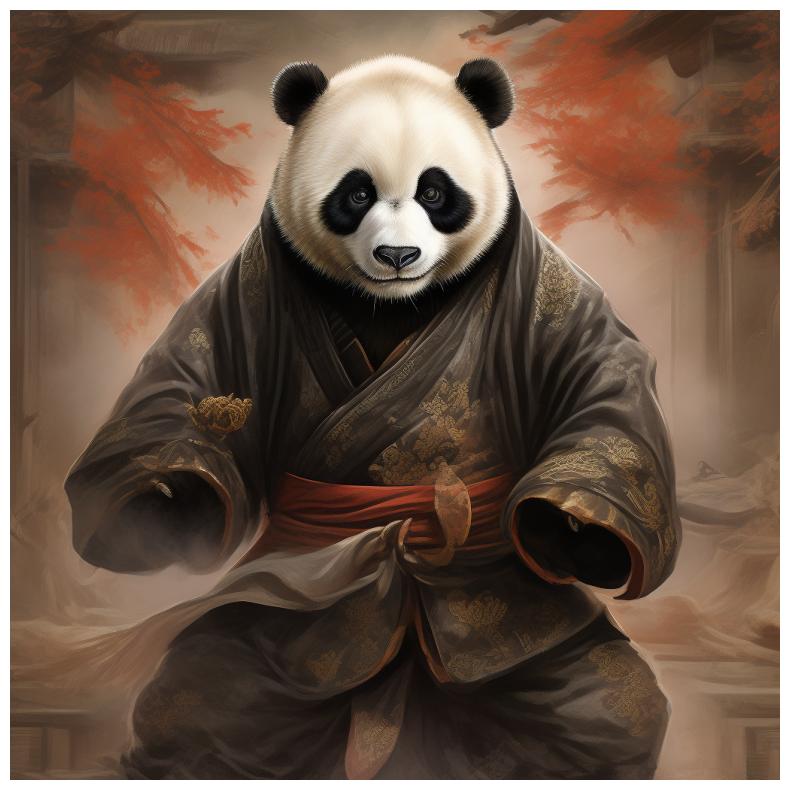}}\hspace{1em}
  \subfloat[\scriptsize \emph{The scene represents a desert composed of red rock resembling planet Mars, there is a cute robot with big eyes feeling alone, It looks straight to the camera looking for friends}]{\includegraphics[width=2.5in]{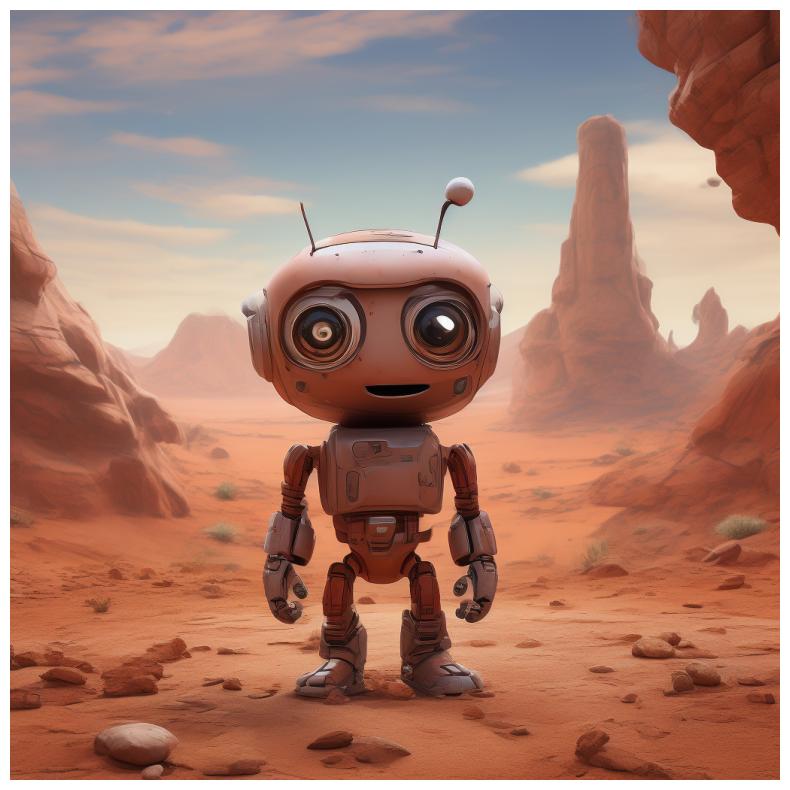}}\hspace{1em}
  \subfloat[\scriptsize \emph{A serving of creamy pasta, adorned with herbs and red pepper flakes, is placed on a white surface, with a striped cloth nearby}]{\includegraphics[width=2.5in]{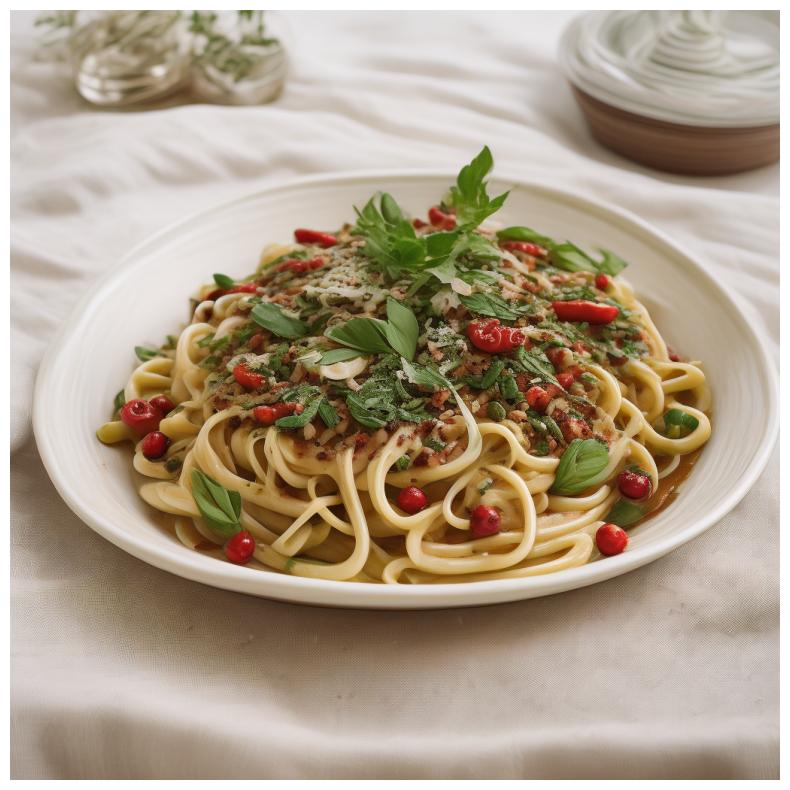}}
  \caption{Application of \emph{Flash Diffusion} to a DiT-based Diffusion model Pixart-$\alpha$.}
  \label{fig:app_dit_2}
\end{figure*}

\begin{figure*}[p]
  \centering
  \captionsetup[subfigure]{position=below, labelformat = empty}
  \subfloat{\includegraphics[width=2.5in]{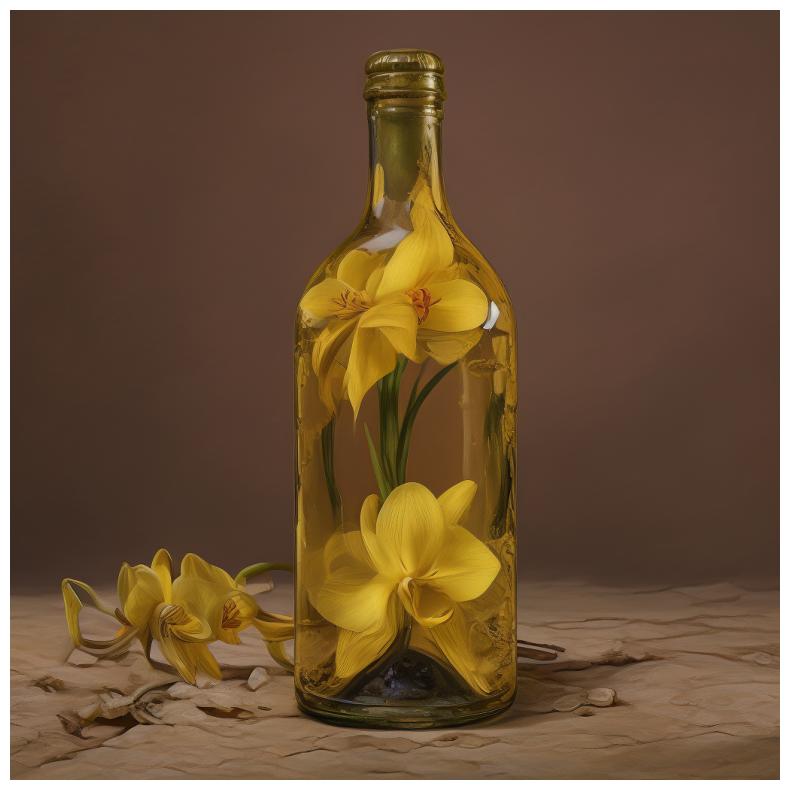}}
  \subfloat{\includegraphics[width=2.5in]{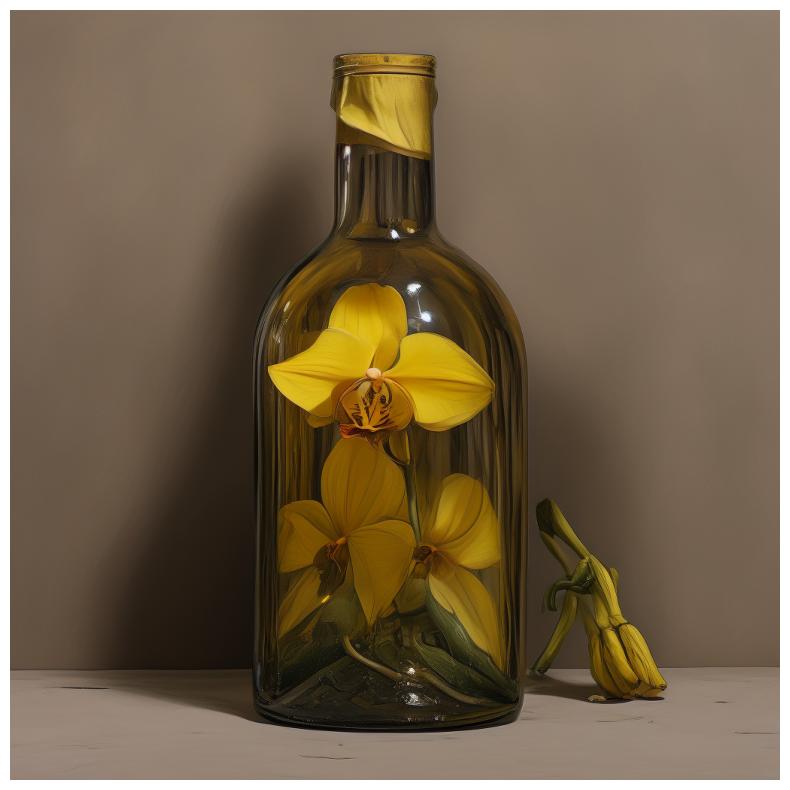}}\vspace{-1.1em}
  \subfloat{\includegraphics[width=2.5in]{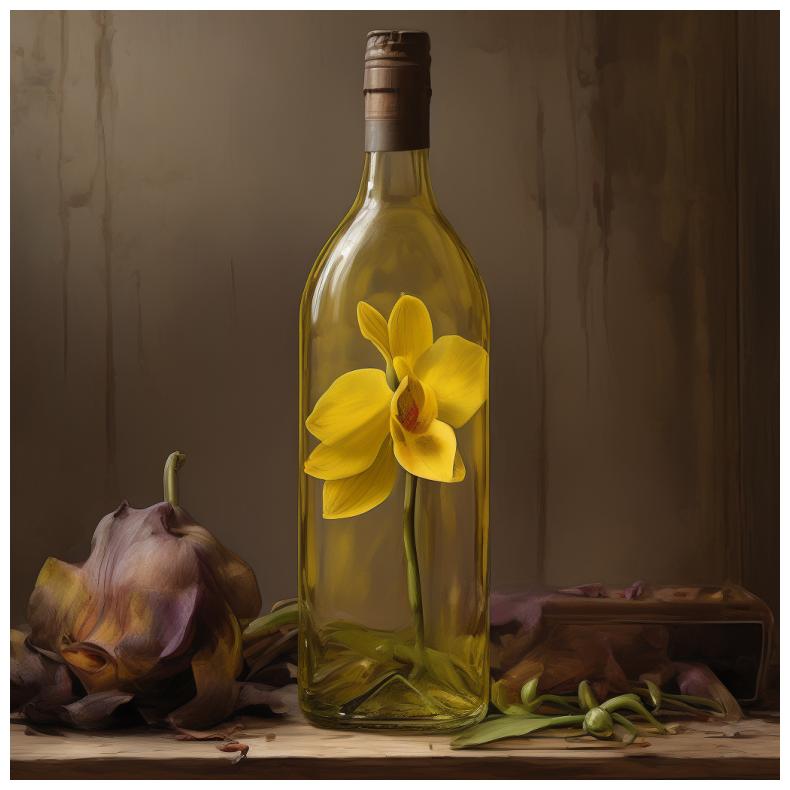}}
      \subfloat{\includegraphics[width=2.5in]{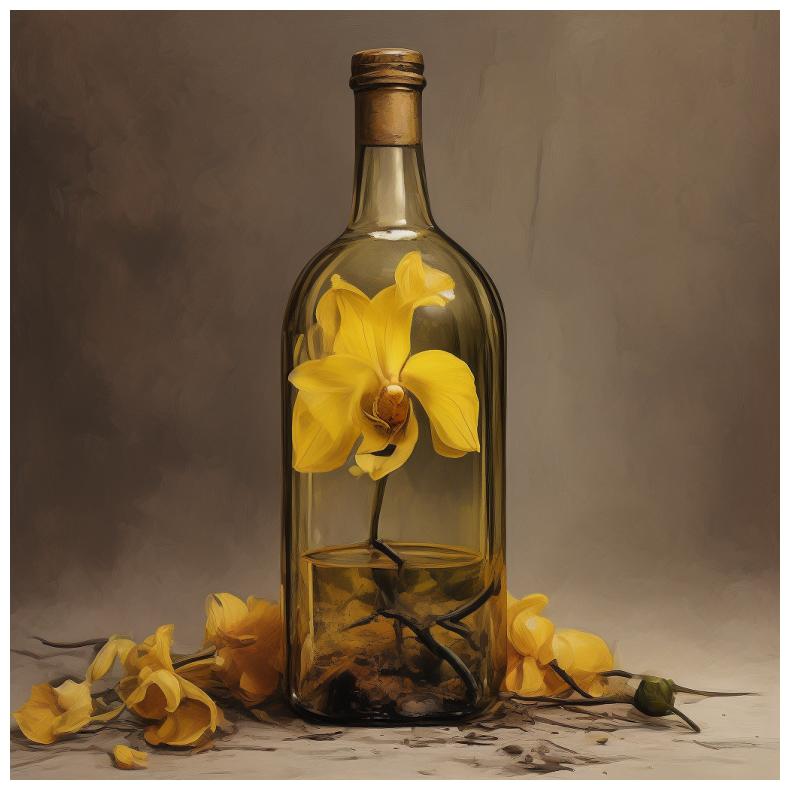}}
  \caption{Generation variation for Flash Pixart with the prompt \emph{A yellow orchid trapped inside an empty bottle of wine}.}
  \label{fig:app_dit_var_1}
\end{figure*}

\begin{figure*}[p]
  \centering
  \captionsetup[subfigure]{position=below, labelformat = empty}
  \subfloat{\includegraphics[width=2.5in]{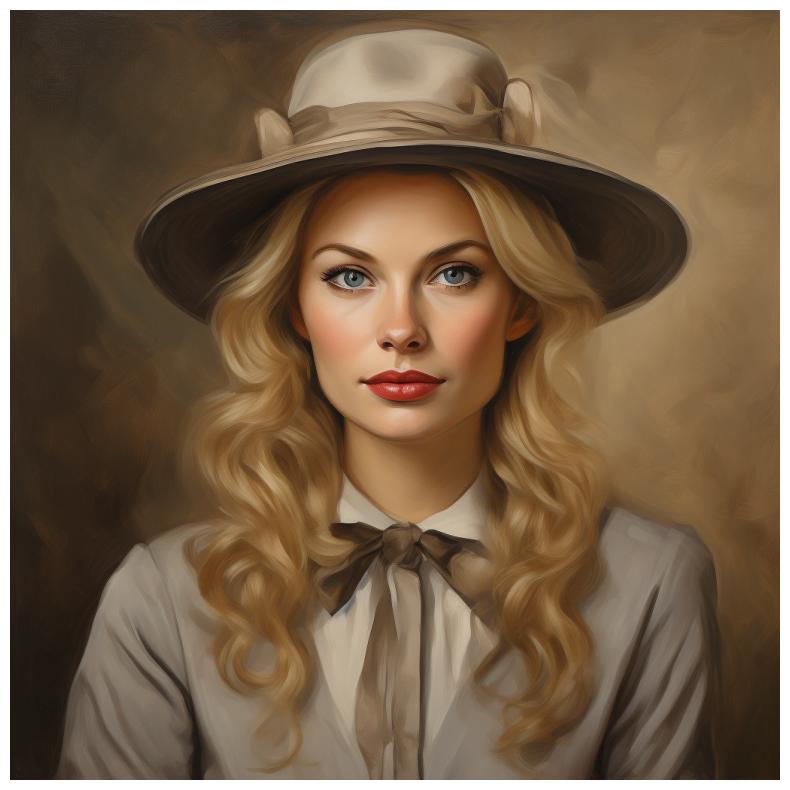}}
  \subfloat{\includegraphics[width=2.5in]{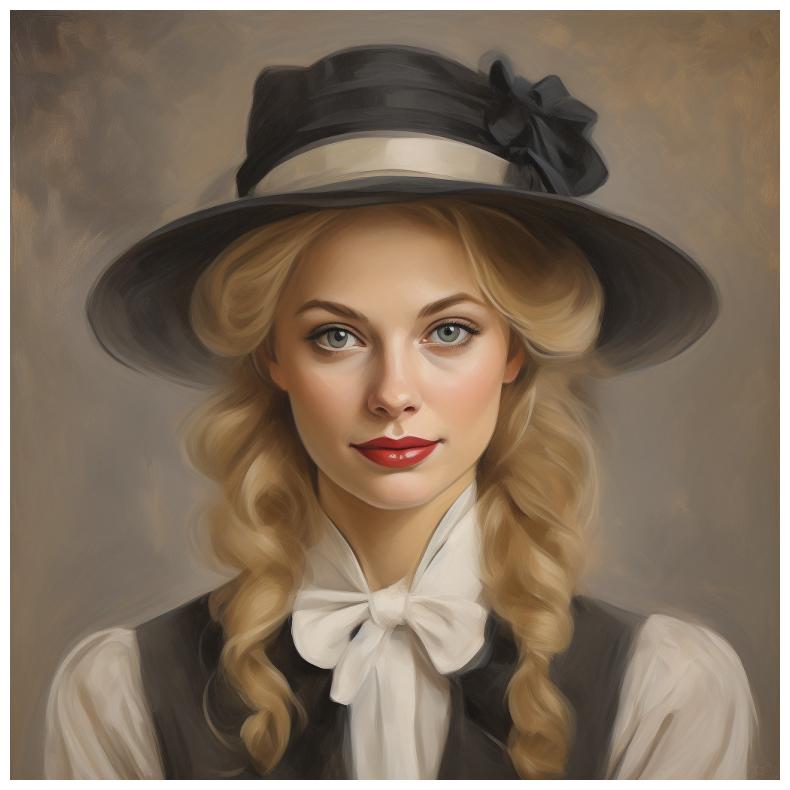}}\vspace{-1.1em}
  \subfloat{\includegraphics[width=2.5in]{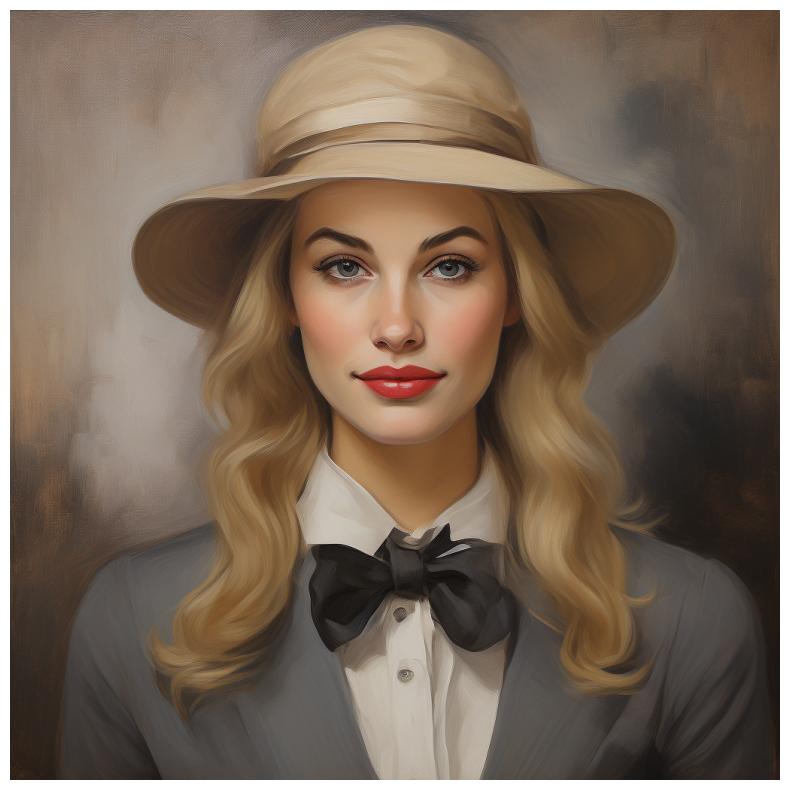}}
  \subfloat{\includegraphics[width=2.5in]{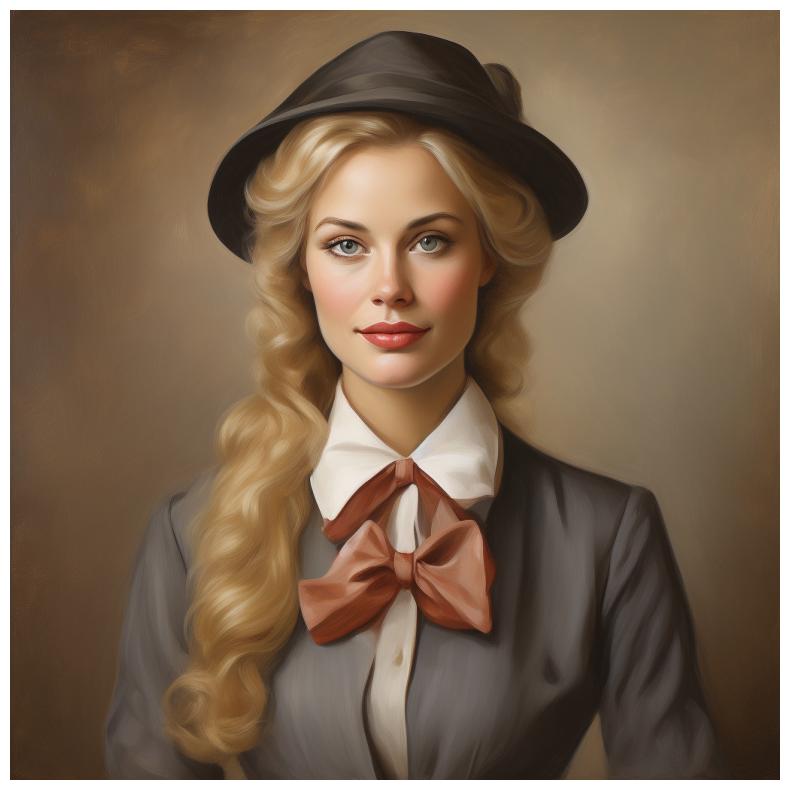}}
  \caption{Generation variation for Flash Pixart with the prompt \emph{An oil painting portrait of an elegant blond woman with a bowtie and hat}.}
  \label{fig:app_dit_var_2}
\end{figure*}

In this section, we provide additional samples using the trained student model using a DiT architecture. In Fig.~\ref{fig:app_dit}, we provide a more complete qualitative comparison with respect to LCM and the teacher model while in Fig.~\ref{fig:app_dit_1} and \ref{fig:app_dit_2}, we show additional samples using the proposed method. In Fig.~\ref{fig:app_dit_var_1} and \ref{fig:app_dit_var_2}, we also show the generation variation with respect to two different prompts: \emph{A yellow orchid trapped inside an empty bottle of wine} and \emph{An oil painting portrait of an elegant blond woman with a bowtie and hat}. The model appears to be able to generate various samples even with a fixed prompt. 

\subsection{Flash Inpainting}\label{sec:app_flash_inpaint}
\begin{figure*}[t]
  \centering
  \captionsetup[subfigure]{position=above, labelformat = empty}
  \subfloat[\scriptsize Original]{\includegraphics[width=1.1in]{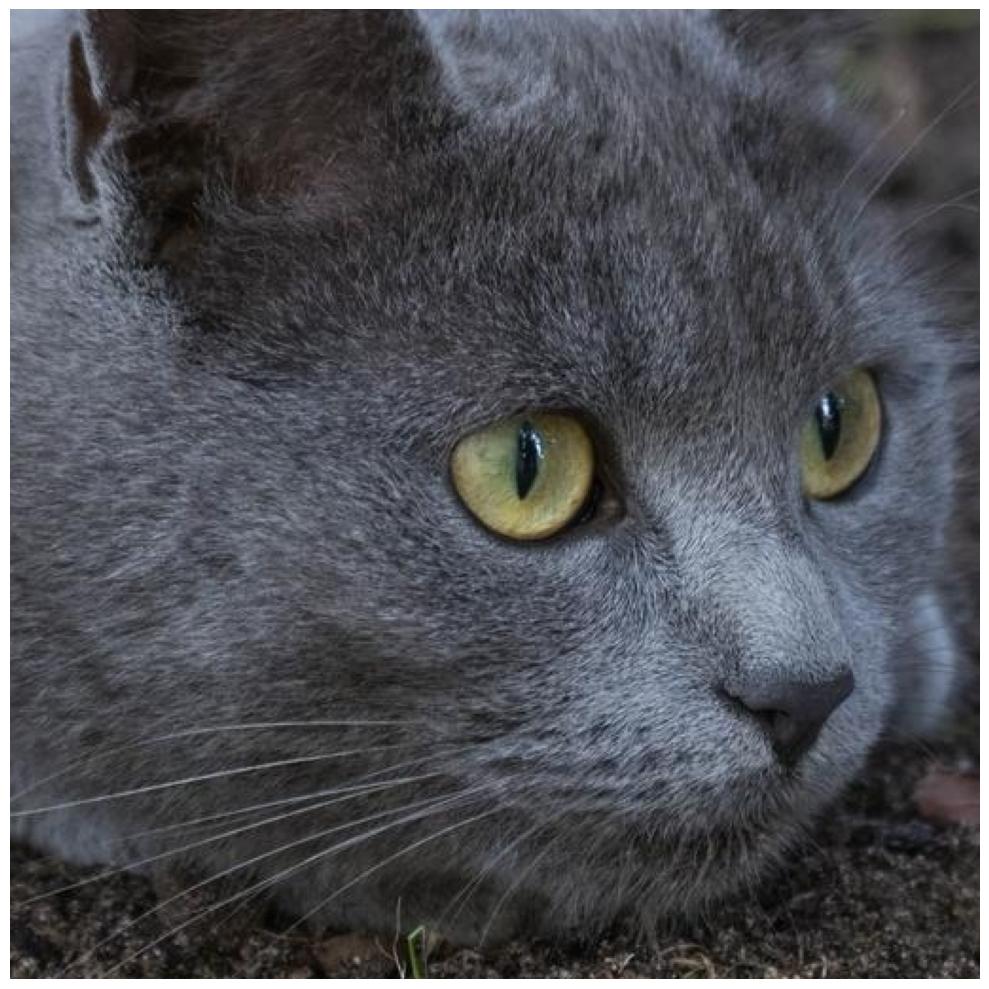}}
    \subfloat[\scriptsize Masked Image]{\includegraphics[width=1.1in]{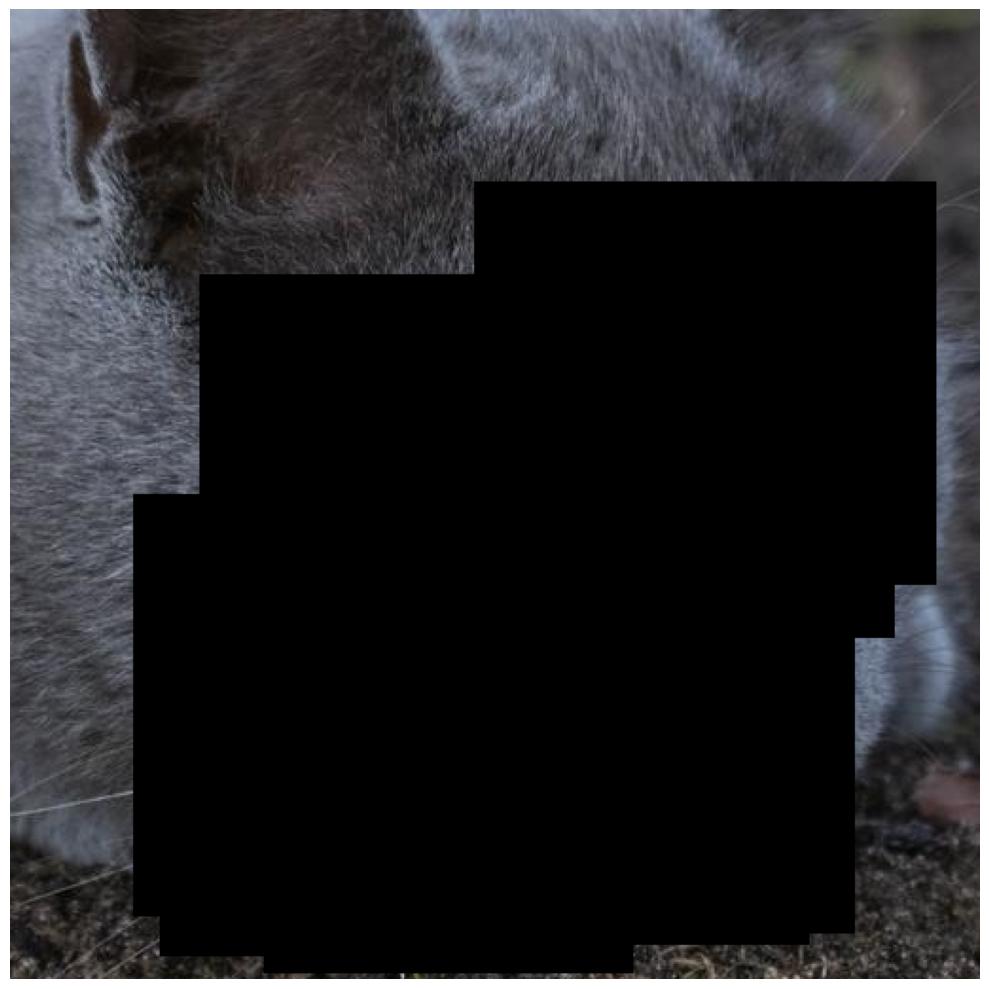}}
    \subfloat[\scriptsize Teacher (8 NFEs)]{\includegraphics[width=1.1in]{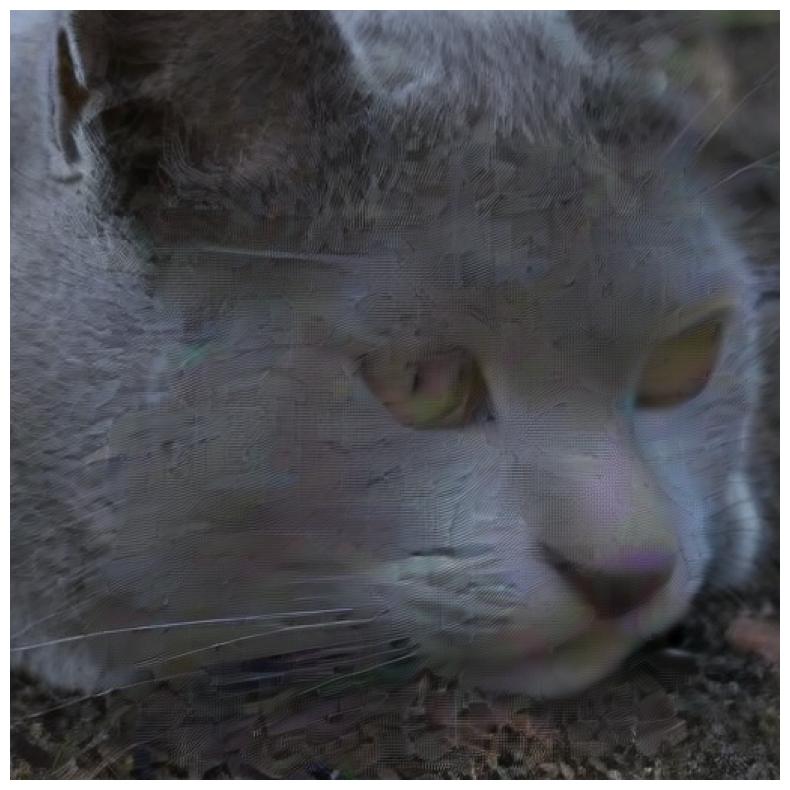}}
    \subfloat[\scriptsize Teacher (40 NFEs)]{\includegraphics[width=1.1in]{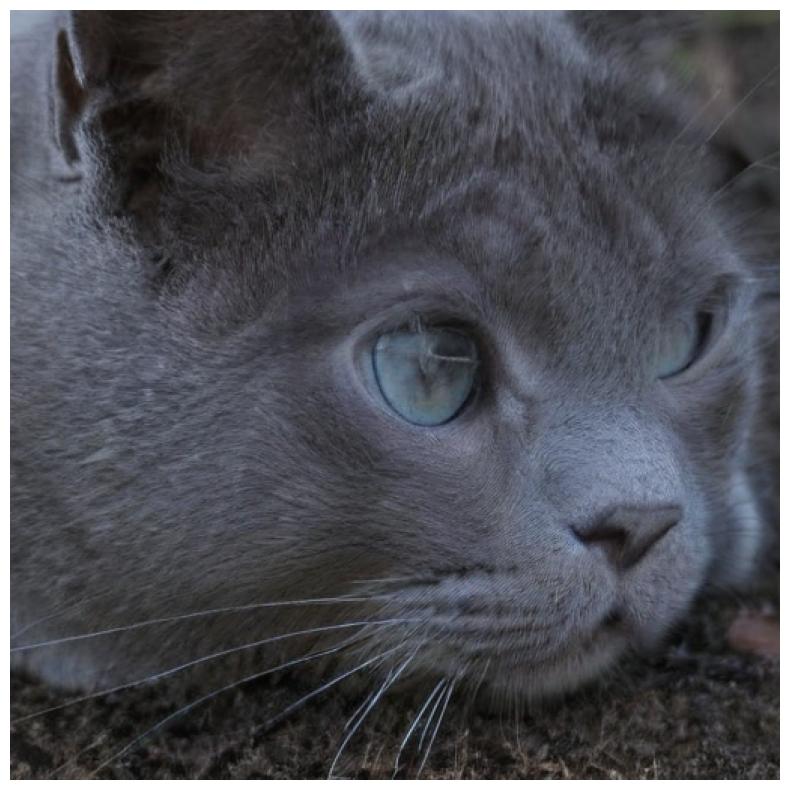}}
    \subfloat[\scriptsize Ours (4 NFEs)]{\includegraphics[width=1.1in]{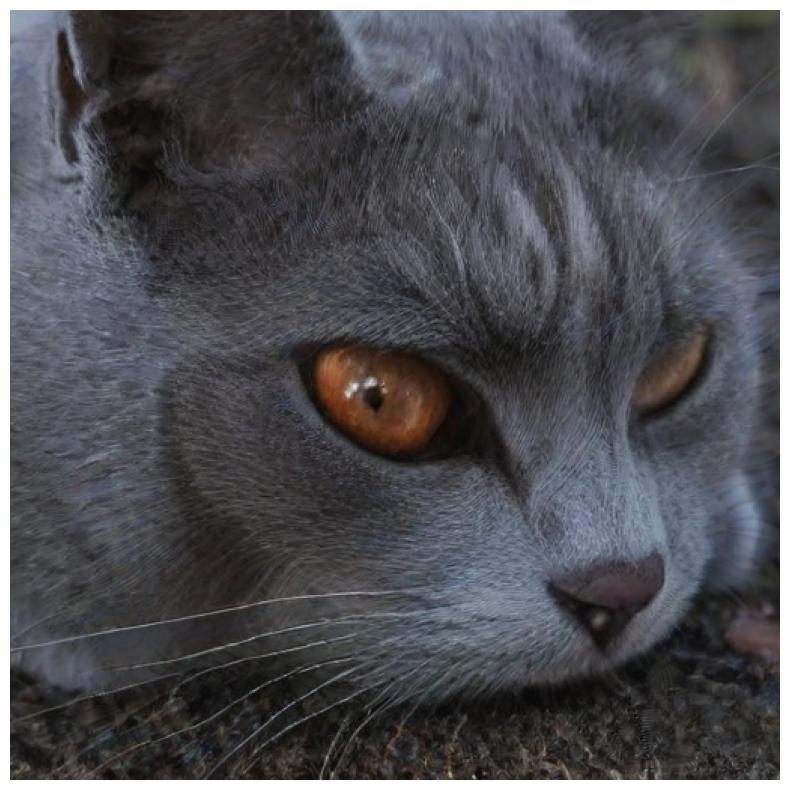}}\vspace{-1.2em}
    \subfloat{\includegraphics[width=1.1in]{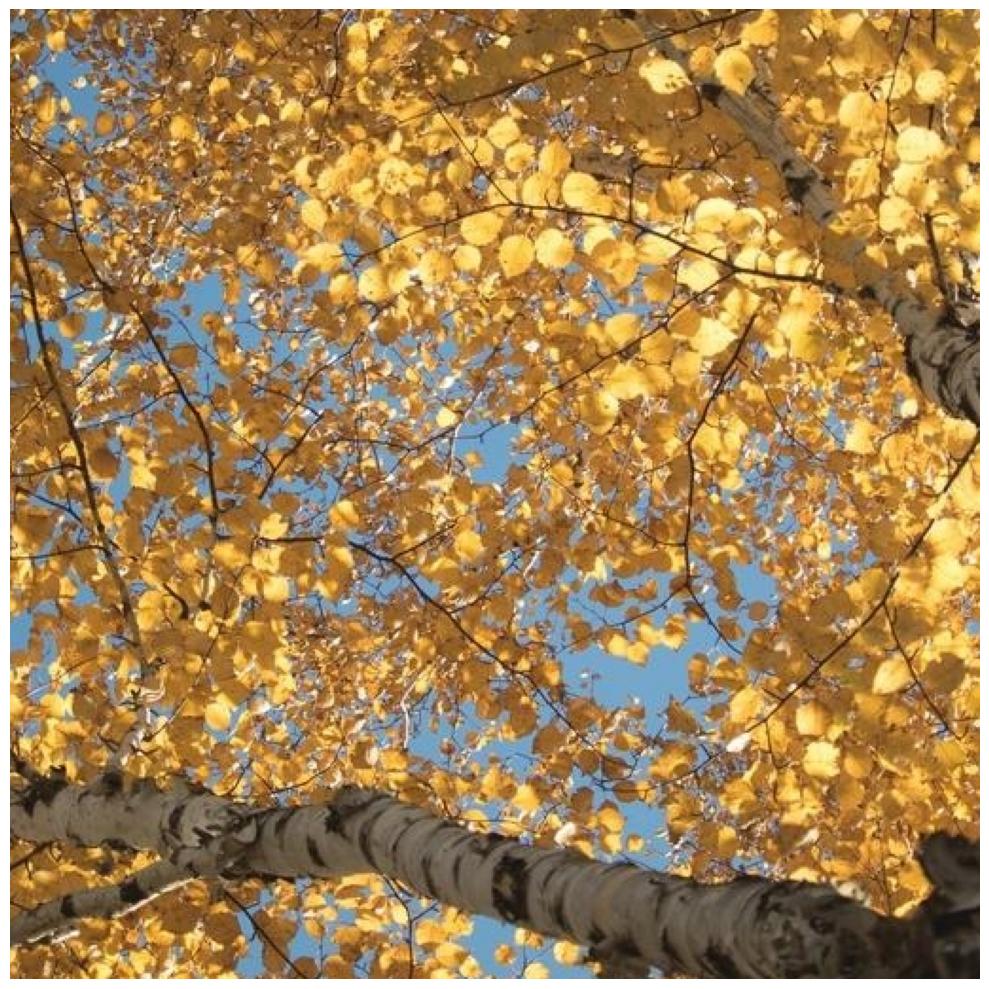}}
  \subfloat{\includegraphics[width=1.1in]{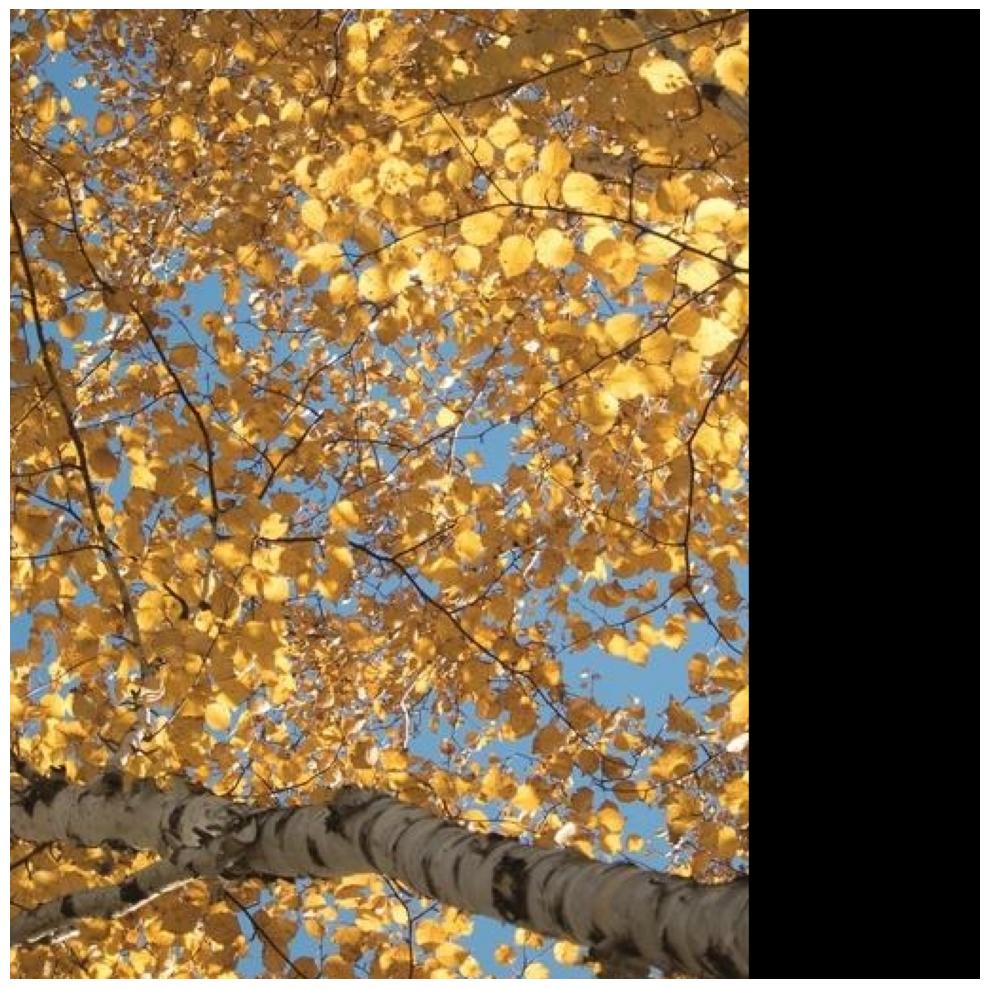}}
  \subfloat{\includegraphics[width=1.1in]{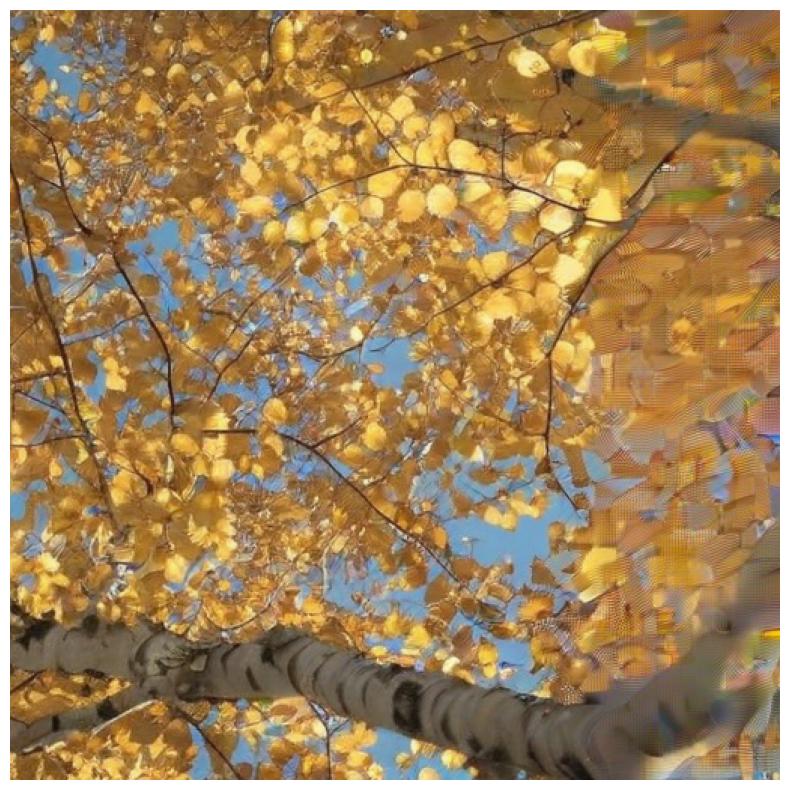}}
  \subfloat{\includegraphics[width=1.1in]{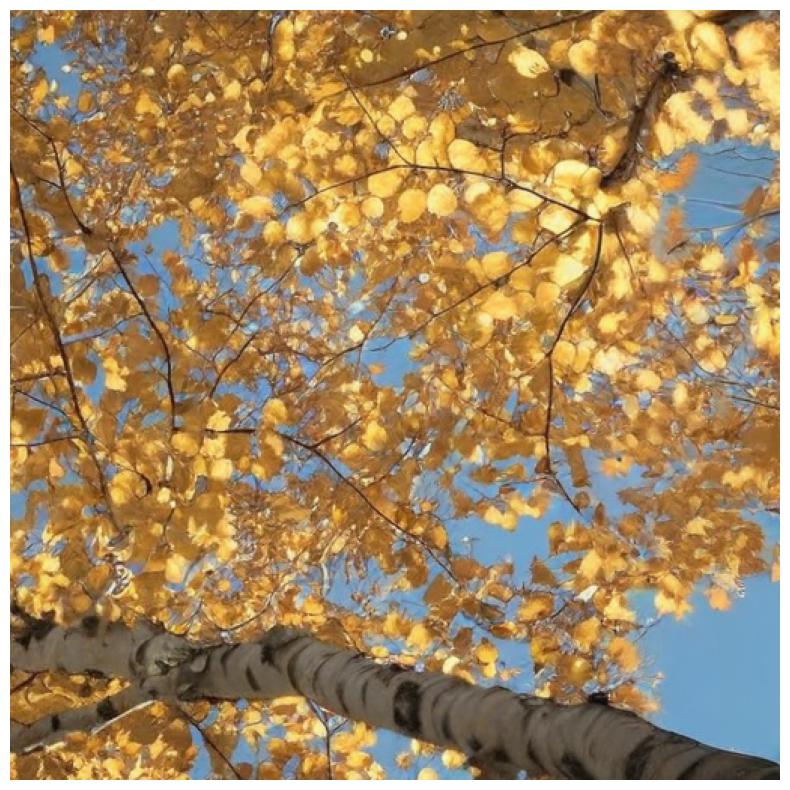}}
  \subfloat{\includegraphics[width=1.1in]{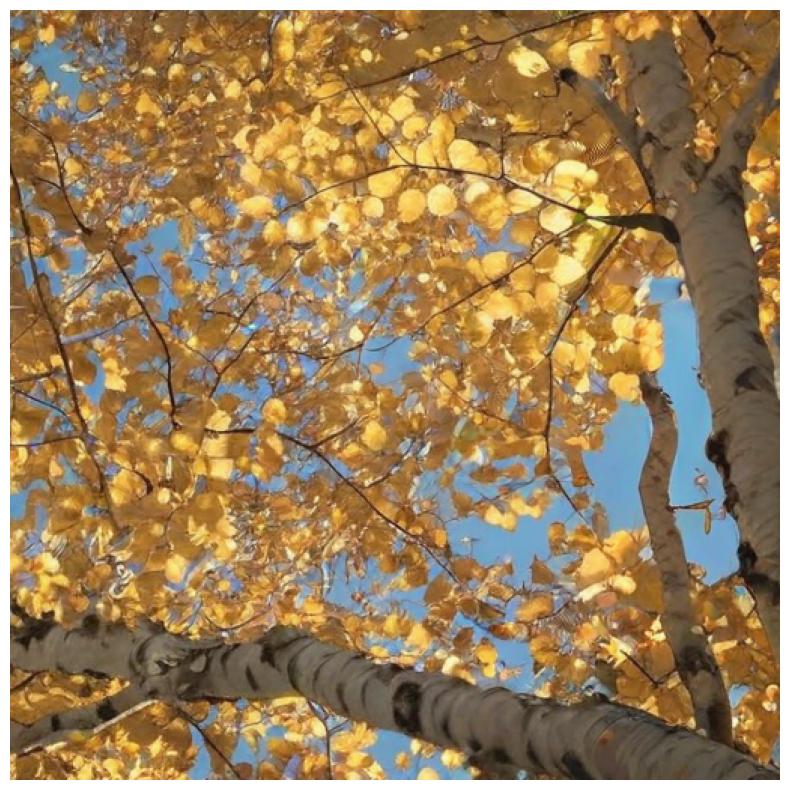}}
  \vspace{-1.2em}
            \subfloat{\includegraphics[width=1.1in]{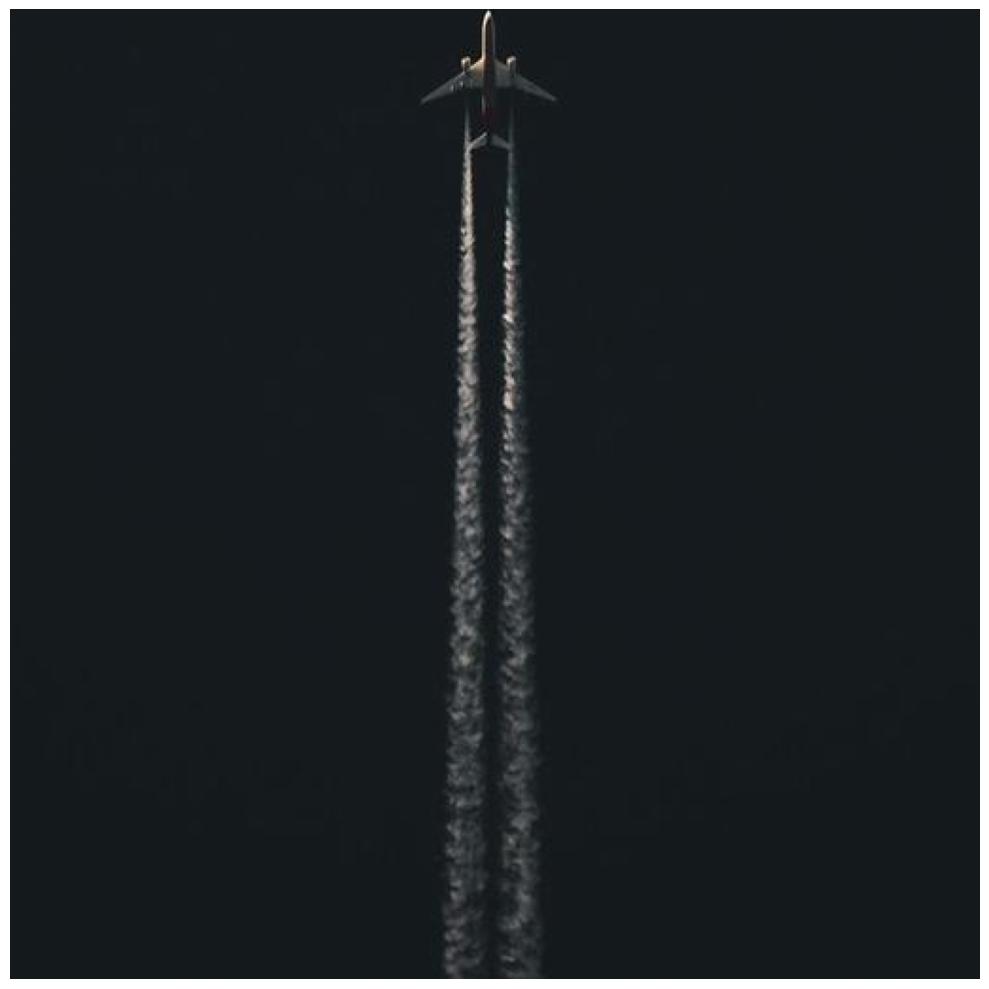}}
    \subfloat{\includegraphics[width=1.1in]{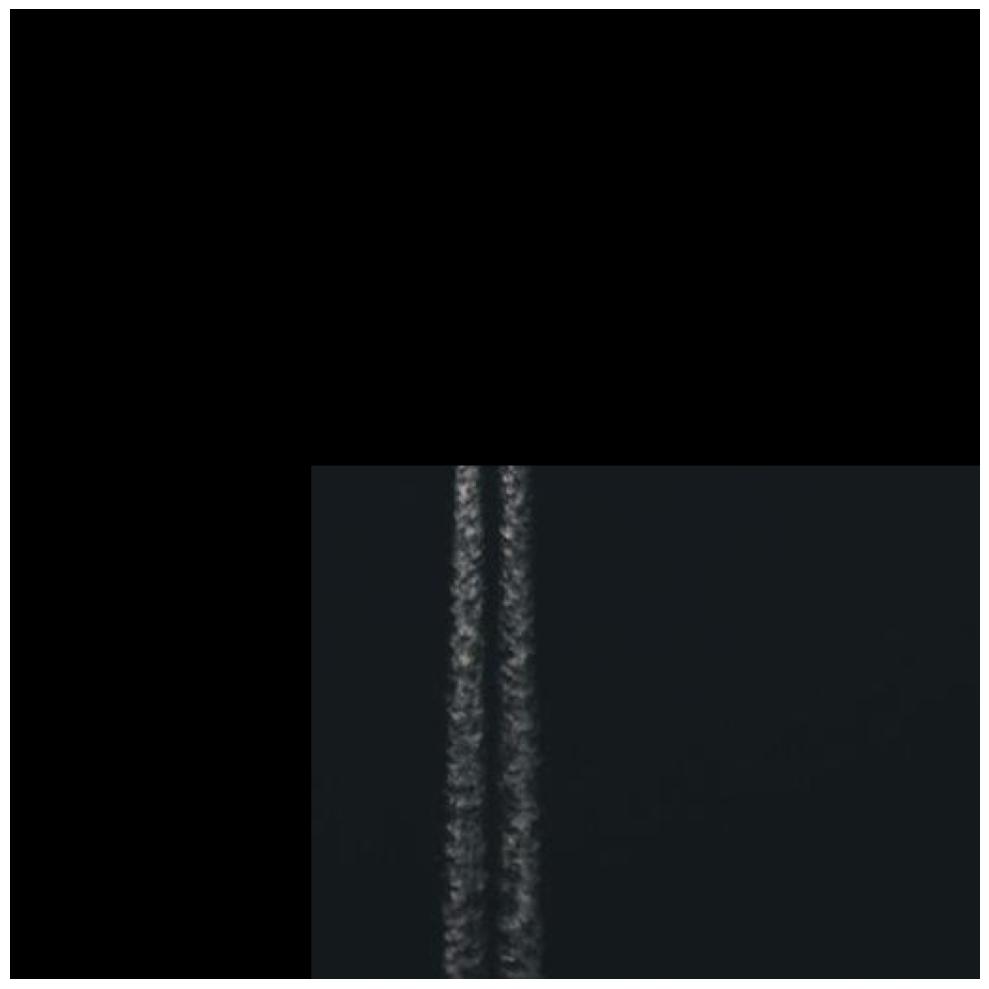}}
     \subfloat{\includegraphics[width=1.1in]{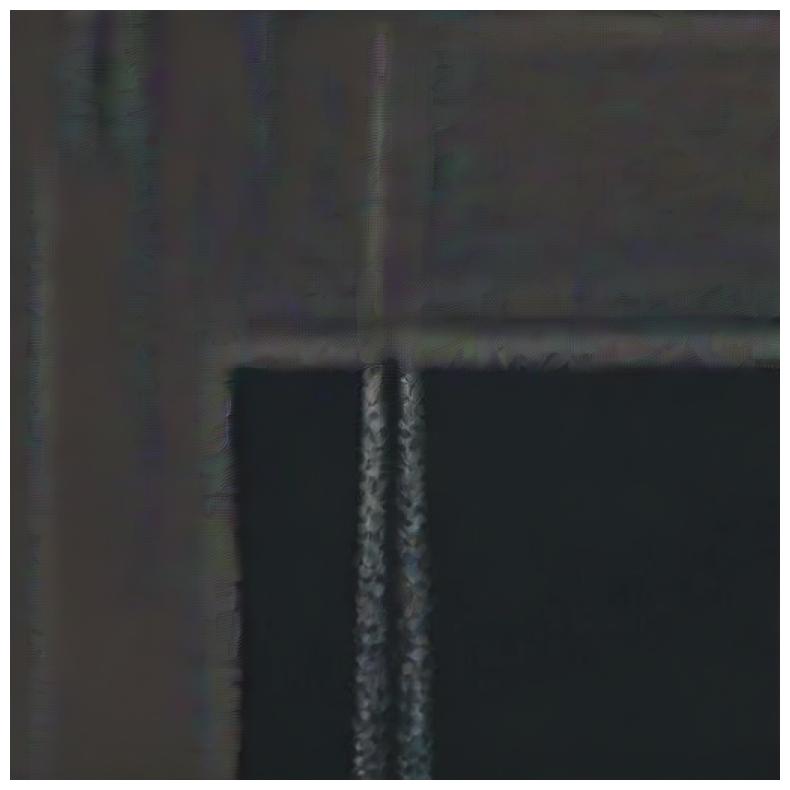}}
    \subfloat{\includegraphics[width=1.1in]{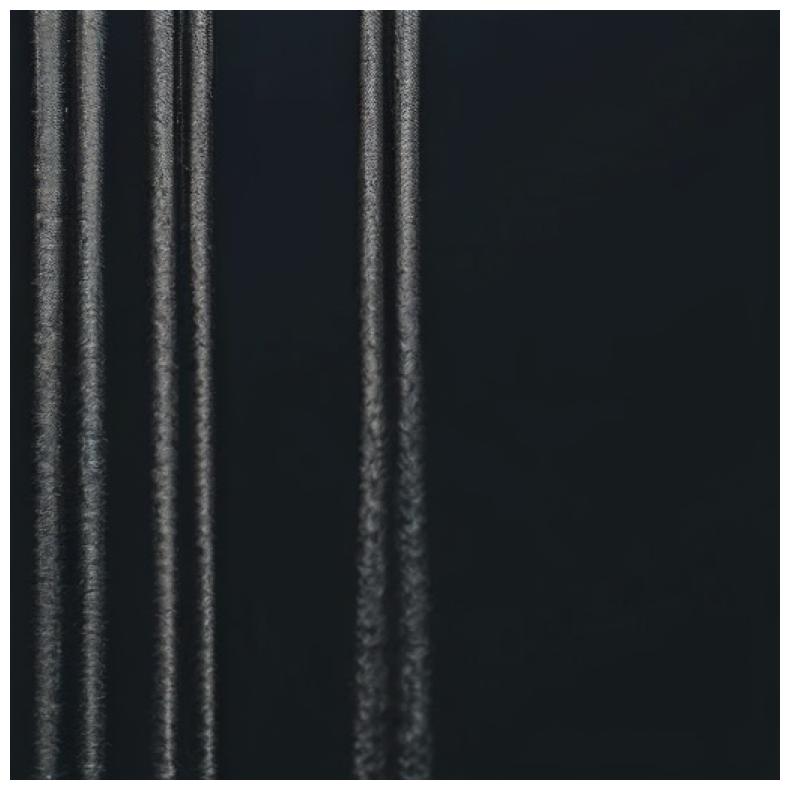}}
    \subfloat{\includegraphics[width=1.1in]{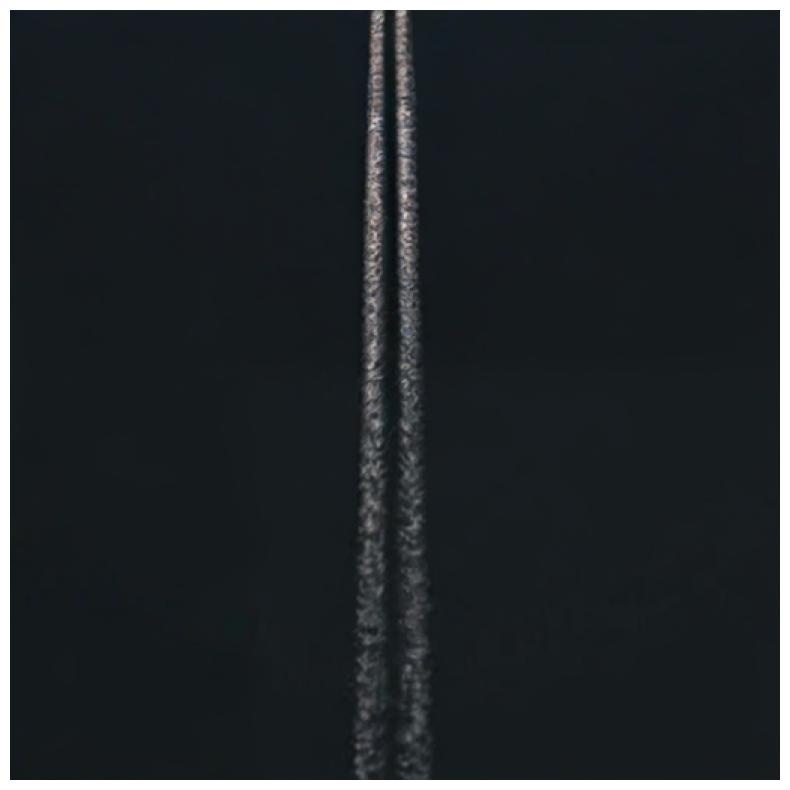}}\vspace{-1.2em}
  \subfloat{\includegraphics[width=1.1in]{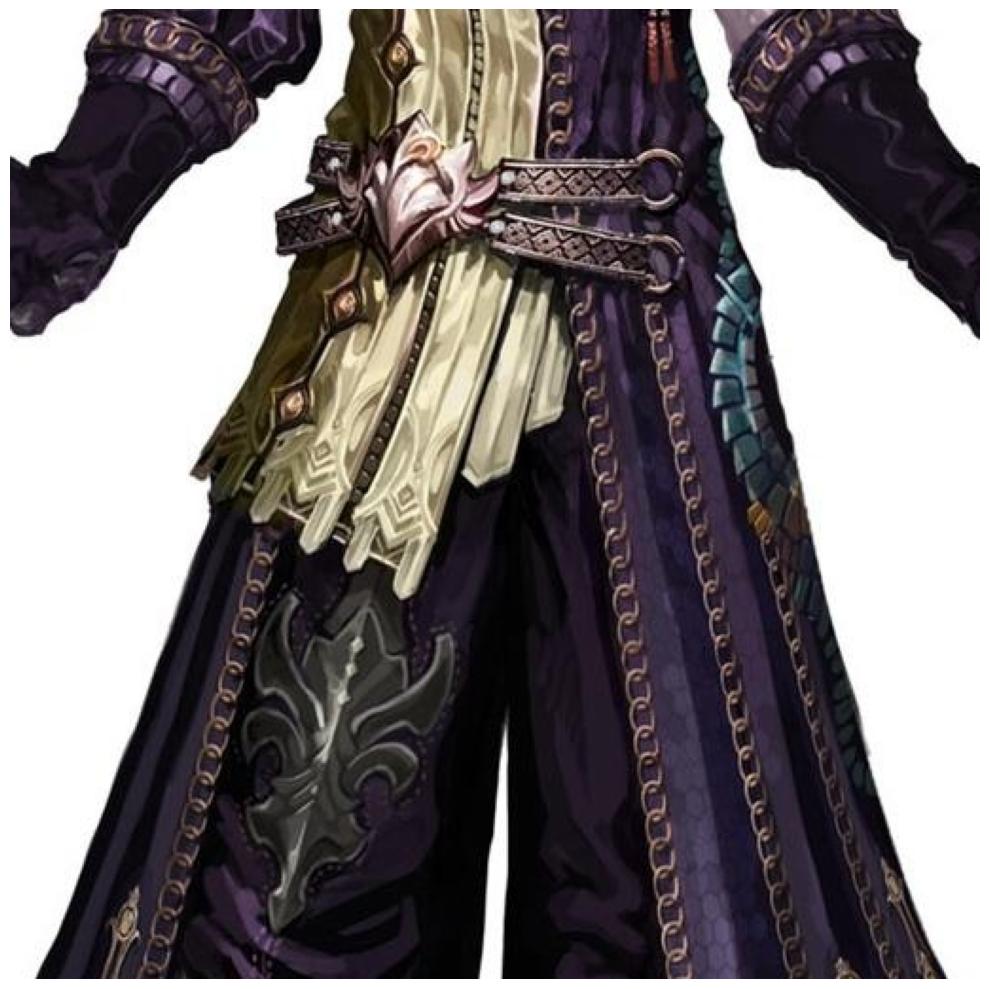}}
  \subfloat{\includegraphics[width=1.1in]{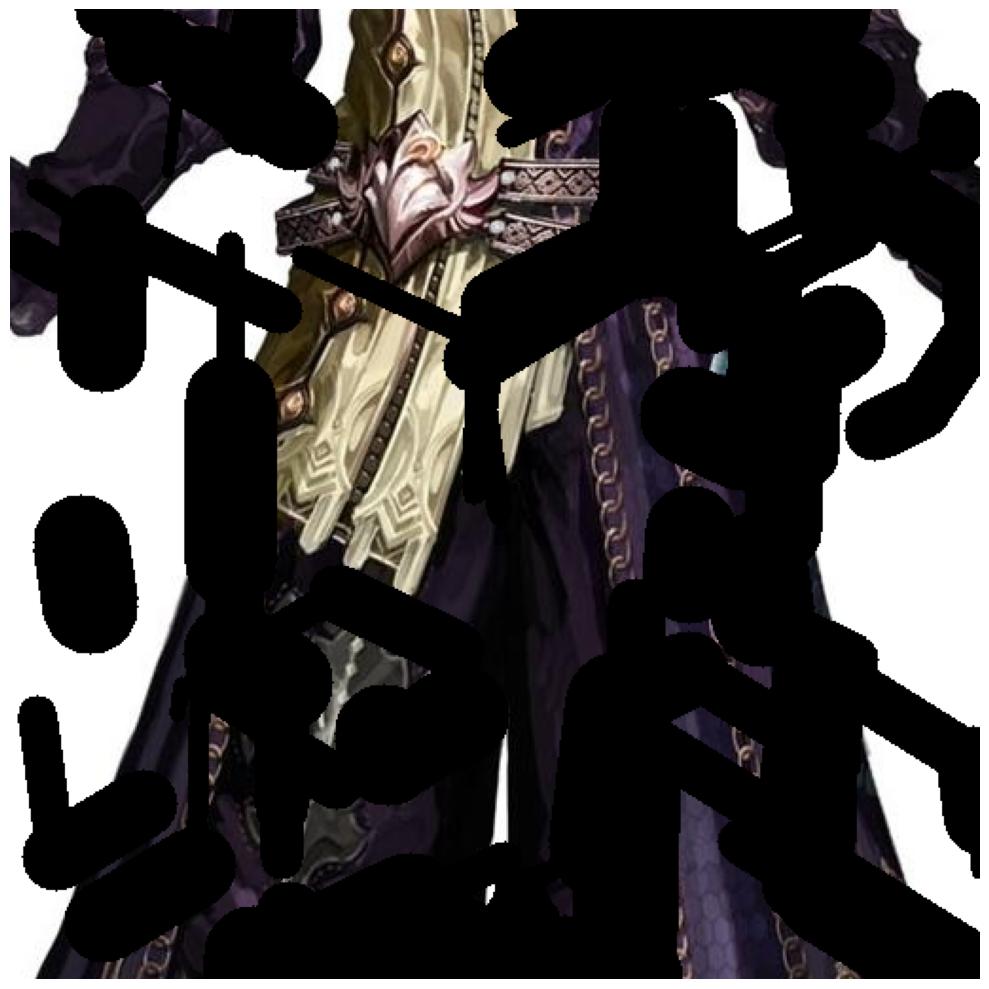}}
  \subfloat{\includegraphics[width=1.1in]{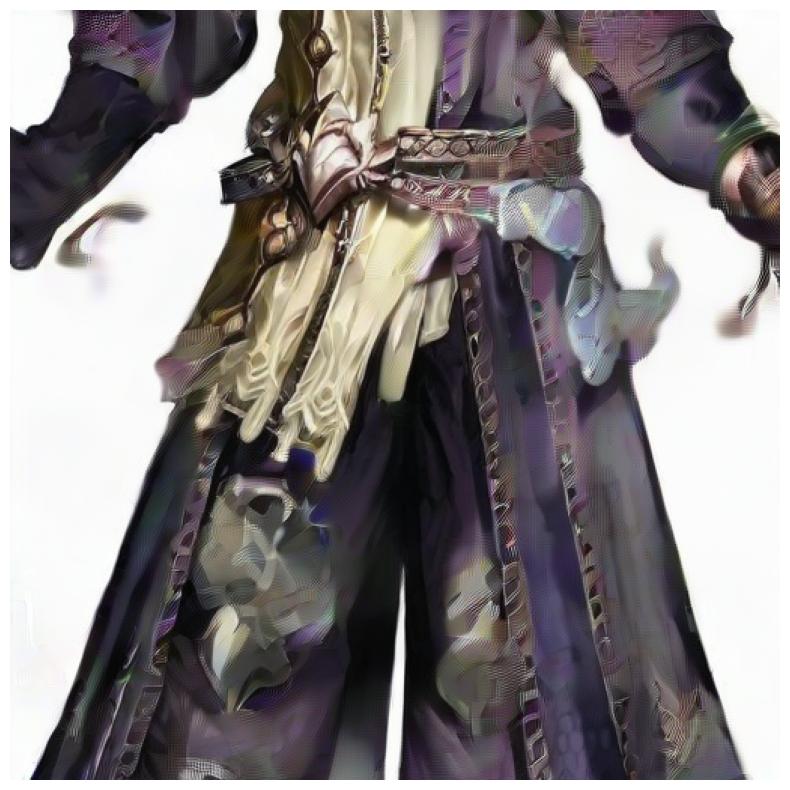}}
  \subfloat{\includegraphics[width=1.1in]{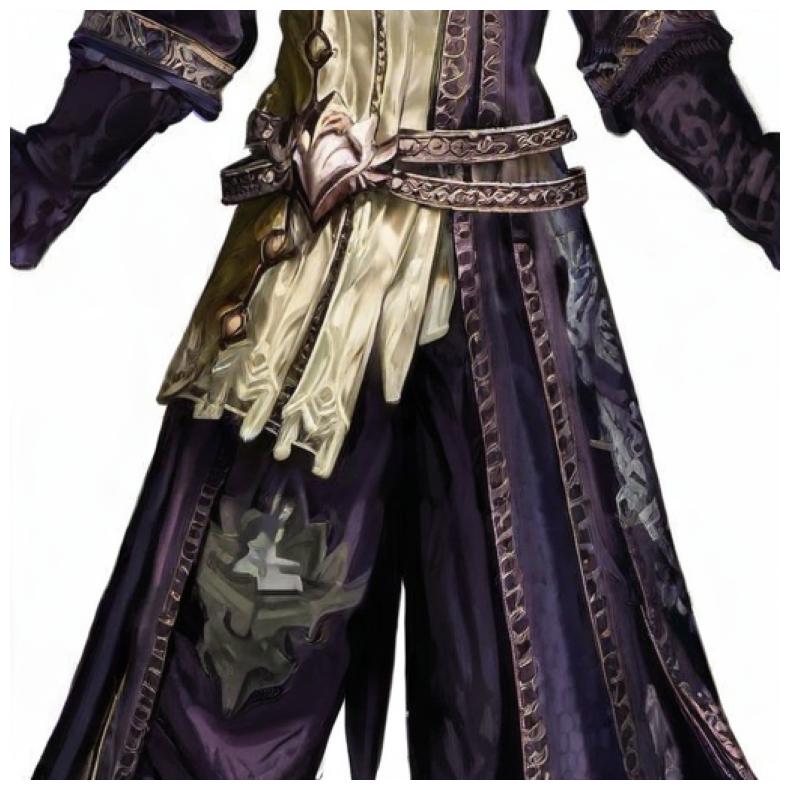}}
  \subfloat{\includegraphics[width=1.1in]{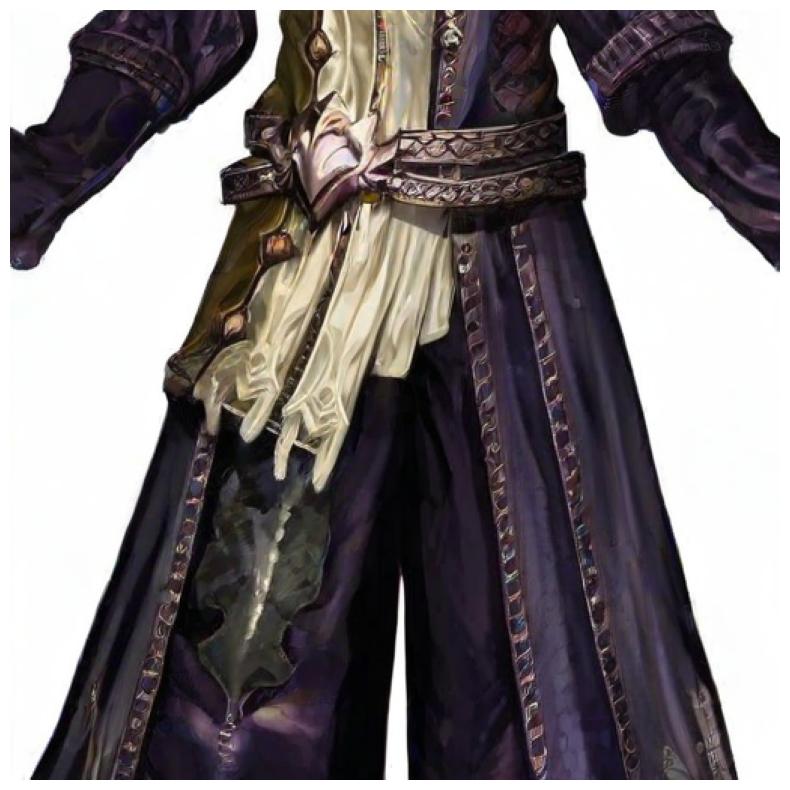}}
  \vspace{-1.2em}
  \subfloat{\includegraphics[width=1.1in]{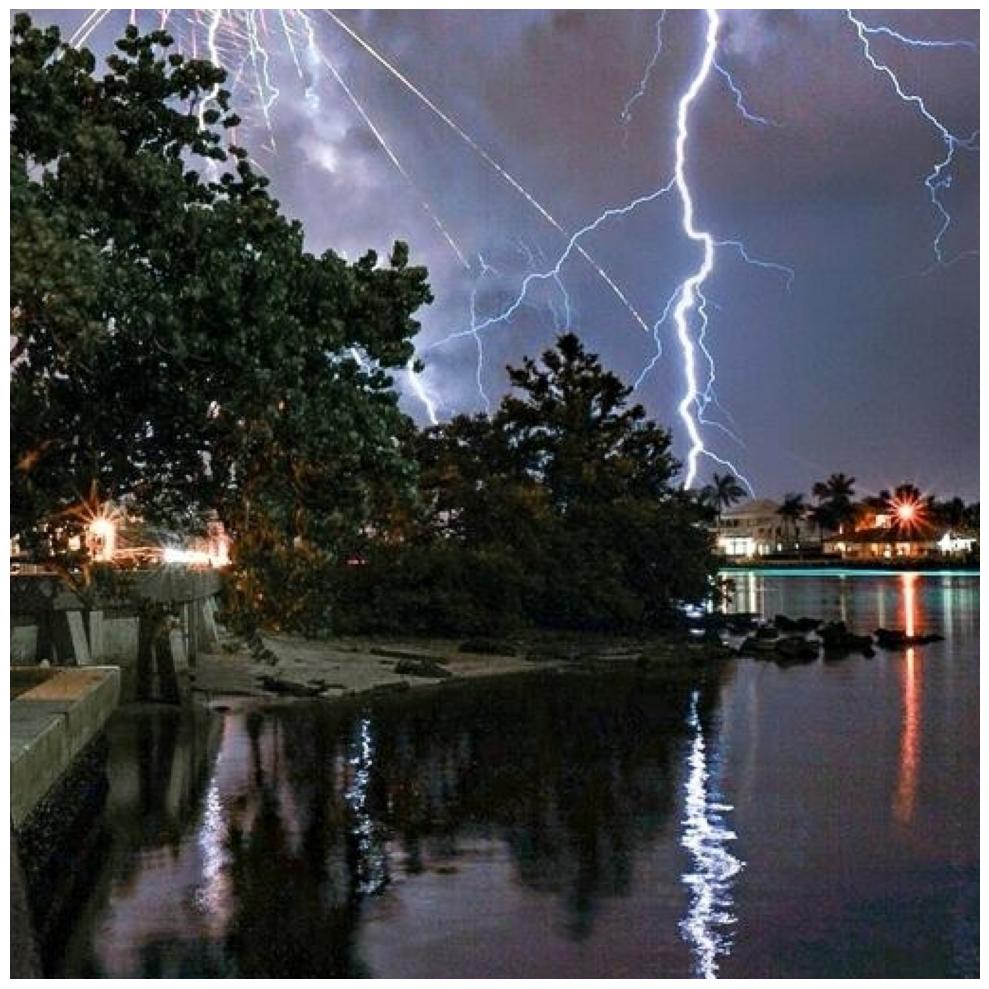}}
  \subfloat{\includegraphics[width=1.1in]{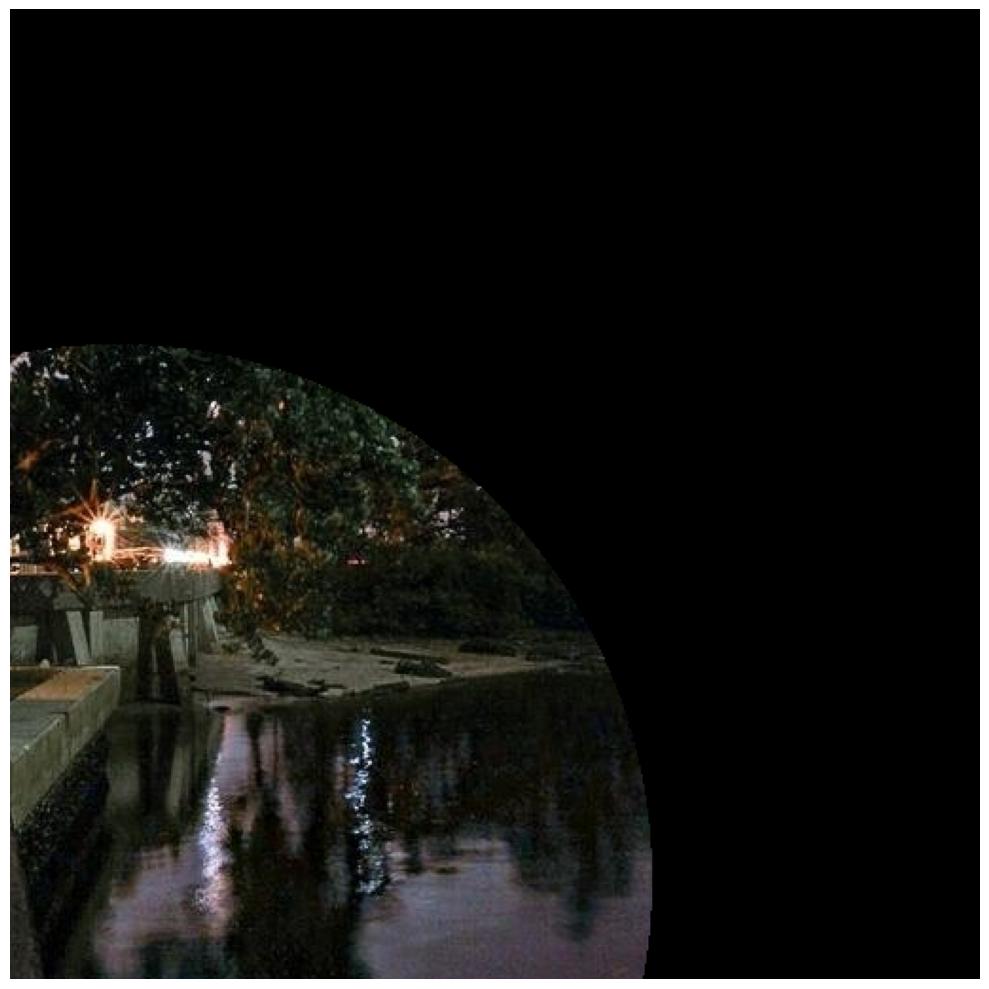}}
  \subfloat{\includegraphics[width=1.1in]{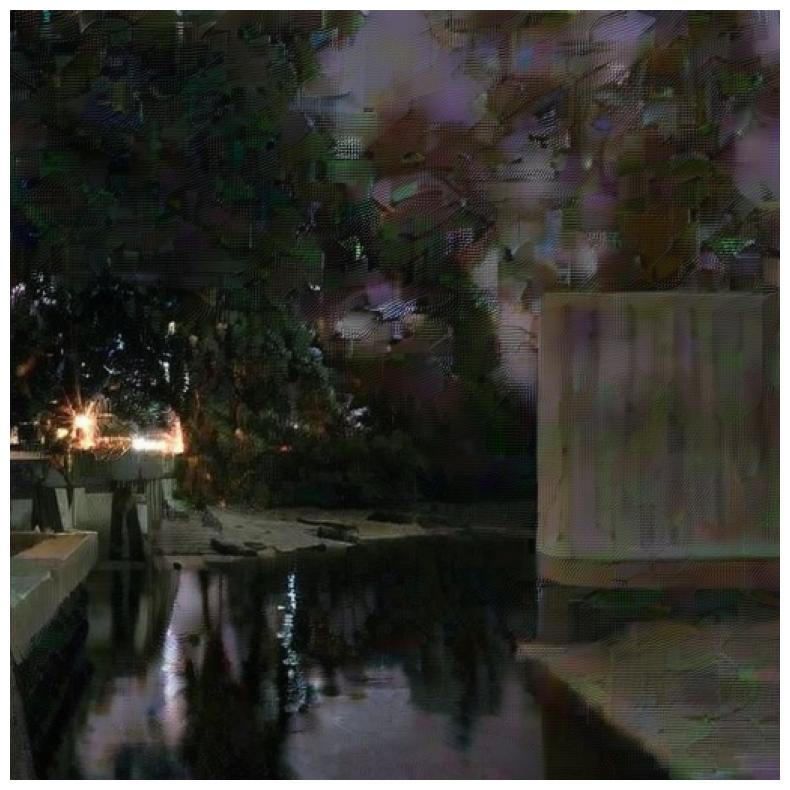}}
  \subfloat{\includegraphics[width=1.1in]{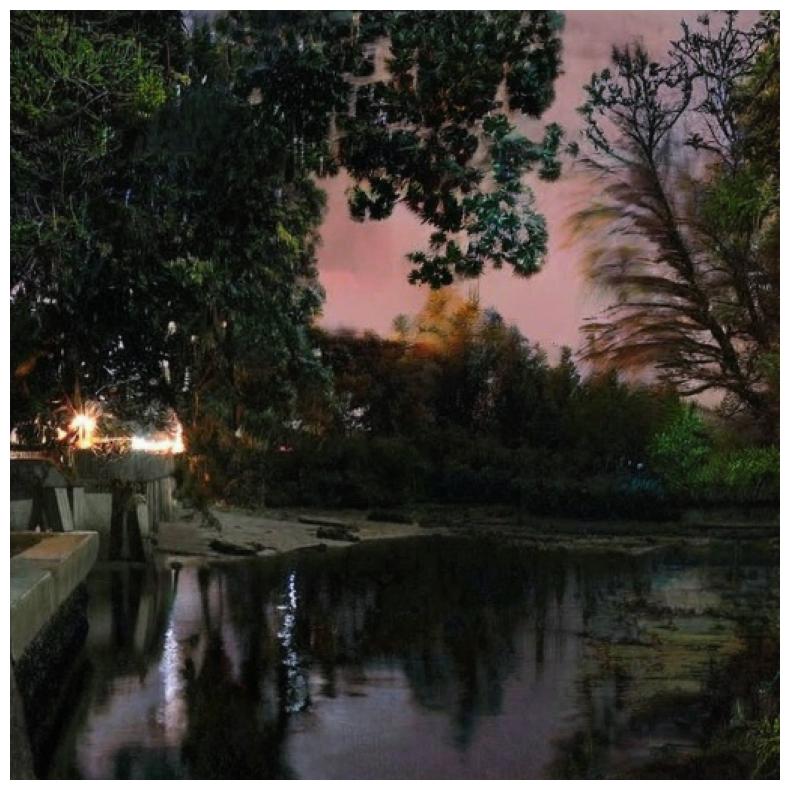}}
  \subfloat{\includegraphics[width=1.1in]{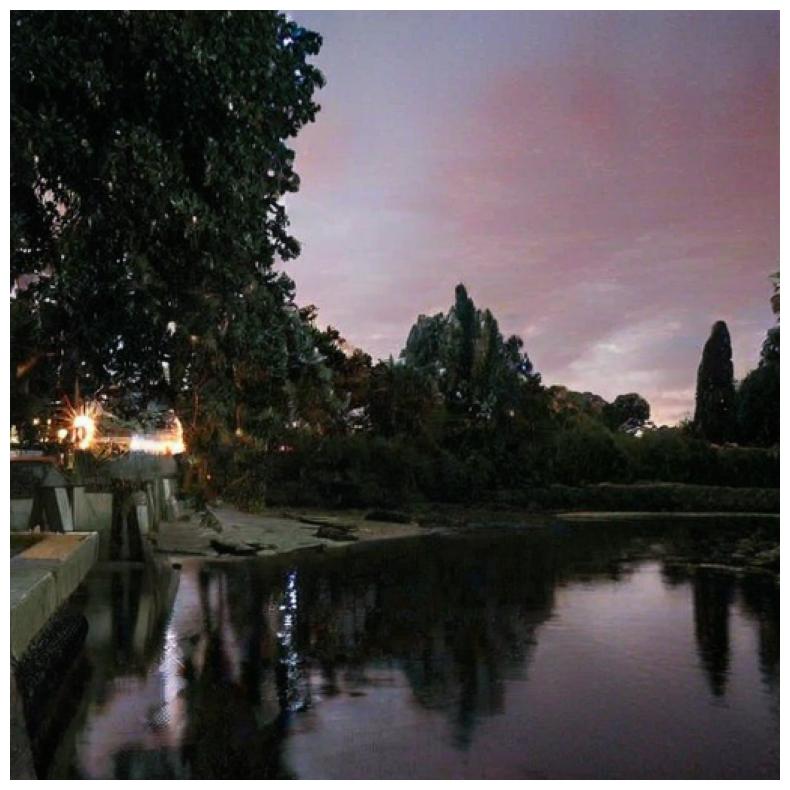}}
  \vspace{-1.2em}
  \subfloat{\includegraphics[width=1.1in]{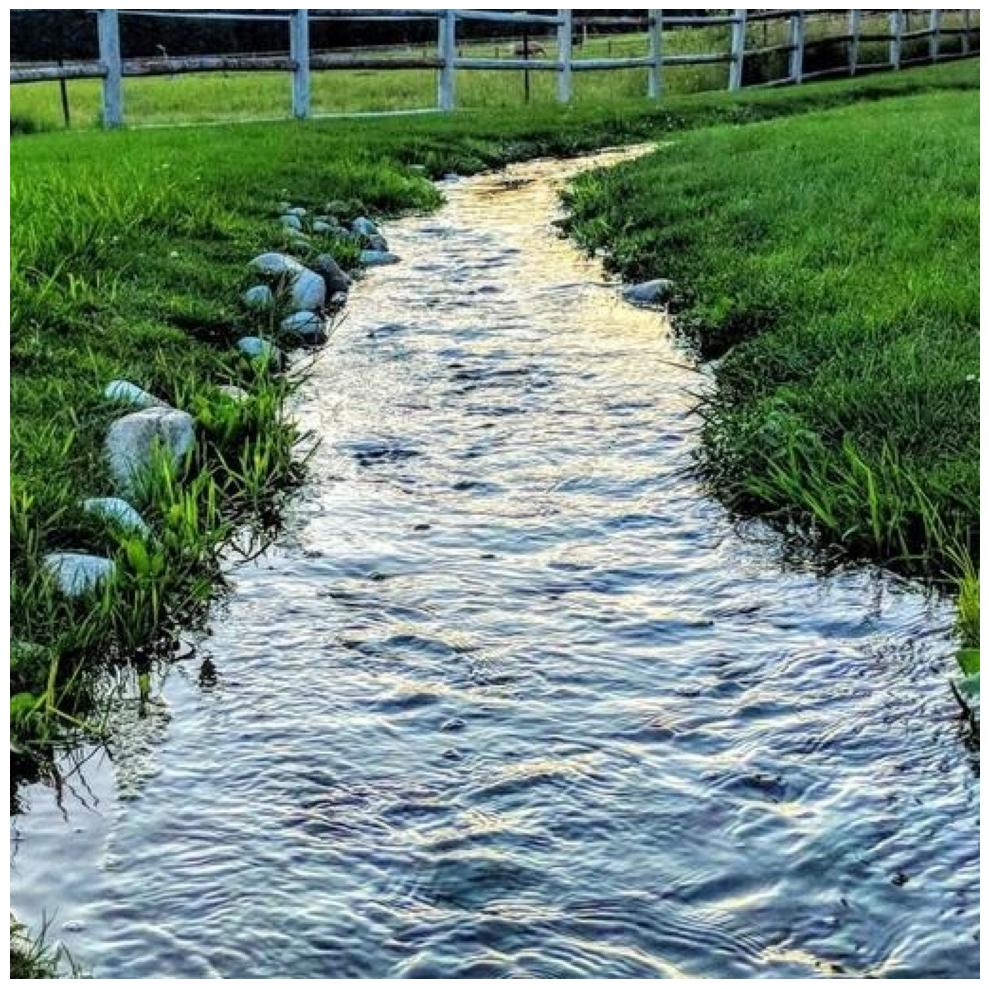}}
  \subfloat{\includegraphics[width=1.1in]{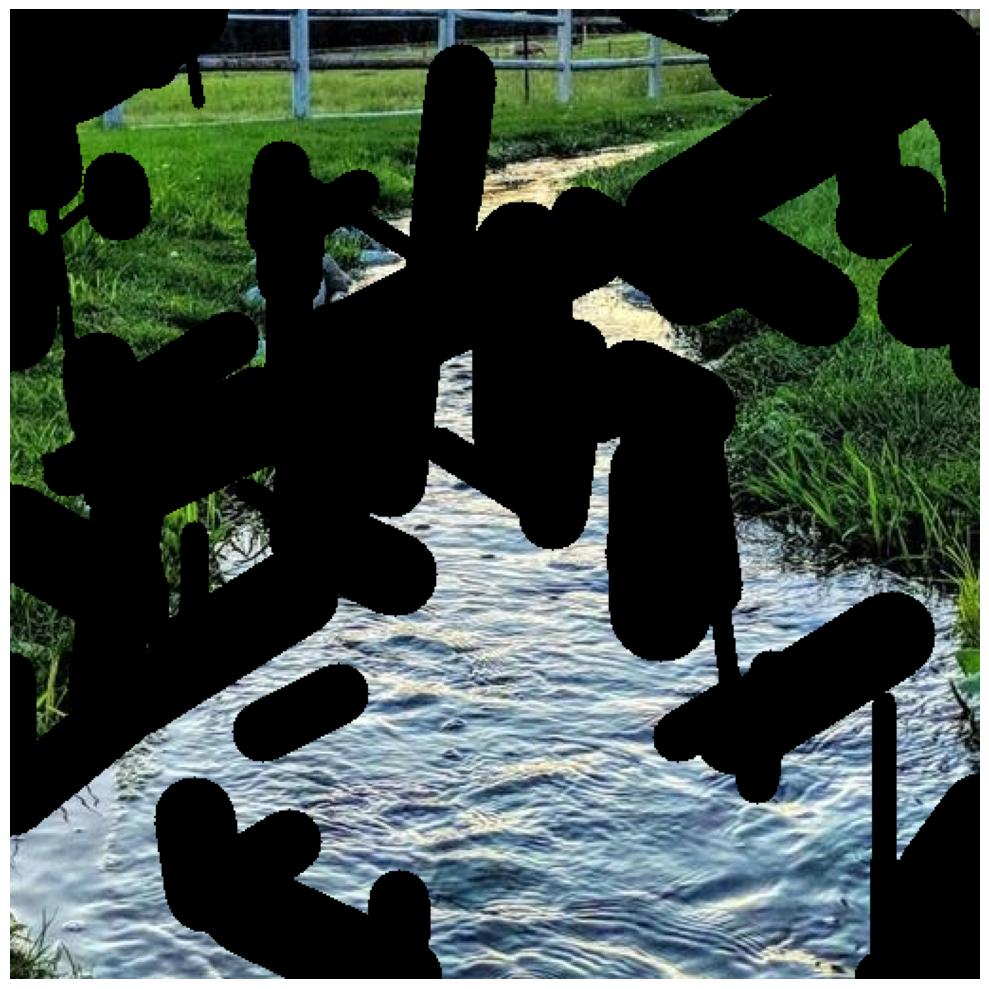}}
  \subfloat{\includegraphics[width=1.1in]{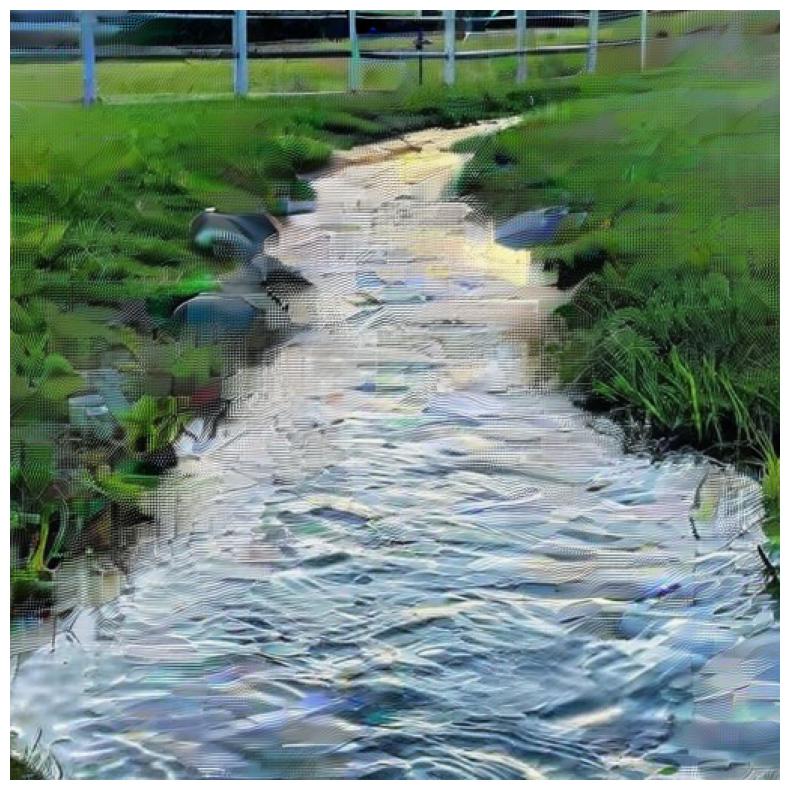}}
  \subfloat{\includegraphics[width=1.1in]{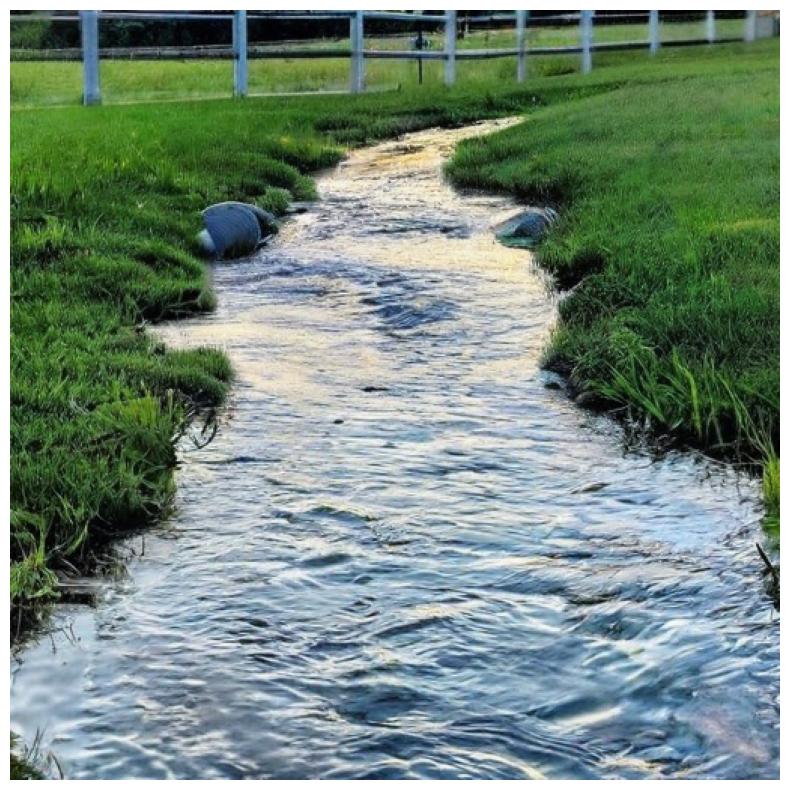}}
  \subfloat{\includegraphics[width=1.1in]{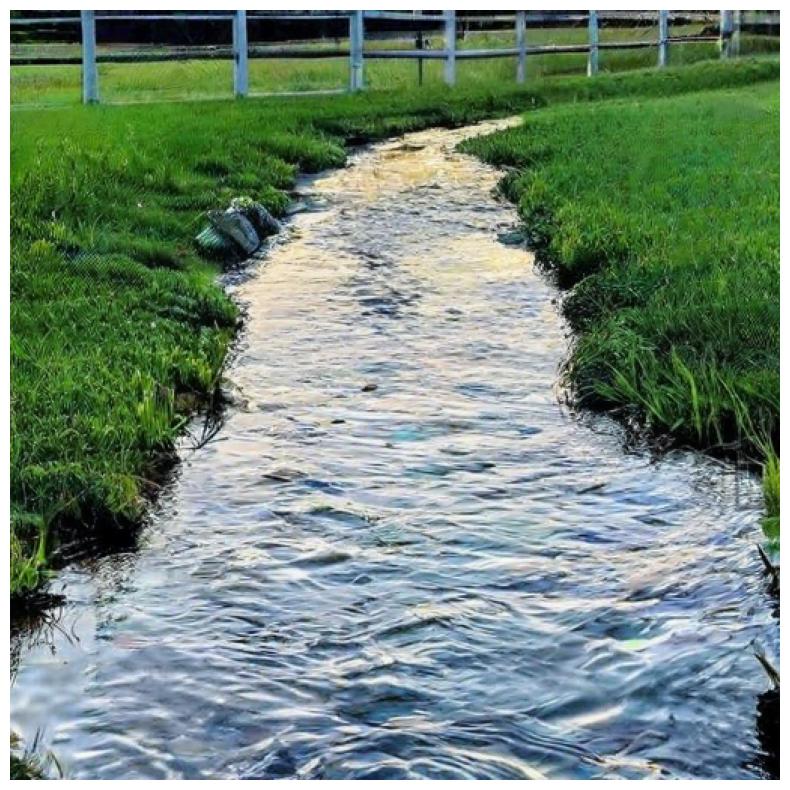}}
  \vspace{-1.2em}
  \subfloat{\includegraphics[width=1.1in]{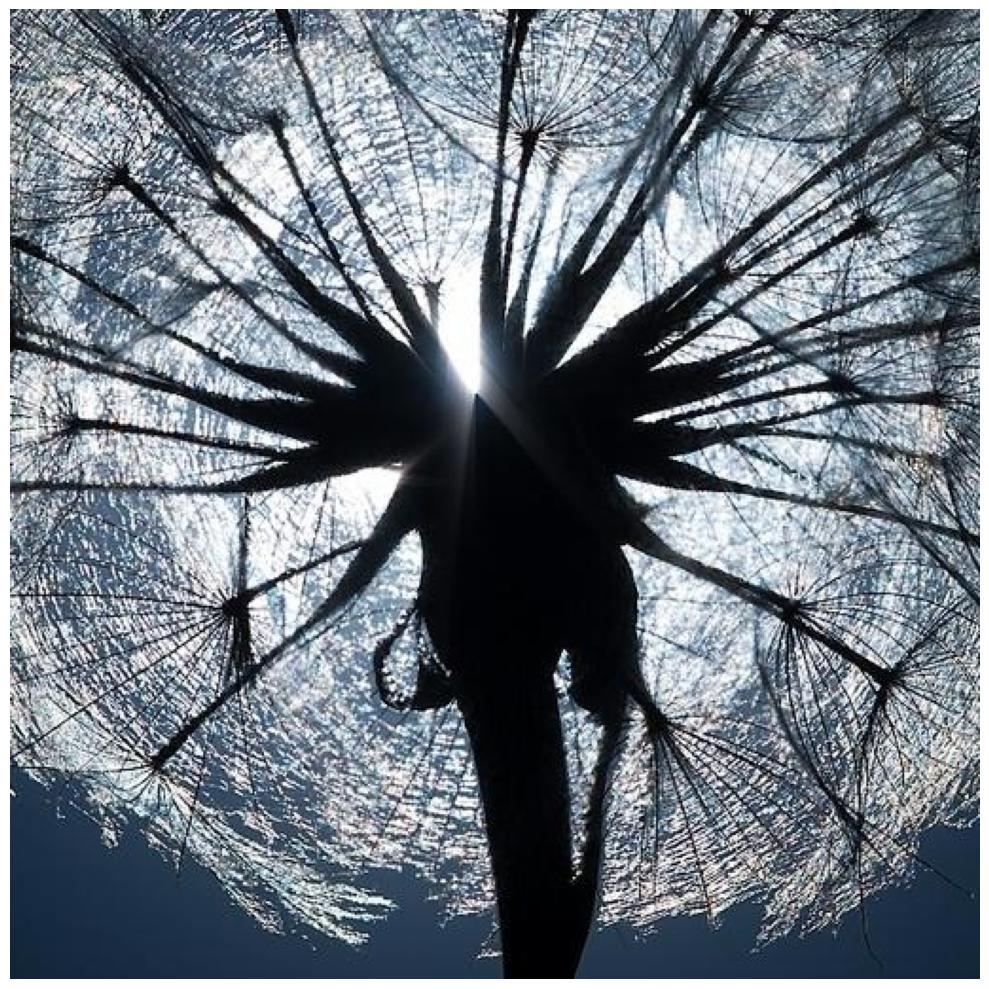}}
  \subfloat{\includegraphics[width=1.1in]{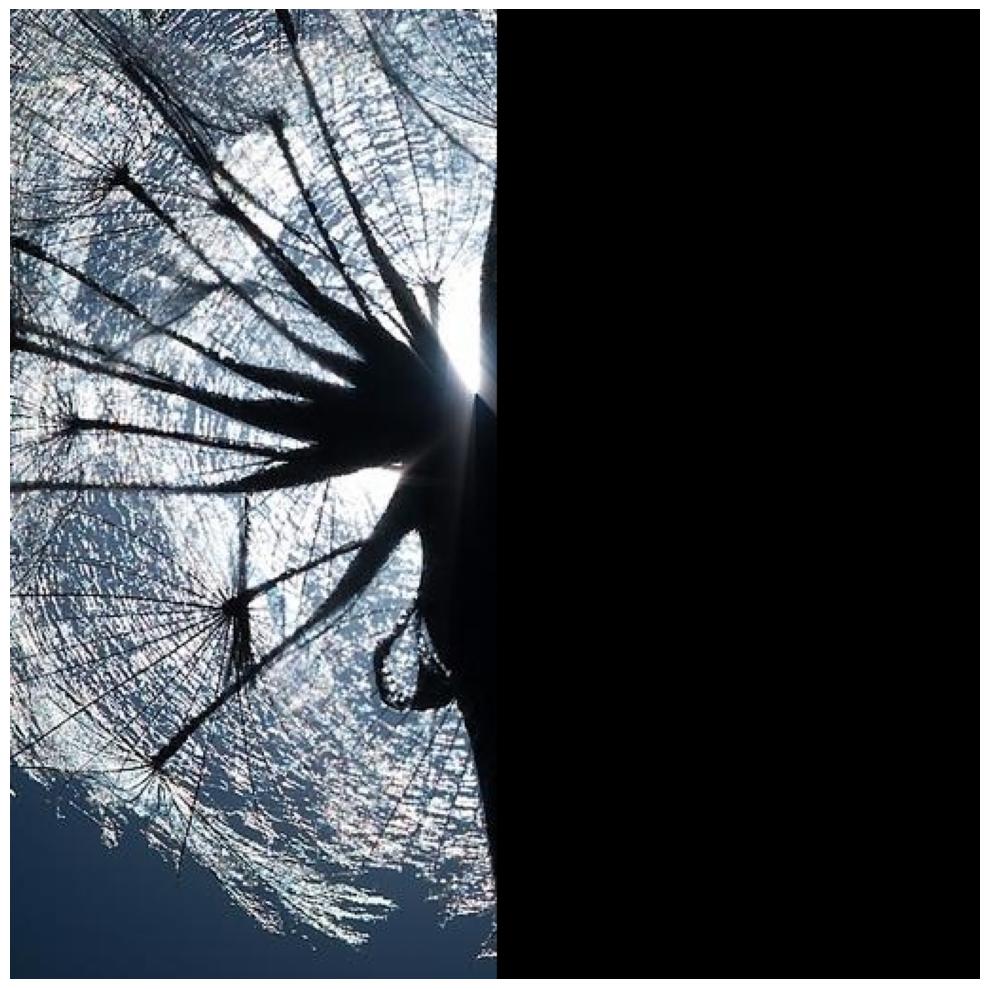}}
  \subfloat{\includegraphics[width=1.1in]{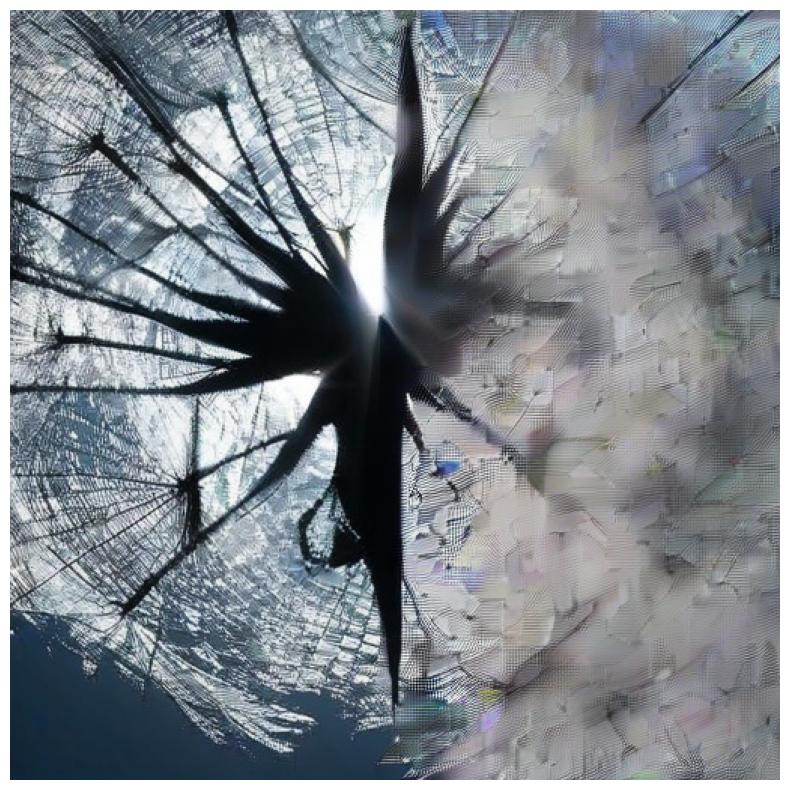}}
  \subfloat{\includegraphics[width=1.1in]{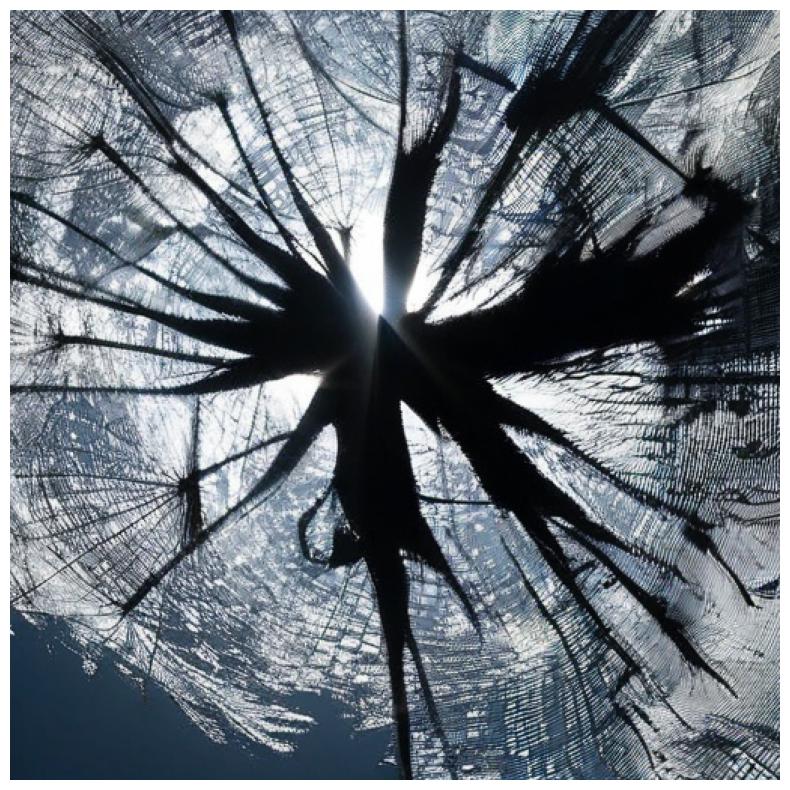}}
  \subfloat{\includegraphics[width=1.1in]{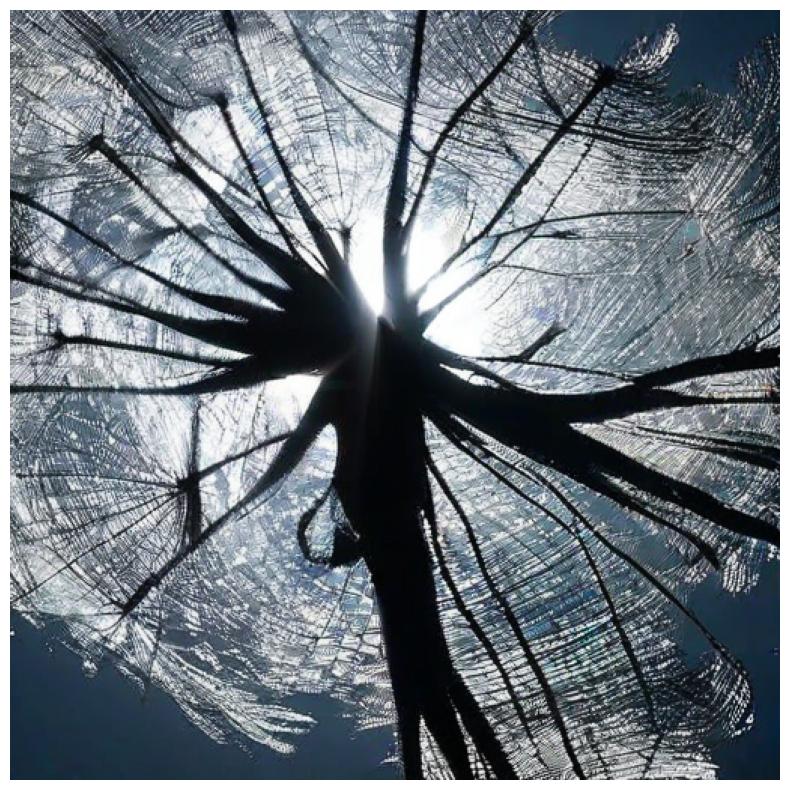}}
  \caption{Application of \emph{Flash Diffusion} to an \emph{in-house} diffusion-based \emph{inpainting} model. Best viewed zoomed in.}
  \label{fig:app_inpainting}
  \end{figure*}

In Fig.~\ref{fig:app_inpainting}, we provide additional samples using the trained \emph{inpainting} student model. We compare the samples generated by the student model using 4 NFEs to the teacher generations using 4 steps (\emph{i.e.} 8 NFEs) and 20 steps (\emph{i.e.} 40 NFEs).

\subsection{Flash Upscaler}\label{sec:app_flash_upscaler}
\begin{figure*}[p]
    \centering
    \captionsetup[subfigure]{position=above, labelformat = empty}
          \subfloat[\scriptsize LR image]{\includegraphics[width=1.36in]{plots/upscaler/4/lr_image.jpg}}
    \subfloat[\scriptsize Teacher (8 NFEs)]{\includegraphics[width=1.36in]{plots/upscaler/4/teacher_4_steps.jpg}}
    \subfloat[\scriptsize \centering Teacher (40 NFEs)]{\includegraphics[width=1.36in]{plots/upscaler/4/teacher_20_steps.jpg}}
    \subfloat[\scriptsize Ours (4 NFEs)]{\includegraphics[width=1.36in]{plots/upscaler/4/student_4_steps.jpg}}\vspace{-1.2em}
    \subfloat{\includegraphics[width=1.36in]{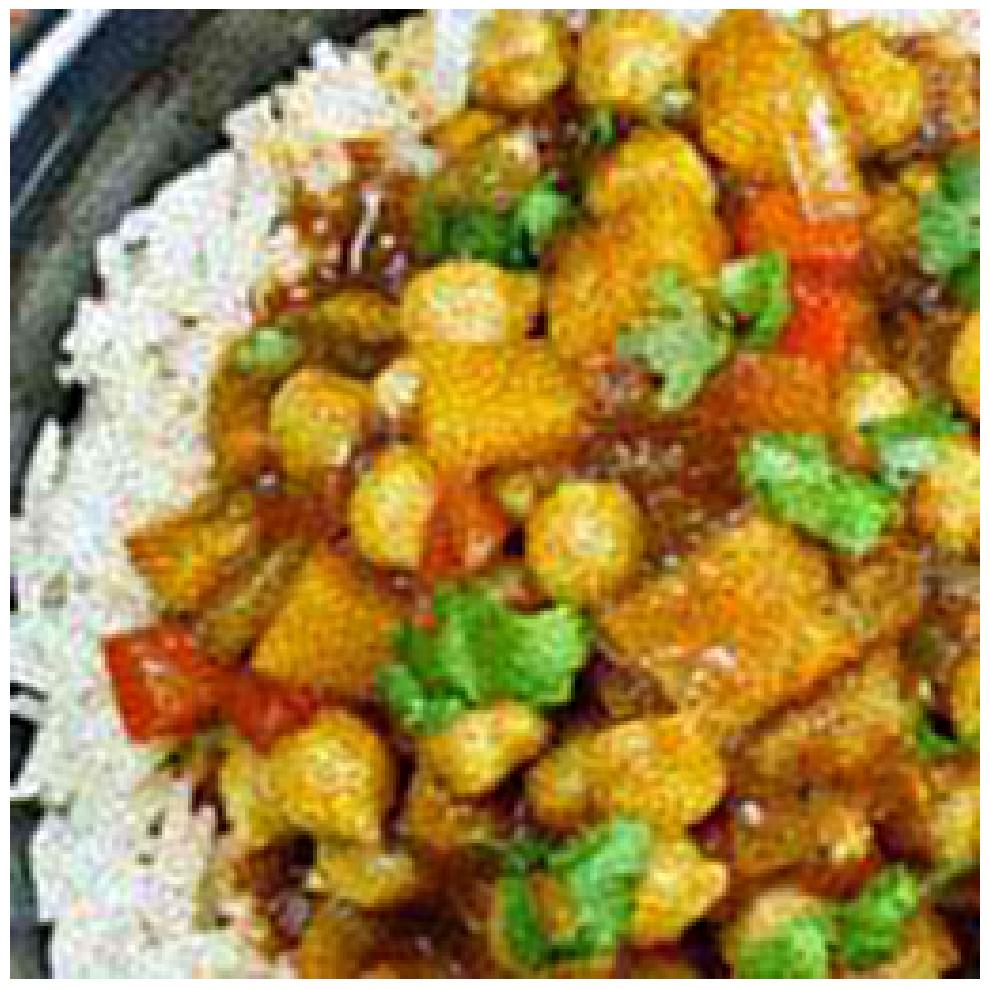}}
    \subfloat{\includegraphics[width=1.36in]{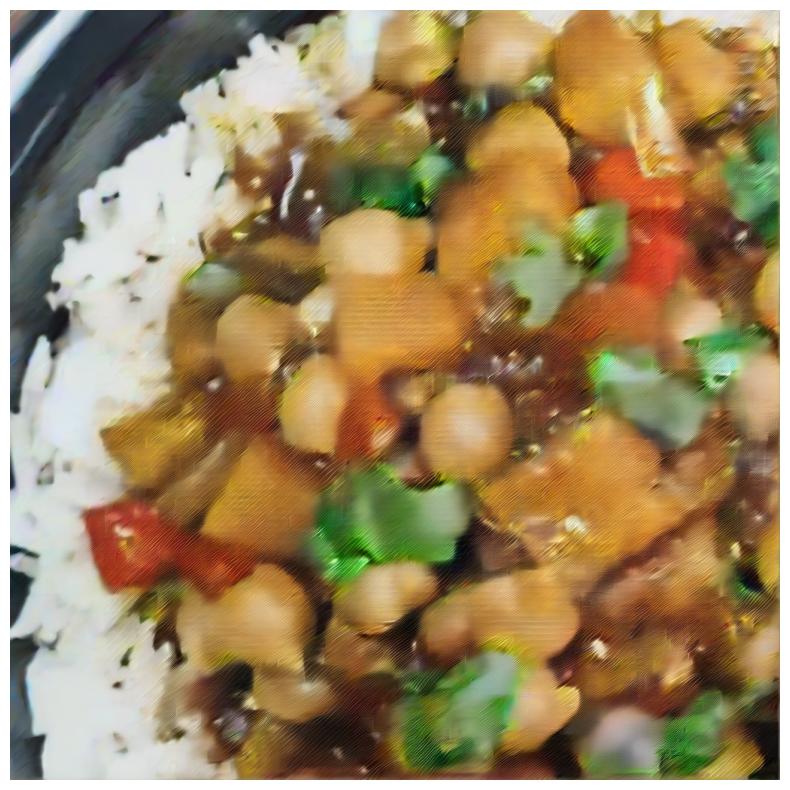}}
    \subfloat{\includegraphics[width=1.36in]{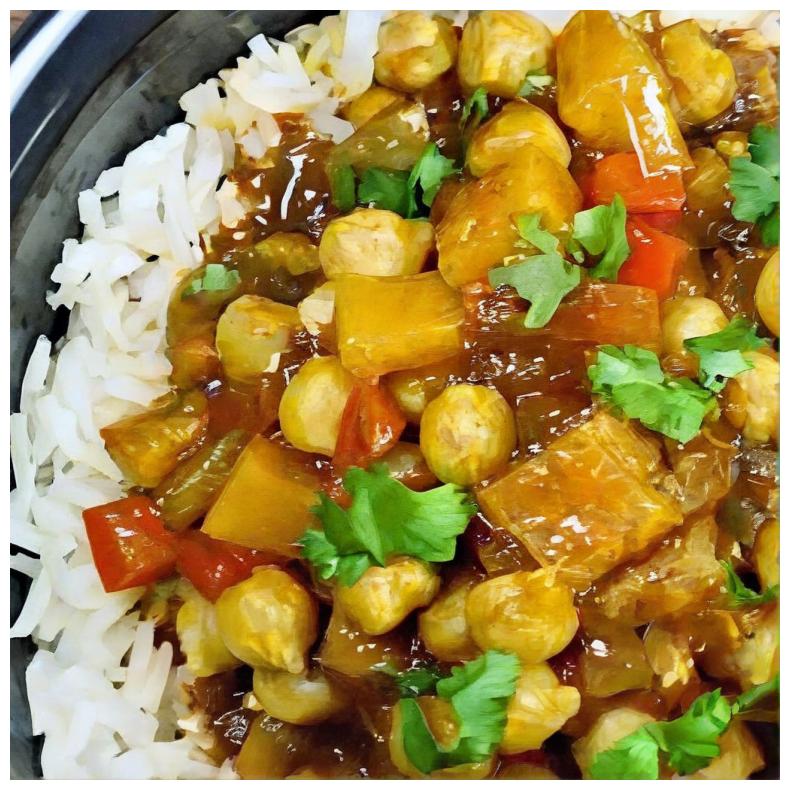}}
    \subfloat{\includegraphics[width=1.36in]{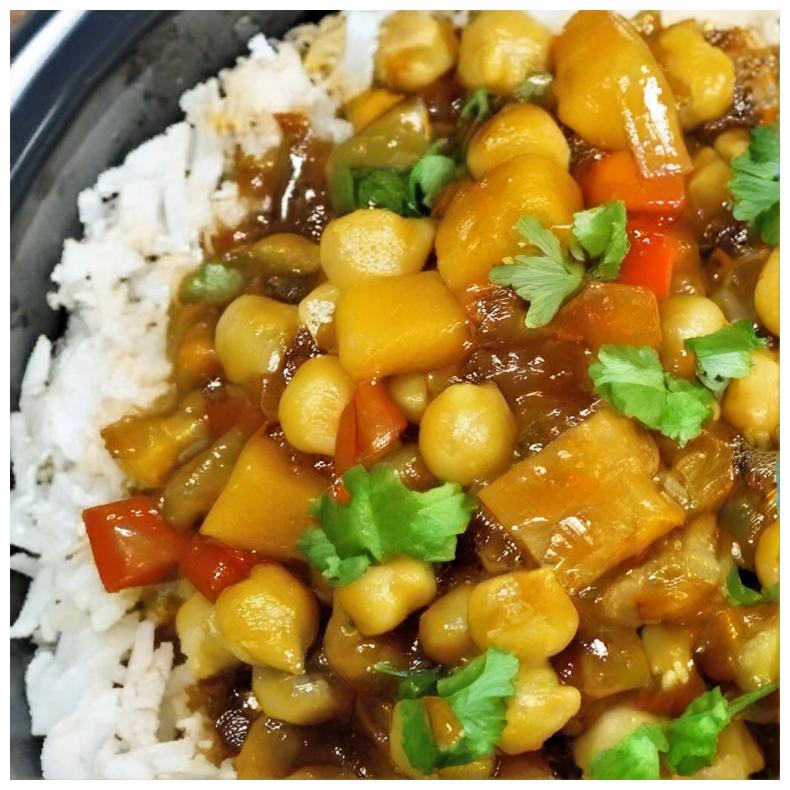}}\vspace{-1.2em}
    \subfloat{\includegraphics[width=1.36in]{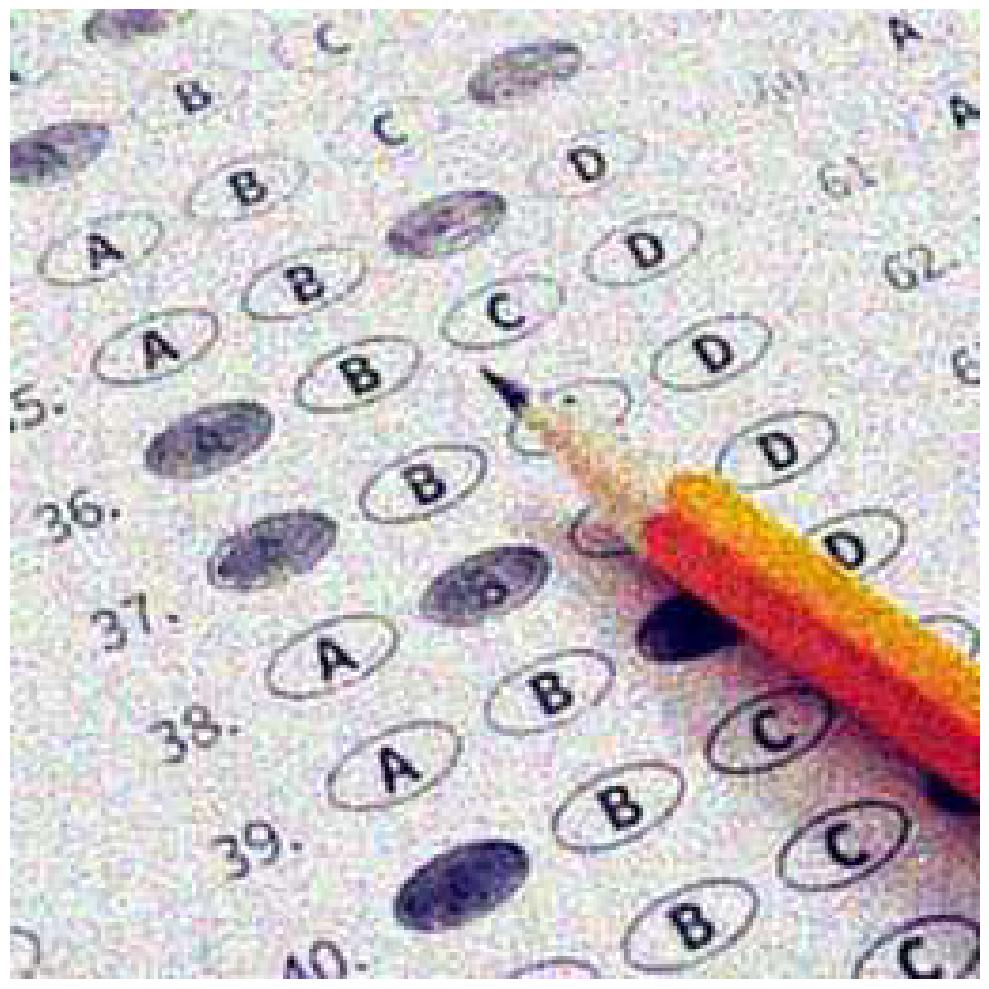}}
    \subfloat{\includegraphics[width=1.36in]{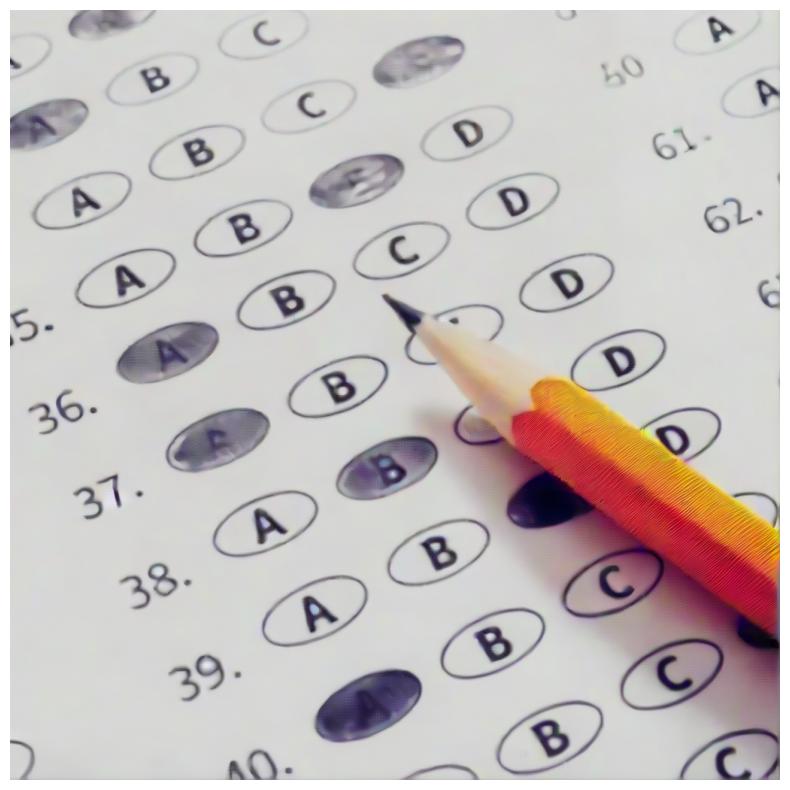}}
    \subfloat{\includegraphics[width=1.36in]{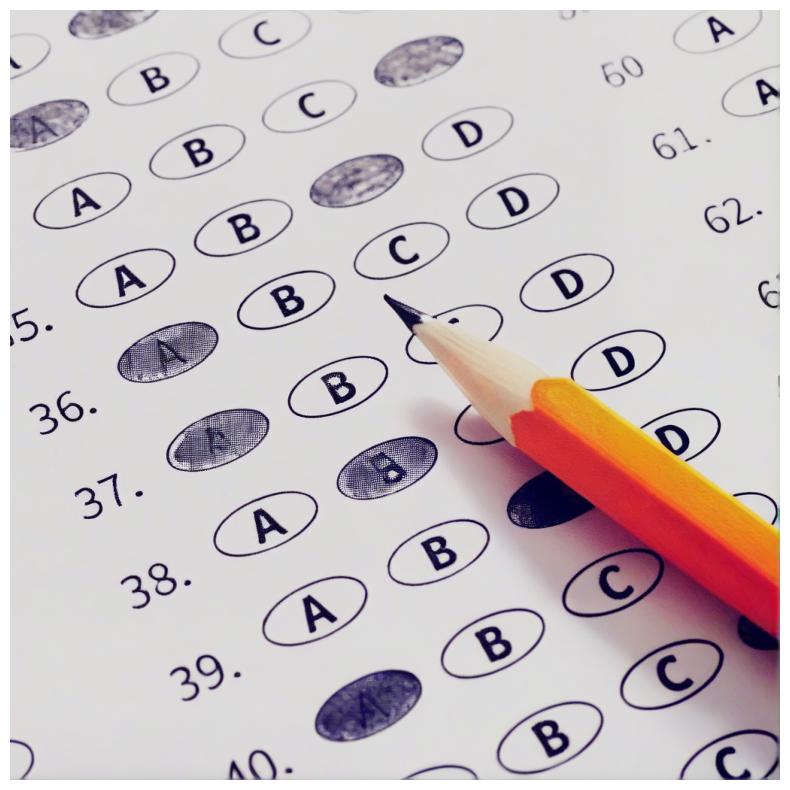}}
    \subfloat{\includegraphics[width=1.36in]{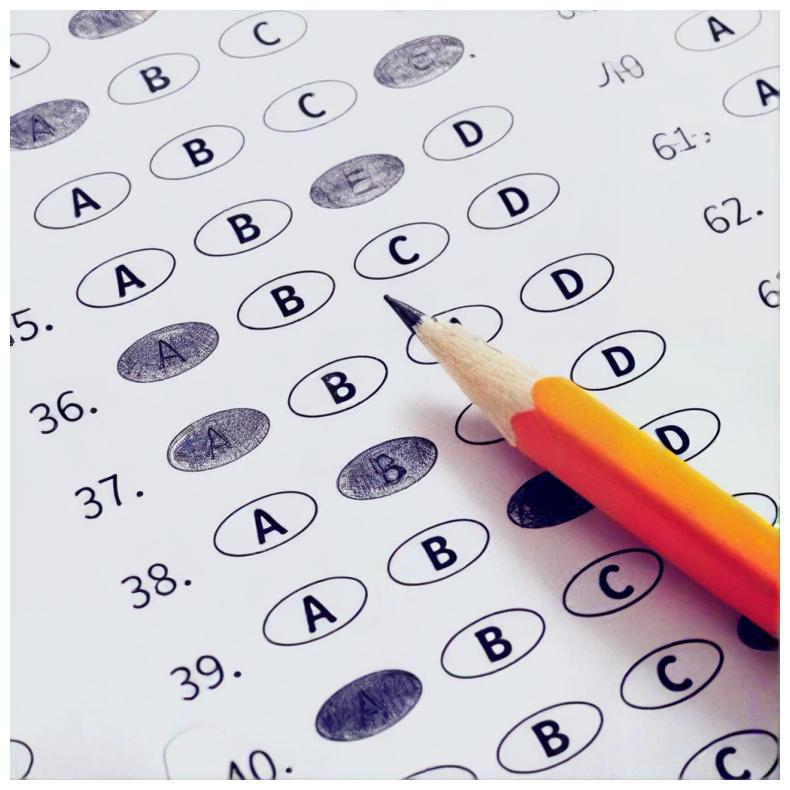}}
    \vspace{-1.2em}
    \subfloat{\includegraphics[width=1.36in]{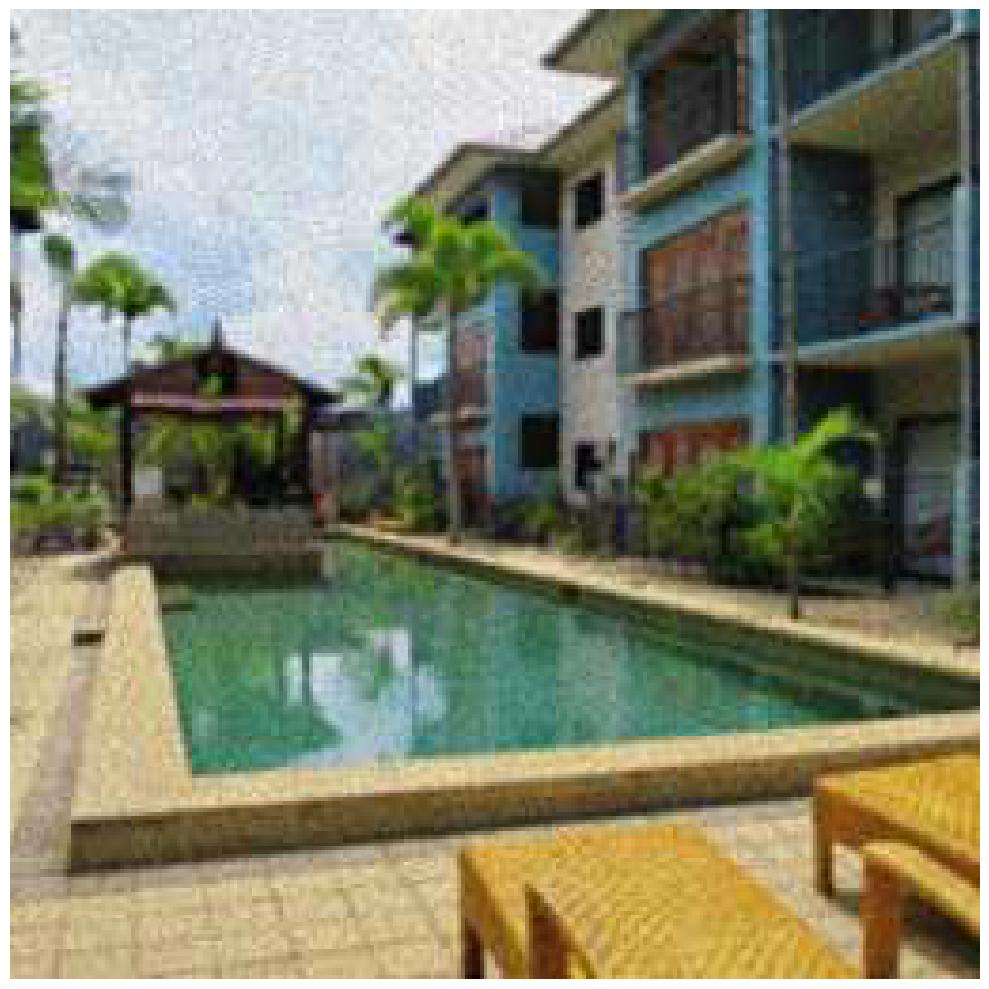}}
    \subfloat{\includegraphics[width=1.36in]{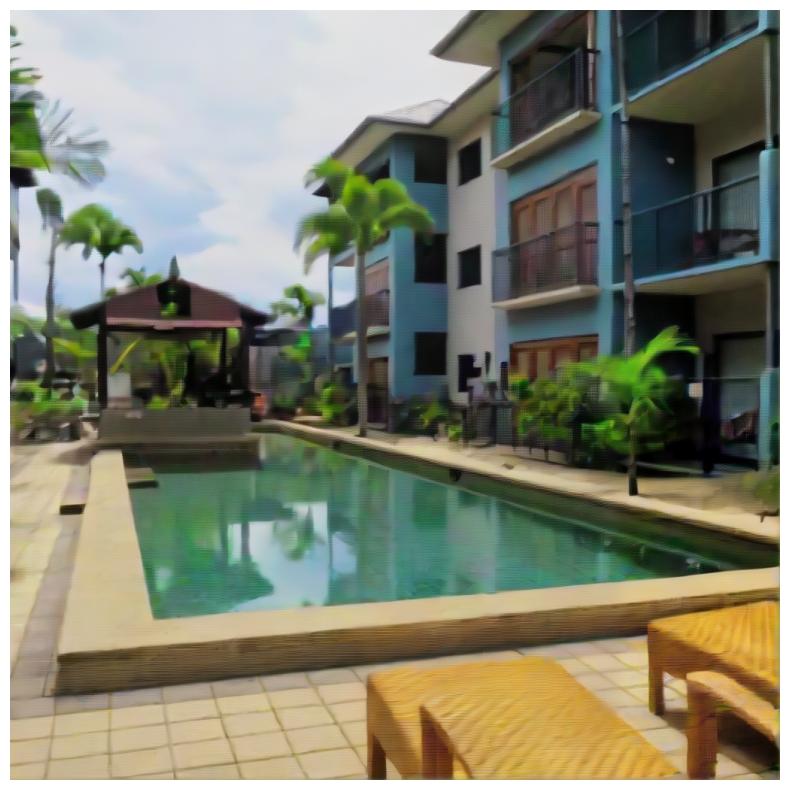}}
    \subfloat{\includegraphics[width=1.36in]{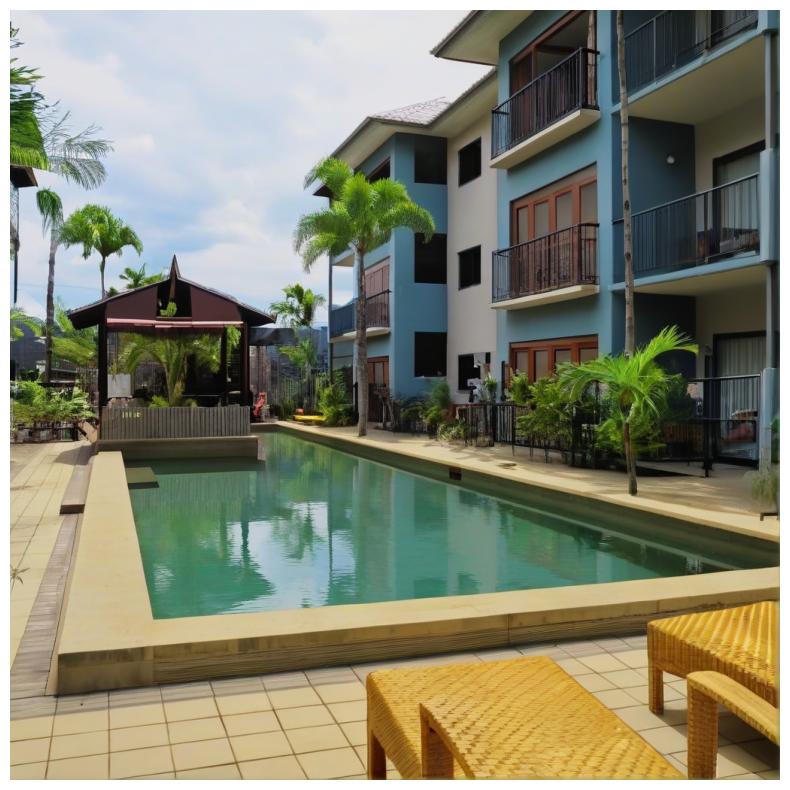}}
    \subfloat{\includegraphics[width=1.36in]{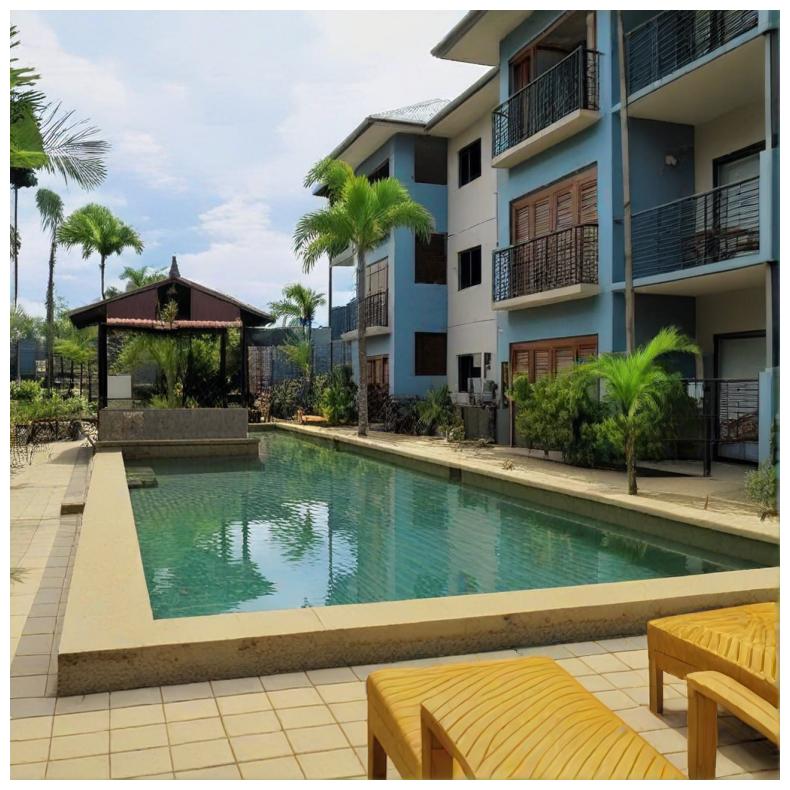}}
    \vspace{-1.2em}
    \subfloat{\includegraphics[width=1.36in]{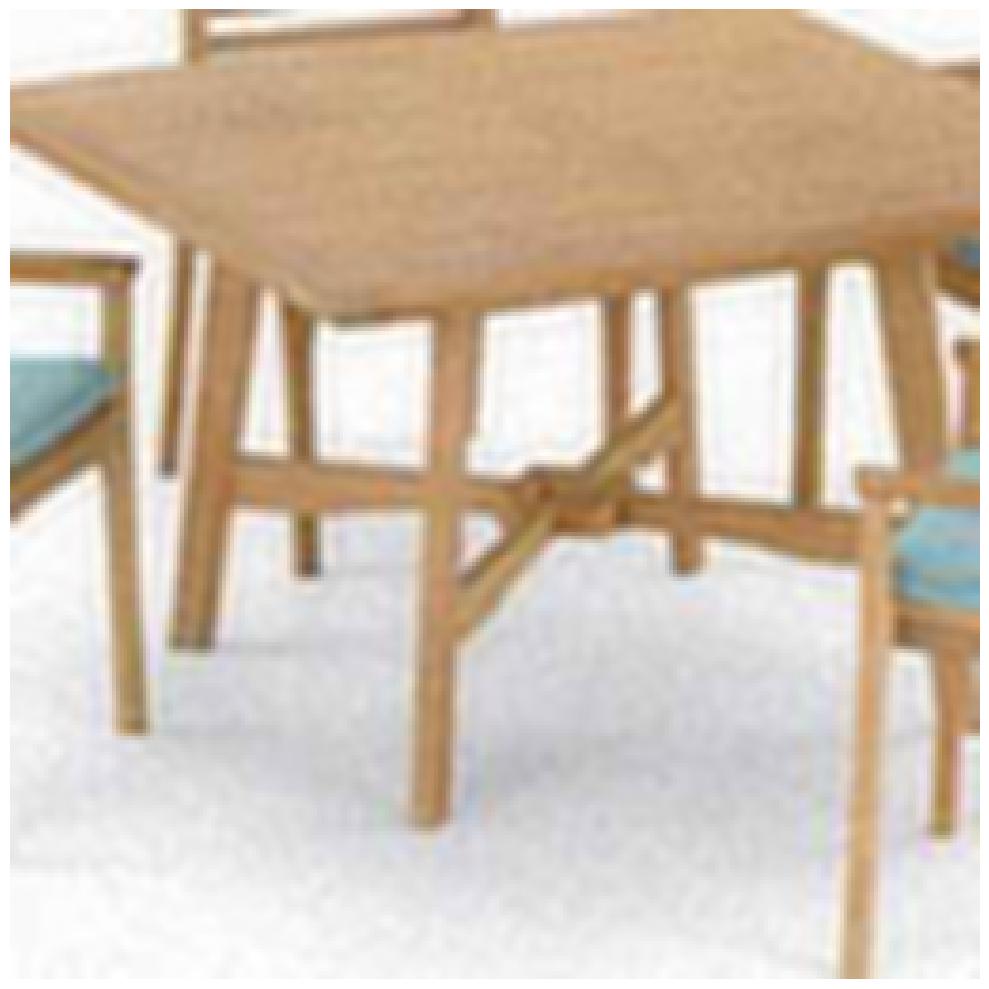}}
    \subfloat{\includegraphics[width=1.36in]{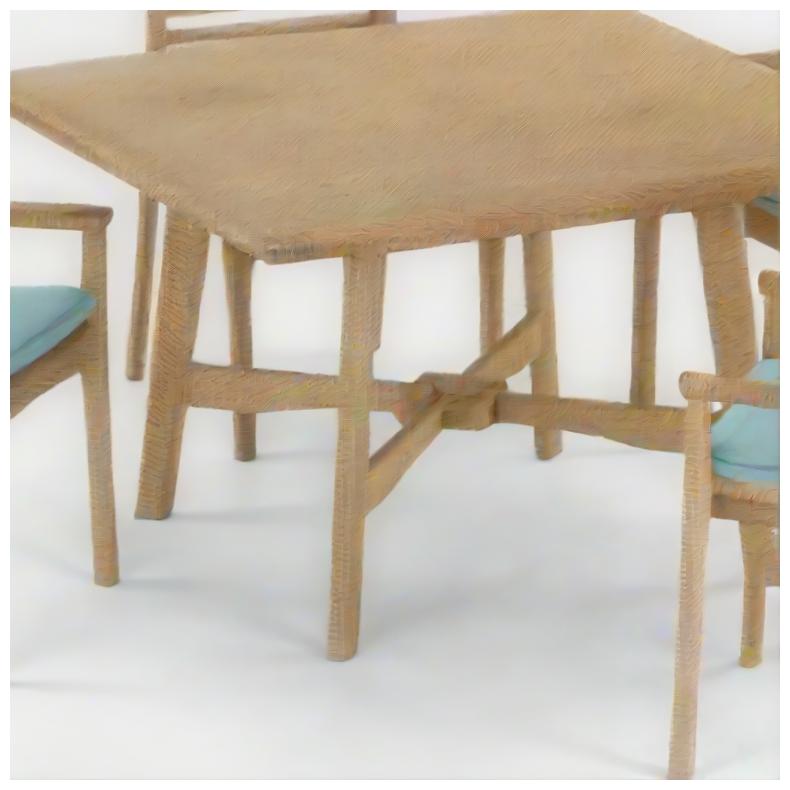}}
    \subfloat{\includegraphics[width=1.36in]{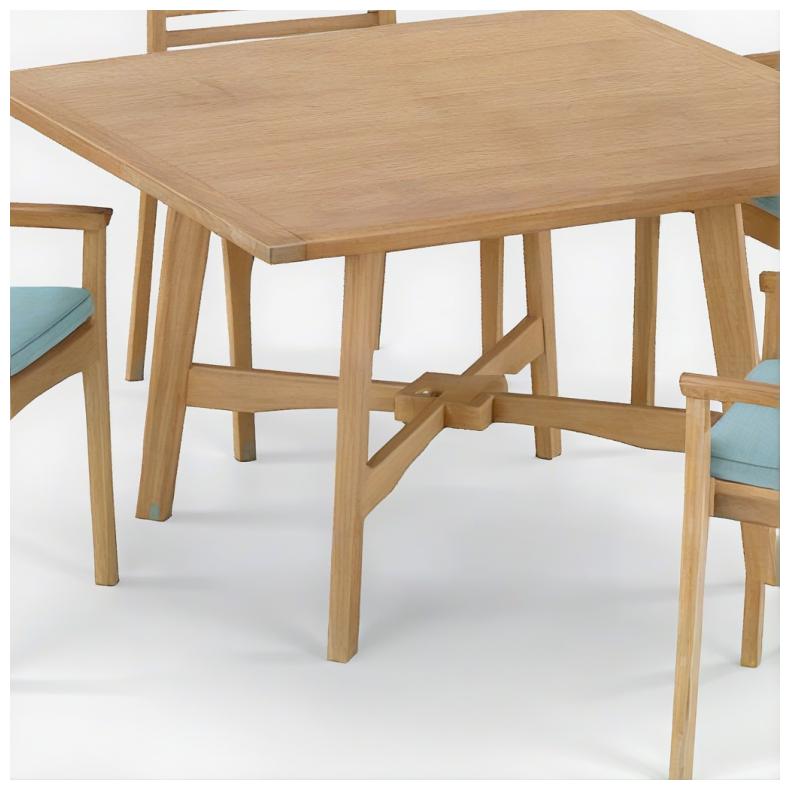}}
    \subfloat{\includegraphics[width=1.36in]{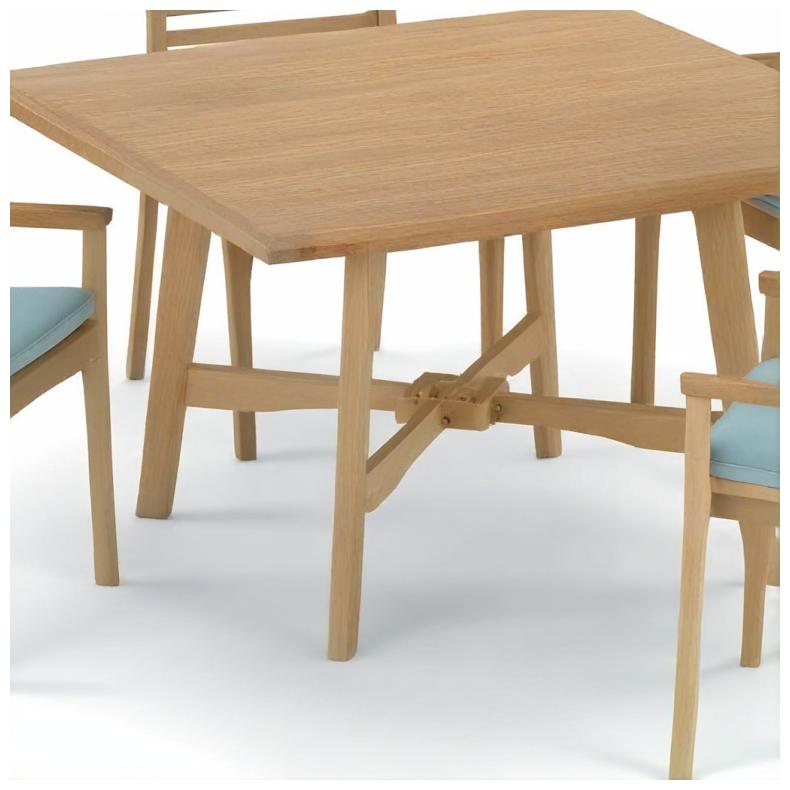}}
    \vspace{-1.2em}
    \subfloat{\includegraphics[width=1.36in]{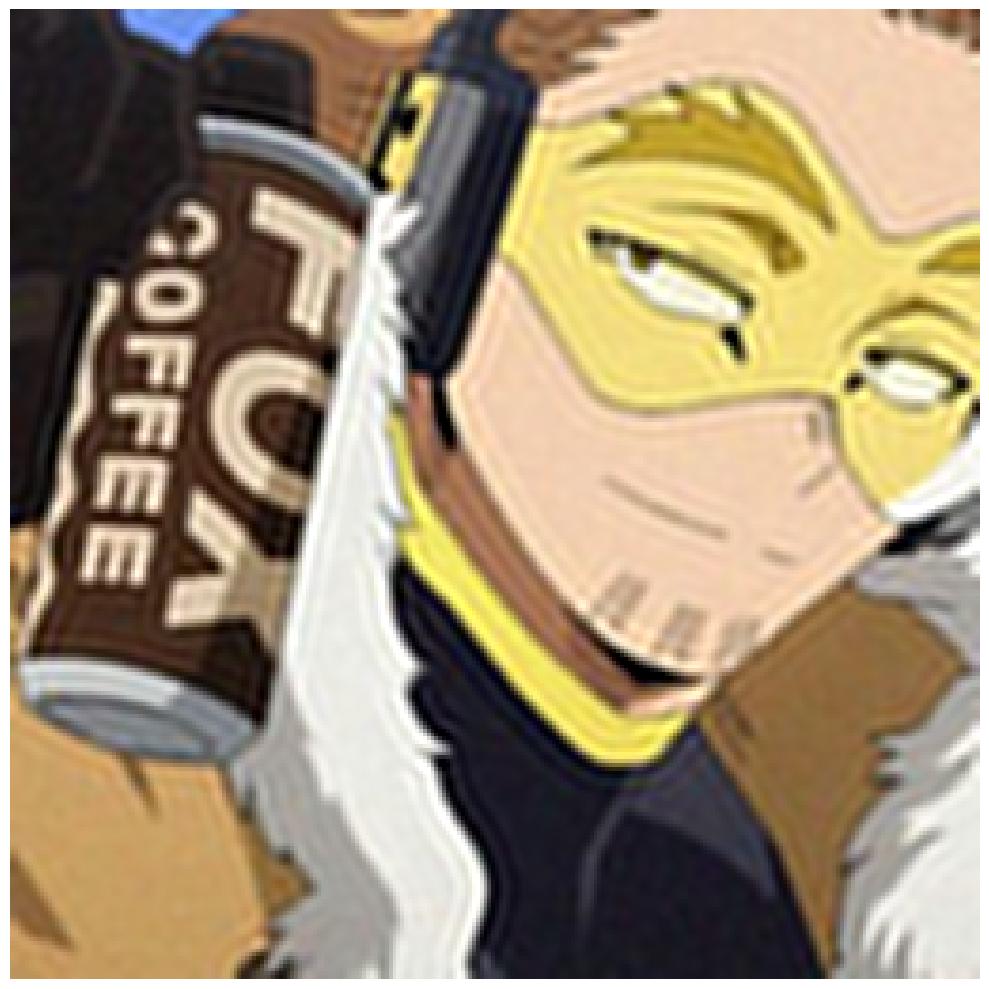}}
    \subfloat{\includegraphics[width=1.36in]{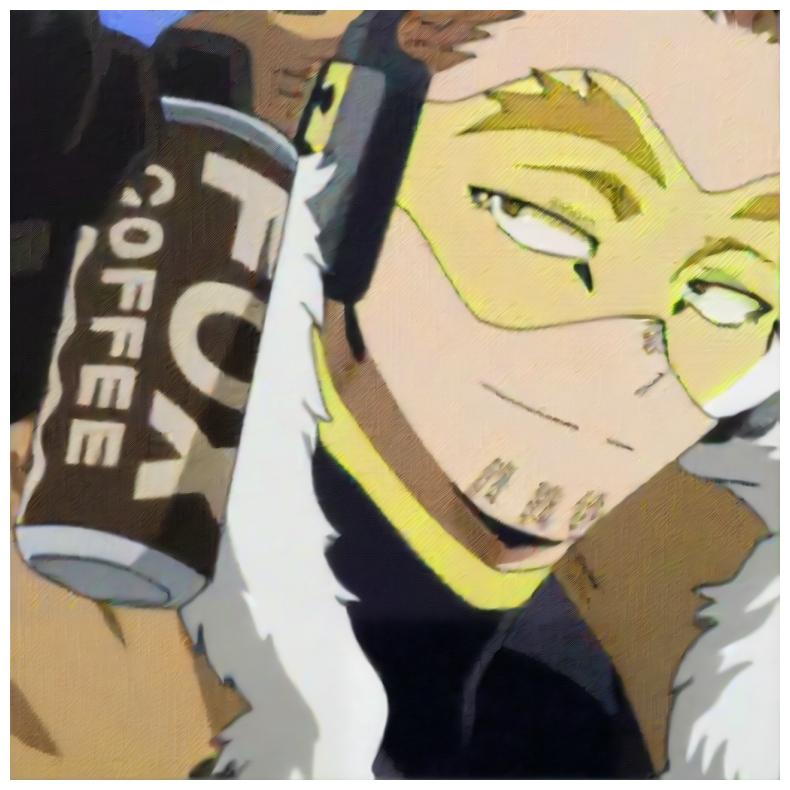}}
    \subfloat{\includegraphics[width=1.36in]{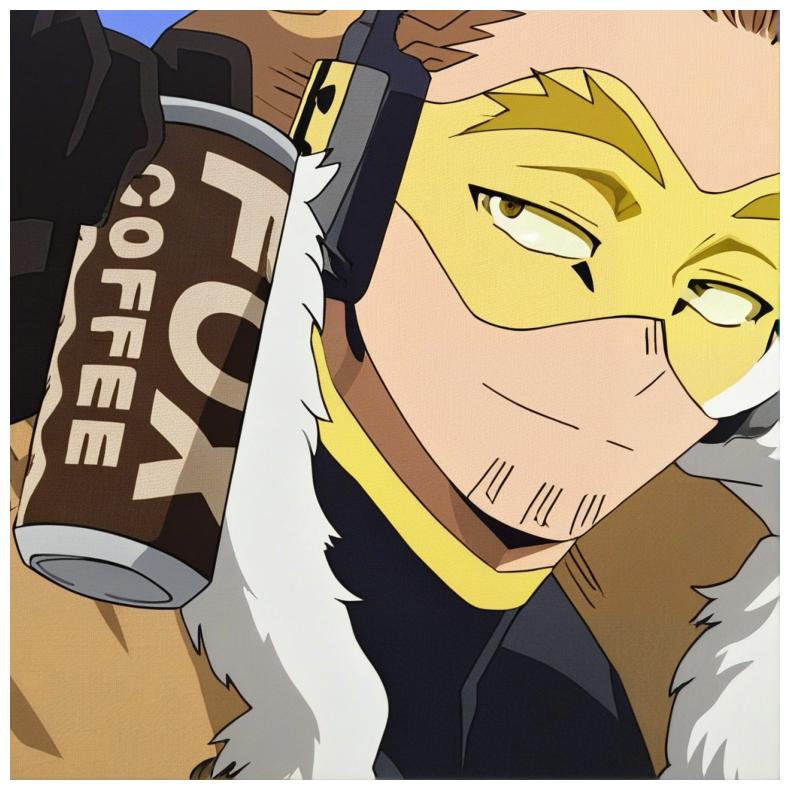}}
    \subfloat{\includegraphics[width=1.36in]{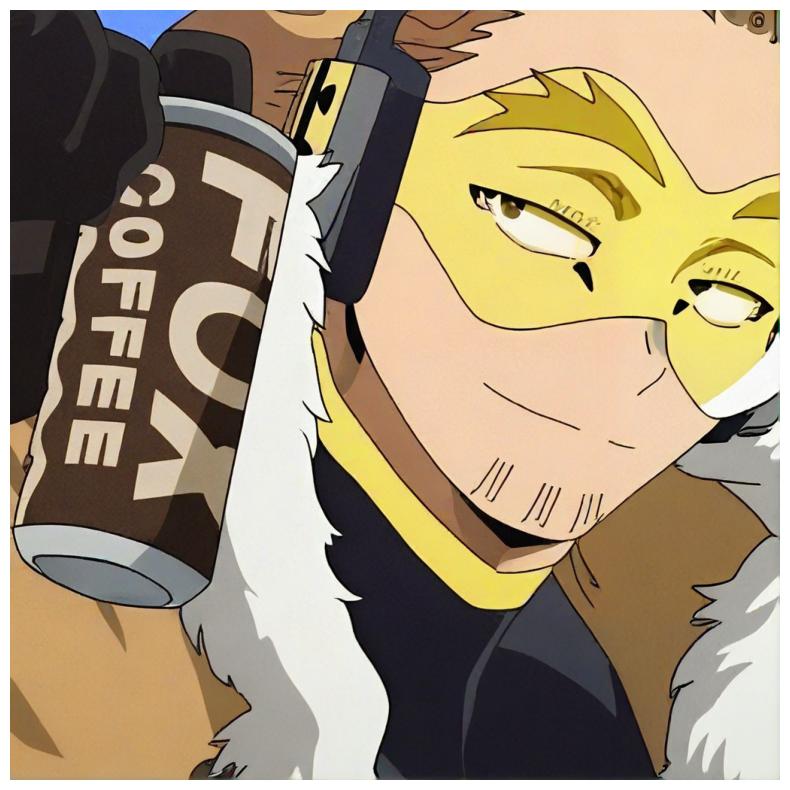}}
    \caption{Application of \emph{Flash Diffusion} to an \emph{in-house} diffusion-based \emph{super-resolution} model. Best viewed zoomed in.}
    \label{fig:app_upscaler}
    \end{figure*}

In Fig.~\ref{fig:app_upscaler}, we provide additional samples using the trained \emph{super-resolution} student model. As in the main paper, the student model is trained to output 1024x1024 images using 256x256 low-resolution images as conditioning. It is compared to the teacher generations using 4 steps (\emph{i.e.} 8 NFEs) and 20 steps (\emph{i.e.} 40 NFEs).

\subsection{Flash Swap}
\begin{figure*}[t]
      \centering
      \captionsetup[subfigure]{position=above, labelformat = empty}
      \subfloat[\scriptsize Source image]{\includegraphics[width=1.3in]{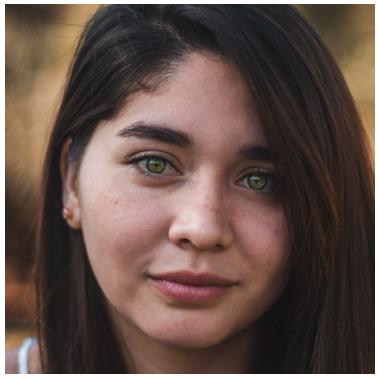}}
      \subfloat[\scriptsize Target image]{\includegraphics[width=1.3in]{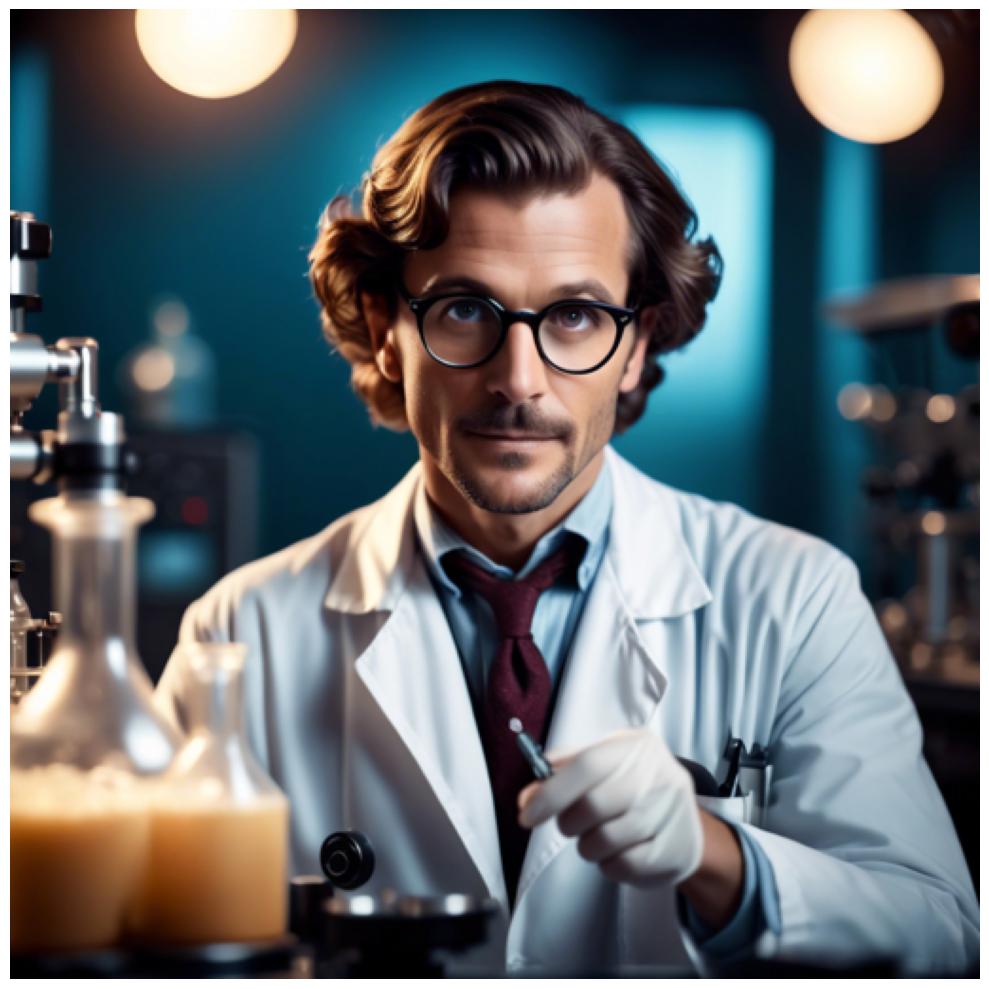}}
      \subfloat[\scriptsize Teacher (8 NFEs)]{\includegraphics[width=1.3in]{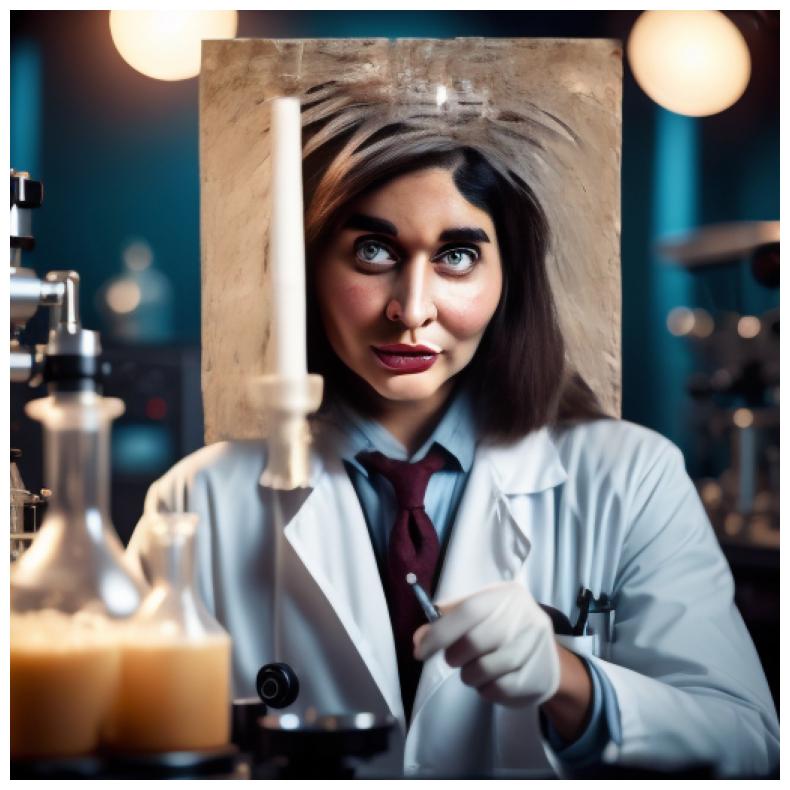}}
      \subfloat[\scriptsize Teacher (40 NFEs)]{\includegraphics[width=1.3in]{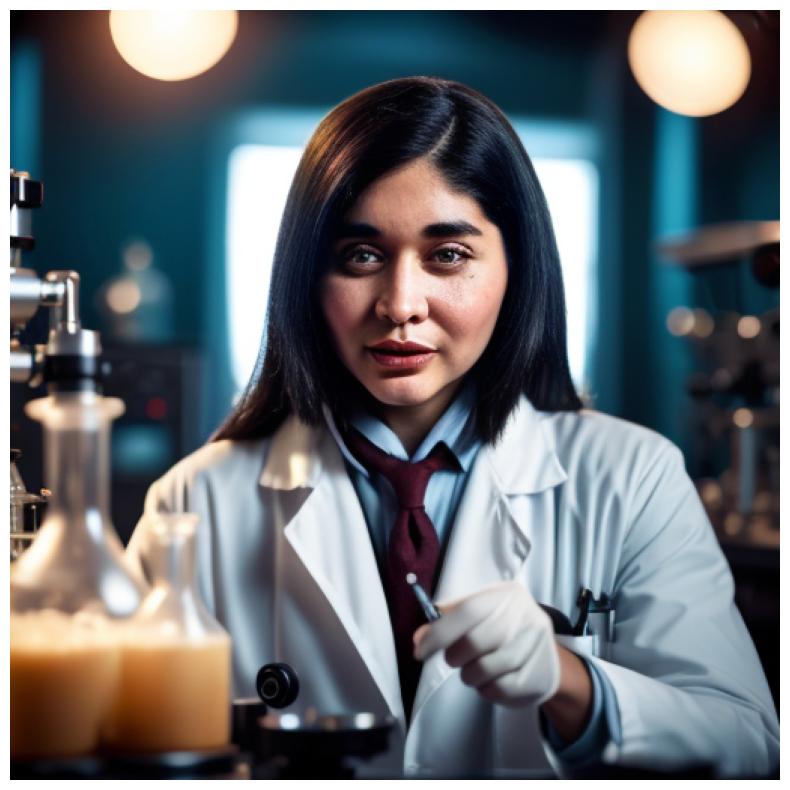}}
      \subfloat[\scriptsize Ours (4 NFEs)]{\includegraphics[width=1.3in]{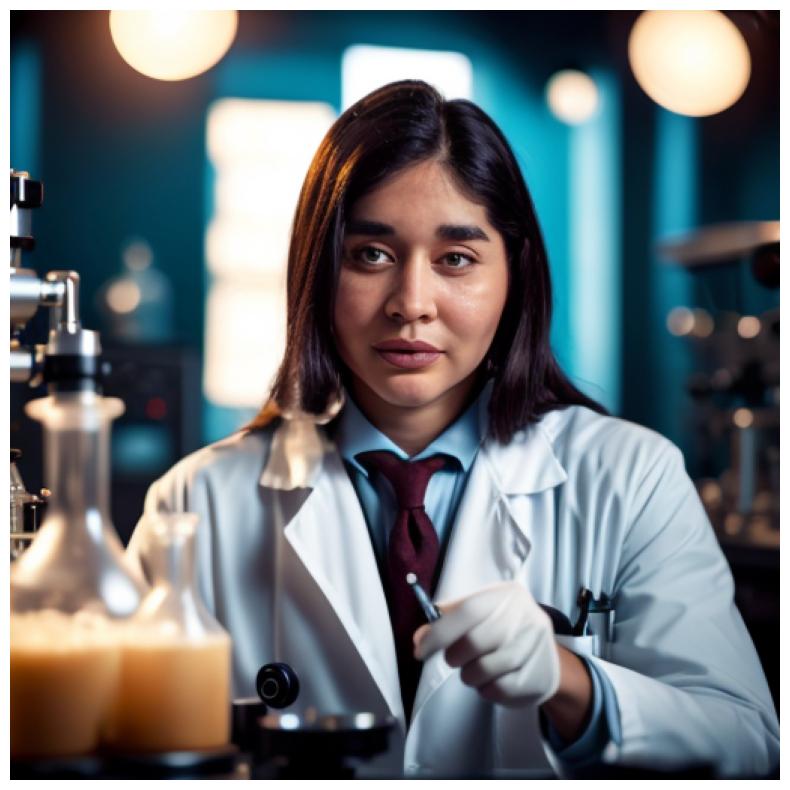}}\vspace{-1.2em}
      \subfloat{\includegraphics[width=1.3in]{plots/swap/1/source_image.jpg}}
      \subfloat{\includegraphics[width=1.3in]{plots/swap/1/target_image_1.jpg}}
       \subfloat{\includegraphics[width=1.3in]{plots/swap/1/teacher_4_steps.jpg}}
      \subfloat{\includegraphics[width=1.3in]{plots/swap/1/teacher_20_steps.jpg}}
      \subfloat{\includegraphics[width=1.3in]{plots/swap/1/student_4_steps.jpg}}
      \vspace{-1.2em}
      \subfloat{\includegraphics[width=1.3in]{plots/swap/3/source_image.jpg}}
      \subfloat{\includegraphics[width=1.3in]{plots/swap/3/target_image.jpg}}
       \subfloat{\includegraphics[width=1.3in]{plots/swap/3/teacher_4_steps.jpg}}
      \subfloat{\includegraphics[width=1.3in]{plots/swap/3/teacher_20_steps.jpg}}
      \subfloat{\includegraphics[width=1.3in]{plots/swap/3/student_4_steps.jpg}}
      \vspace{-1.2em}
        \subfloat{\includegraphics[width=1.3in]{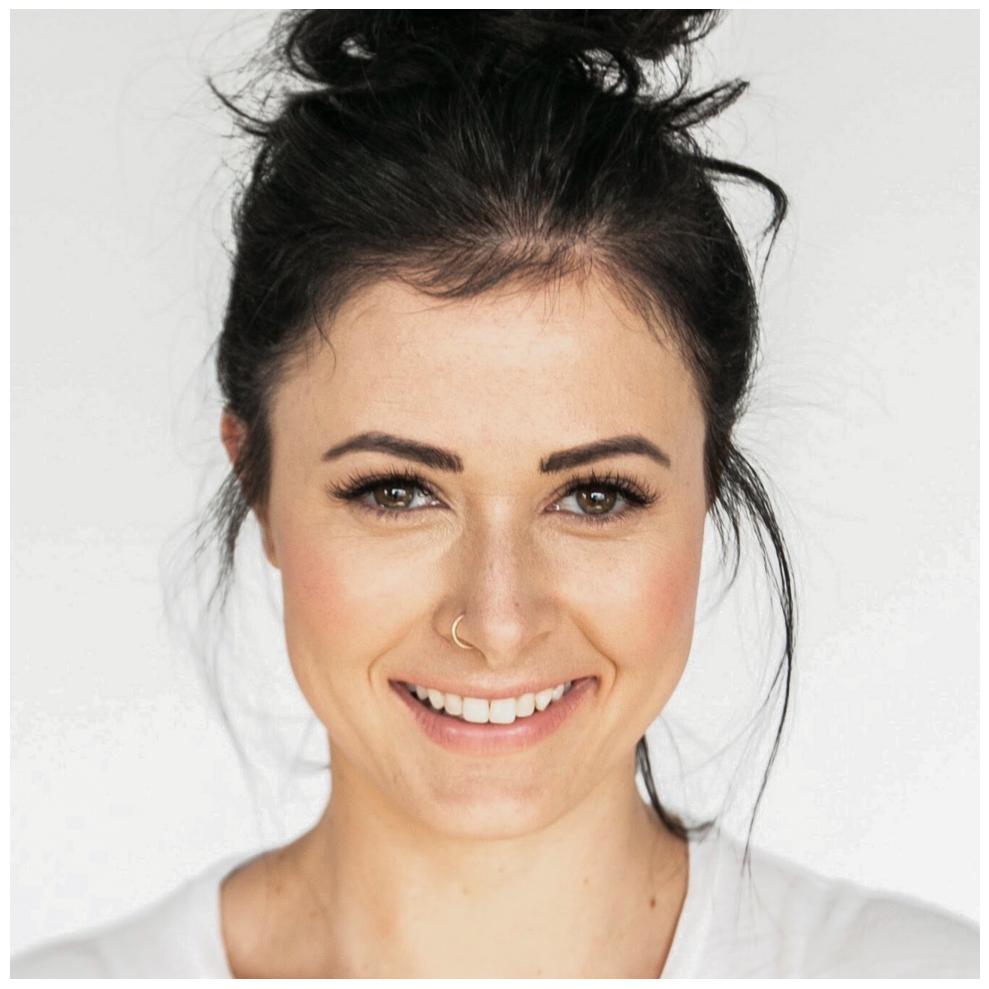}}
      \subfloat{\includegraphics[width=1.3in]{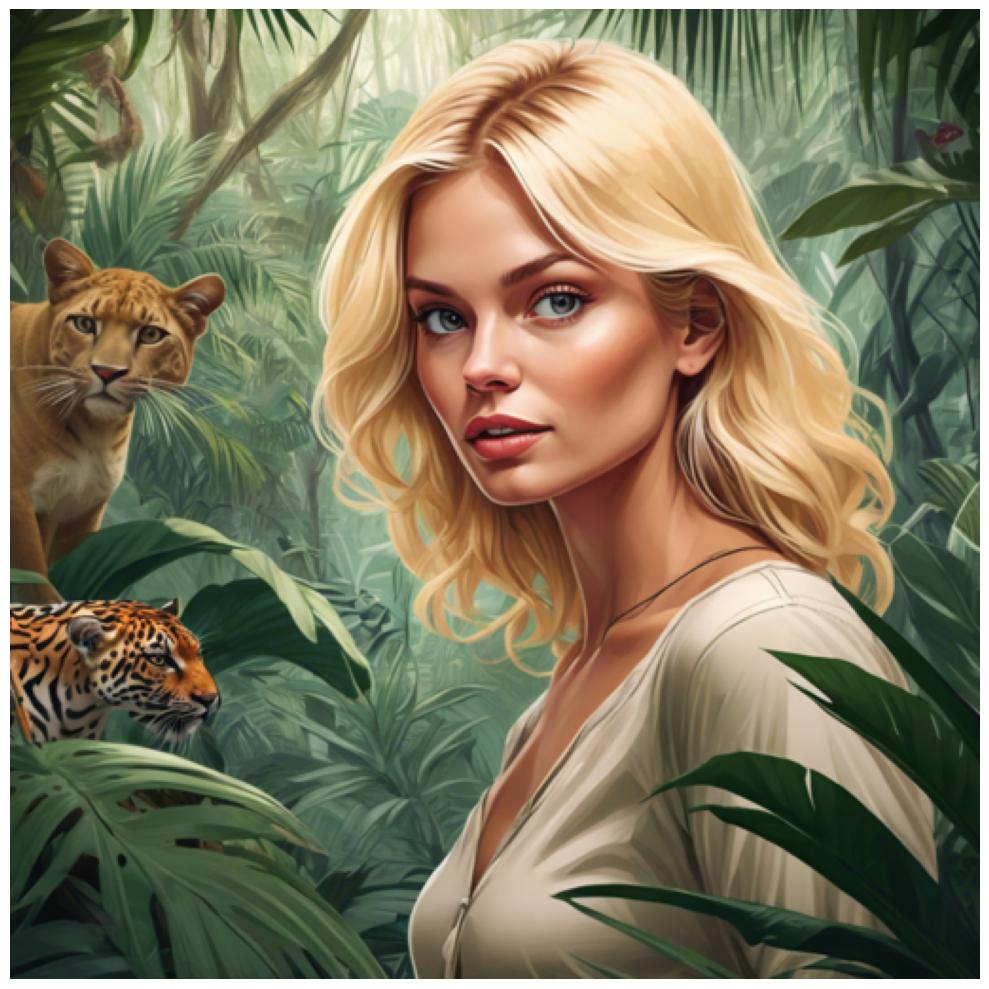}}
       \subfloat{\includegraphics[width=1.3in]{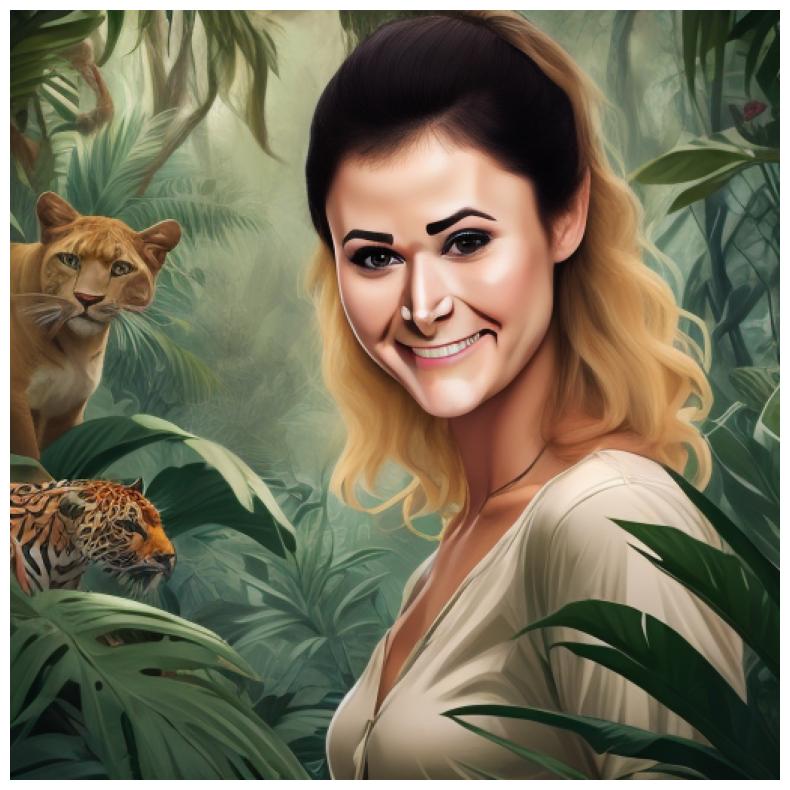}}
      \subfloat{\includegraphics[width=1.3in]{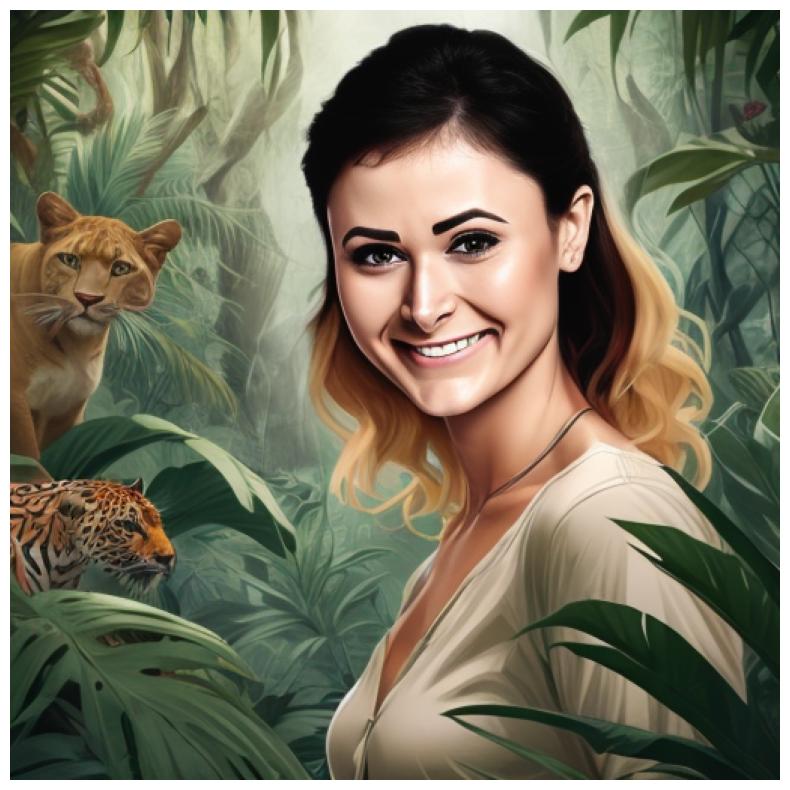}}
      \subfloat{\includegraphics[width=1.3in]{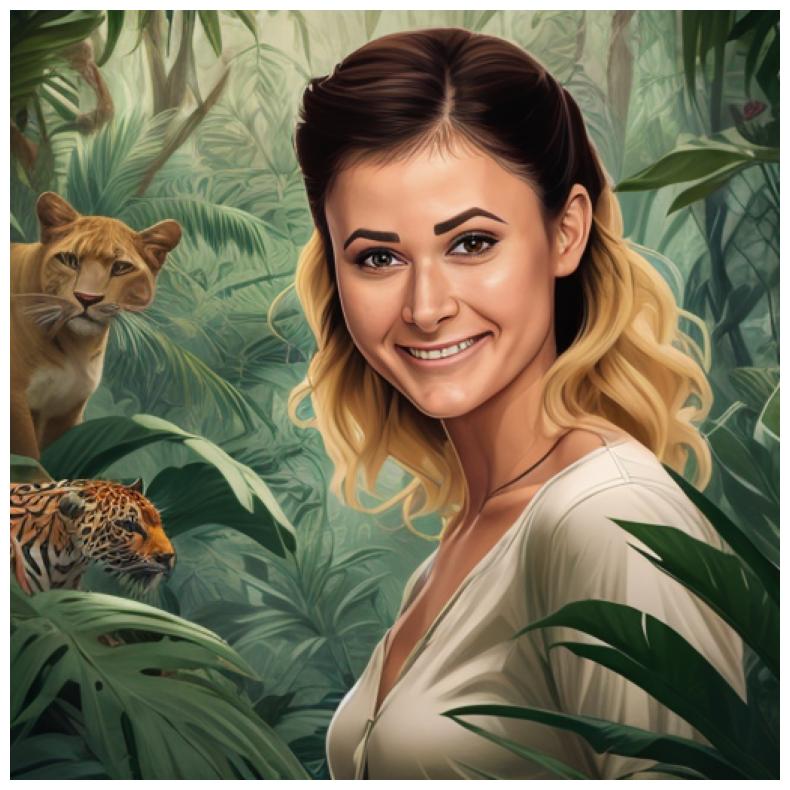}}
      \vspace{-1.2em}
      \subfloat{\includegraphics[width=1.3in]{plots/swap/5/source_image.jpg}}
      \subfloat{\includegraphics[width=1.3in]{plots/swap/5/target_image_5.jpg}}
       \subfloat{\includegraphics[width=1.3in]{plots/swap/5/teacher_4_steps.jpg}}
      \subfloat{\includegraphics[width=1.3in]{plots/swap/5/teacher_20_steps.jpg}}
      \subfloat{\includegraphics[width=1.3in]{plots/swap/5/student_4_steps.jpg}}
      \caption{Application of \emph{Flash Diffusion} to an \emph{in-house} diffusion-based \emph{face-swapping} model. Best viewed zoomed in.}
      \label{fig:swap}
      \end{figure*}

In Fig.~\ref{fig:swap}, we provide additional samples using the trained \emph{face-swapping} student model. The model is trained to replace the face of the person in the target image by the one of the person in the source image. It is compared to the teacher generations using 4 steps (\emph{i.e.} 8 NFEs) and 20 steps (\emph{i.e.} 40 NFEs).

\end{document}